%% file: main.tex
\documentclass[10pt,twocolumn,letterpaper]{article}

\usepackage[pagenumbers]{cvpr} % To force page numbers, e.g. for an arXiv version

\input{preamble}
\definecolor{cvprblue}{rgb}{0.21,0.49,0.74}
\usepackage[pagebackref,breaklinks,colorlinks,allcolors=cvprblue]{hyperref}

\usepackage{amssymb}
\usepackage{graphicx}
\usepackage{float}
\usepackage{pifont}
\usepackage{amsmath}
\usepackage{makecell}
\usepackage{cuted}      % provides the strip environment
\usepackage{multirow}
\usepackage[utf8]{inputenc}
\usepackage{colortbl} % For table coloring
\usepackage{xcolor} % For text coloring
\usepackage{tcolorbox} % For block boxes
\usepackage[table]{xcolor}
\usepackage[dvipsnames]{xcolor}
\usepackage[symbol]{footmisc}
\renewcommand{\thefootnote}{\fnsymbol{footnote}}
\usepackage{listings}
\usepackage{placeins}
\usepackage{dblfloatfix}
\usepackage{caption}
\usepackage{microtype}

\def\paperID{*****} % *** Enter the Paper ID here
\def\confName{CVPR}
\def\confYear{2026}

\title{Attention is Case-Sensitive}

\author{
  Maximilian Dillitzer$^{1,3,*}$ \qquad
  Tin Stribor Sohn$^{2,3,*}$ \qquad
  Jason J. Corso$^4$ \qquad
  Michael Auerbach$^1$ \\[0.6em]
  $^1$University of Applied Science Esslingen \qquad
  $^2$Karlsruhe Institute of Technology \\[0.2em]
  $^3$Dr. Ing. h.c. F. Porsche AG \qquad
  $^4$University of Michigan \\[0.5em]
  {\small $^*$Equal contribution} \\
}

\begin{document}

\twocolumn[{%
  \renewcommand\twocolumn[1][]{#1}% Prevents nested \twocolumn error from \maketitle
  \maketitle
  \begin{center}
    \centering
    \includegraphics[width=0.98\linewidth]{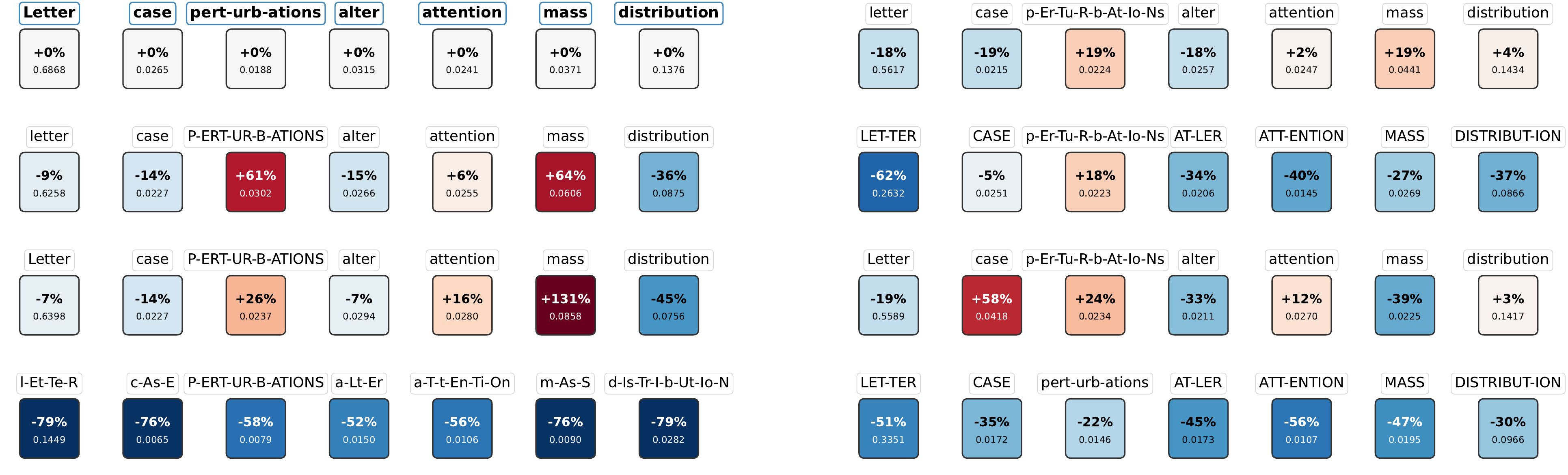}
    \captionof{figure}{\textbf{Letter case as an attentional attractor.} LLMs (here, Qwen2.5-7B-Instruct) systematically shift internal attention weights based on typographic casing, even when semantic content remains identical. We show this property is inherent to the model's pretraining and impacts internal state dynamics. Visualizations represent \textit{relative attention mass} compared to a baseline reference sentence (top-left), which is calculated as the \textit{mean attention weight per token} aggregated at the word level.}
    \label{fig:we_can_steer_attention}
  \end{center}
  \vspace{1.5em} % Adjust spacing between caption and abstract
}]

\input{02_Main/01_abstract}    
\input{02_Main/02_intro}
\input{02_Main/03_relwork}
\input{02_Main/04_method}
\input{02_Main/05_experiments}

\input{02_Main/06_conclusion}

{
    \small
    \bibliographystyle{ieeenat_fullname}
    \bibliography{main}
}

\input{02_Main/X_supplementary}

\end{document}

%% file: 02_Main/01_abstract.tex
%%%%%%%%%%%%%%%%%%%%%%%%%%%%%%%%%%%%%%%%%%% ABSTRACT %%%%%%%%%%%%%%%%%%%%%%%%%%%%%%%%%%%%%%%%%%%
\begin{abstract}
    In human visual perception, uppercase lettering serves as a natural salience cue that captures attention within lowercase text. In this paper, we present a systematic empirical characterization study revealing that Large Language Models (LLMs) exhibit an analogous property: letter casing modulates internal attention allocation. Through analysis across 13 models---nine LLMs and four Vision-Language Models (VLMs)---with diverse tokenization schemes, we show that formatting target information in alternating or uppercase against a lowercase context concentrates attention on those textual spans. In text this effect is universal, holding across every evaluated non-reasoning model. We frame it as a previously under-explored latent property of pretrained transformers rather than a prescriptive method. Our investigation reveals a central attention–performance divergence: while this ``casing effect'' robustly shifts attention, its impact on downstream accuracy is non-trivial---increased concentration does not inherently improve task accuracy and, in high-entropy contexts like alternating case, can degrade it. We further identify a boundary condition: the deliberative ``thinking'' phase in reasoning models acts as a semantic buffer that mitigates typographic sensitivity in text. Extending the study to VLMs, we find the effect transfers partially: the same prompt-side casing reorganizes cross-modal attention along two coupled axes---predominantly a macroscopic disengagement from the image toward the text prompt, and secondarily a concentration of the residual visual attention on the target region. By isolating casing as a zero-shot mechanism for attention steering that requires no model access or fine-tuning, we provide a new foundational understanding of how pretraining internalizes typographic emphasis.
\end{abstract}

%% file: 02_Main/02_intro.tex
%%%%%%%%%%%%%%%%%%%%%%%%%%%%%%%%%%%%%%%%%%% INTRODUCTION %%%%%%%%%%%%%%%%%%%%%%%%%%%%%%%%%%%%%%%%%%%
\section{Introduction}
\label{sec:intro}

Human visual perception exhibits a well-documented sensitivity to typographic variation: uppercase lettering naturally draws attention within predominantly lowercase text~\cite{Inhoff_2000_Springer_Uppercase,Slattery_2011_APA_UppercaseFixation,Cutter_2020_APA_Capitalization}. Electrophysiological evidence further supports the pre-lexical nature of case-based attentional modulation, establishing case as a surface feature that guides human focus~\cite{Vergara_2020_Lettercase,Fournet_2022_APA_Effects}. This phenomenon, rooted in the visual salience of capital letters against their lowercase counterparts, has long been exploited in typography, advertising, and interface design to manipulate cognitive priority without altering semantic content~\cite{Klinke_2024_Advertising}.

This human perceptual mechanism raises a compelling question for artificial language understanding: do Large Language Models (LLMs), despite operating on abstract token representations rather than visual stimuli, exhibit analogous case-sensitive attention patterns? While case variation is conventionally treated as a passive artifact of tokenization, we hypothesize that LLMs internalize typographic case as a latent importance signal during pretraining. In this paper, we present a systematic empirical study revealing that letter casing is not merely a formatting choice but a fundamental property that modulates internal attention allocation within the transformer architecture.

Our investigation focuses on characterizing this ``casing effect'' by isolating letter case as the \textit{independent variable} at the pre-tokenization level while strictly preserving word order and semantic context (\cf \cref{fig:we_can_steer_attention}). By analyzing seven non-reasoning and two reasoning LLMs alongside two non-reasoning and two reasoning Vision-Language Models (VLMs) with diverse scales and tokenizer types, we observe that formatting target information in alternating or uppercase against a lowercase context consistently concentrates internal attention weights on those text spans and corresponding image regions. 

Importantly, we frame this discovery as an inherent behavioral characteristic of pretrained LLMs and VLMs rather than a prescriptive steering method intended to optimize benchmark performance. Our results show that while the shift in attention is robust and universal, its impact on downstream performance is non-trivial: increased attention concentration can lead to both performance gains and degradations. This suggests that while case acts as a powerful attentional attractor, ``more attention'' does not inherently equate to better performance. Notably, we find that the deliberative reasoning process in reasoning models acts as a semantic realignment layer in text that filters out these typographic cues in favor of logical coherence. Crucially, we approach experiments with zero model access---no fine-tuning, adapter insertion, or runtime introspection---operating entirely through prompt engineering (\ie, only changing letter case).
Our work makes the following contributions to the foundational understanding of LLM and VLM properties: 
\begin{itemize}
    \item We provide empirical evidence that LLMs and VLMs exhibit a ``casing-salience'' property, where attention is meaningfully modulated by typographic case, mirroring human visual salience effects across diverse architectures.
    \item We demonstrate the robustness of this property across seven non-reasoning LLMs, proving the effect persists regardless of tokenization scheme or scale.
    \item We observe that this pattern also applies to multimodal settings, where image attention in VLMs is steered by letter casing of the textual description.
    \item We conduct a causal analysis using adversarial de-emphasis (\eg, uppercasing context while lowercasing targets), confirming that the observed attention shifts are a direct response to orthographic variation.
    \item We reveal a complex relationship between attention allocation and downstream performance, showing that while case can be utilized to influence model focus, the resulting impact on task accuracy is highly context-dependent.
    \item We identify the ``reasoning buffer'' as a boundary condition in LLMs, where reasoning mitigates the influence of surface-level typographic attractors.
\end{itemize}
By characterizing these latent attentional levers, we move beyond viewing case as a tokenization nuance and establish it as a surface-level property that can be utilized for future research in attention steering and model interpretability.

%% file: 02_Main/03_relwork.tex
%%%%%%%%%%%%%%%%%%%%%%%%%%%%%%%%%%%%%%%%%%% RELATED WORK %%%%%%%%%%%%%%%%%%%%%%%%%%%%%%%%%%%%%%%%%%%
\section{Related Work}
\label{sec:sota}

The internal attention mechanisms of LLMs and VLMs are the primary drivers of their reasoning and generative capabilities. Consequently, a vast body of research has emerged dedicated to ``attention steering''---the intentional redirection of a model's focus to improve safety, factual recall, or task performance. Literature typically treats attention as a variable manipulated through external interventions. We categorize these into training-time, inference-time, and activation-level modifications, highlighting the gap our empirical study addresses.  

%%%%%%%%%%%%%%%%%%%%%%%%%%%%%%%%%%%%%%%%%%%%%%%%%%%% RELATED WORK SUBSECTION %%%%%%%%%%%%%%%%%%%%%%%%%%%%%%%%%%%%%%%%%%%%%%%%%%%%
\subsection{Engineered Attention Steering}

Existing strategies for controlling attention typically require deep architectural or computational interventions:

\paragraph{Training-Time and Structural Modifications.}
Parameter-level interventions such as ROME~\cite{Meng_2022_NIPS_ROME}, MEMAT~\cite{Tamayo_2024_ACL_MEMAT}, MEND~\cite{Mitchell_2022_arXiv_MEND}, or REMEDI~\cite{Hernandez_2024_arXiv_REMEDI} attempt to ``hard-code'' attention patterns by editing weights to alter factual recall. Likewise, bias-only adaptation techniques train per-layer steering vectors via reinforcement learning, effectively tuning the model for specific reasoning tasks without full fine-tuning~\cite{Sinii_2025_arXiv_SteeringLLMReasonging}. Other approaches, such as Value-State Gated Attention (VGA)~\cite{Bu_2026_arXiv_VGA}, propose architectural redesigns to mitigate pathological attention behaviors. While effective, these studies treat attention as a structural feature that must be modified during training to achieve desired behaviors.

\paragraph{Inference-Time Computational Interventions.}
Recent works like PASTA~\cite{Zhang_2024_arXiv_PASTA}, AutoPASTA~\cite{Zhang_2024_arXiv_AutoPASTA}, and InstABoost~\cite{Guardieiro_2025_arXiv_InstABoost} introduce mechanisms to rescale attention scores during the forward pass. These methods ``steer'' the model by manually amplifying the weights of specific tokens (\eg, instruction tokens). Similarly, SEKA~\cite{Anonymous_2026_OpenReview_SEKA} edits key-embeddings before the attention computation occurs and Hsieh \etal{}~\cite{Hsieh_2024_ACL_FoundInTheMiddle} propose positional calibration to subtract a learned baseline from positional attention bias. While these operate on frozen models, they treat attention as a computational resource that requires white-box hooks and runtime recomputation to be redirected.

\paragraph{Activation Engineering.}
A third category involves injecting ``steering vectors'' into the model's residual stream to encourage specific traits~\cite{Turner_2024_arXiv_ActAdd,vonRuette_2024_arXiv_ActivationSteering,Lee_2025_arXiv_CAST,Stolfo_2025_arXiv_InstructionSteering,Venkateswaran_2026_arXiv_SpotLight}. These techniques rely on extracting activation patterns from contrastive datasets and adding them to internal layers during inference~\cite{Davarmanesh_2026_arXiv_EfficientSteering}, focussing on manipulating the \textit{content} of the hidden states to influence attention, rather than exploring the model's natural response to input variations.

%%%%%%%%%%%%%%%%%%%%%%%%%%%%%%%%%%%%%%%%%%%%%%%%%%%% RELATED WORK SUBSECTION %%%%%%%%%%%%%%%%%%%%%%%%%%%%%%%%%%%%%%%%%%%%%%%%%%%%
\subsection{Input-Level Influence and Surface Form}

Studies on how input formatting affects LLMs and VLMs has largely focused on semantic or structural perturbations. Researchers have documented positional bias~\cite{Wu_2025_arXiv_PositionBias}, sink token effects~\cite{Li_2025_arXiv_CTR_SinkTokens}, and other token-level effects~\cite{Phan_2024_arXiv_TokenBias,Sinclair_2022_TACL_StructuralPriming}, or used XML-like delimiters to provide structural cues for instructions~\cite{Alpay_2025_arXiv_XLMPrompting}. These works generally operate on the level of \textit{semantic priming}---adding tokens or changing order to influence the model but omitting the effects analysis on vision modalities entirely.
In contrast, our work focuses on \textit{orthographic variation} that preserves 100\% of the lexical and semantic content alongside unchanged visual inputs in VLMs. While prior work treats casing as a variable to be normalized away during tokenization, our study reveals it as a potent, inherent attentional attractor.

%%%%%%%%%%%%%%%%%%%%%%%%%%%%%%%%%%%%%%%%%%%%%%%%%%%% RELATED WORK SUBSECTION %%%%%%%%%%%%%%%%%%%%%%%%%%%%%%%%%%%%%%%%%%%%%%%%%%%%
\subsection{The Gap: Characterizing Case Sensitivity}

The critical gap in current research is the lack of understanding regarding the \textit{inherent} sensitivity of pretrained transformers to typographic surface forms. While the aforementioned steering methods aim to \textit{improve performance} through complex interventions, they overlook a fundamental property of models themselves:
\begin{enumerate}
    \item \textbf{Attention vs. Performance}: Prior works assume that redirecting attention to relevant tokens will monotonically improve accuracy. Our empirical analysis of the ``casing effect'' provides a more nuanced view, showing that while attention is steered by casing, the resulting task performance depends on the ``readability'' of the pattern (\eg, uppercase vs. alternating case).
    \item \textbf{Latent Properties}: Most steering research focuses on \textit{prescriptive methods} (how to force the model to look elsewhere). We provide a \textit{descriptive characterization} of a property that already exists in pretrained weights across nine diverse LLMs and four VLMs.
\end{enumerate}
By documenting how letter casing (\ie, the minimal orthographic perturbation) systematically reshapes attention flow, we bridge cognitive typography and transformer interpretability. This work shifts the perspective from ``how can we force attention to move'' to ``what existing features of text naturally attract it''.

%% file: 02_Main/04_method.tex
\section{Research Methodology}
\label{sec:methodology}

Inspired by human visual salience effects where uppercase lettering naturally attracts attention within lowercase text, we conduct a \emph{controlled empirical study} to characterize whether letter case functions as an inherent attentional modulator in pretrained LLMs and VLMs. We deliberately adopt a ground-truth-guided protocol to isolate the pure effect of typographic case from confounding factors, such as span prediction errors. Our objective is the foundational validation of this model property; we treat the observed attention shifts as a behavioral characteristic rather than a prescriptive method. Consequently, while we demonstrate how this property can be utilized for steering, a fully automated pipeline for deployment is outside the scope of this characterization study (\cf \cref{sec:conclusion}).

Our protocol is entirely training-free and input-driven. We define static capitalization interventions applied \emph{before tokenization}, measure their causal impact on internal attention distribution (textual and visual attention) and downstream task performance under controlled conditions, and verify the universality of these effects across diverse model families.

%%%%%%%%%%%%%%%%%%%%%%%%%%%%%%%%%%%%%%%%%%%%%%%%%%%% METHOD SUBSECTION %%%%%%%%%%%%%%%%%%%%%%%%%%%%%%%%%%%%%%%%%%%%%%%%%%%%
\subsection{Characterization Scope and Causal Isolation}

Given raw input text $X = (x_1, \dots, x_N)$ as a sequence of characters and either a \emph{ground-truth answer span} or a \textit{bounding box label} $A \subseteq X$ provided by benchmark annotations, we construct a typographically transformed version $\tilde{X}$ such that:
\begin{enumerate}
    \item $\tilde{X}$ differs from $X$ \emph{only} in letter case, preserving word order and semantic content of the text exactly;
    \item $\tilde{X}$ applies a \emph{deterministic rule} formatting $A$ in a target case pattern (\eg, uppercase) while the surrounding context follows a complementary pattern;
    \item when processed by a frozen LLM or VLM $f_\theta$, $\tilde{X}$ yields measurable shifts in cross-layer attention concentration on $A$ (which can be text or text-corresponding image regions) and changes in task accuracy relative to naturally cased $X$.
\end{enumerate}
This defines a \emph{pre-tokenization causal isolation protocol} which similarly applies for vision tasks incorporating text captions. The ground-truth dependency is intentional: it eliminates noise from auxiliary models or heuristics, allowing us to attribute any observed changes in model state specifically to the typographic variation.

%%%%%%%%%%%%%%%%%%%%%%%%%%%%%%%%%%%%%%%%%%%%%%%%%%%% METHOD SUBSECTION %%%%%%%%%%%%%%%%%%%%%%%%%%%%%%%%%%%%%%%%%%%%%%%%%%%%
\subsection{Target Span Identification Protocol}
\label{sec:span_identification}

To ensure reproducibility, target spans are derived strictly from benchmark-provided labels using a two-stage matching protocol:
\begin{itemize}
    \item \textbf{Exact matching (98.2\% of samples)}: We locate $A$ in $X$ via case-insensitive matching. The transformation is applied to the matched character span while preserving non-alphabetic characters (punctuation, digits, whitespace).
    \item \textbf{Fuzzy fallback (1.8\% of samples)}: If exact matching fails, we apply Levenshtein-distance alignment (threshold $\leq 2$ character edits), ensuring alignment fidelity through a 100-sample subset manual verification.
\end{itemize}
Crucially, these transformations preserve token boundaries; we do not alter whitespace or lexical structure. For inherently mixed-case entities (\eg, ``iPhone''), we apply the transformation to the entire span (\eg, ``IPHONE'') to maintain consistency in the experimental signal.

%%%%%%%%%%%%%%%%%%%%%%%%%%%%%%%%%%%%%%%%%%%%%%%%%%%% METHOD SUBSECTION %%%%%%%%%%%%%%%%%%%%%%%%%%%%%%%%%%%%%%%%%%%%%%%%%%%%
\subsection{Typographic Intervention Formalism}
\label{sec:case_transformation}

Let $\mathcal{C}: \Sigma \rightarrow \{ \text{upper}, \text{lower}, \text{title}, \text{alternating} \}$ denote a case assignment function over character alphabet $\Sigma$. Our intervention constructs:
\begin{equation}
    \tilde{X} = \mathcal{T}_{\mathcal{C}}(X) = \big( c_1(x_1),\; c_2(x_2),\; \dots,\; c_N(x_N) \big),
    \label{eq:case_transform}
\end{equation}
where $c_i \in \{\text{upper}, \text{lower}, \text{title}, \text{alternating}\}$ is applied per character, satisfying the semantic preservation constraint:
\begin{equation}
    \forall i,\quad \texttt{lower}\big(c_i(x_i)\big) = \texttt{lower}(x_i).
    \label{eq:semantic_preservation}
\end{equation}
This ensures lexical identity remains invariant, even if the underlying tokenizer produces a different sequence of token IDs for $\tilde{X}$ compared to $X$.

Our study tests the hypothesis that pretrained LLMs and VLMs exhibit an inherent sensitivity to case as an importance signal. Specifically, we investigate the following relationships:
\begin{equation}
\begin{aligned}
    \text{Att}_A(f_{\theta}(\tilde{X})) &\neq \text{Att}_A(f_{\theta}(X)) \\
    \text{and in text} \quad \text{Acc}(f_{\theta}(\tilde{X})) &\neq \text{Acc}(f_{\theta}(X)),
\end{aligned}
\label{eq:sensitivity_hypothesis}
\end{equation}
where $\text{Att}_A(\cdot)$ denotes the mean attention weight over target $A$ across all layers/heads, and $\text{Acc}(\cdot)$ is the task accuracy of the frozen model $f_{\theta}$. Crucially, in VLMs, the target $A$ generalizes from discrete textual tokens to spatial image regions, specifically mapping to corresponding patches or pixels. Unlike traditional method-driven papers, we do not assume a monotonic improvement; rather, we characterize the \textit{directionality} and \textit{magnitude} of these shifts to reveal the model's internal response to typographic emphasis.

%%%%%%%%%%%%%%%%%%%%%%%%%%%%%%%%%%%%%%%%%%%%%%%%%%%% METHOD SUBSECTION %%%%%%%%%%%%%%%%%%%%%%%%%%%%%%%%%%%%%%%%%%%%%%%%%%%%
\subsection{Cross-Architecture Robustness Protocol}
\label{sec:generalization}

To evaluate the generality of this property, we apply identical interventions across nine non-reasoning models and four reasoning models spanning five families (LLaMA-3~\cite{Grattafiori_2024_arXiv_LLaMA3}, Gemma-2/3/4~\cite{Gemma2_arXiv_TechnicalReport,Gemma3_arXiv_TechnicalReport,google_gemma_core_en}, Mistral~\cite{Jiang_2023_arXiv_Mistral7b}, Qwen2.5/3~\cite{Qwen_2024_arXiv_Qwen2,Qwen_2025_arXiv_Qwen3,bai2025qwen3vltechnicalreport}, GPT~\cite{openai_2025_gptoss20b}) and various tokenization schemes. No per-model tuning is performed. By measuring relative changes against each model's naturally cased baseline (as in the benchmarks' protocol), we determine if case sensitivity is a universal emergent property of the transformer architecture or a training regime artifact.

%%%%%%%%%%%%%%%%%%%%%%%%%%%%%%%%%%%%%%%%%%%%%%%%%%%% METHOD SUBSECTION %%%%%%%%%%%%%%%%%%%%%%%%%%%%%%%%%%%%%%%%%%%%%%%%%%%%
\subsection{Principles of the Empirical Study}

Our experimental protocol is built on three principles:
\begin{enumerate}
    \item \textbf{Causal Attribution}: Isolation of letter case as the independent variable ensures effects are not due to semantic, structural, or visual changes.
    \item \textbf{Zero-Inference Observation}: We analyze the behavior of frozen, off-the-shelf models to identify properties already present in the weights.
    \item \textbf{Phenomenological Analysis}: We report both the successes (performance gains) and the failures (performance degradation) of attention concentration to provide a complete picture of the ``casing effect'' in text.
\end{enumerate}
By framing our research methodology as a characterization study, we provide the groundwork for future work to utilize these latent attentional levers in more complex, automated steering applications.

%% file: 02_Main/05_experiments.tex
%%%%%%%%%%%%%%%%%%%%%%%%%%%%%%%%%%%%%%%%%%%%%%%%%%%% EXPERIMENTS %%%%%%%%%%%%%%%%%%%%%%%%%%%%%%%%%%%%%%%%%%%%%%%%%%%%
\section{Experiments}
\label{sec:exp}

This section details our systematic empirical investigation into the property of case-sensitive attention in LLMs and VLMs. Following the protocol in~\Cref{sec:methodology}, our evaluation follows a two-stage paradigm: we first perform phenomenological discovery on pure text, as text serves as the highest-leverage modality for isolating core transformer mechanics~\cite{wu2025textdominance}, and subsequently transfer these foundational insights to cross-modal attention steering experiments in the vision domain. Accordingly, our experiments are designed to: (i)~characterize how diverse typographic interventions modulate internal attention allocation; (ii)~quantify the causal relationship between case contrast and attention dynamics through systematic ablations; and (iii)~validate the universality and cross-modal generalizability of this property across different modalities, model families, parameter scales, and tokenization schemes. Importantly, we treat changes in task performance as a behavioral readout of this property, acknowledging that redirected attention does not always result in improved task accuracy.

%%%%%%%%%%%%%%%%%%%%%%%%%%%%%%%%%%%%%%%%%%%%%%%%%%%% EXPERIMENTAL SETTING %%%%%%%%%%%%%%%%%%%%%%%%%%%%%%%%%%%%%%%%%%%%%%%%%%%%
\subsection{Experimental Setting}
\label{sec:exp_setting}

Our study utilizes a strictly controlled protocol where inputs are transformed \emph{only} via letter case manipulation. Lexical content, word order, punctuation, and spacing are preserved identically to the original benchmark distributions to ensure that any observed shifts in model behavior are attributable solely to orthographic variation. For the cross-modal vision grounding experiments, we instantiate this protocol by embedding target object labels into generic template sentences to form an image description prompt. The label string within this prompt functions as the target span undergoing typographic alteration. We then isolate and quantify the internal cross-modal attention mass allocated by the model within the spatial region bounded by the ground-truth bounding box annotation corresponding to that label.

\paragraph{Models.}
We evaluate 13 models across five representative model families to ensure the property is not an artifact of a specific training regime or tokenizer:

\begin{itemize}
    \item \textbf{LLaMA-3 Family}~\cite{Grattafiori_2024_arXiv_LLaMA3}: We use LLaMA-3.2-3B-Instruct~\cite{llama3.2_3b_instruct} and LLaMA-3.1-8B-Instruct~\cite{llama3.1_8b_instruct}. These models utilize the Byte-Pair Encoding (BPE)-based \texttt{tiktoken} tokenizer~\cite{tiktoken}.
    \item \textbf{Gemma Family}~\cite{Gemma2_arXiv_TechnicalReport,Gemma3_arXiv_TechnicalReport,google_gemma_core_en}: We evaluate Gemma-3-1B-IT~\cite{gemma_3_1b_it}, Gemma-2-2B-IT~\cite{gemma_2_2b_it}, Gemma-3-4B-IT~\cite{gemma_3_4b_it}, and Gemma-4-E4B-IT~\cite{gemma_4_e4b_it}. These models employ a SentencePiece tokenizer with split digits, preserved whitespace, and byte-level encodings~\cite{Kudo_2018_arXiv_SentencePiece}.
    \item \textbf{Mistral}~\cite{Jiang_2023_arXiv_Mistral7b}: We use Mistral-7B-Instruct-v0.3~\cite{mistral_7b_instruct_v03}, a widely adopted mid-scale model with a distinct SentencePiece~\cite{Kudo_2018_arXiv_SentencePiece} implementation that preserves case-related linguistic features.
    \item \textbf{Qwen Family}~\cite{Qwen_2024_arXiv_Qwen2,Qwen_2025_arXiv_Qwen3}: We use Qwen2.5-7B-Instruct~\cite{qwen2.5_7b_instruct}, Qwen2.5-14B-Instruct~\cite{qwen2.5_14b_instruct}, Qwen3-4B-Thinking-2507~\cite{qwen3_4b_thinking_2507}, Qwen3-VL-4B-Instruct~\cite{qwen3vl_4b_instruct}, and Qwen3-VL-2B-Thinking~\cite{qwen3vl_2b_thinking}, all featuring byte-level BPE (BBPE) tokenization~\cite{Bai_2023_arXiv_QwensTokenizer}.
    \item \textbf{GPT}: We use the gpt-oss-20B model~\cite{openai_2025_gptoss20b}, a model specifically designed for advanced reasoning using the BPE-based \texttt{o200k\_harmony} tokenizer.
\end{itemize}
This selection enables controlled comparison across tokenizer sensitivity to case, parameter scale, pretraining corpus composition, and between textual and visual modalities, critical for isolating whether case steering generalizes beyond a single model.

\paragraph{Benchmarks.}
We select three text understanding benchmarks and one vision-language benchmark that provide ground-truth answer spans or spatial bounding box annotations, allowing us to precisely target the typographic interventions. We evaluate factual and scientific reasoning using \textbf{MMLU-Pro}~\cite{Wang_2024_arXiv_MMLUPro} and \textbf{ARC-Challenge}~\cite{Chollet_2019_arXiv_ARC}, reading comprehension with \textbf{SQuADv2}~\cite{Rajpurkar_2018_ArXiv_SQuADv2}, and cross-modal visual grounding via \textbf{RefCOCOg}~\cite{Mao_2016_RefCOCO}. Multilingual and code generation evaluations on \textbf{XQuAD}~\cite{Artetxe_2020_XQuAD} and \textbf{HumanEval}~\cite{Chen_2021_HumanEval} are detailed in Appendix~\ref{appendix:vision_code_language_results}.

%%%%%%%%%%%%%%%%%%%%%%%%%%%%%%%%%%%%%%%%%%%%%%%%%%%% CASE SCHEME DEFINITIONS %%%%%%%%%%%%%%%%%%%%%%%%%%%%%%%%%%%%%%%%%%%%%%%%%%%%
\subsection{Taxonomy of Typographic Interventions}
\label{sec:exp_schemes}

To characterize the boundaries of LLM and VLM case sensitivity, we define a taxonomy of \emph{static, deterministic interventions}. These interventions manipulate the contrast between the \textbf{target sequence} (the answer-relevant span or bounding box label) and the \textbf{context} (the remainder of the textual input). By varying the intensity and style of this contrast, we can determine whether the model responds to absolute casing (\eg, uppercase) or relative salience (\eg, case mismatch). All schemes produce strings $\tilde{X}$ with identical semantic content to the original input $X$, differing solely in surface-form typography.

\paragraph{Global Uniformity (U).}
These serve as baselines to measure the impact of removing all typographic cues.
\begin{itemize}
    \item \textbf{U1: All-Caps}. Entire input in uppercase.
    \item \textbf{U2: All-Lowercase}. Entire input in lowercase.
    \item \textbf{U3: Title-Case}. Every word capitalized.
\end{itemize}

\paragraph{Target Emphasis (TE).}
These test the hypothesis that uppercase acts as an attentional attractor when placed against a de-emphasized background.
\begin{itemize}
    \item \textbf{TE1 (Max-Contrast)}: Target in uppercase; context in lowercase.
    \item \textbf{TE2 (Pattern-Conflict)}: Target in uppercase; context in alternating case (\eg, \texttt{wRoNg}).
    \item \textbf{TE3 (Natural-Context)}: Target in uppercase; context retains the original casing.
\end{itemize}

\paragraph{Target Title Case (TT).}
These probe whether subtler, grammatically conventional emphasis is sufficient to shift attention.
\begin{itemize}
    \item \textbf{TT1 (Inverse-Contrast)}: Target in title case; context in uppercase.
    \item \textbf{TT2 (Standard-Emphasis)}: Target in title case; context in lowercase.
    \item \textbf{TT3 (Standard-Natural)}: Target in title case; context retains the original casing.
\end{itemize}

\paragraph{Target Alternating Case (TA).}
These test the model's response to unnatural, high-entropy typographic patterns.
\begin{itemize}
    \item \textbf{TA1 (Alt-Upper)}: Target in alternating case (\eg, \texttt{CoRrEcT}); context in uppercase.
    \item \textbf{TA2 (Alt-Lower)}: Target in alternating case; context in lowercase.
    \item \textbf{TA3 (Alt-Natural)}: Target in alternating case; context retains the original casing.
\end{itemize}

\paragraph{Adversarial De-Emphasis (ADE).}
Critical for causal validation, these interventions deliberately suppress the target span while emphasizing the context.
\begin{itemize}
    \item \textbf{ADE1 (Target-Suppression)}: Target in lowercase; context in uppercase.
    \item \textbf{ADE2 (Target-Conflict)}: Target in lowercase; context in alternating case.
    \item \textbf{ADE3 (Target-Lower-Natural)}: Target in lowercase; context retains the original casing.
\end{itemize}

\paragraph{Latency and Tokenization Dynamics.} 
The computational overhead of these surface-form variations is negligible (mean $\approx 0.0015$ ms; see \cref{tab:pattern_comparison}, App.~\ref{appendix:tokenization_analysis}). While different casing can occasionally alter token counts due to tokenizer behavior, the consistency of results across diverse tokenizers confirms that the ``casing effect'' is a representational property rather than a byproduct of sequence length. By applying these interventions, we isolate letter case as a causal variable. This setup allows us to observe how the model's internal state reacts to surface-level changes without modifying model weights or lexical and visual input.

%%%%%%%%%%%%%%%%%%%%%%%%%%%%%%%%%%%%%%%%%%%%%%%%%%%% RESULTS SECTION %%%%%%%%%%%%%%%%%%%%%%%%%%%%%%%%%%%%%%%%%%%%%%%%%%%%
\subsection{Characterizing Textual Attention Allocation Dynamics}
\label{sec:results_attention}

\begin{figure*}[!t]
    \centering
    \begin{subfigure}{0.45\textwidth} % width of the subfigure
        \centering
        \includegraphics[width=\textwidth]{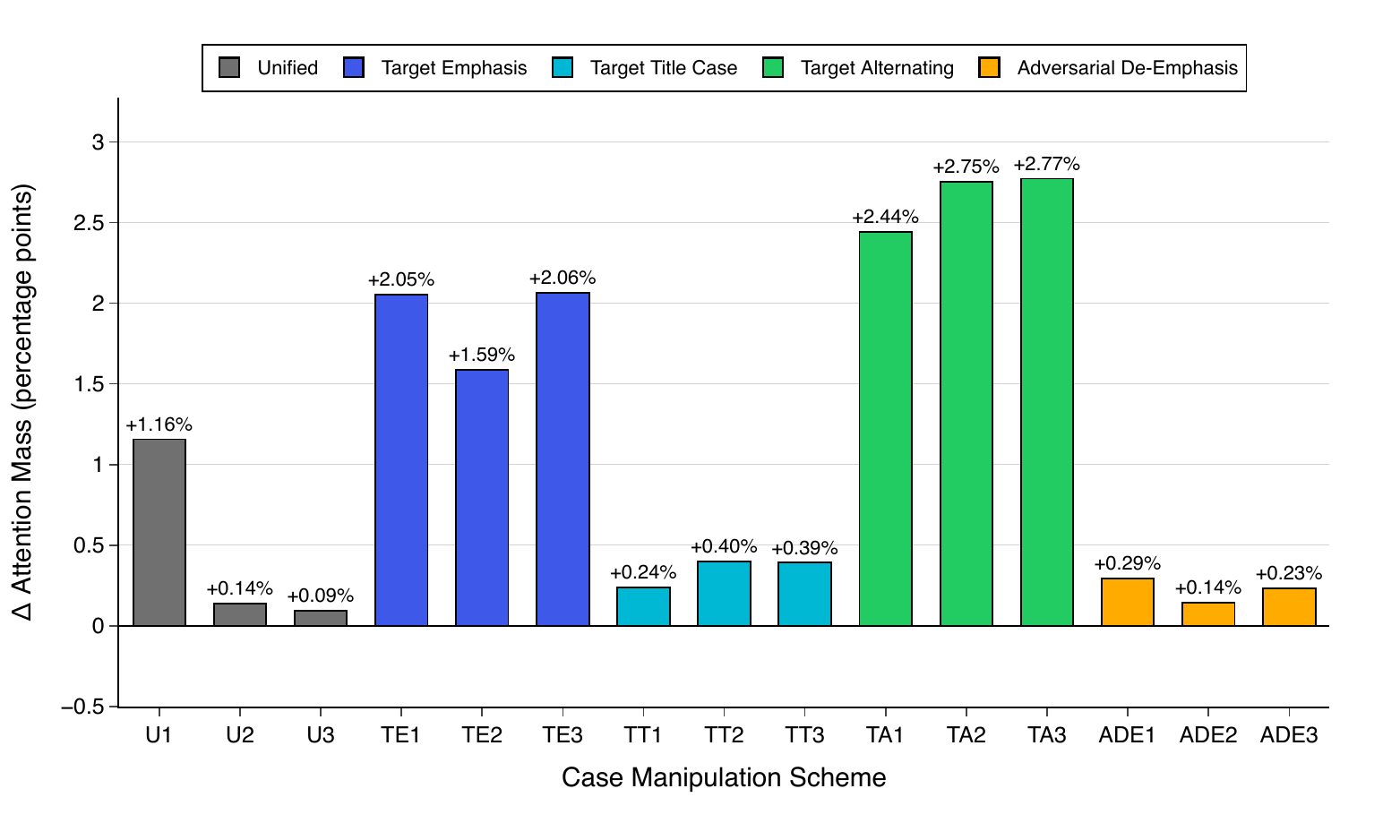}
        \caption{Relative Attention Mass on Target}
         \label{fig:qwen7_attention}
    \end{subfigure}% 
    \hspace{2em}
    \begin{subfigure}{0.45\textwidth} % width of the subfigure
        \centering
        \includegraphics[width=\textwidth]{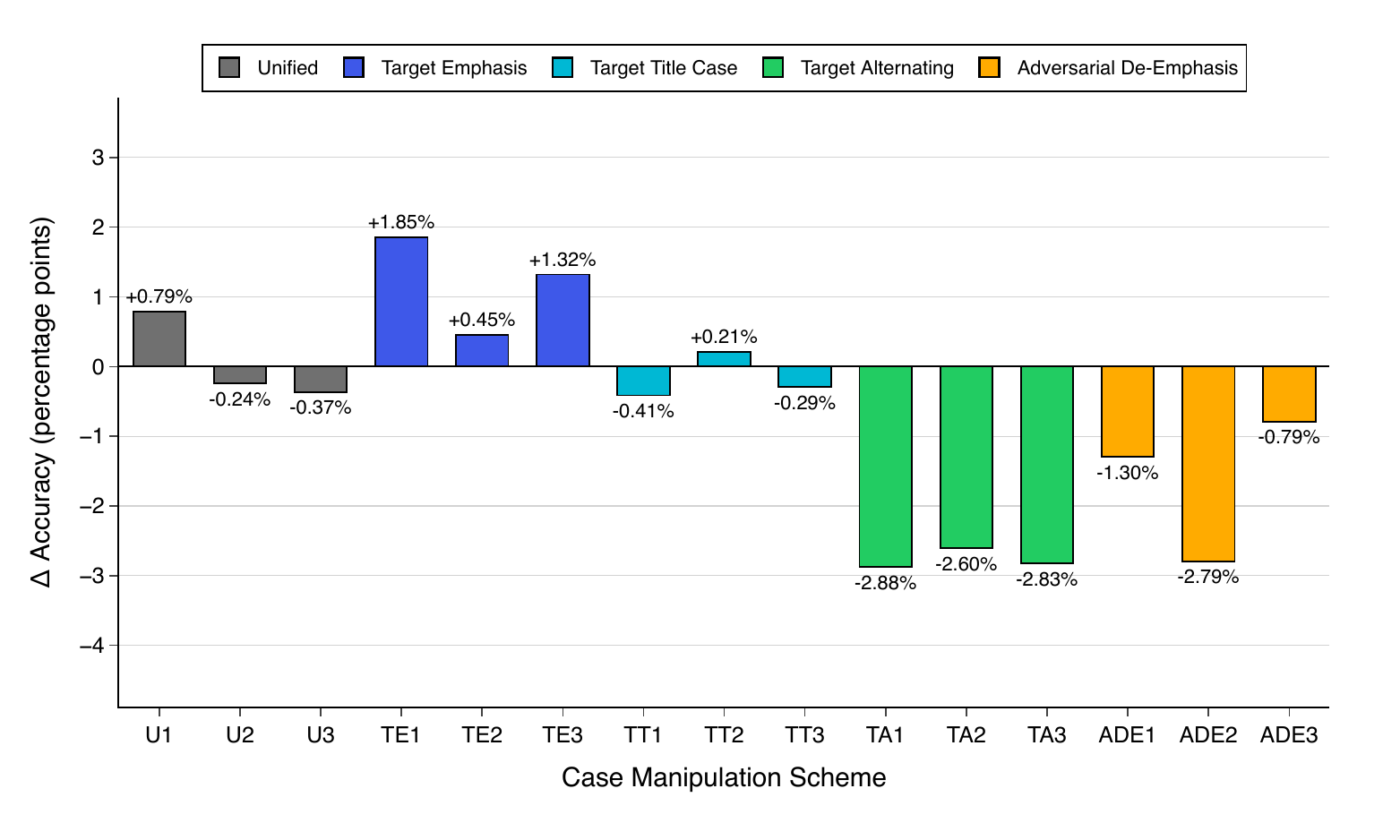}
        \caption{Relative Task Accuracy}
    \end{subfigure}% 
    \label{fig:qwen7_accuracy}
    \caption{\textbf{Impact of typographic interventions on model behavior.} Comparative analysis of relative attention mass allocated to the target span and downstream task accuracy across different case-steering schemes using Qwen2.5-7B-Instruct.}
    \label{fig:qwen7_in_paper}
\end{figure*}

Our empirical analysis begins by characterizing how typographic interventions modulate internal attention mass. We use Qwen2.5-7B-Instruct~\cite{qwen2.5_7b_instruct} as the discovery model. Results show that each intervention alters attention allocation, with magnitude varying by pattern (\cf \cref{fig:qwen7_in_paper,fig:attentionmass_per_scheme}).

The most substantial shifts in attention toward the target span are elicited by \textbf{Target Alternating (TA)} schemes. Specifically, TA3 (alternating target, natural context) and TA2 (alternating target, lowercased context) yield the highest gains of $+2.77$ pp and $+2.75$ pp in mean attention mass, respectively. The negligible difference between TA2 and TA3 suggests that since natural English context is predominantly lowercase, the model responds primarily to the high-entropy alternating pattern of the target. Even when the context is uppercased (TA1), the target attracts a $+2.44$ pp gain, marking alternating case as the most potent attentional attribution shifter in our taxonomy.

\textbf{Target Emphasis (TE)} schemes, involving full uppercasing of the target, also yield notable shifts. TE3 (uppercase target, natural context) and TE1 (uppercase target, lowercase context) increase attention allocation by $+2.06$ pp and $+2.05$ pp, respectively. This effect is attenuated when the context is formatted in alternating case (TE2), dropping to $+1.59$ pp. This suggests that while uppercase is a strong attractor, its salience is partially diminished when the background context itself possesses high typographic entropy.

In contrast, unified patterns (U2, U3), title-case schemes (TT), and adversarial de-emphasis (ADE) interventions exert minimal influence, with attention shifts typically remaining below $+0.4$ pp. A notable exception is global uppercasing (U1), which induces a general gain of $+1.16$ pp, suggesting that the model treats all-caps text with a higher baseline of attentional intensity.

%%%%%%%%%%%%%%%%%%%%%%%%%%%%%%%%%%%%%%%%%%%%%%%%%%%% RESULTS SECTION %%%%%%%%%%%%%%%%%%%%%%%%%%%%%%%%%%%%%%%%%%%%%%%%%%%%
\subsection{The Attention-Performance Divergence}
\label{sec:results_accuracy}

A central finding of our study is the non-trivial relationship between attention allocation and downstream task performance. While alternating case (TA) most effectively concentrates attention, it serves as a \textit{destructive attractor}: across all benchmarks, TA schemes consistently degrade downstream accuracy, with drops of up to $-2.88$ pp on the discovery model (\cf \cref{fig:accuracy_per_scheme}). This suggests that the high-entropy nature of alternating case, while salient to the attention mechanism, likely disrupts the model's ability to extract semantic meaning from the tokens, seen in Appendix~\ref{appendix:mechanistic_insights}.

Conversely, \textbf{Target Emphasis (TE)} via uppercasing acts as a \textit{productive attractor}. TE1 (uppercase target, lowercase context) achieves a performance gain of $+1.85$ pp. This indicates that the attention directed by uppercase lettering is utilized by the model, aligning with the hypothesis that LLMs internalize uppercase as a signal for importance rather than just a visual anomaly. Absolute values and detailed per-scheme breakdowns are provided in \Cref{tab:schemes_absolutevals_qwen_7b}.

%%%%%%%%%%%%%%%%%%%%%%%%%%%%%%%%%%%%%%%%%%%%%%%%%%%% RESULTS SECTION %%%%%%%%%%%%%%%%%%%%%%%%%%%%%%%%%%%%%%%%%%%%%%%%%%%%
\subsection{Case Sensitivity in Vision-Language Models}
\label{sec:exp:vlms}
Having established the casing effect on purely textual streams, we ask whether typographic salience crosses the modality boundary. We stress at the outset that we do \emph{not} expect the multimodal setting to reproduce the text-only phenomenon verbatim: in text, casing concentrates attention \emph{within} a single stream, and the resulting attention--performance divergence is an intra-modal effect. In VLMs, the same surface-form perturbation instead operates as a \emph{cross-modal routing lever}. We therefore frame our multimodal results as a \emph{partial} replication---the \emph{family-level ordering} of attractors transfers across modalities, while its mechanism and downstream consequences differ in ways specific to multimodal processing. Concretely, a casing-induced steering effect is clearly present in VLMs, but it is less consistent than in the text-only regime: whereas the textual effect holds uniformly across every model and tokenizer family (\cf \cref{sec:results_generalization}), the multimodal effect is robust only at the level of pattern-family ordering and the macroscopic modality axis, and becomes fragmented across individual models at the fine-grained spatial level.

\begin{figure*}[!t]
    \centering
    % Row 1
    \begin{subfigure}{0.31\textwidth}
        \centering
        \includegraphics[width=\textwidth]{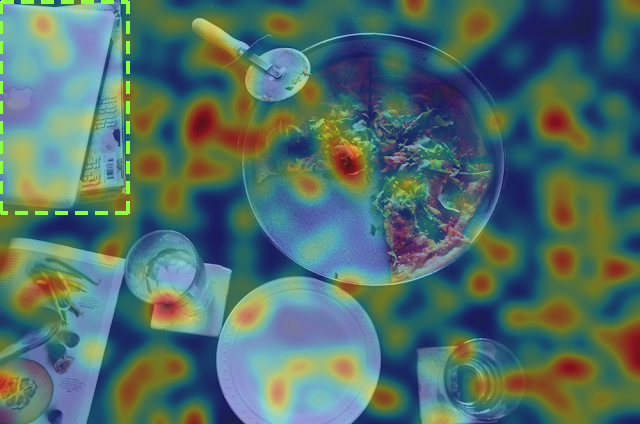}
        \caption{Baseline}
    \end{subfigure}\hfill
    \begin{subfigure}{0.31\textwidth}
        \centering
        \includegraphics[width=\textwidth]{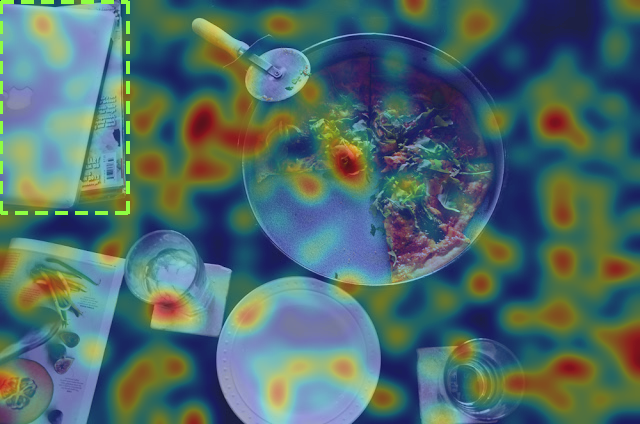}
        \caption{U2}
    \end{subfigure}\hfill
    \begin{subfigure}{0.31\textwidth}
        \centering
        \includegraphics[width=\textwidth]{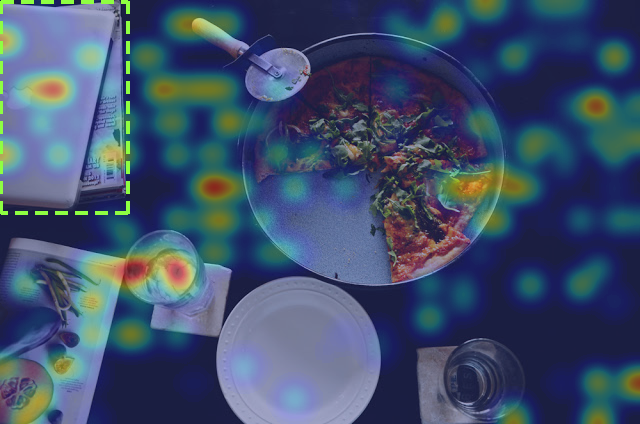}
        \caption{TE3}
    \end{subfigure}

    \vspace{0.4em} % Blank line above forces line break before vspace

    % Row 2
    \begin{subfigure}{0.31\textwidth}
        \centering
        \includegraphics[width=\textwidth]{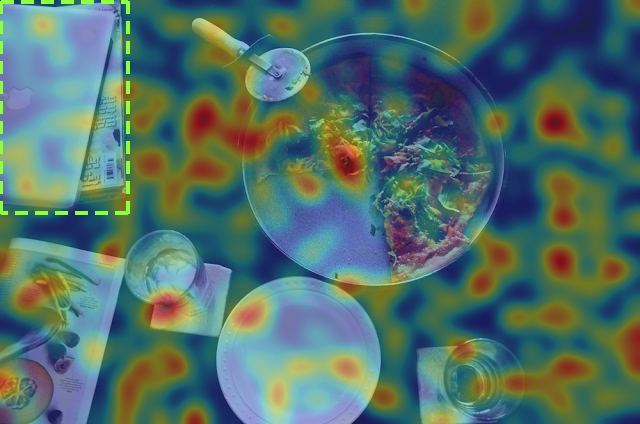}
        \caption{TT3}
    \end{subfigure}\hfill
    \begin{subfigure}{0.31\textwidth}
        \centering
        \includegraphics[width=\textwidth]{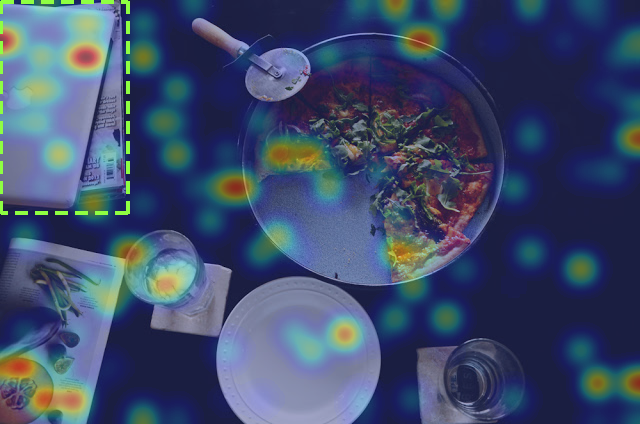}
        \caption{TA1}
    \end{subfigure}\hfill
    \begin{subfigure}{0.31\textwidth}
        \centering
        \includegraphics[width=\textwidth]{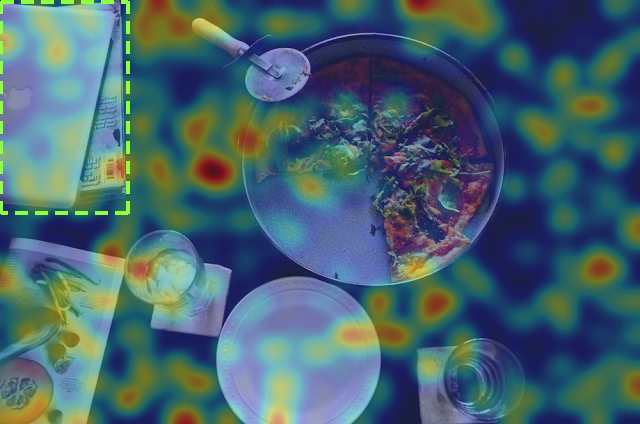}
        \caption{ADE1}
    \end{subfigure}

    \caption{\textbf{Cross-modal attention steering via text-based typographic salience in VLMs} shown with Grad-CAM visualizations of the Vision Transformer cross-attention maps on the RefCOCOg grounding benchmark with Qwen3-VL-2B-Thinking.}
    \label{fig:computervision}
\end{figure*}

We evaluate our intervention taxonomy on the RefCOCOg visual grounding benchmark~\cite{Mao_2016_RefCOCO} across four VLMs spanning two families and both inference regimes: Qwen3-VL-4B-Instruct~\cite{qwen3vl_4b_instruct}, Qwen3-VL-2B-Thinking~\cite{qwen3vl_2b_thinking}, Gemma-3-4B-IT~\cite{gemma_3_4b_it}, and Gemma-4-E4B-IT (Thinking)~\cite{gemma_4_e4b_it}. Crucially, the intervention is applied \emph{only} to the textual description prompt; the image is never altered. Across the dataset, non-standard casing induces a coupled, dual-axis response (\cf \cref{fig:computervision}): a \emph{microscopic} concentration of the residual visual budget onto the target region (a mean $+1.55$ pp absolute shift, a relative $+23.08\%$ increase in target-region attention alignment), together with a \emph{macroscopic} re-allocation away from the image as a whole (a mean $-1.85$ pp drop in whole-image attention mass, \ie, a $12.41\%$ relative disengagement from the visual stream in favour of the textual prompt). Consistent with reports of textual dominance in multimodal transformers~\cite{wu2025textdominance}, the dominant effect is macroscopic: casing primarily decides \emph{how much} of the image is attended to, and secondarily \emph{where}.

\paragraph{The family hierarchy transfers.}
Aggregating by pattern family reproduces the ordering observed on text. The Target Alternating (TA) family is again the strongest attractor, yielding the largest microscopic pull ($+2.84$ pp on the target box) and the largest macroscopic shift ($+24.22\%$ disengagement); Target Emphasis (TE) follows ($+1.77$ pp; $+21.12\%$), while uniform (U), title-case (TT), and de-emphasis (ADE) families remain weak. This cross-modal persistence of the \mbox{TA $>$ TE $>$ rest} ordering indicates that the model treats high-entropy and uppercase prompt tokens as importance signals \emph{regardless of modality}. We caution, however, that the hierarchy is robust only at the family-aggregate level: as detailed in Appendix~\ref{appendix:multimodal_results} (\cref{tab:appendix_micro_spatial_steering}), the per-model microscopic ordering is noisy---on the Qwen3-VL models, uniform uppercasing (U1), title-case contrast (TT1), and even de-emphasis (ADE1) rival TA1, whereas Gemma-3-4B-IT exhibits a sharper response (peaks of $+6.23$ pp under TA1 and $+5.74$ pp under TE2). The \emph{macroscopic} axis is sign-consistent across all four models and constitutes the more reliable multimodal signature of the ``casing effect'' in VLMs.

\paragraph{Pattern-conflict drives modality disengagement.}
At the individual-pattern level, the pattern-conflicting configuration TE2 (uppercase target inside an alternating-case context) is the primary driver of modality disruption, producing the largest whole-image drain ($-4.11$ pp) and the strongest cross-model disengagement (a mean of $33.35\%$, and up to $43.74\%$ on Qwen3-VL-2B-Thinking). When the surrounding prompt carries high orthographic entropy, processing concentrates on linguistic tokens at the expense of the visual feature maps (mechanistic detail in App.~\ref{appendix:mechanistic_insights}). Uniform lowercasing (U2) minimises these shifts, perturbing spatial allocation by only $+0.48$ pp.

\paragraph{Reasoning reverses its role across modalities.}
The most salient multimodal-specific finding is a reversal of the ``reasoning buffer'' we identified on text. Whereas extended reasoning \emph{attenuates} typographic sensitivity in text-only LLMs (\cf App.~\ref{appendix:reasoning_models}), the visual reasoning phase \emph{amplifies} macroscopic reliance on the text stream under typographic stress. Reasoning VLMs disengage from vision most strongly (Qwen3-VL-2B-Thinking: mean $14.88\%$, peak $43.74\%$ under TE2; Gemma-4-E4B-IT: mean $12.92\%$, peak $28.83\%$ under TA1), while the direct-inference models re-allocate more modestly (Gemma-3-4B-IT $11.20\%$, Qwen3-VL-4B-Instruct $10.64\%$). Conversely, direct-inference models are the more vulnerable to fine-grained \emph{microscopic} steering. We interpret this as a division of labor: reasoning loops bias the macro-level modality choice toward textual coherence, while standard models remain exposed to raw spatial salience cues.

%%%%%%%%%%%%%%%%%%%%%%%%%%%%%%%%%%%%%%%%%%%%%%%%%%%% RESULTS SECTION %%%%%%%%%%%%%%%%%%%%%%%%%%%%%%%%%%%%%%%%%%%%%%%%%%%%
\subsection{Cross-Model Generalization and Robustness}
\label{sec:results_generalization}

To verify that these behaviors are inherent properties of the transformer architecture rather than model-specific artifacts, we evaluate the intervention taxonomy across 13 models. Detailed quantitative results, including extended evaluations across diverse languages and specialized tasks, can be seen in \Cref{tab:schemes_absolutevals_llama_3b,tab:schemes_absolutevals_llama_8b,tab:schemes_absolutevals_gemma_1b,tab:schemes_absolutevals_gemma_2b,tab:schemes_absolutevals_mistral_7b,tab:schemes_absolutevals_qwen_7b,tab:schemes_absolutevals_qwen_14b,tab:schemes_absolutevals_qwen_thinking_4b,tab:schemes_absolutevals_gptoss_20b} as well as in Appendix~\ref{appendix:vision_code_language_results} and~\ref{appendix:tokenizer_details}. A consistent picture emerges across both modalities, but with an important asymmetry in consistency: in the text-only setting the casing effect is \emph{universal}, applying across all models and tokenizer families, whereas in VLMs the same effect arises only \emph{partially}---robust at the family-aggregate and macroscopic level, yet fragmented across individual models.

\begin{figure*}[!ht]
    \centering
    \begin{subfigure}{0.45\textwidth} % width of the subfigure
        \centering
        \includegraphics[width=\textwidth]{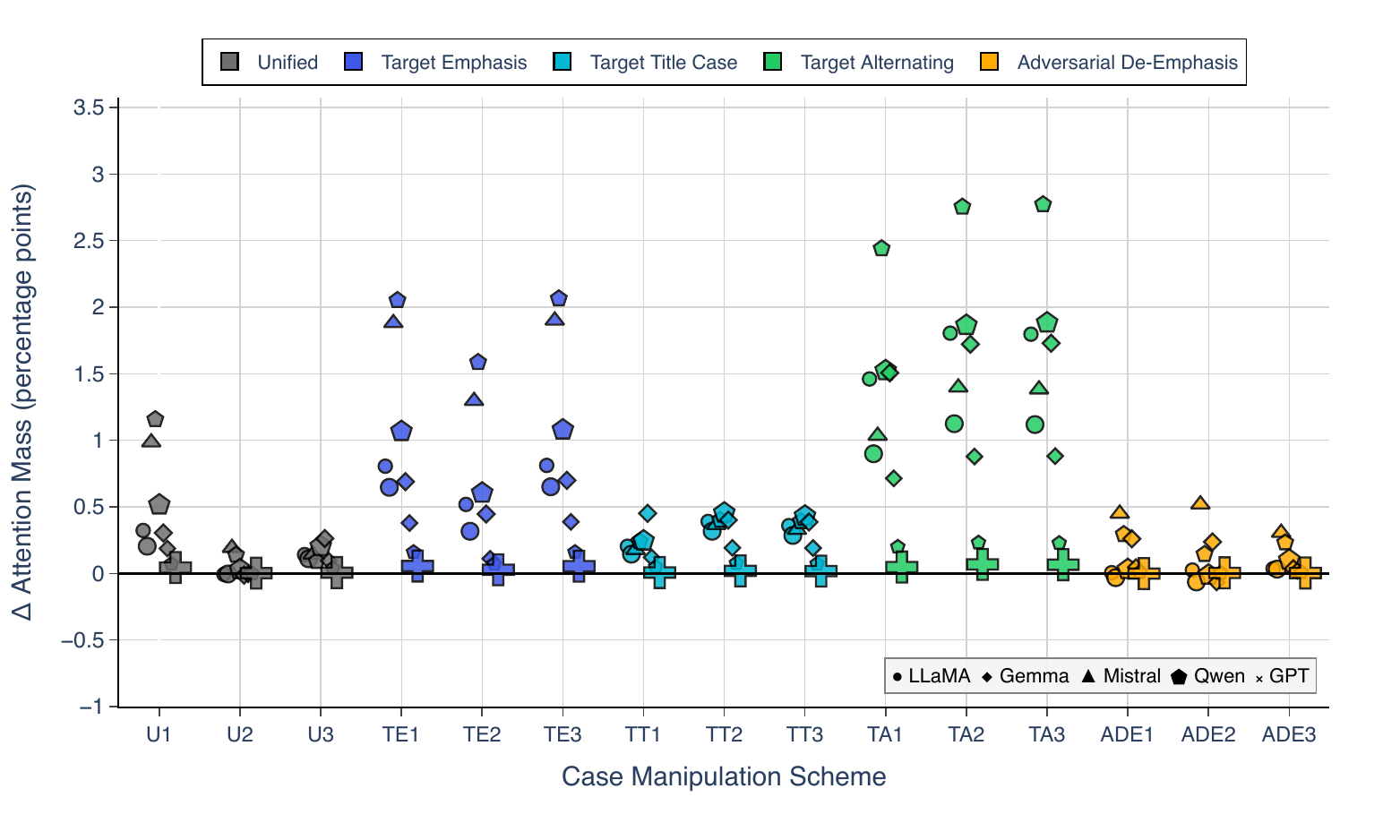}
        \caption{Relative Attention Mass on Target}
         \label{fig:allmodels_attention}
    \end{subfigure}% 
    \hspace{2em}
    \begin{subfigure}{0.45\textwidth} % width of the subfigure
        \centering
        \includegraphics[width=\textwidth]{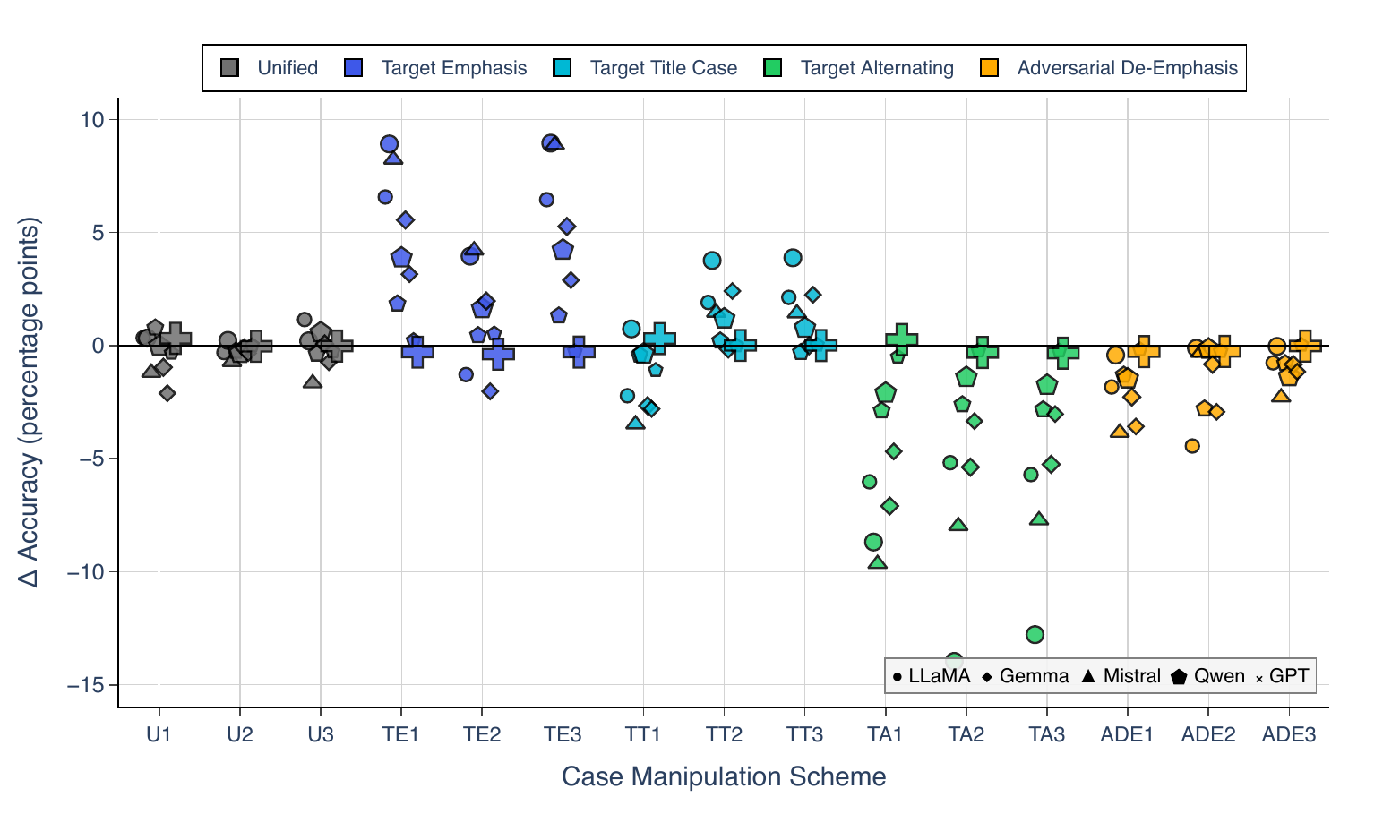}
        \caption{Relative Task Accuracy}
    \end{subfigure}% 
    \label{fig:allmodels_accuracy}
    \caption{\textbf{Generalization of typographic interventions across different models.} Comparative analysis of relative attention mass allocated to the target span and downstream task accuracy across different case-steering schemes and model architectures. Symbols encode the primary model family, while their size encodes model size.}
    \label{fig:allmodels_in_paper}
\end{figure*}

\begin{itemize}
    \item \textbf{Textual Observation (Universal Effect)}: As shown in \Cref{fig:allmodels_in_paper}, non-reasoning LLMs exhibit consistent patterns of attention shift and corresponding task-accuracy variations across all schemes, albeit with differing magnitudes. This consistency is \emph{universal} in non-reasoning models: it holds across every evaluated model and across all three tokenizer families (BPE, SentencePiece, and BBPE; \cf App.~\ref{appendix:tokenizer_details}), confirming that the effect is a representational property rather than a tokenizer artifact. In contrast, reasoning models deviate from this trend, showing near-zero sensitivity in both attention allocation and downstream task accuracy (\cf App.~\ref{appendix:reasoning_models}).

    \item \textbf{Reasoning Influence}: The deliberative reasoning process in text-only LLMs largely mitigates shifts in attention mass and exhibits minimal variation in downstream task accuracy (typically below $\pm0.5$ pp). This suggests that the ``thinking'' phase acts as a semantic realignment layer that filters out typographic salience in favor of logical coherence (\cf App.~\ref{appendix:reasoning_models}).

    \item \textbf{Consistent Failure Modes}: In non-reasoning models, high-entropy casing acts as a destructive attractor; alternating case (TA) consistently degrades performance, reaching a peak loss of $-13.96$ pp in LLaMA-3.1-8B-Instruct. Adversarial de-emphasis (ADE) schemes similarly confirm that misdirecting attention through casing systematically lowers task accuracy.

    \item \textbf{Utility of Uppercasing}: Across all standard architectures, TE1 and TE3 remain the only interventions that reliably yield performance gains. These schemes achieve mean accuracy increases of up to $+8.95$ pp, suggesting that standard LLMs have internalized capitalization as valid importance signal.

    \item \textbf{Partial Transfer to VLMs}: The textual hierarchy reproduces \emph{partially} in the visual domain. Text-level casing acts as a zero-shot cross-modal lever, increasing the attention mass that falls inside the target bounding box by a relative $+23.08\%$ (a mean $+1.55$ pp absolute shift), with the family ordering (TA $>$ TE $>$ rest) preserved in aggregate. Unlike the universal textual effect, however, the per-model spatial response is fragmented: direct-inference pipelines are the most locally sensitive (peaking at a $+6.23$ pp spatial attention surge in Gemma-3-4B-IT), whereas the Qwen3-VL models exhibit no clear family ordering at the fine-grained level (\cf \cref{tab:appendix_micro_spatial_steering,tab:appendix_macro_modality_shifts}).

    \item \textbf{Macroscopic Modality Shifts}: The axis that is sign-consistent across all four VLMs is macroscopic. Non-standard casing redirects attention away from the image toward the text prompt, yielding a global $+12.41\%$ disengagement from the visual stream (a $-1.85$ pp whole-image drop). This shift is amplified by visual reasoning---reversing the text-only ``reasoning buffer''---and peaks at a $43.74\%$ total visual attention drop in Qwen3-VL-2B-Thinking.
\end{itemize}

%% file: 02_Main/06_conclusion.tex
%%%%%%%%%%%%%%%%%%%%%%%%%%%%%%%%%%%%%%%%%%% CONCLUSION %%%%%%%%%%%%%%%%%%%%%%%%%%%%%%%%%%%%%%%%%%%
\section{Conclusion}
\label{sec:conclusion}
In this paper, we presented a systematic empirical study characterizing letter case as a fundamental modulator of internal attention in both LLMs and VLMs. By isolating typographic variation from semantic and lexical content, we established that casing is not merely a tokenization artifact but a latent property inherent to the transformer's pretraining. Our results across seven non-reasoning LLMs demonstrate that interventions such as formatting target spans in uppercase reliably concentrate attention mass and improve downstream performance by up to $+8.95$ pp without model modification, an effect that holds across every evaluated non-reasoning model and tokenizer family. Extending this paradigm to the visual domain, we find that text-level casing transfers across the modality boundary partially: the family-level ordering of attractors is preserved in aggregate, but casing reorganizes cross-modal attention along two coupled axes---shifting overall allocation away from the image (the dominant effect), and secondarily concentrating the residual visual budget on the target region.

A central contribution is characterizing the non-trivial relationship between attention allocation and task utility. We identified a divergence between \textit{productive} attractors (uppercase), which enhance downstream performance, and \textit{destructive} attractors (alternating case), which elicit the strongest attention shifts but significantly degrade performance. Crucially, our analysis uncovers a boundary condition that \emph{reverses} across modalities: whereas the text-only ``thinking'' phase acts as a semantic buffer that filters out typographic salience, the visual reasoning phase instead amplifies macro-level reliance on the textual stream. Under typographic stress, reasoning VLMs exhibit a pronounced shift away from visual features, whereas direct-inference models remain vulnerable to localized spatial steering.

\paragraph{Limitations and Future Work.}
Our evaluation focused on languages where letter casing is defined; exploring this property in non-Latin scripts and applying causal mediation analysis would provide deeper mechanistic insights. Additionally, developing automated pipelines that leverage this property in textual and visual contexts represents a promising path toward efficient, black-box attention steering.  Ultimately, this work establishes typographic salience as a training-free attention-steering lever across text and vision domains.

\section*{Acknowledgments}
This research was funded, in part, by the U.S. Government under ARPA-H contract 1AY2AX000062 and DARPA contract HR0011-24-9-0429. The views and conclusions contained in this document are those of the authors and should not be interpreted as representing the official policies, either expressed or implied, of the U.S. Government.

%% file: 02_Main/X_supplementary.tex
\clearpage
\onecolumn
\appendix

\begin{center}
  \Large \textbf{Attention is Case-Sensitive} \\[0.4em]
  \Large \textbf{(Supplementary Material)}
\end{center}
\vspace{1.5em}

\section*{Appendix}

This appendix provides extended data and technical documentation supporting our empirical study of case-sensitivity in Large Language Models (LLMs) and Vision-Language Models (VLMs). The following sections offer a deeper characterization of the ``casing effect'' and provide the necessary details to ensure the reproducibility of our findings.

\begin{itemize}
    % \item \textbf{Appendix~\ref{appendix:impactstatement}}: The \textit{Impact Statement} discusses the broader implications of our discovery for model interpretability, accessibility, and robustness.
    
    \item \textbf{Appendix~\ref{appendix:mechanistic_insights}}: \textit{Mechanistic Insights and Internal Dynamics} provides a granular structural analysis of internal network routing, tracking layer-head dynamics across both textual and visual modalities, softmax context-starvation, and latent semantic drift alongside our proposed architectural mitigation.

    \item \textbf{Appendix~\ref{appendix:additional_results}}: \textit{Additional Empirical Results and Analyses} presents extended quantitative and qualitative performance and attention mass tables for all standard and reasoning models across text and vision benchmarks. This section details our \textit{Extended Multimodal Experimental Evaluation} alongside qualitative Grad-CAM cross-attention maps, a study on \textit{Self-Reported Attentional Focus: Bridging Mass and Steering}, \textit{Qualitative Observations from Active Interrogation} with relative attention maps for each letter casing scheme, and outlines our \textit{Attention Extraction and Normalization Protocol}.
    
    \item \textbf{Appendix~\ref{appendix:vision_code_language_results}}: \textit{Robustness and Generalization across Languages and Tasks} establishes the domain generalization of typographic salience, evaluating its impact across diverse cased scripts and functional programming benchmarks.
    
    \item \textbf{Appendix~\ref{appendix:tokenization_analysis}}: \textit{Tokenization Dynamics and Computational Overhead} provides an analysis of latency induced by letter casing schemes alongside the impact on token fragmentation.
    
    \item \textbf{Appendix~\ref{appendix:tokenizer_details}}: \textit{Tokenizer-Specific Letter Casing Effects} offers a qualitative and statistical analysis of how various tokenizer families respond to letter case interventions, supporting our claim that the casing effect is an inherent representational property.
\end{itemize}

\clearpage \newpage
\begin{table}[!ht]
    \caption{\textbf{Summary of letter casing schemes used in this work}, with operational rules and experimental rationale. All transformations are static, deterministic, and applied prior to tokenization.}
    \centering
    \small
    \begin{tabular}{>{\centering\arraybackslash}m{1.5cm} | m{10cm}}
        \toprule
        \textbf{Scheme} & \textbf{Description} \\
        \midrule
        U1 & Full uppercase (\eg \texttt{HELLO WORLD}). Tests degradation under maximal case uniformity. \\
        \midrule
        U2 & Full lowercase (\eg \texttt{hello world}). Tests impact of complete case removal on attention focus. \\
        \midrule
        U3 & Global title case (\eg \texttt{Hello World}). Probes sensitivity to structural formatting without target emphasis. \\
        \midrule
        TE1 & Target uppercase; context lowercase (\eg \texttt{TARGET; context}). Maximizes contrast to test attentional attraction to emphasized spans. \\
        \midrule
        TE2 & Target uppercase; context alternating case (\eg \texttt{TARGET; cOnTeXt}). Tests steering robustness under unnatural context patterns. \\
        \midrule
        TE3 & Target uppercase; context original case (\eg \texttt{TARGET; context}). Tests emphasis effects under natural typography. \\
        \midrule
        TT1 & Target title case; context uppercase (\eg \texttt{Target; CONTEXT}). Isolates relative contrast (not absolute case) as steering driver. \\
        \midrule
        TT2 & Target title case; context lowercase (\eg \texttt{Target; context}). Tests whether conventional emphasis suffices for attention guidance. \\
        \midrule
        TT3 & Target title case; context original case (\eg \texttt{Target; context}). Measures partial emphasis under natural context. \\
        \midrule
        TA1 & Target alternating case; context uppercase (\eg \texttt{tArGeT; CONTEXT}). Probes attention to case regularity independent of semantics. \\
        \midrule
        TA2 & Target alternating case; context lowercase (\eg \texttt{tArGeT; context}). Tests boundaries of unnatural yet deterministic patterns. \\
        \midrule
        TA3 & Target alternating case; context original case (\eg \texttt{tArGeT; context}). Evaluates robustness under natural context variation. \\
        \midrule
        ADE1 & Target lowercase; context uppercase (\eg \texttt{target; CONTEXT}). Inverts TE1 to validate causal misdirection of attention. \\
        \midrule
        ADE2 & Target lowercase; context alternating case (\eg \texttt{target; cOnTeXt}). Tests de-emphasis without active distraction. \\
        \midrule
        ADE3 & Target lowercase; context original case (\eg \texttt{target; context}). Measures vulnerability to target suppression. \\
        \bottomrule
    \end{tabular}
    \label{tab:casing_schemes}
\end{table}

\clearpage \newpage
\section{Mechanistic Insights and Internal Dynamics}
\label{appendix:mechanistic_insights}

\begin{figure}[!t]
    \centering
    \begin{subfigure}{0.24\linewidth} % width of the subfigure
        \centering
        \includegraphics[width=\textwidth]{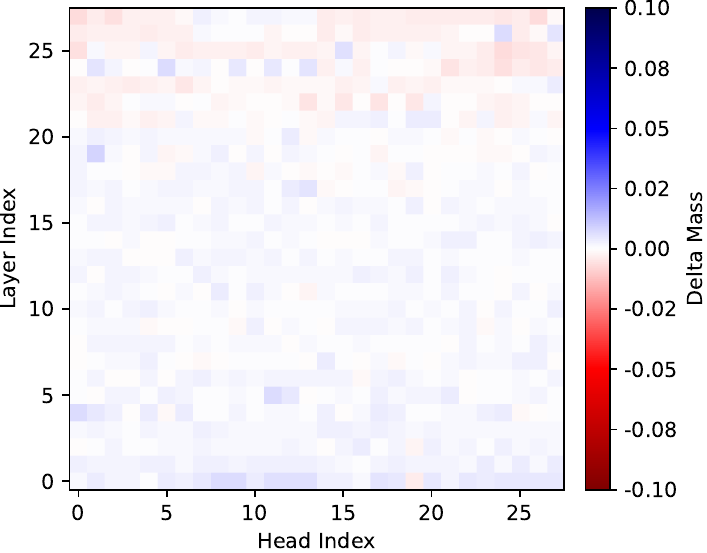}
        \caption{U1}
    \end{subfigure}% 
    \hfill
    \begin{subfigure}{0.24\linewidth} % width of the subfigure
        \centering
        \includegraphics[width=\textwidth]{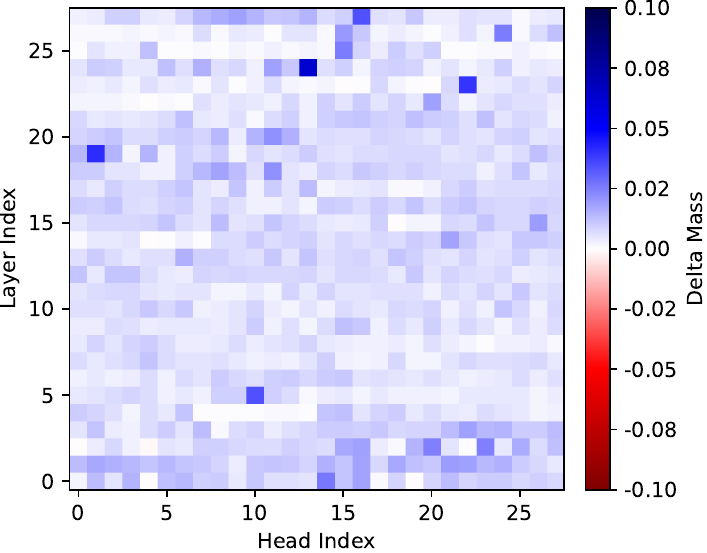}
        \caption{TE1}
    \end{subfigure}%
    \hfill
    \begin{subfigure}{0.24\linewidth} % width of the subfigure
        \centering
        \includegraphics[width=\textwidth]{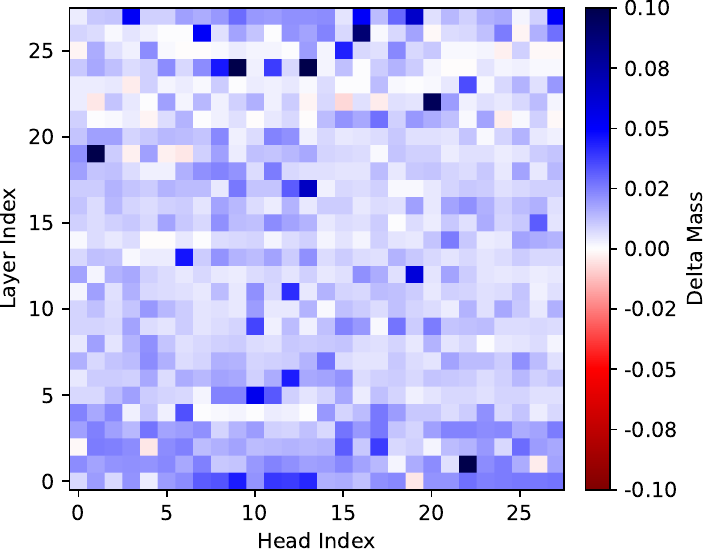}
        \caption{TA1}
    \end{subfigure}% 
    \hfill
    \begin{subfigure}{0.24\linewidth} % width of the subfigure
        \centering
        \includegraphics[width=\textwidth]{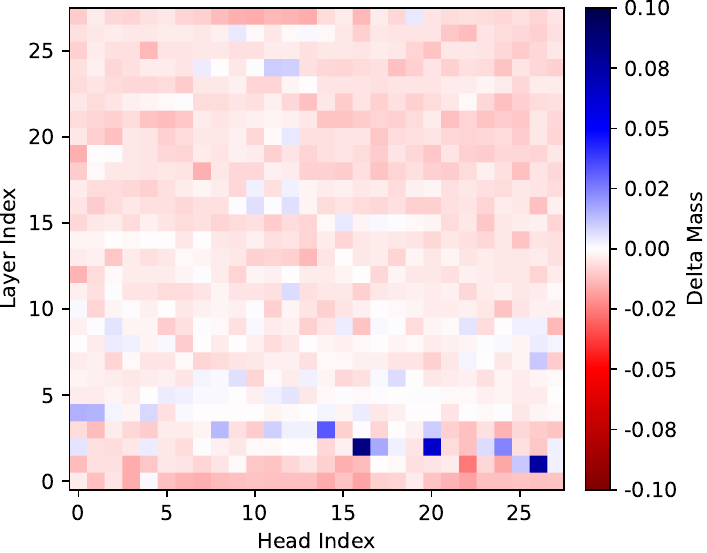}
        \caption{ADE2}
    \end{subfigure}%
    \caption{\textbf{Layer- and head-wise breakdown of relative target attention allocation mass across intervention schemes on Qwen2.5-7B-Instruct.} Attention shifts are plotted across all layers and heads relative to a standard casing baseline. Blue regions denote a localized increase in attention mass; red regions signify a reduction.}
    \label{fig:mechanistic_instights}
\end{figure}

To uncover the underlying neural mechanisms driving the typographic casing phenomenon, we analyze the internal routing of attention mass and track the corresponding shifts in the model's representation space. This section provides a detailed breakdown of layer-wise and head-wise dynamics, a structural examination of attention ``hijacking'' via the softmax operator, an empirical evaluation of latent semantic drift, and an architectural hypothesis for future mitigation.

\subsection{Layer-Wise and Head-Wise Attention Dynamics in Text}
To localize where typographic salience alters internal processing, we analyze layer-head attention heatmaps constructed relative to a baseline of standard casing (\cf \cref{fig:mechanistic_instights}). In these visualizations, target attention allocation is plotted across all layers and heads; blue regions denote an increase in attention mass compared to the baseline, while red regions signify a reduction. 

Our empirical tracking reveals that the casing effect is highly localized rather than distributed uniformly across the network architecture. Both productive Target Emphasis (TE) and high-entropy Target Alternating (TA) patterns exhibit a prominent attention allocation spike that peaks sharply at Layer 24. Within the context of deep Transformer architectures, Layer 24 corresponds structurally to the late semantic aggregation stage. This is the critical phase where token representations transition from localized lexical features to synthesized contextual concepts, making it sensitive to typographic modulations.

\subsection{Softmax Hijacking and Context Starvation}
We demonstrate semantic disruption structurally by examining the mathematical constraints of the softmax attention mechanism. Because softmax produces a bounded probability distribution, the total attention mass allocated by any single head across a sequence is fundamentally a zero-sum resource. Consequently, when an anomalous typographic feature draws an disproportionate share of attention, it directly deprives the remaining tokens of processing capacity.

\paragraph{Destructive Attractors (TA):} The Target Alternating (TA) pattern operates as an extreme, destructive attractor. At the semantic aggregation stage (Layer 24), TA triggers an overall layer-wide attention surge of $+2.78$ pp. More granular inspection reveals that individual routing heads bear the brunt of this effect; for example, head L24H13 exhibits an attention spike of $+18.9$ pp. Because the distribution is bounded, this over-allocation effectively ``hijacks'' the head's attention capacity. The surrounding textual context is mathematically starved of the attention required for coherence synthesis. In late layers, which are responsible for generating the final representations used in downstream predictions, this context-starvation disrupts reading comprehension, providing a structural explanation for the observed drops in model accuracy.

\paragraph{Productive Attractors (TE):} Conversely, Target Emphasis (TE) functions as a controlled and stable attractor. It provides a localized attention boost of $+1.1$ pp to the target token. This increase is sufficient to highlight the target's relative salience for downstream processing, yet small enough that it avoids eclipsing or starving the surrounding textual context. By preserving contextual awareness across the sequence, TE allows the model to leverage the targeted token without compromising broad comprehension.

\paragraph{Adversarial De-emphasis (ADE):} The Adversarial De-emphasis (ADE) pattern operates on the opposite spectrum. ADE actively suppresses late-layer attention directed at the target token. Instead of allowing semantic aggregation, it forces the model to shift minor attention focus back down to shallow, lexical layers (specifically peaking around Layer 2). As a result, the model semantically bypasses the target token during deep contextual synthesis, failing to integrate its meaning into the final output representations.

\subsection{Layer-Wise and Head-Wise Attention Dynamics in Vision}
\label{app:vlm_dynamics}
\begin{figure}[!t]
    \centering
    \begin{subfigure}{0.24\linewidth} % width of the subfigure
        \centering
        \includegraphics[width=\textwidth]{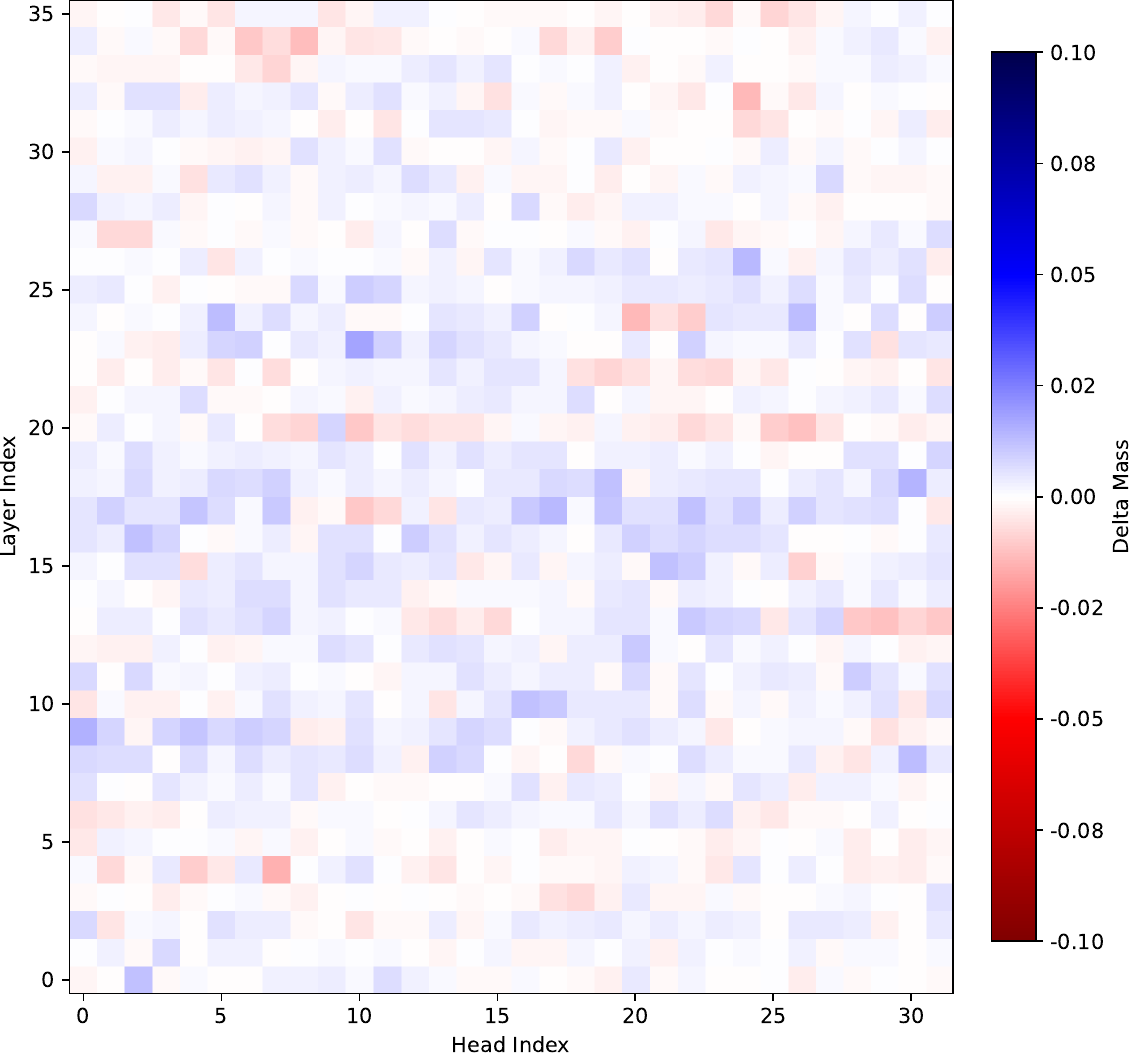}
        \caption{U2}
    \end{subfigure}% 
    \hfill
    \begin{subfigure}{0.24\linewidth} % width of the subfigure
        \centering
        \includegraphics[width=\textwidth]{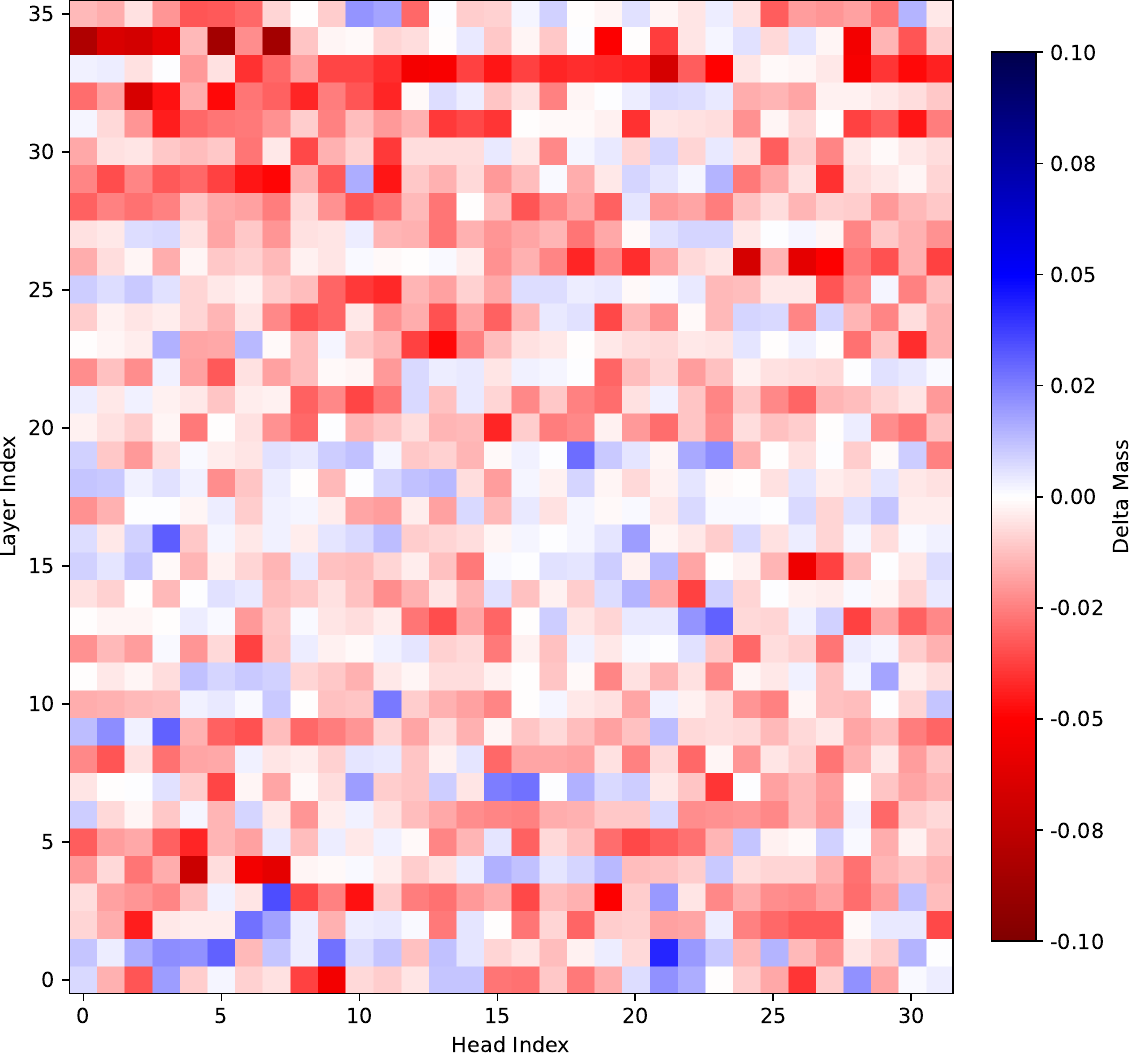}
        \caption{TE2}
    \end{subfigure}%
    \hfill
    \begin{subfigure}{0.24\linewidth} % width of the subfigure
        \centering
        \includegraphics[width=\textwidth]{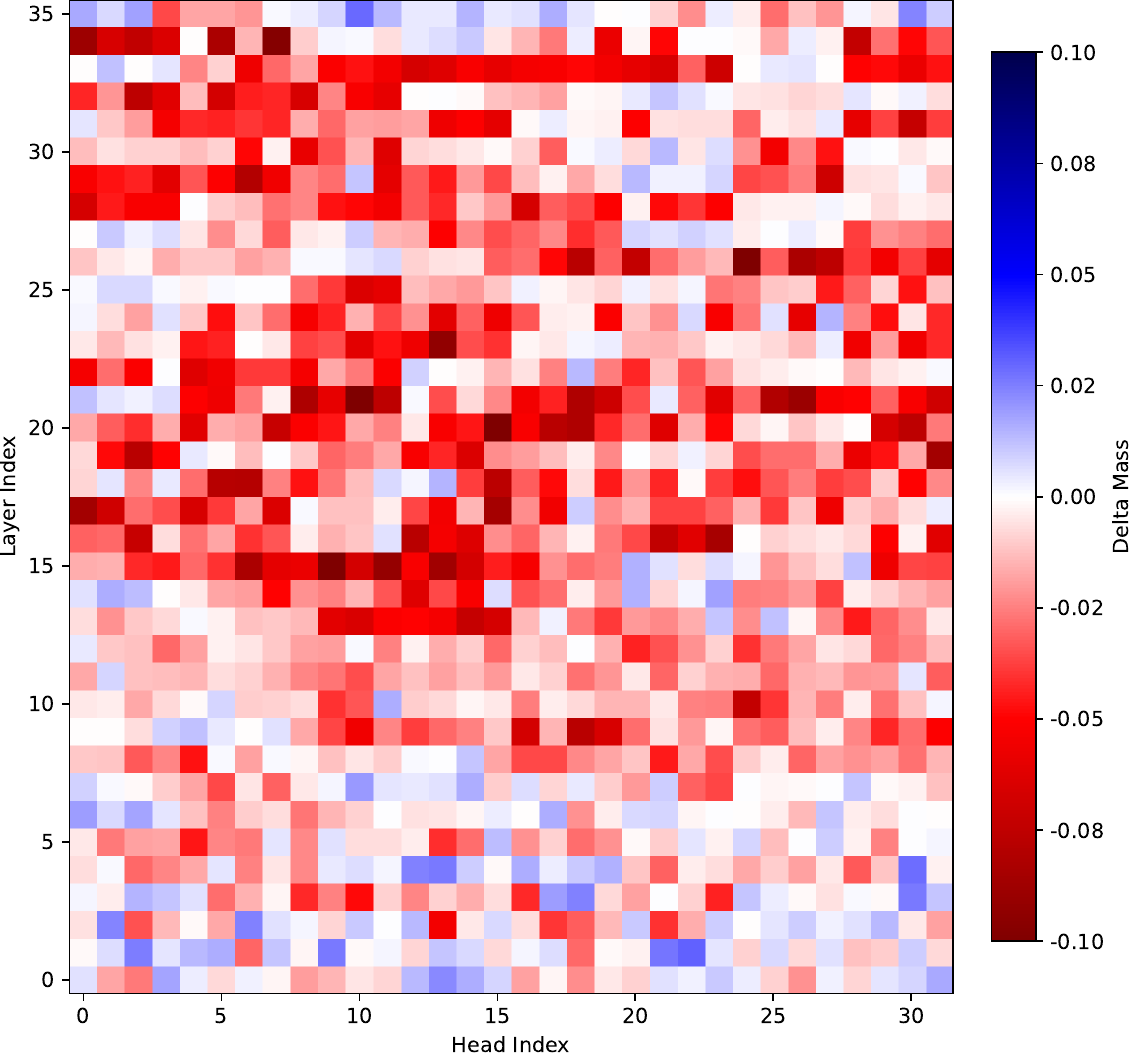}
        \caption{TA2}
    \end{subfigure}% 
    \hfill
    \begin{subfigure}{0.24\linewidth} % width of the subfigure
        \centering
        \includegraphics[width=\textwidth]{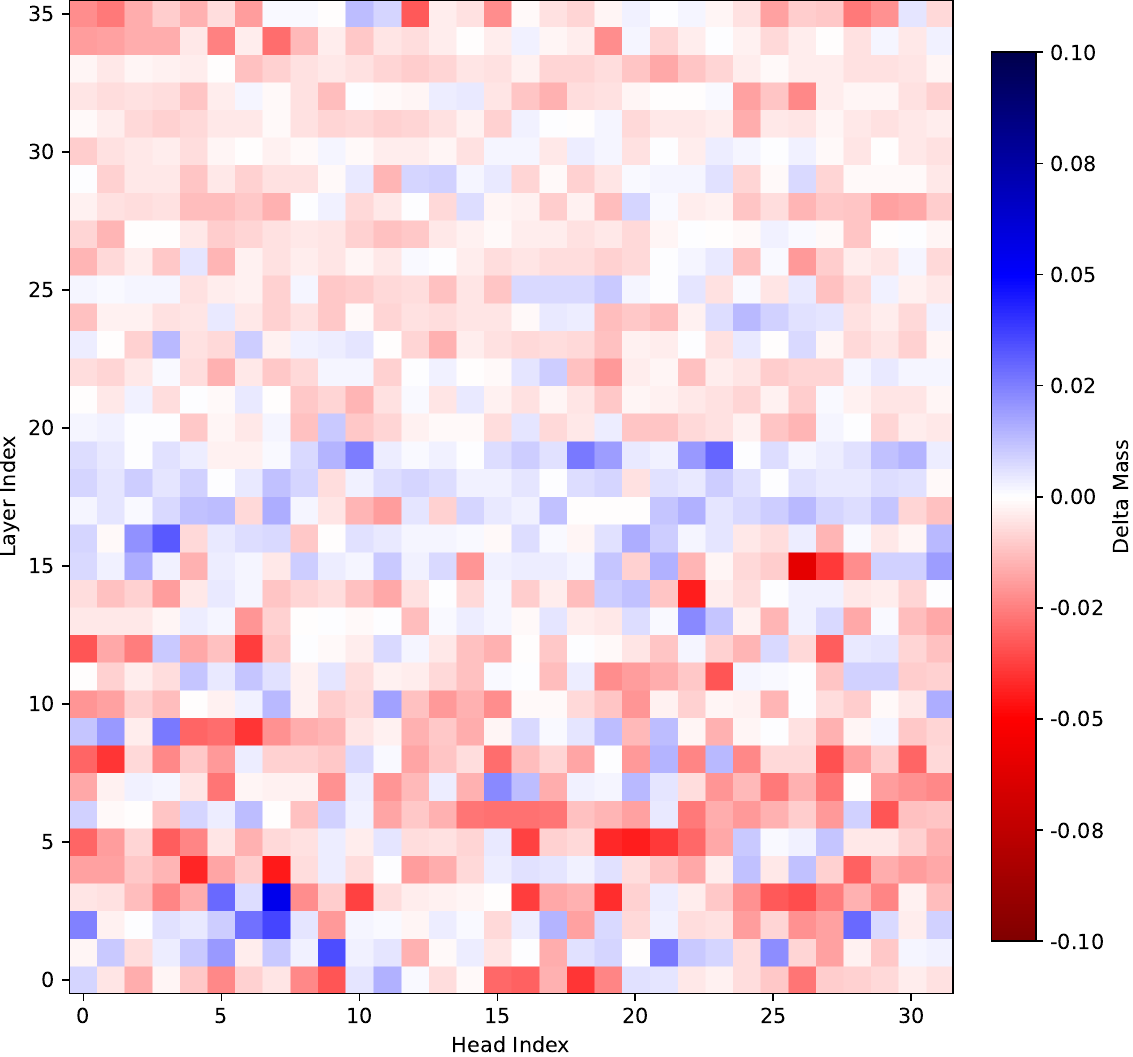}
        \caption{ADE2}
    \end{subfigure}%
    \caption{\textbf{Layer- and head-wise breakdown of relative visual attention allocation mass across intervention schemes on Qwen3-VL-4B-Instruct.} Attention shifts in modality are plotted across all layers and heads relative to a standard casing baseline. Blue regions denote a localized increase in image attention mass; red regions signify a reduction (\ie, redirection toward the text modality).}
    \label{fig:mechanistic_instights_VLM}
\end{figure}

The layer--head analysis of Qwen3-VL-4B-Instruct in \Cref{fig:mechanistic_instights_VLM} offers a mechanistic view of the macroscopic visual disengagement reported in \Cref{sec:exp:vlms}. As in the text-only analysis, the effect is \emph{localized in pattern but distributed in depth}; yet the localization differs. Rather than a single late-layer spike on the target token, casing induces a broad, head-spanning \emph{withdrawal} of visual attention. Relative to the standard-casing baseline, uniform lowercasing (U2) produces only weak, diffuse changes, consistent with its limited spatial and modality-level effects. In contrast, TE2 and TA2 induce widespread reductions in visual attention mass across many heads, becoming most pronounced in the middle and late layers where visual and linguistic representations are fused for grounding. The pattern-conflicting TE2 configuration therefore does not merely perturb a few specialized heads; it broadly redirects processing away from the visual feature maps and toward the typographically salient prompt stream. TA2 exhibits an even denser, model-wide suppression signature, mirroring at the head level the family-aggregate finding that alternating casing is the strongest cross-modal driver.

\paragraph{From local steering to visual disengagement.}
Reduced whole-image attention does not imply that the remaining visual attention is uninformative. The behavioral results indicate a \emph{simultaneous} concentration of the diminished visual budget onto the target bounding box: casing first shifts overall modality allocation toward text, while the residual visual attention becomes more spatially selective. Typographic salience can consequently concentrate a larger share of the (reduced) visual attention on the target region even as the model disengages from the image globally---an increase in attention routed to the target. This is the dual mechanism that distinguishes the multimodal regime from the purely textual one. ADE2 produces a more heterogeneous signature, combining shallow- and late-layer suppression with a localized band of increased attention around Layers~16--19, indicating an intermediate-stage redistribution rather than the broad withdrawal seen for TE2 and TA2.

\paragraph{De-emphasis does not suppress in the cross-modal setting.}
A notable departure from the text regime is that the Adversarial De-Emphasis (ADE) family ceases to act as a clean negative control. In text, lowercasing the target reliably suppresses attention to it; in the cross-modal setting, the target \emph{label} nonetheless occupies a fixed slot in the description prompt, and because casing routes attention via prompt-side salience \emph{contrast} rather than the absolute case of the target, the labeled image region continues to receive residual visual attention (\eg, ADE1 yields a mean $+1.44$ pp microscopic steering). This asymmetry is itself evidence for the text-dominance interpretation: what primarily is steered is the \emph{prompt-conditioned} routing of cross-modal attention.

\subsection{Representation Drift via Hidden State Analysis}
To confirm that these attention shifts introduce lasting alterations to the model's internal understanding, we analyze the model's hidden states using token-wise cosine similarity metrics. By comparing the latent embedding vectors produced under perturbed casing patterns against their standard baseline counterparts, we quantify the exact degree of semantic drift induced across the network.

The hidden state analysis directly corroborates our structural attention findings. As summarized in Table~\ref{tab:cosine_similarity}, the Target Alternating (TA) pattern induces a severe, systemic semantic drift, yielding a low mean cosine similarity of $0.341$ relative to the baseline. We observe maximum localized divergence gaps reaching up to $0.381$, proving that the attention hijacking at Layer 24 actively distorts the geometry of the latent vector space. In contrast, the productive Target Emphasis (TE) pattern maintains a significantly more stable representation profile, yielding a higher mean cosine similarity of $0.496$. This confirms that TE refines token salience without causing the underlying representation to drift into different regions of the embedding space.

\subsection{Architectural Mitigation Strategy}
Based on these mechanistic insights, we hypothesize that the model's vulnerability to destructive typographic attractors stems from unconstrained scale propagation through late-stage normalization layers. When a specific head over-allocates its probability mass due to an anomalous string token, the sudden scalar variance can distort downstream multi-head feature aggregation.

To mitigate this behavior, we hypothesize that a promising conceptual direction for neutralizing unwanted attractors in future architectures could be the introduction of \textit{case-agnostic Layer Normalization} immediately preceding late-stage semantic blocks. While verifying this mechanism remains an open avenue for future empirical work, we conjecture that by calculating normalization statistics across features structurally decoupled from orthographic case-salience indicators, the network could theoretically rescale anomalous attention surges. This architectural adjustment might effectively prevent high-entropy typographic perturbations from hijacking the bounded softmax distribution, allowing late synthesis layers to remain resilient against context-starvation and subsequent semantic drift.

\begin{table}[t]
    \centering
    \caption{\textbf{Latent representation drift within late semantic aggregation layers (Layer 24).} Token-wise mean cosine similarity metrics are computed against naturally cased baseline hidden states.}
    \label{tab:cosine_similarity}
    \begin{tabular}{lc}
        \toprule
        \textbf{Pattern} & \textbf{Mean Cosine Similarity} \\ 
        \midrule
        Target Emphasis (TE)        & 0.496 \\
        Target Alternating (TA)     & 0.341 \\ 
        \bottomrule
    \end{tabular}
\end{table}

\clearpage \newpage
\section{Additional Empirical Results and Analyses}
\label{appendix:additional_results}

The extended results provide a multi-dimensional view of how LLMs and VLMs process typographic signals across different tasks. For letter casing scheme reference, see \Cref{tab:casing_schemes}.

\subsection{Case Sensitivity in Non-Reasoning Models}
\paragraph{Benchmark-Specific Sensitivity.}
While the directionality of attention shifts is largely consistent across architectures, the magnitude and subsequent impact on performance vary by task type. Downstream task accuracy depends on the underlying task format, even when internal attention redirection patterns remain highly uniform across models. 

\begin{figure*}[!ht]
    \centering
    \begin{subfigure}{0.425\textwidth} % width of the subfigure
        \centering
        \includegraphics[width=\textwidth]{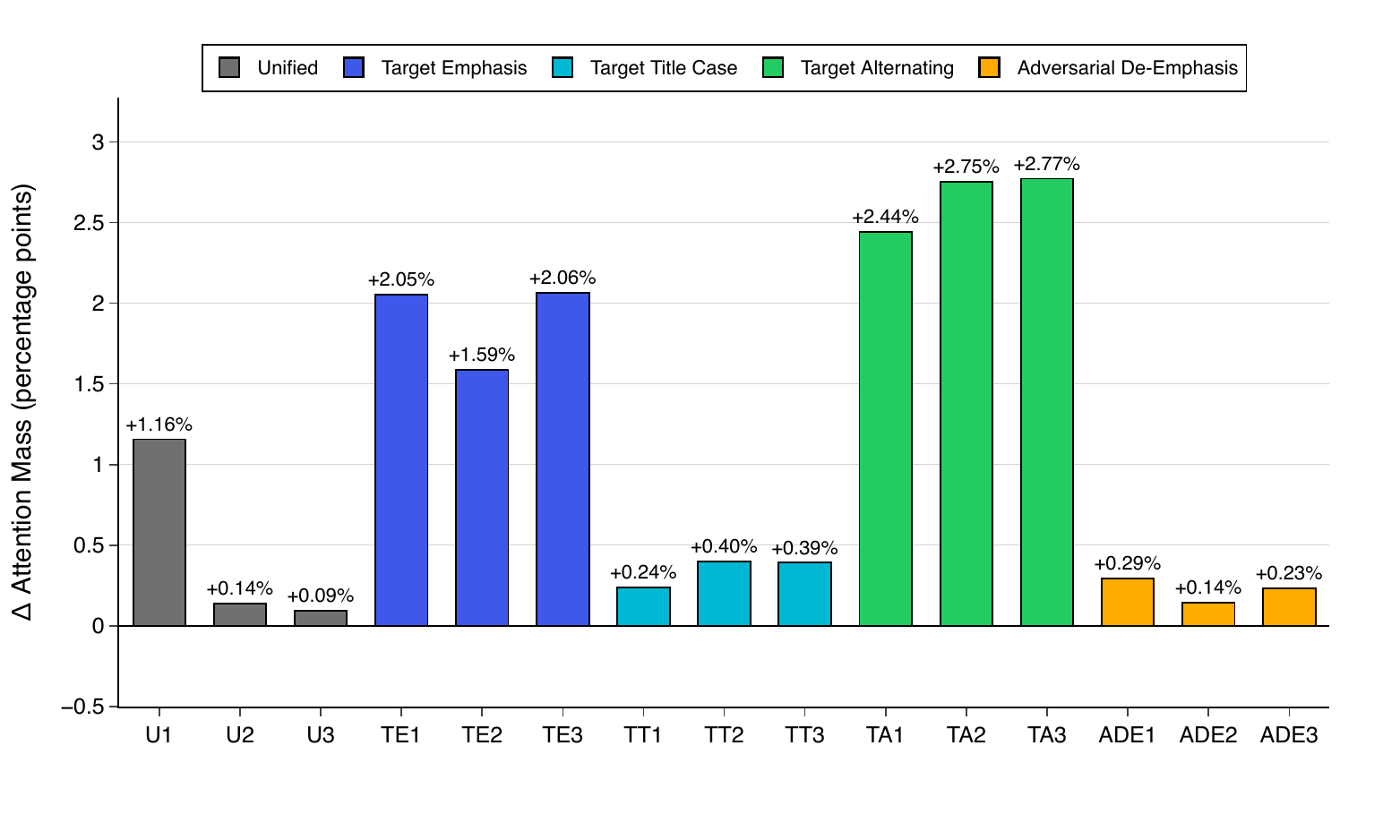}
        \caption{MMLU-Pro Attention Mass}
    \end{subfigure}% 
    \hspace{2em}
    \begin{subfigure}{0.425\textwidth} % width of the subfigure
        \centering
        \includegraphics[width=\textwidth]{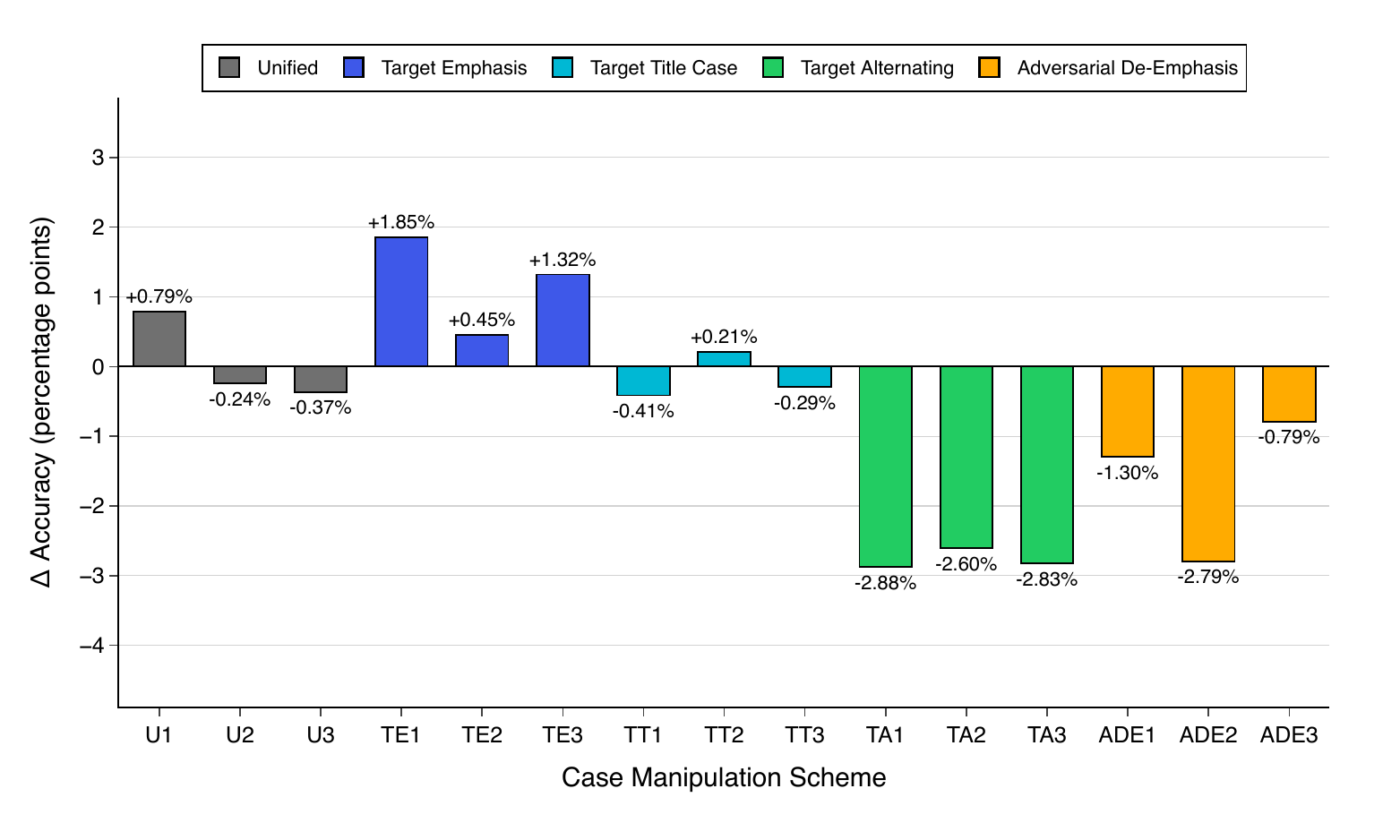}
        \caption{MMLU-Pro Task Accuracy}
    \end{subfigure}% 
    \vspace{0.5em}
    \begin{subfigure}{0.425\textwidth} % width of the subfigure
        \centering
        \includegraphics[width=\textwidth]{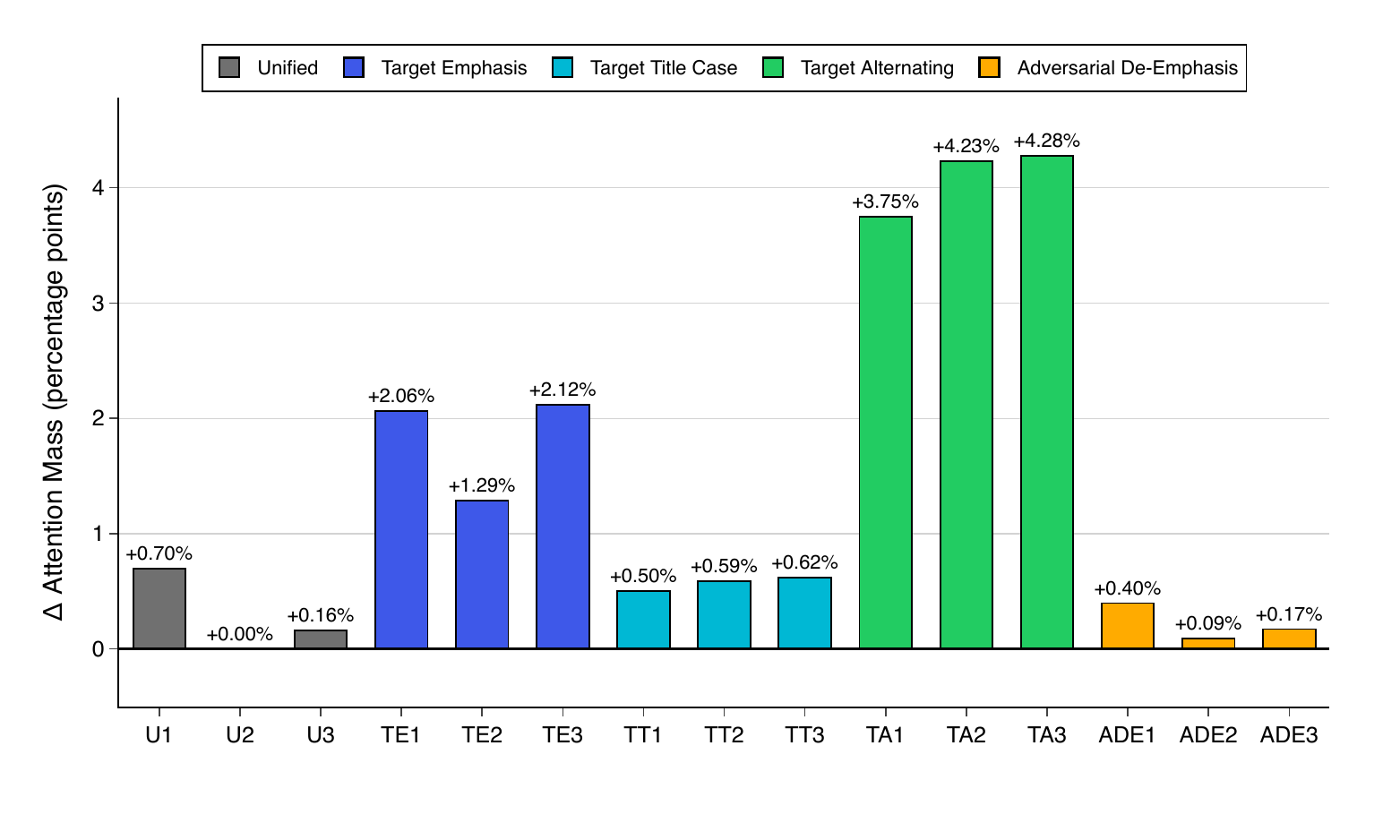}
        \caption{ARC-Challenge Attention Mass}
    \end{subfigure}% 
    \hspace{2em}
    \begin{subfigure}{0.425\textwidth} % width of the subfigure
        \centering
        \includegraphics[width=\textwidth]{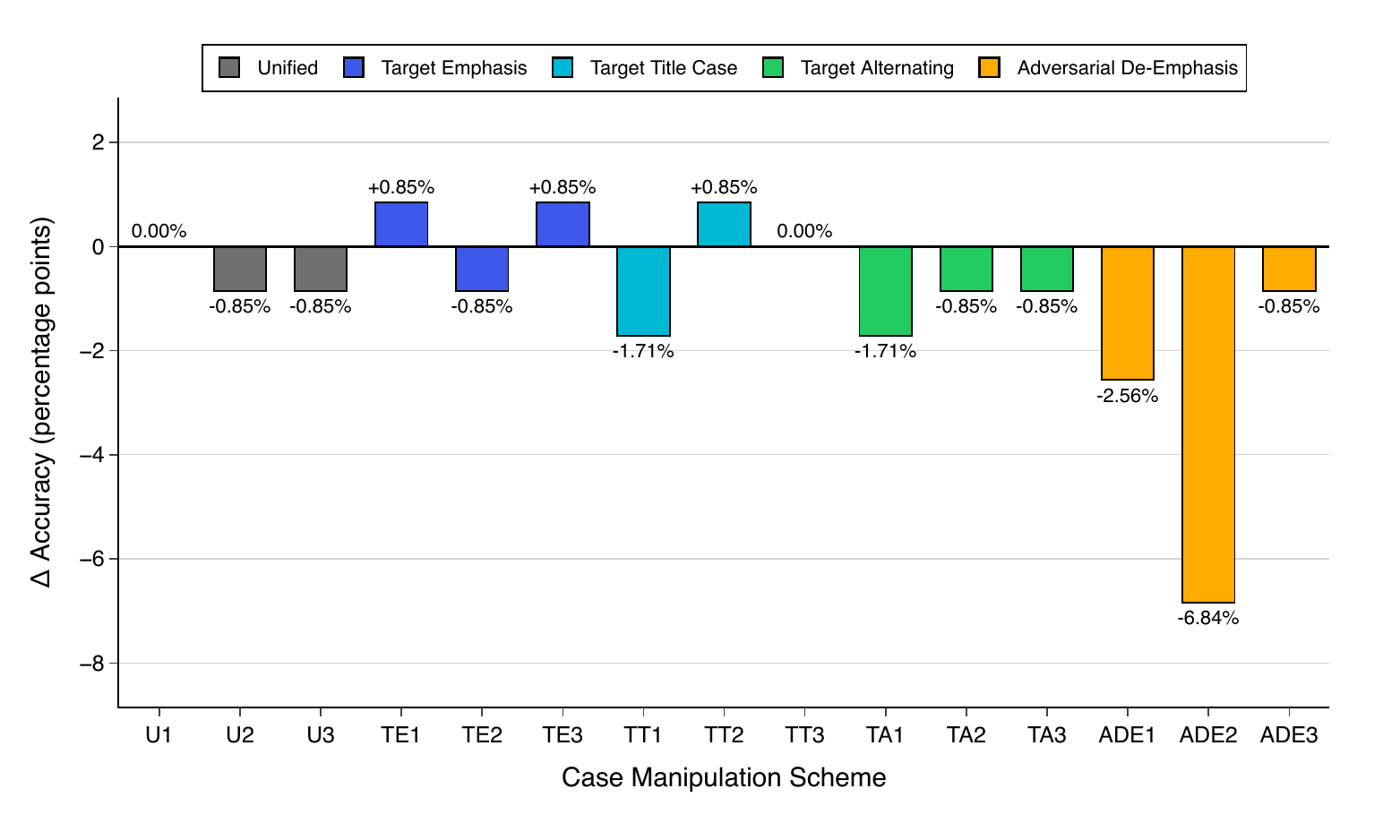}
        \caption{ARC-Challenge Task Accuracy}
    \end{subfigure}% 
    \vspace{0.5em}
    \begin{subfigure}{0.425\textwidth} % width of the subfigure
        \centering
        \includegraphics[width=\textwidth]{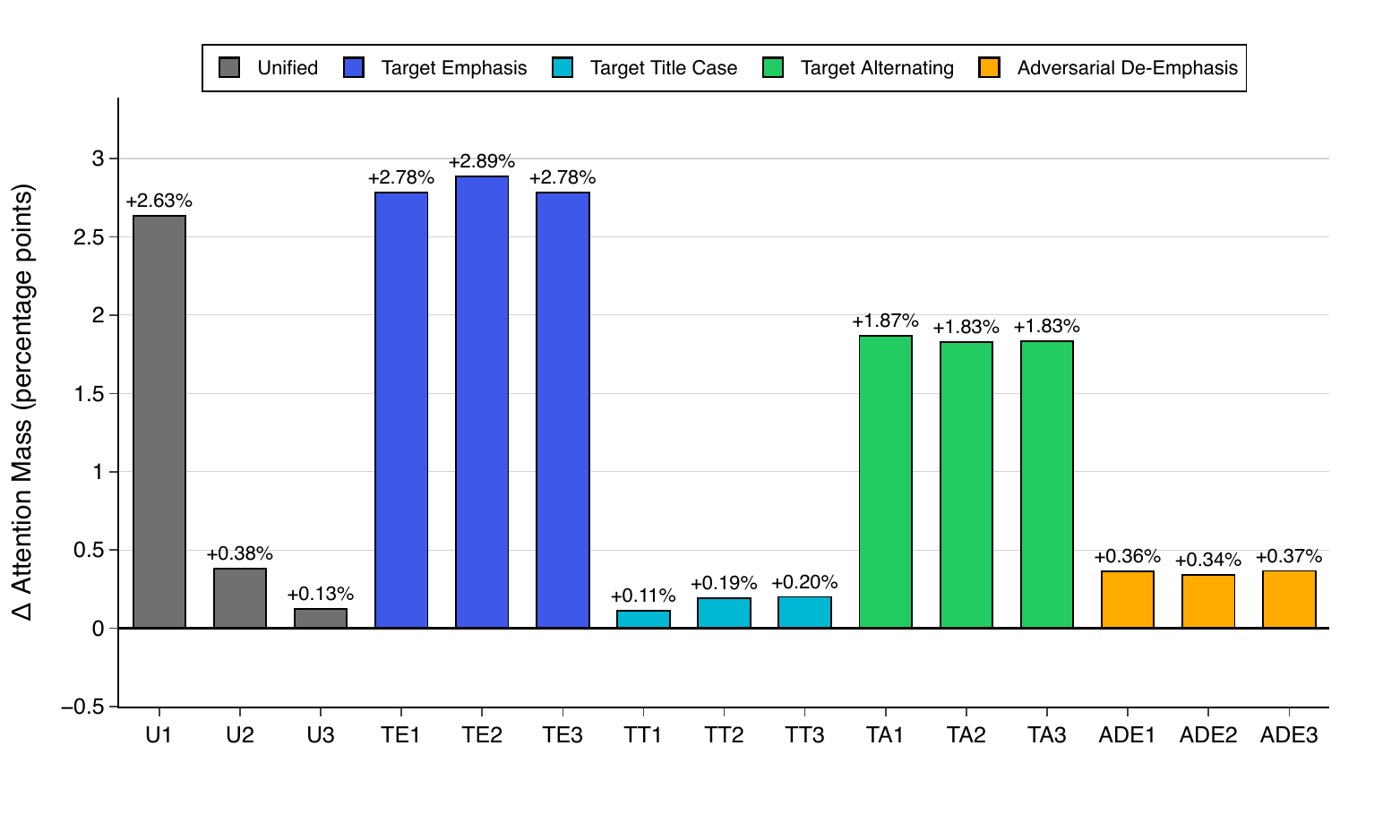}
        \caption{SQuADv2 Attention Mass}
    \end{subfigure}% 
    \hspace{2em}
    \begin{subfigure}{0.425\textwidth} % width of the subfigure
        \centering
        \includegraphics[width=\textwidth]{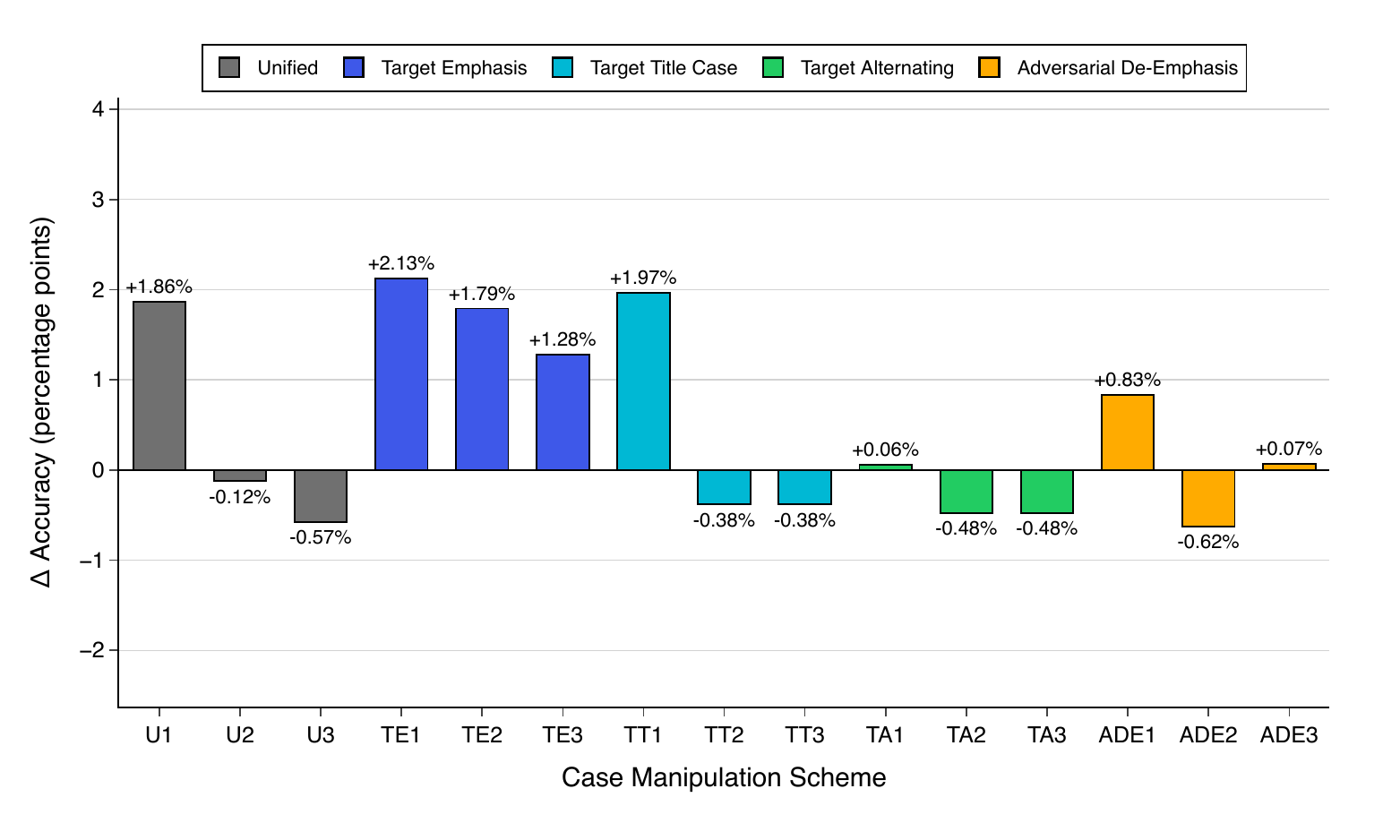}
        \caption{SQuADv2 Task Accuracy}
    \end{subfigure}% 
    \caption{\textbf{Per-benchmark breakdown of relative attention mass and task accuracy on the Qwen2.5-7B-Instruct model.} Bar charts illustrate the mean percentage change to the models' naturally cased baseline.}
    \label{fig:per_benchmark_qwen_breakdown}
\end{figure*}

\begin{itemize}
    \item \textbf{MMLU-Pro and ARC-Challenge}: These benchmarks, focusing on factual knowledge and scientific reasoning with multiple-choice answers, demonstrate the highest sensitivity to Target Emphasis (TE) schemes. As shown in \Cref{tab:schemes_absolutevals_llama_8b}, LLaMA-3.1-8B exhibits a substantial accuracy jump in MMLU-Pro from $29.59\%$ to $44.39\%$ ($+14.8$ pp) under TE1, accompanied by a $+29.4$ pp relative increase in attention mass. Conversely, in multiple-choice questions (MCQs), high-entropy Target Alternating (TA) patterns sharply disrupt the semantic coherence required for option-matching. This structural disruption directly penalizes accuracy (degrading by $-2.88$ pp on MMLU) despite generating a much higher localized attention concentration ($+4.28$ pp on ARC). Furthermore, as illustrated in the Qwen2.5-7B-Instruct breakdown (\cref{fig:per_benchmark_qwen_breakdown}), these tasks can diverge; \eg, in the TE2 setting, MMLU-Pro performance increases while ARC decreases, suggesting that complex scientific reasoning may be more sensitive to context-level typographic noise.
    
    \item \textbf{SQuADv2}: Reading comprehension exhibits unique attentional dynamics under case manipulation. In the Qwen2.5-7B-Instruct analysis, TE patterns exert a higher impact on attention mass in SQuADv2 compared to TA patterns or other benchmarks. Furthermore, global text interventions like U1 (All-Caps) induce a significant attention mass gain of $+2.63$ pp in this context. While extractive token matching is generally more robust to typographic variation, SQuADv2 shows notable performance gains of up to $2.13$ pp in the TE1 setting and a $1.97$ pp gain under TT1 as the clean orthographic contrast helps localize target sequences. However, within these extractive and abstention-driven environments, high-entropy TA setups similarly mislead the model's extraction logic, resulting in a performance degradation of $-0.48$ pp.
\end{itemize}

The detailed breakdown in \Cref{fig:per_benchmark_qwen_breakdown} further confirms the risk of structural misdirection, with the most severe performance degradation observed in the ADE2 setting on MMLU-Pro ($-6.84$ pp). Ultimately, our cross-benchmark analysis proves that productive TE formatting preserves necessary readability, consistently yielding robust performance gains across these text tasks ($+2.58$ pp MMLU, $+2.13$ pp SQuAD), whereas adversarial de-emphasis (ADE) systematically degrades generation accuracy. This suggests that while case acts as a helpful localizing anchor, it can simultaneously act as a powerful distractor when manipulated out-of-distribution.

\paragraph{The Destructive Attractor: Target Alternating (TA).}
The absolute values confirm that \textbf{TA2 and TA3} (Alternating Case) are the most potent attention attractors across every single model tested. In the Qwen2.5-7B model (\cf \cref{tab:schemes_absolutevals_qwen_7b}), TA2 reaches an attention mass of $0.104$ on ARC-Challenge~\cite{Chollet_2019_arXiv_ARC}; nearly double the baseline ($0.061$). However, this same intervention causes a significant accuracy drop in nearly all LLaMA and Qwen variants. This empirical divergence supports our hypothesis that high-entropy casing disrupts the semantic coherence of the token embeddings while simultaneously triggering attention.

\paragraph{Scaling and Architecture Robustness.}
Comparing the 3B and 8B LLaMA models, as well as the 7B and 14B Qwen models, we observe that larger parameter counts do not ``wash out'' the casing effect. If anything, larger models (like Qwen2.5-14B) show more refined utilization of productive attractors (TE1/TE3), achieving significant gains in MMLU-Pro ($+9.06$ pp) while maintaining high baseline performance. This suggests that case sensitivity is a property that scales with model capacity rather than an artifact of smaller, under-trained architectures.

\paragraph{Adversarial Validation.}
The Adversarial De-Emphasis (ADE) results serve as a consistent negative control. Across all tables, ADE schemes generally result in attention mass values that are either equal to or lower than the baseline. This confirms that the attention mechanism is not merely reacting to \textit{any} change in the target span, but specifically to \textit{salient typographic contrast} (\eg, uppercase).

\subsection{Case Sensitivity in Reasoning Models}
\label{appendix:reasoning_models}

To further characterize the boundaries of the ``casing effect,'' we extended our study to include two state-of-the-art reasoning models: Qwen3-4B-Thinking-2507~\cite{qwen3_4b_thinking_2507} and gpt-oss-20B~\cite{openai_2025_gptoss20b}. These models are distinguished by their \textit{Chain-of-Thought} or \textit{Thinking} phases, where the model performs internal deliberation before generating a final response. Our results reveal a strikingly different behavior compared to standard instruction-tuned models.

\begin{figure*}[!ht]
    \centering
    \begin{subfigure}{0.425\textwidth} % width of the subfigure
        \centering
        \includegraphics[width=\textwidth]{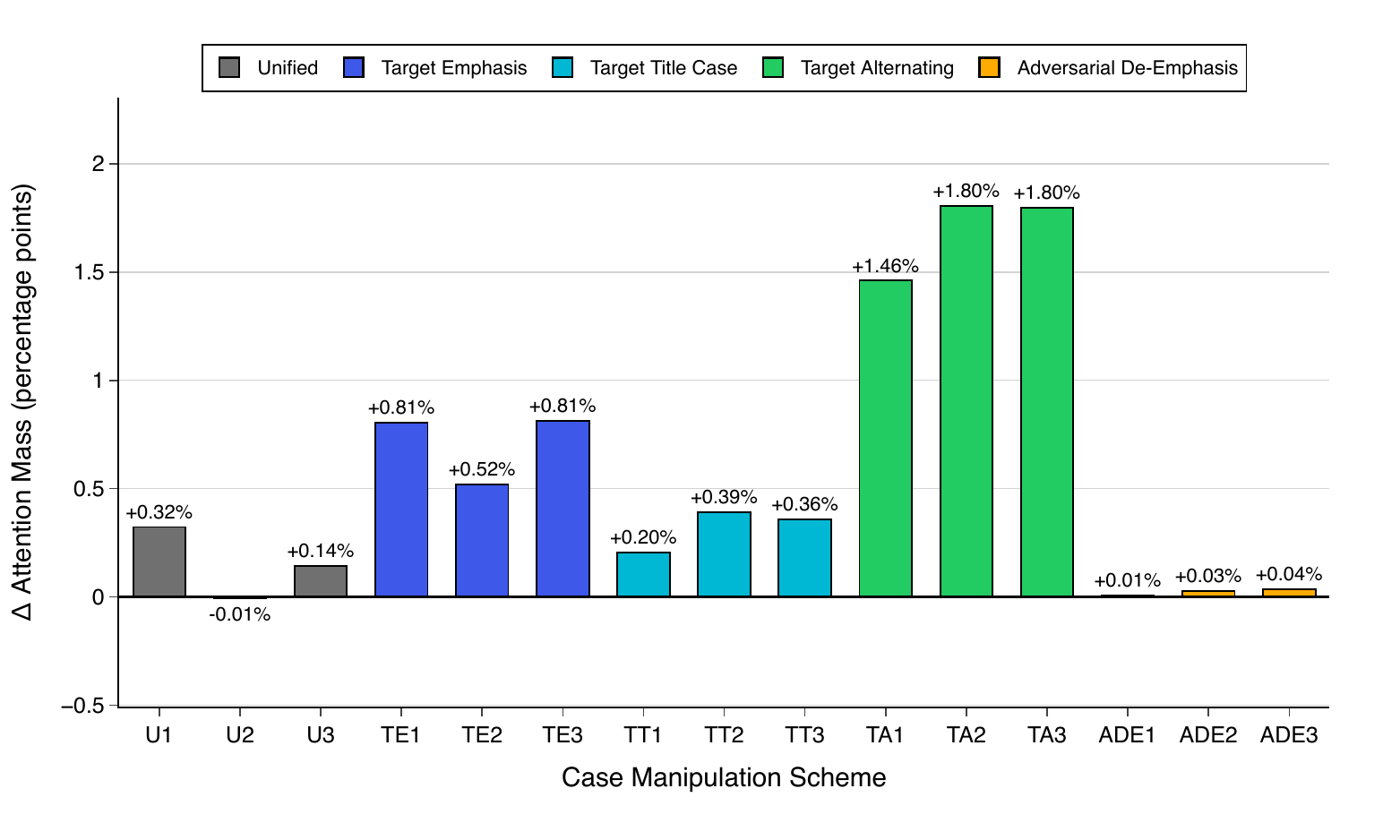}
        \caption{LLaMA-3.2-3B-Instruct}
    \end{subfigure}% 
    \hspace{2em}
    \begin{subfigure}{0.425\textwidth} % width of the subfigure
        \centering
        \includegraphics[width=\textwidth]{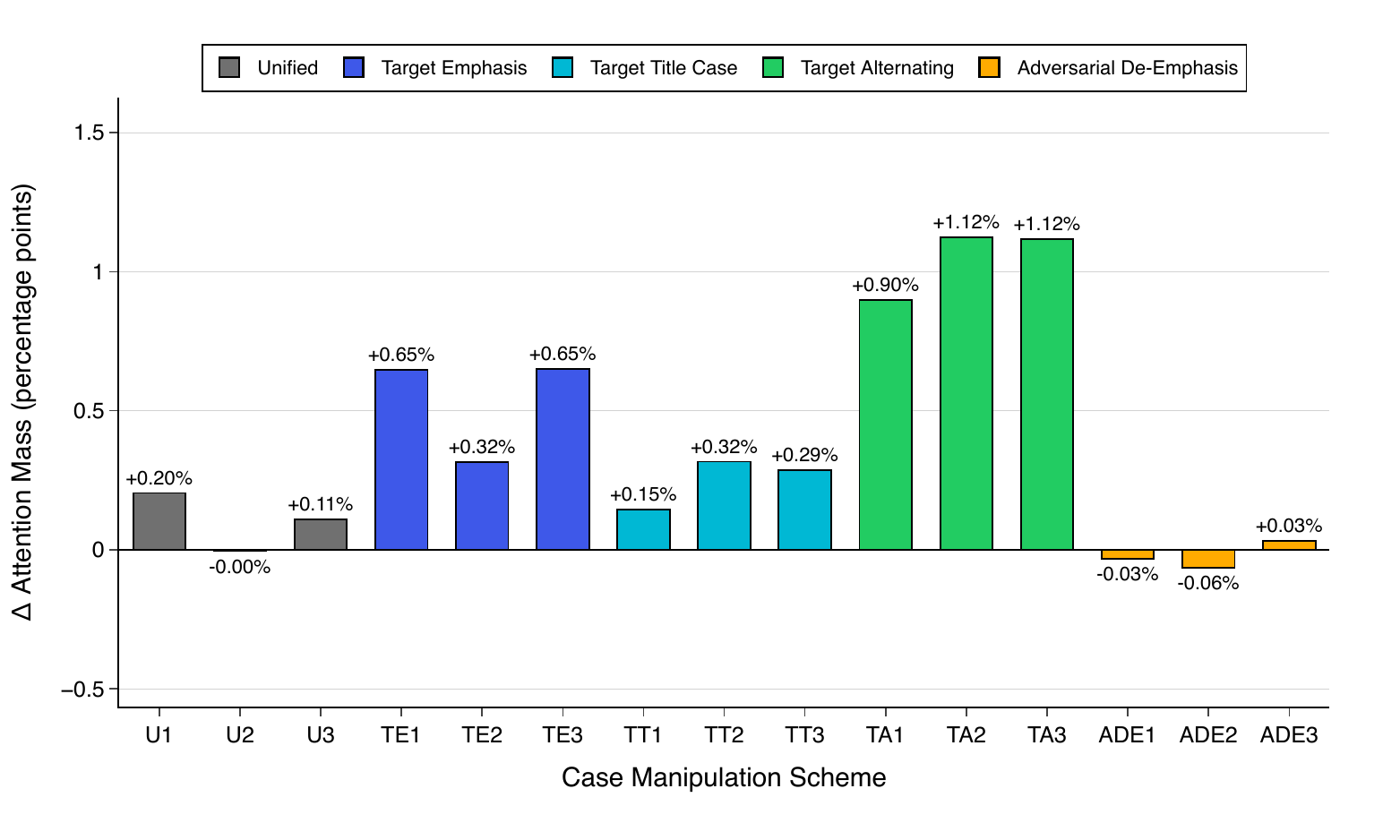}
        \caption{LLaMA-3.1-8B-Instruct}
    \end{subfigure}% 
    \vspace{0.5em}
    \begin{subfigure}{0.425\textwidth} % width of the subfigure
        \centering
        \includegraphics[width=\textwidth]{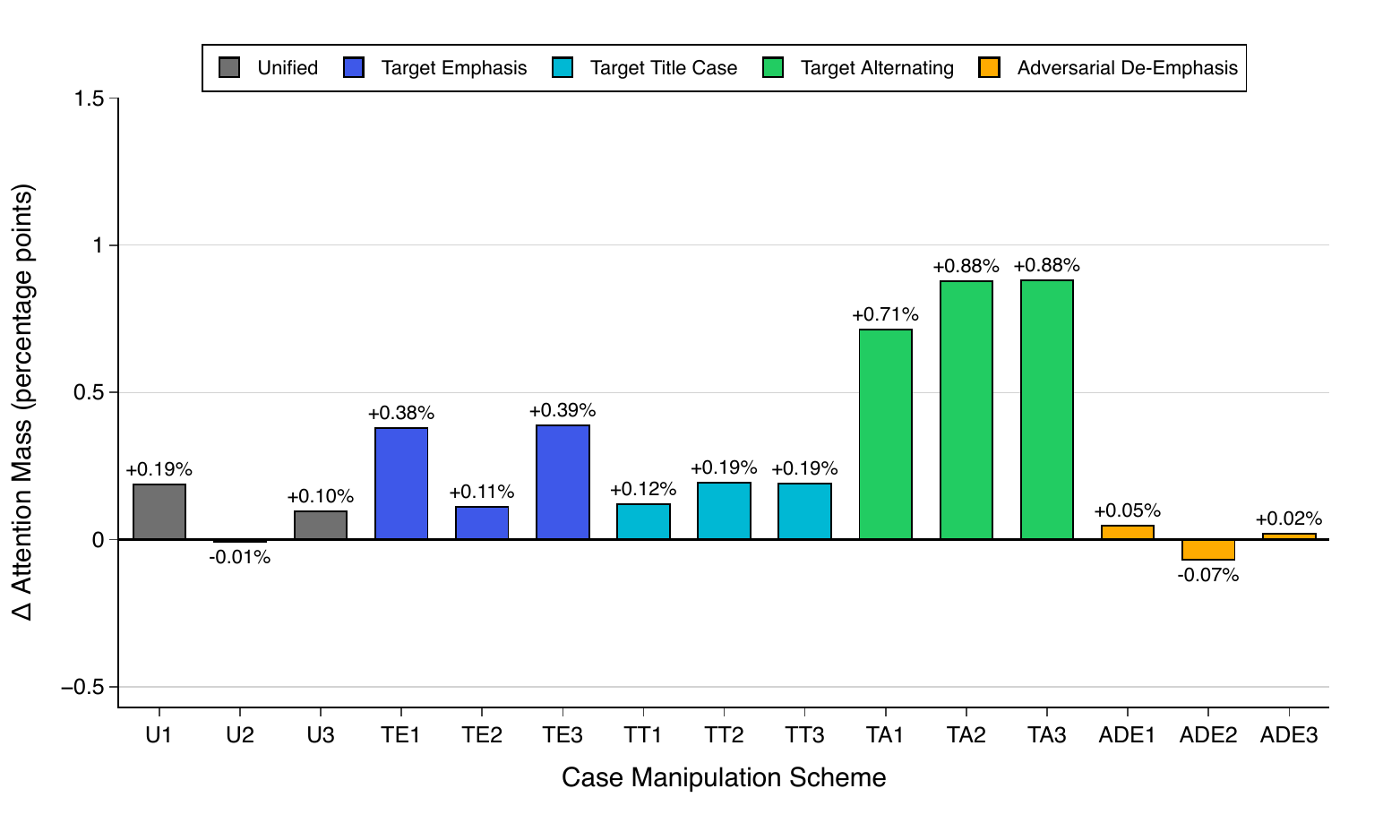}
        \caption{Gemma-3-1B-IT}
    \end{subfigure}% 
    \hspace{2em}
    \begin{subfigure}{0.425\textwidth} % width of the subfigure
        \centering
        \includegraphics[width=\textwidth]{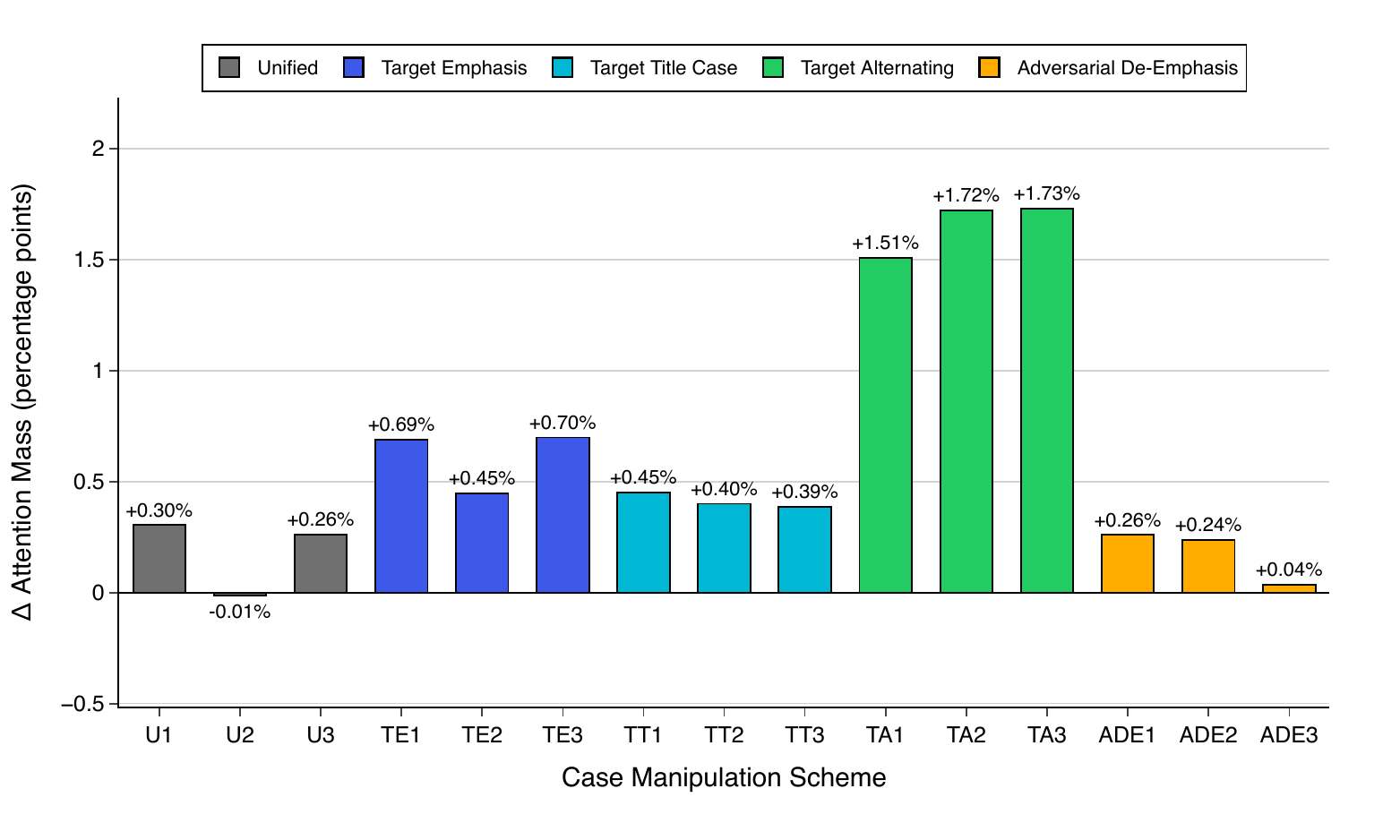}
        \caption{Gemma-2-2B-IT}
    \end{subfigure}% 
    \vspace{0.5em}
    \begin{subfigure}{0.425\textwidth} % width of the subfigure
        \centering
        \includegraphics[width=\textwidth]{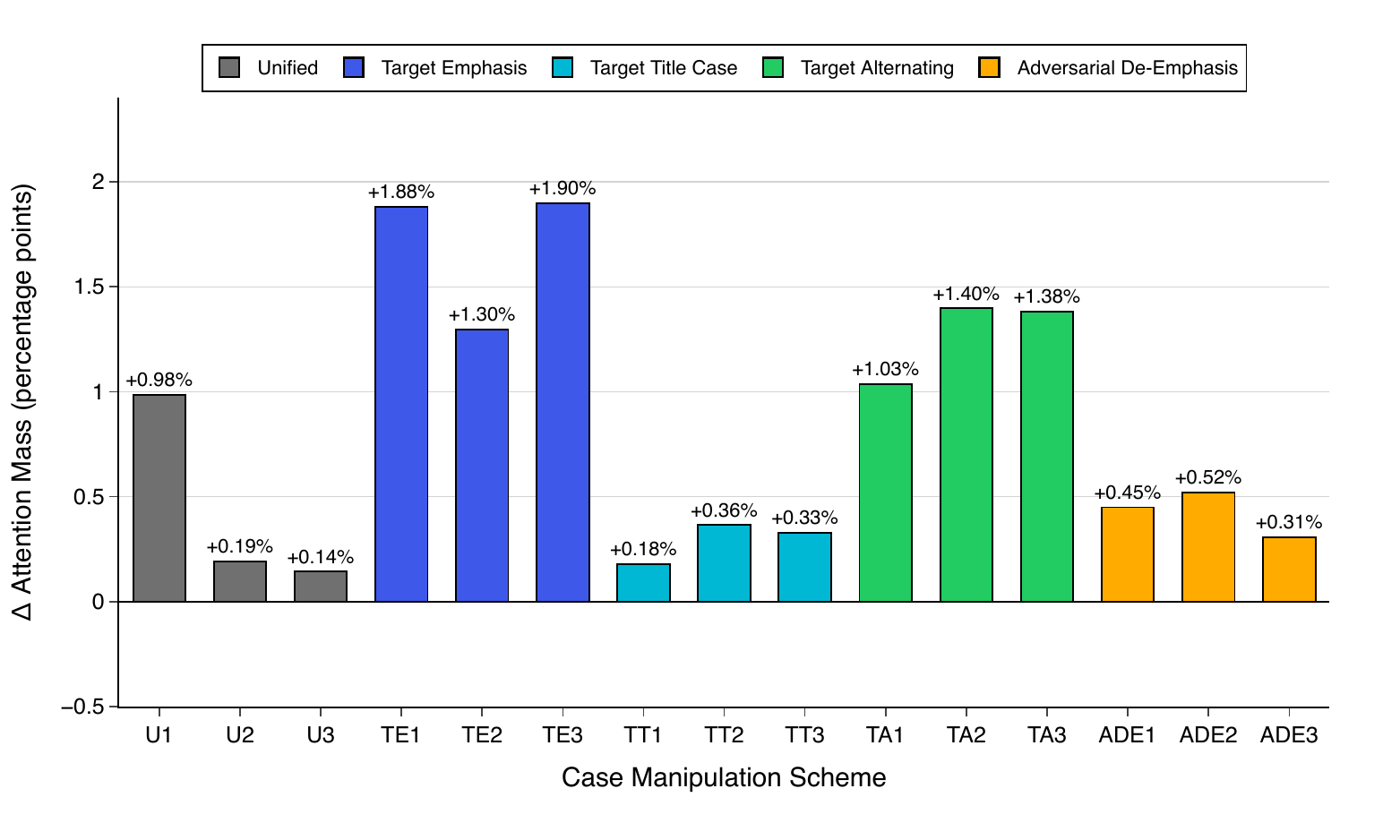}
        \caption{Mistral-7B-Instruct-v0.3}
    \end{subfigure}% 
    \hspace{2em}
    \begin{subfigure}{0.425\textwidth} % width of the subfigure
        \centering
        \includegraphics[width=\textwidth]{Figures/case_steering_attention_mass_Qwen_7b.pdf}
        \caption{Qwen2.5-7B-Instruct}
    \end{subfigure}% 
    \vspace{0.5em}
    \begin{subfigure}{0.425\textwidth} % width of the subfigure
        \centering
        \includegraphics[width=\textwidth]{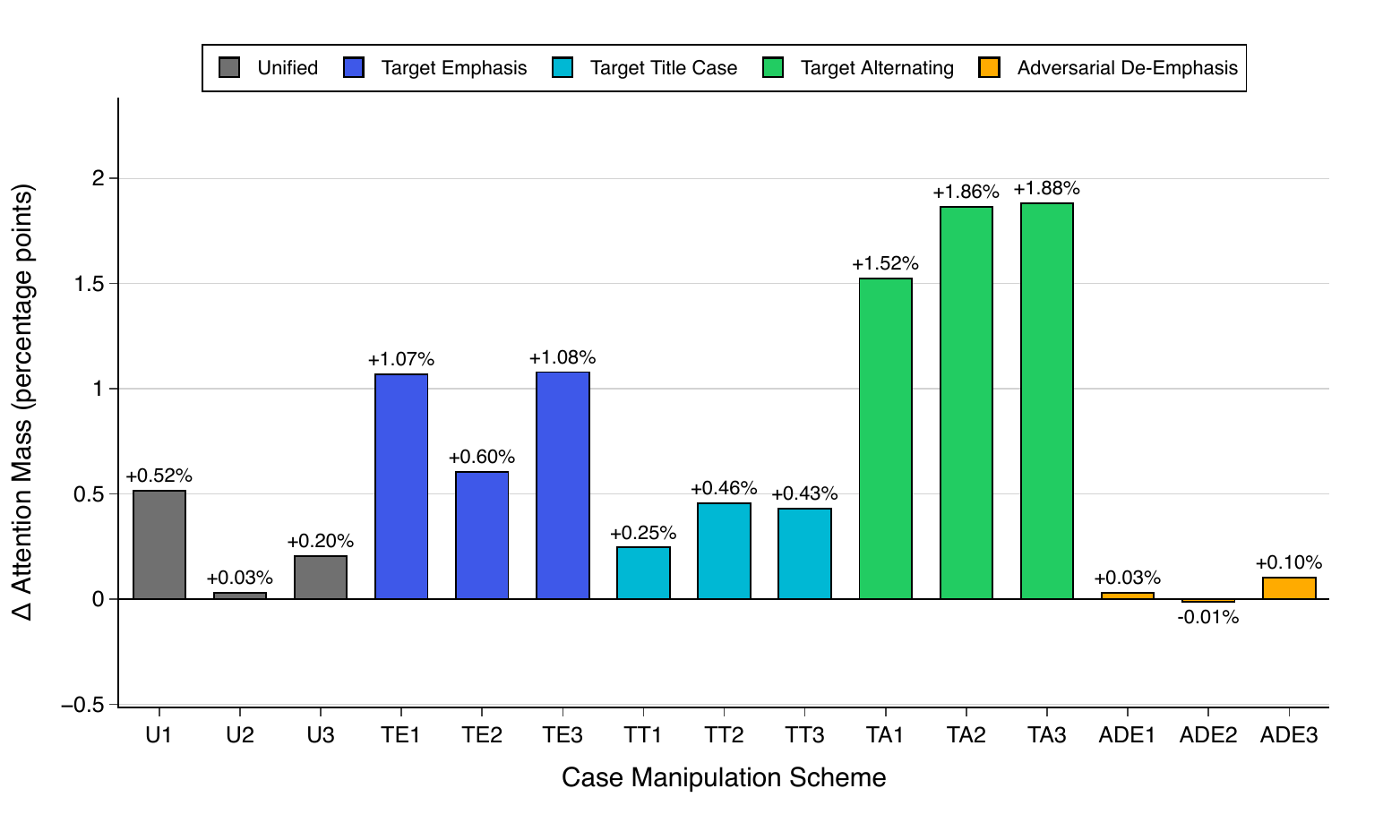}
        \caption{Qwen2.5-14B-Instruct}
    \end{subfigure}% 
    \caption{\textbf{Relative attention mass across model families and intervention schemes.} Bar charts illustrate the mean percentage change in attention allocation relative to each model's naturally cased baseline.}
    \label{fig:attentionmass_per_scheme}
\end{figure*}

\begin{figure*}[!ht]
    \centering
    \begin{subfigure}{0.425\textwidth} % width of the subfigure
        \centering
        \includegraphics[width=\textwidth]{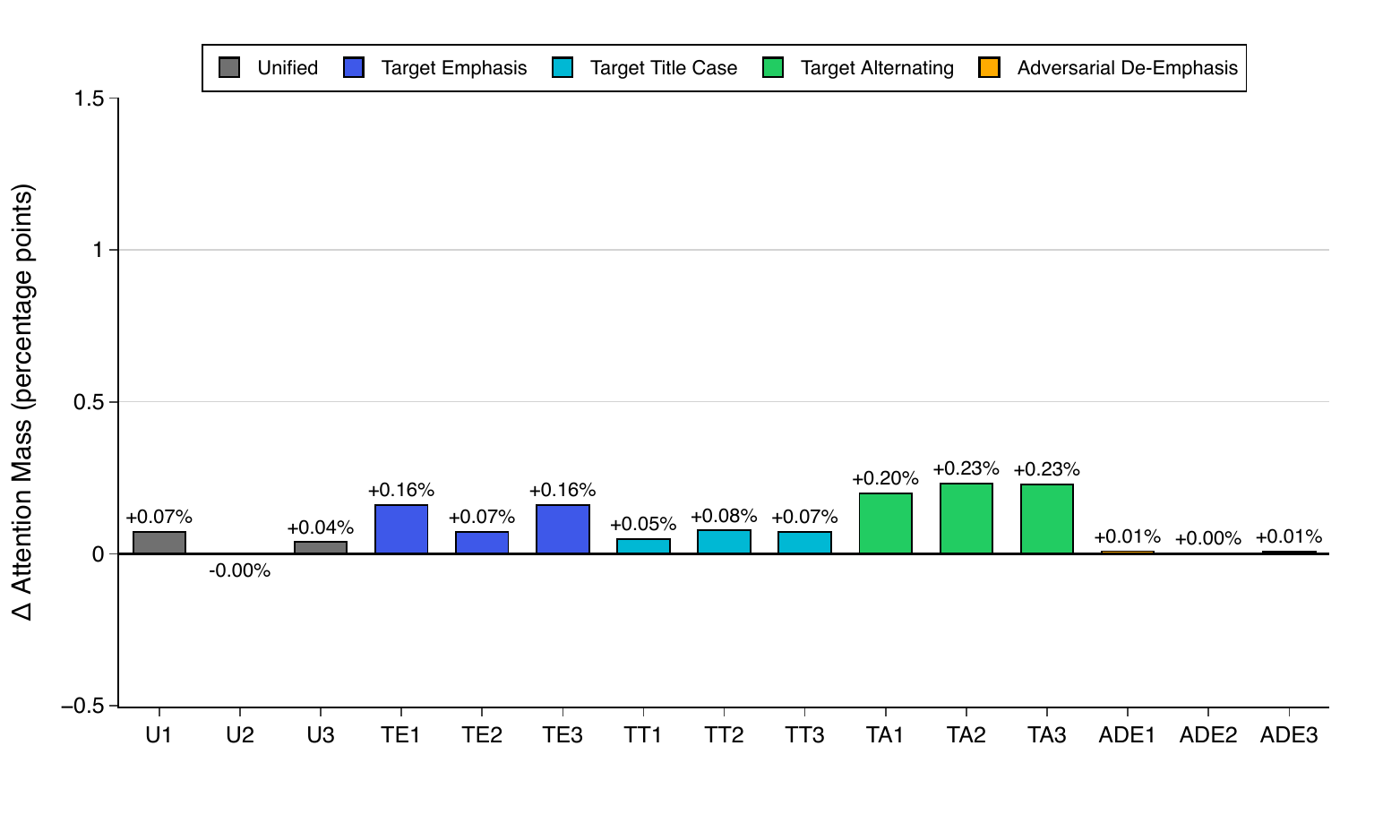}
        \caption{Qwen3-4B-Thinking-2507}
    \end{subfigure}% 
    \hspace{2em}
    \begin{subfigure}{0.425\textwidth} % width of the subfigure
        \centering
        \includegraphics[width=\textwidth]{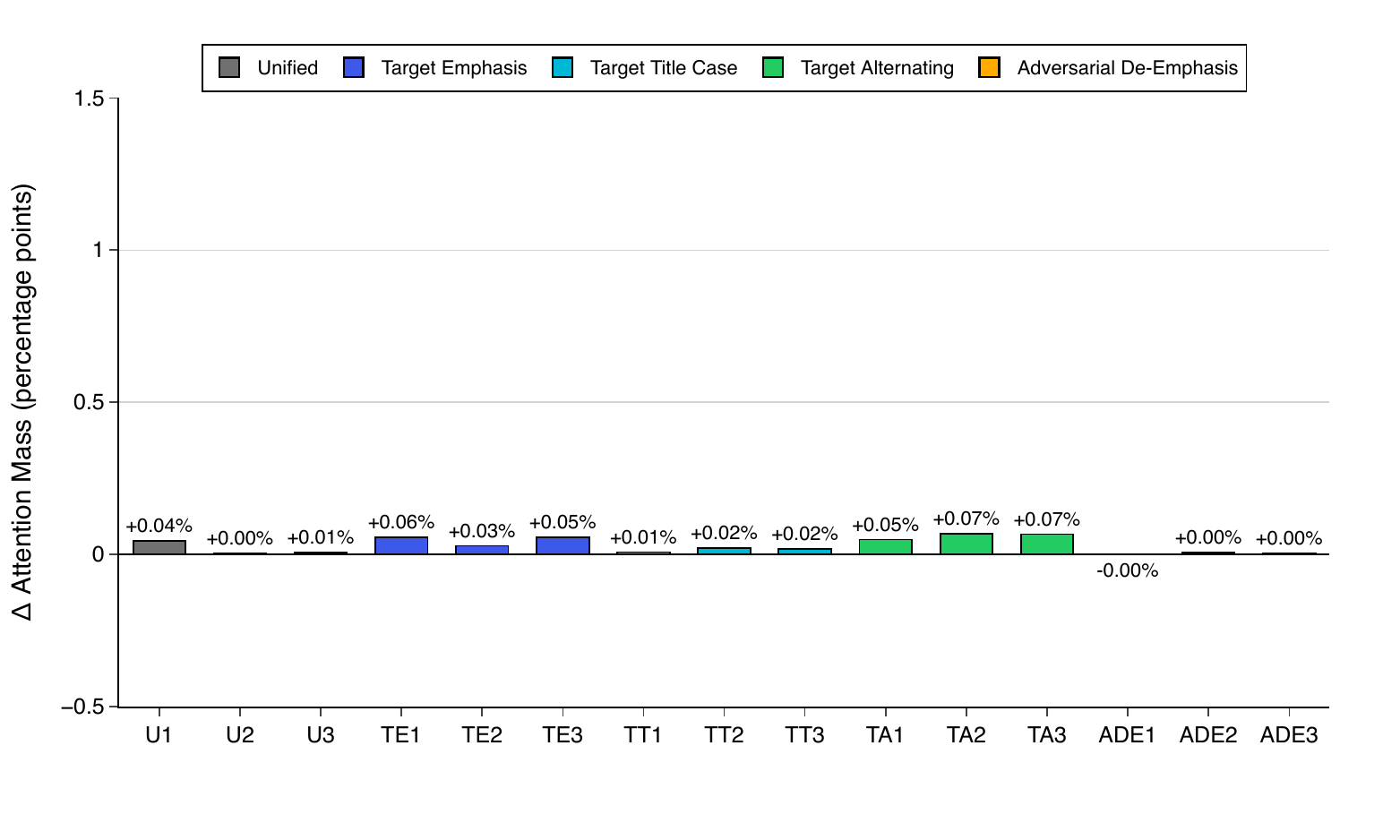}
        \caption{gpt-oss-20B}
    \end{subfigure}% 
    \caption{\textbf{Relative attention mass across two reasoning models and intervention schemes.} Bar charts illustrate the mean percentage change in attention allocation relative to each model's naturally cased baseline.}
    \label{fig:attentionmass_reasoning_per_scheme}
\end{figure*}

\paragraph{Mitigation of Attentional Shifts.}
A comparative analysis between standard architectures (\cf \cref{fig:attentionmass_per_scheme}) and reasoning-oriented models (\cf \cref{fig:attentionmass_reasoning_per_scheme}) indicates that the deliberative reasoning process nearly completely mitigates the shifts in attention mass. For Qwen3-4B-Thinking, the highest observed attention shifts remain near zero, with TA and TE schemes yielding marginal gains of only $+0.23$ pp and $+0.16$ pp, respectively. This mitigation is even more pronounced in gpt-oss-20B, where the peak shifts for TA and TE are limited to $+0.07$ pp and $+0.05$ pp. While the relative hierarchy of attractors remains identical (TA and TE schemes still exert the most influence), the magnitude of this influence is reduced by magnitudes.

\begin{figure*}[!ht]
    \centering
    \begin{subfigure}{0.425\textwidth} % width of the subfigure
        \centering
        \includegraphics[width=\textwidth]{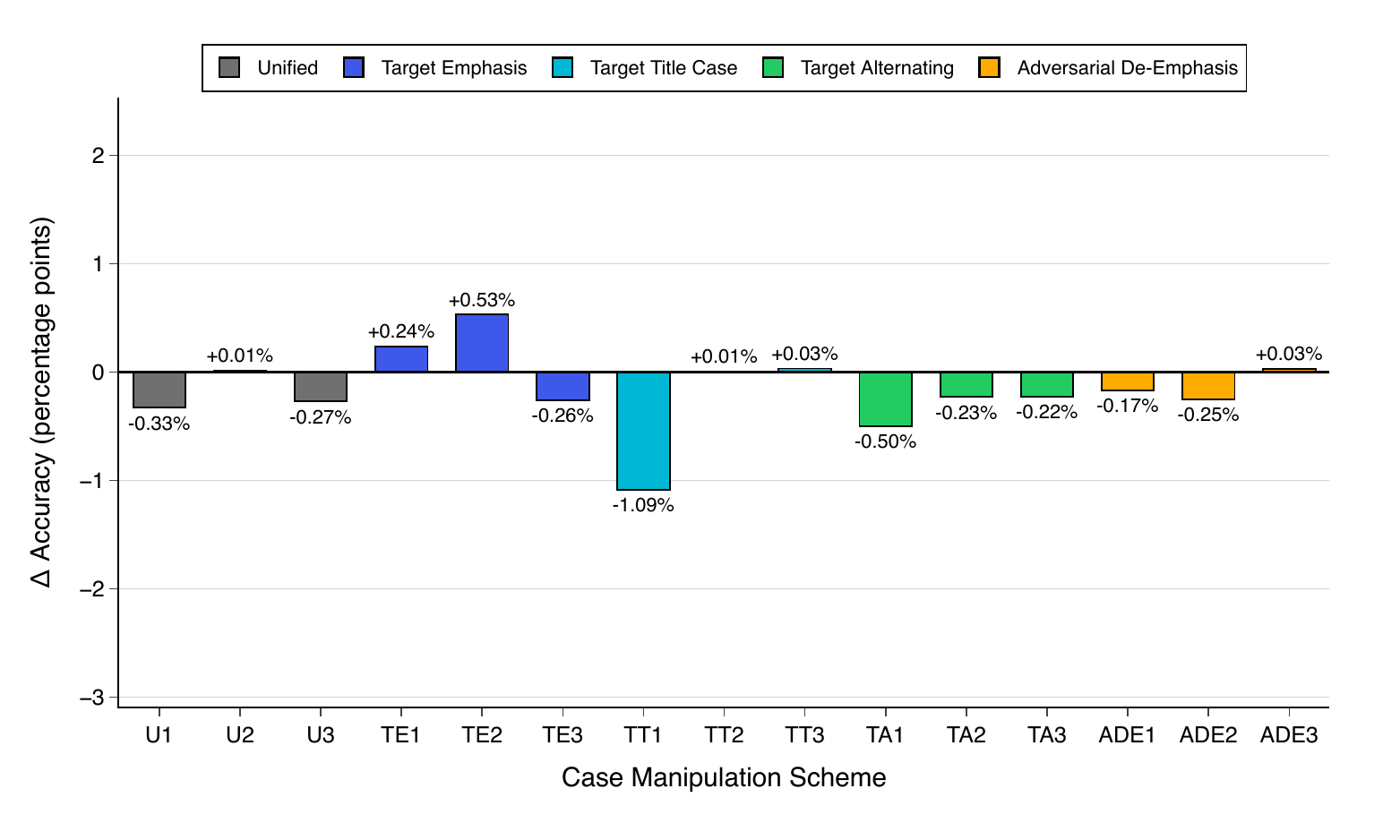}
        \caption{Qwen3-4B-Thinking-2507}
    \end{subfigure}% 
    \hspace{2em}
    \begin{subfigure}{0.425\textwidth} % width of the subfigure
        \centering
        \includegraphics[width=\textwidth]{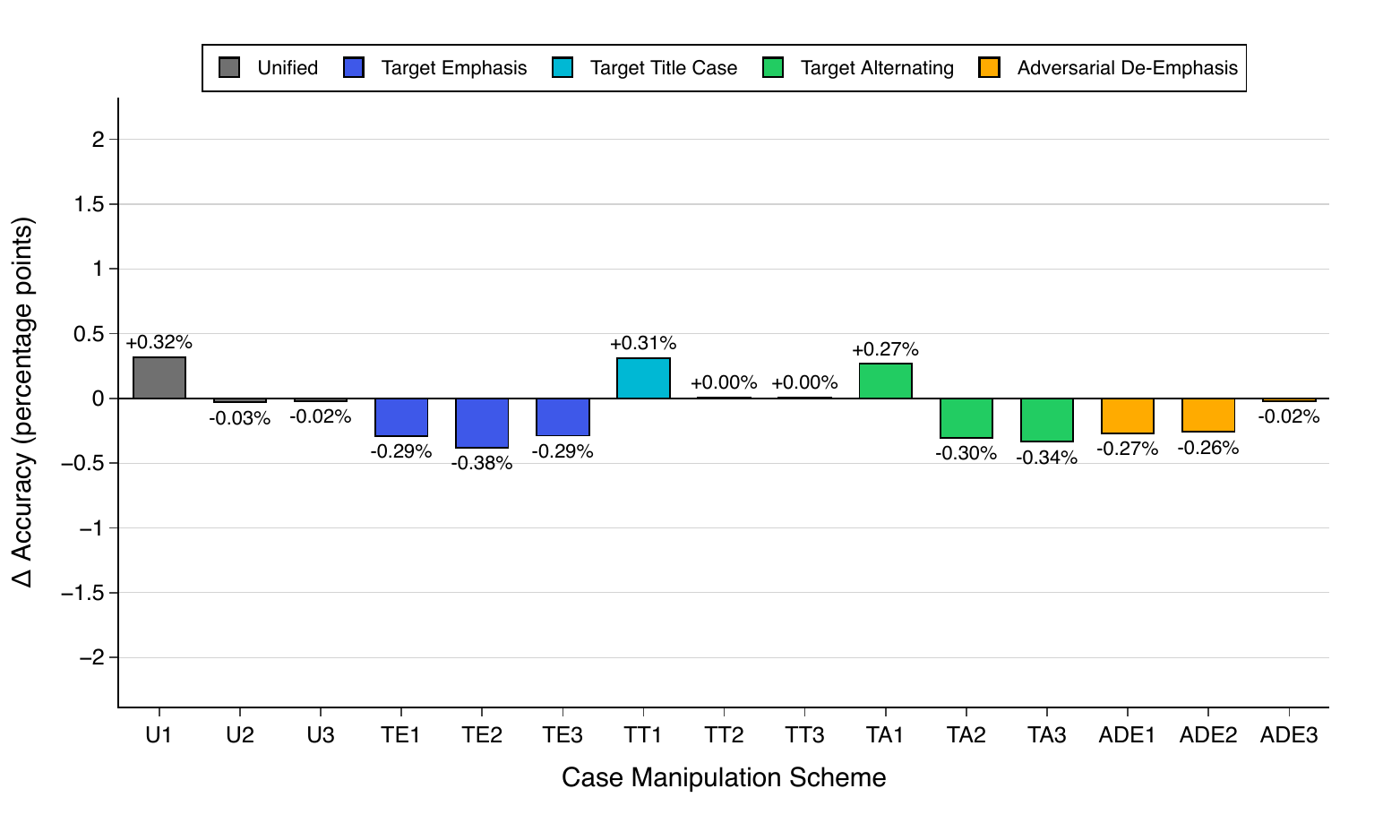}
        \caption{gpt-oss-20B}
    \end{subfigure}% 
    \caption{\textbf{Impact of typographic interventions on downstream task accuracy for two reasoning models.} We report the relative change in mean accuracy across three benchmarks compared to naturally cased baselines.}  
    \label{fig:accuracy_reasoning_per_scheme}
\end{figure*}

\paragraph{Performance Stability.}
Downstream task accuracy similarly exhibits negligible sensitivity to typographic variation in reasoning models (\cf \cref{fig:accuracy_reasoning_per_scheme}). Relative accuracy fluctuations remained consistently within a $\pm0.5$ pp range across all letter casing schemes. Minor exceptions were noted in the Qwen3-4B-Thinking model, specifically a $+0.53$ pp increase in TE2 and a $-1.09$ pp decrease in TT1, yet these remain outliers compared to the double-digit performance swings observed in standard models (\eg, LLaMA-3.1-8B).

\paragraph{Analysis: The Reasoning Buffer.}
We hypothesize that this lack of sensitivity stems from the \textit{internal re-alignment} inherent in reasoning models. During the ``Thinking'' phase, these models typically map input tokens into a higher-level abstract space to perform logical planning. This intermediate step likely acts as a semantic buffer, where surface-level orthographic features are normalized or discarded in favor of core lexical identities. Consequently, the ``casing effect''---while inherent to the underlying transformer weights---is notably decreased by the model's deliberative focus on logical coherence over surface-form salience. This suggests that case-sensitive attention steering is a property most applicable to models operating in a direct-prediction regime rather than those employing extended reasoning trajectories.

\subsection{Self-Reported Attentional Focus: Bridging Mass and Steering}
\label{sec:active_interrogation}

To connect attention mass shifts with behavioral effects, we introduce an \textit{active interrogation} study. The goal is to test whether case-induced attention concentration translates into an explicitly identifiable dominant signal.

\paragraph{Experimental Protocol.}
We construct 500 semantically neutral sentences{\renewcommand{\thefootnote}{\fnsymbol{footnote}}\footnote[2]{The full CSV with all sentences and anchor annotations is provided in the supplementary material.}} generated with GPT-5.2~\cite{singh2025openaigpt5card} to minimize factual or logical priors (\eg, \textit{``A cautious driver measured the warm blanket in the garden, but nobody noticed at first''}). Each sentence contains a single \textbf{ground-truth anchor}. We apply our casing taxonomy by formatting the anchor in the target scheme and the remaining tokens in the complementary context scheme. Instead of measuring downstream accuracy, we prompt Qwen2.5-7B-Instruct~\cite{qwen2.5_7b_instruct} with a meta-cognitive query: \textit{``Identify the single word from the following sentence that received your primary focus.''} This system prompt enforces a single-word output of the identified word.

\paragraph{Evaluation Metric: Alignment Score.}
For each scheme, we compute an \textbf{Alignment Score}: a trial is scored $1$ if the self-reported word matches the ground-truth anchor, and $0$ otherwise. Scores are averaged over all 500 sentences. High alignment indicates that a casing pattern acts as the model's dominant processing cue rather than merely attracting numerical attention mass.

This analysis contextualizes the performance--attention divergence observed in \Cref{sec:results_accuracy}. In particular, we test whether \textbf{Target Alternating (TA)} schemes (high mass, low accuracy) induce coherent focus or disruptive saliency, contrasted with \textbf{Target Emphasis (TE)} schemes (moderate mass, high accuracy). The resulting alignment patterns clarify whether specific typographic signals function as effective guides or high-entropy distractors.

\paragraph{Results of the Active Interrogation.}
The interrogation results quantify how casing schemes affect explicit focus identification (\cf \cref{tab:qualitative_500_study}). Under natural casing, the anchor is reported as primary focus in $12.6\%$ of trials. All \textbf{Unified (U)} schemes remain within this range ($10.8$--$12.8\%$), indicating that global formatting changes do not induce localized focus shifts. \textbf{Target Title Case (TT)} schemes yield moderate alignment (TT2: $56.2\%$, TT3: $59.6\%$), demonstrating that conventional capitalization alone can function as a local attractor.

\begin{figure*}[!t]
    \centering
    \begin{subfigure}{0.19\textwidth} % width of the subfigure
        \centering
        \includegraphics[width=\textwidth]{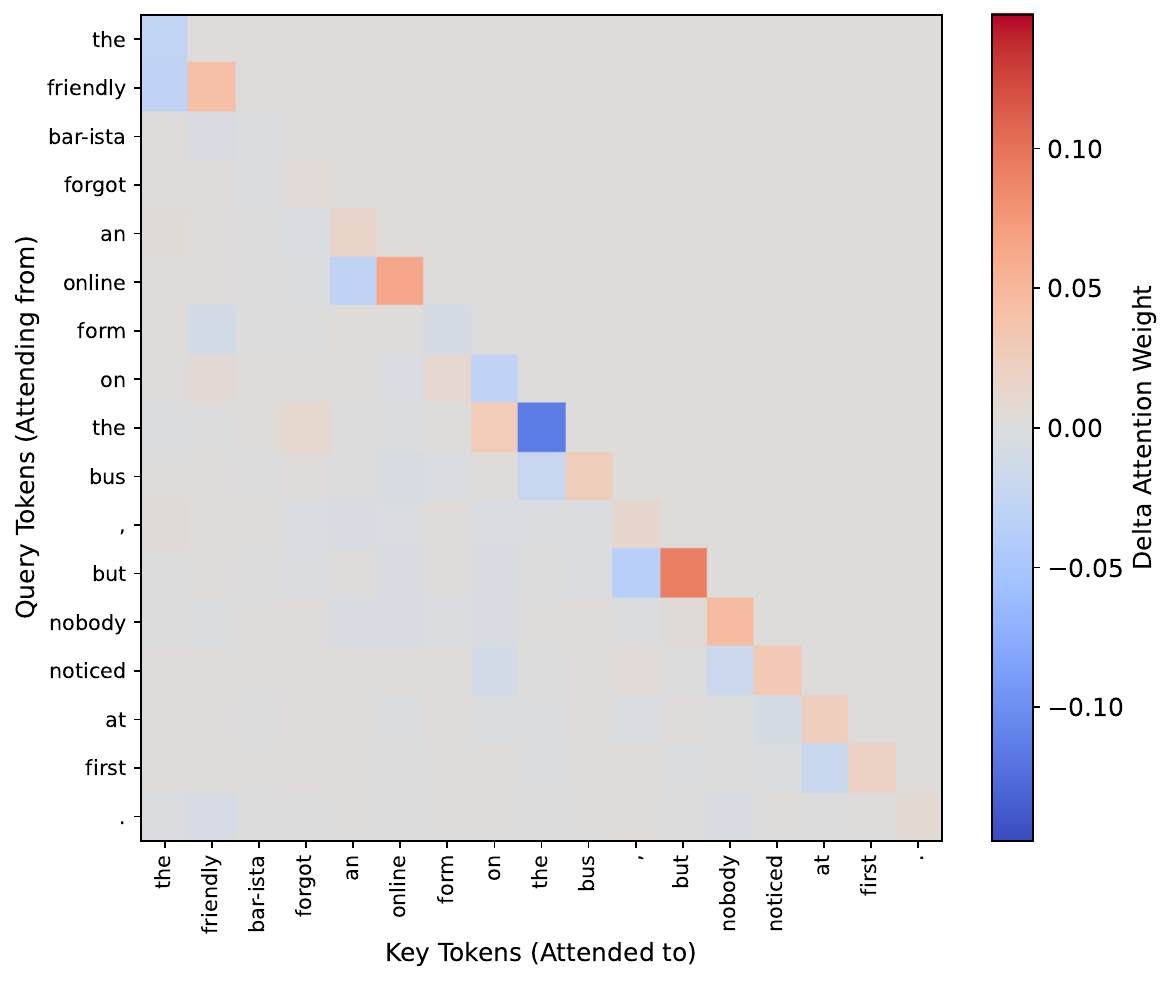}
    \end{subfigure}% 
    \hfill
    \begin{subfigure}{0.19\textwidth} % width of the subfigure
        \centering
        \includegraphics[width=\textwidth]{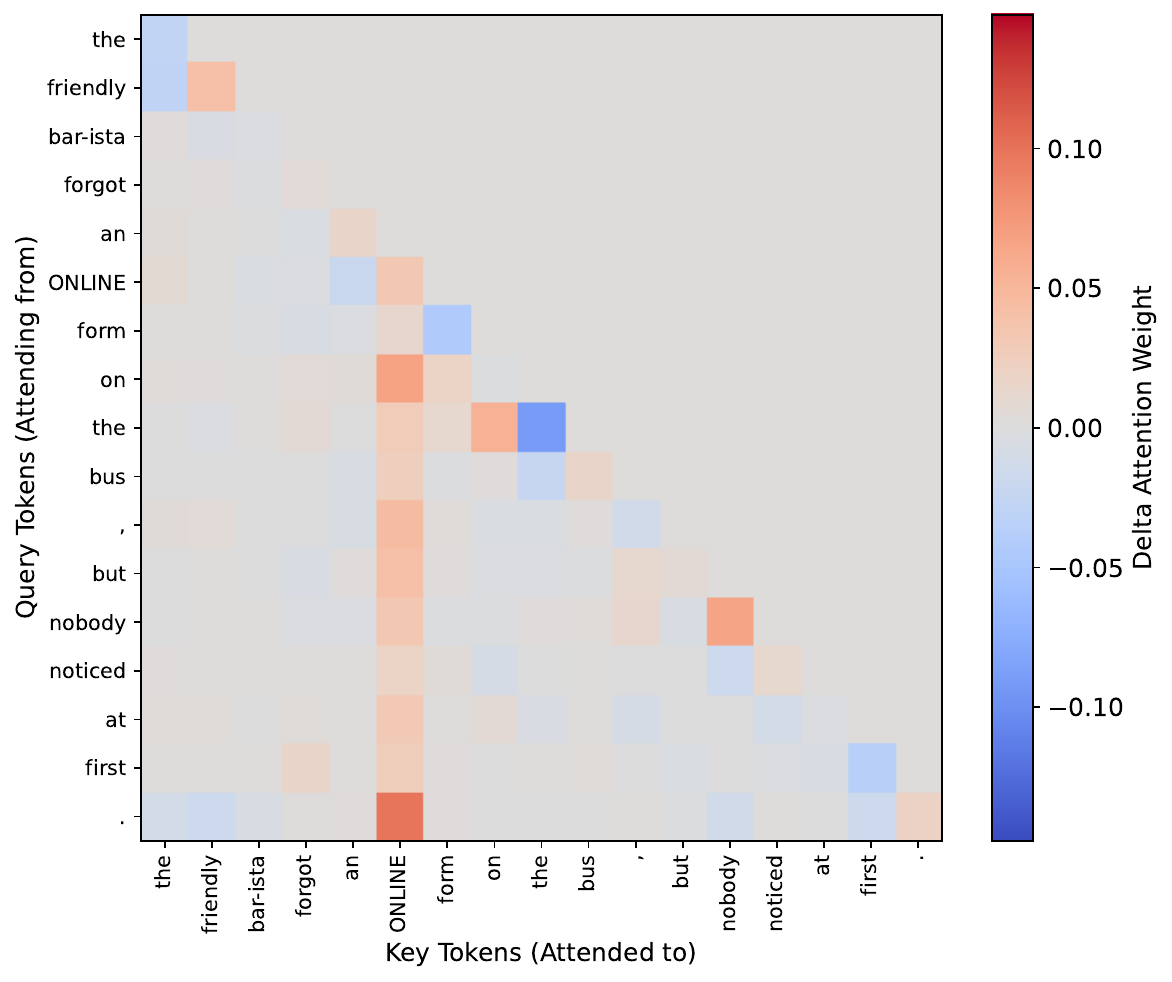}
    \end{subfigure}% 
    \hfill
    \begin{subfigure}{0.19\textwidth} % width of the subfigure
        \centering
        \includegraphics[width=\textwidth]{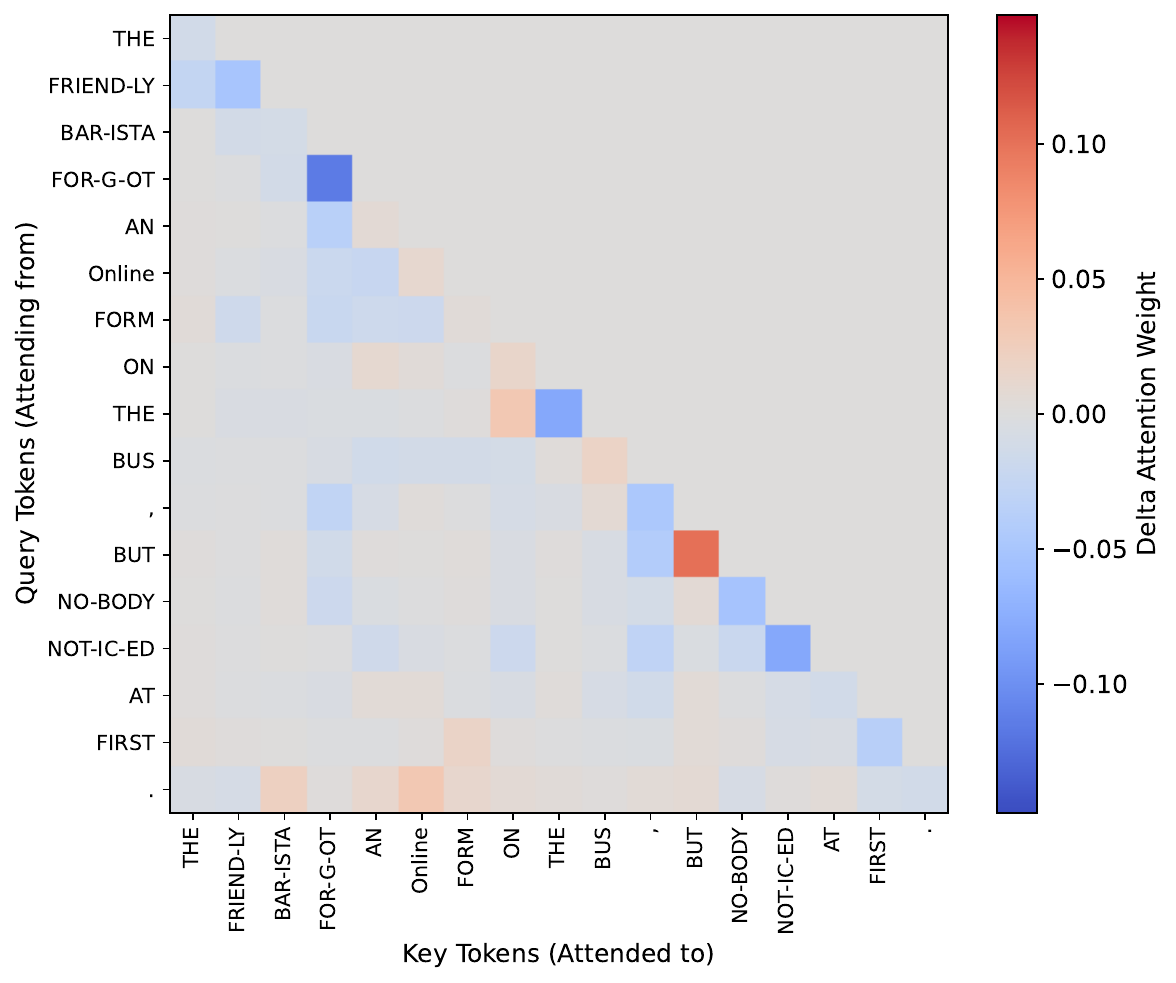}
    \end{subfigure}% 
    \hfill
    \begin{subfigure}{0.19\textwidth} % width of the subfigure
        \centering
        \includegraphics[width=\textwidth]{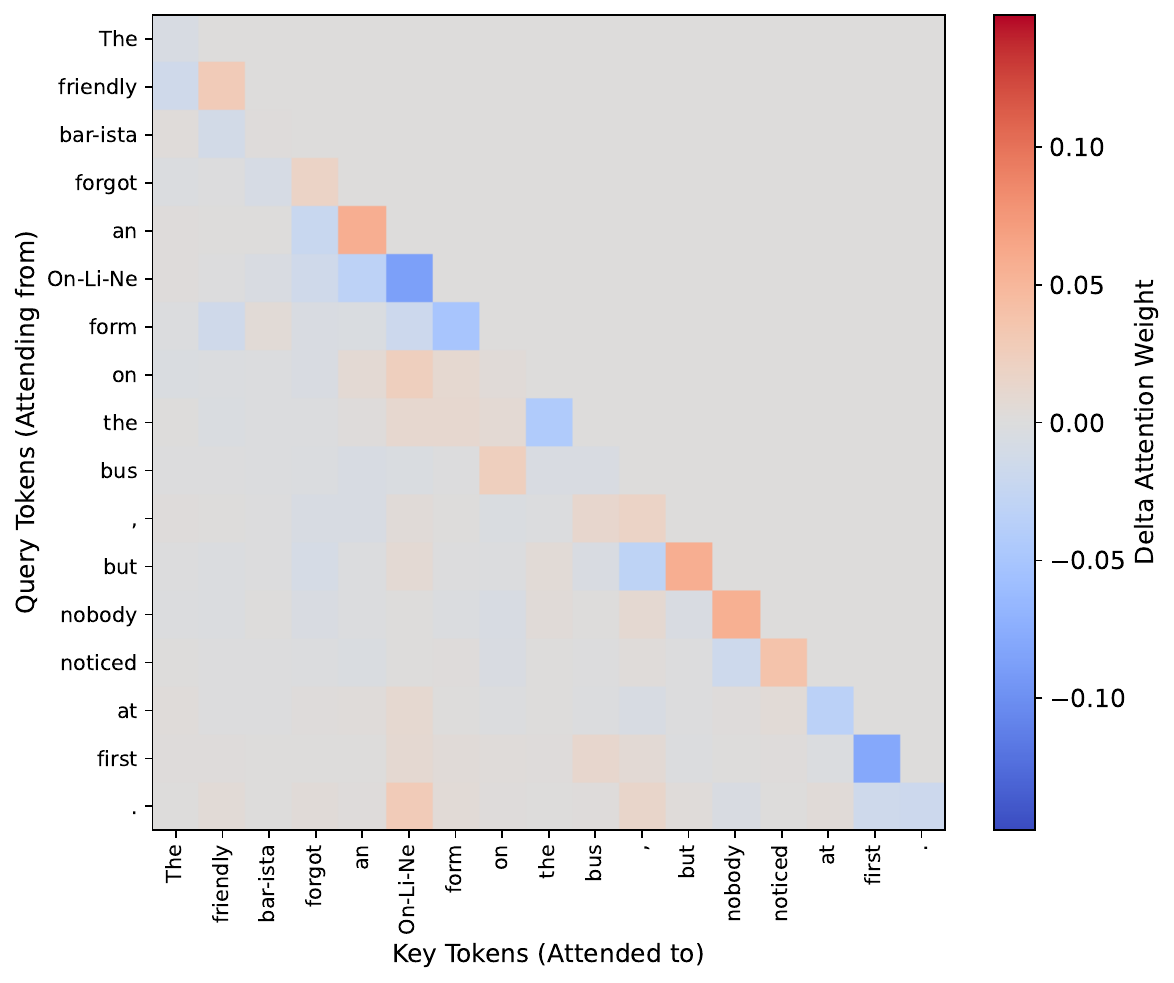}
    \end{subfigure}% 
    \hfill
    \begin{subfigure}{0.19\textwidth} % width of the subfigure
        \centering
        \includegraphics[width=\textwidth]{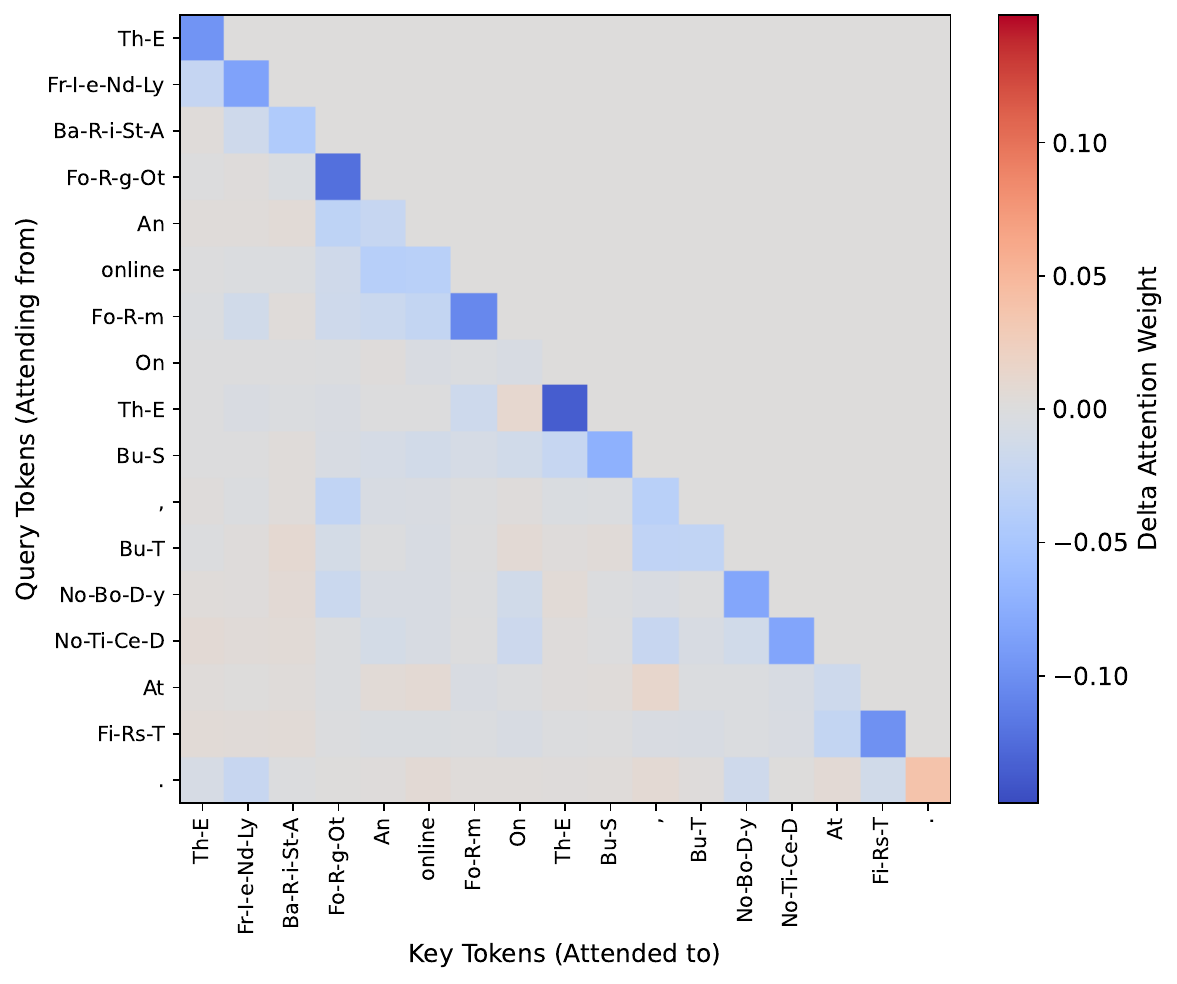}
    \end{subfigure}% 
    \caption{\textbf{Relative attention shifts for the 500-sentence dataset.} The target word in this example is \textbf{\textit{online}}. Increases in relative attention mass compared to the natural cased baseline sentence are shown in red, while decreases are shown in blue. The panels, from left to right, correspond to schemes U2, TE1, TT1, TA3, and ADE2.}
    \label{fig:attention_maps_355_inpaper}
\end{figure*}

\textbf{Target Emphasis (TE)} schemes produce the strongest and most consistent alignment (TE1: $93.0\%$, TE3: $92.6\%$). The corresponding delta attention maps in \Cref{fig:attention_maps_355_inpaper} show a concentrated vertical band of increased attention on the target span (red), confirming that uppercase emphasis induces both numerical attention concentration and explicit focus selection. Thus, for TE patterns, attention mass and behavioral identification are tightly coupled.

\textbf{Target Alternating (TA)} schemes also achieve high alignment (TA2: $76.8\%$, TA3: $81.8\%$), reflected by localized attention increases in \Cref{fig:attention_maps_355_inpaper}. While consistent with our broader benchmark findings, the attention shift in this study is less pronounced than in long-form datasets. This attenuation is likely a function of \textit{attentional saturation} in shorter sequences; where the baseline focus is already concentrated among few tokens, the relative delta induced by casing is mathematically constrained compared to long-form contexts. This supports a critical distinction in the performance--attention divergence: while high Alignment Scores prove the model is aware of the alternating-case span, the high-entropy orthography disrupts the semantic coherence of the resulting embeddings, decoupling attentional priority from downstream utility.

Finally, \textbf{Adversarial De-Emphasis (ADE)} schemes substantially suppress anchor selection (\eg, ADE2: $2.6\%$), confirming causal directionality of the casing effect; the model consistently reports lower focus on the anchor word when it is de-emphasized against more salient contexts. Further qualitative results can be seen in \Cref{fig:attention_maps_355,fig:attention_maps_384,fig:attention_maps_119}.

\begin{table}[!t]
    \centering
    \caption{\textbf{Experimental results on self-reported attentional focus.} This study bridges the gap between quantitative attention mass and true attention steering.}
    \begin{tabular}{lc p{0.5cm} lc}
        \toprule
        \textbf{Scheme} & \textbf{Alignment Score} $\uparrow$ && \textbf{Scheme} & \textbf{Alignment Score} $\uparrow$ \\
        \midrule
        \addlinespace[0.5em]
        \multicolumn{2}{l}{\textit{Global Uniformity (U)}} && \multicolumn{2}{l}{\textit{Target Emphasis (TE)}} \\
        \midrule
        U1 & 10.8\%               && TE1 & 93.0\% \\
        U2 & 12.8\%               && TE2 & 35.0\% \\
        U3 & 11.4\%               && TE3 & 92.6\% \\

        \addlinespace[1em]
        \multicolumn{2}{l}{\textit{Target Title Case (TT)}} && \multicolumn{2}{l}{\textit{Target Alternating Case (TA)}} \\
        \midrule
        TT1 & 19.4\%               && TA1 & 24.0\% \\
        TT2 & 56.2\%               && TA2 & 76.8\% \\
        TT3 & 59.6\%               && TA3 & 81.8\% \\

        \addlinespace[1em]
        \multicolumn{5}{l}{\textit{Adversarial De-Emphasis (ADE)}} \\
        \midrule
        ADE1 & 9.0\%     && ADE3 & 12.6\% \\
        ADE2 & 2.6\% \\
        \bottomrule
    \end{tabular}
    \label{tab:qualitative_500_study}
\end{table}

\subsection{Qualitative Observations from Active Interrogation}
\label{sec:qual_obs_benchmarks}

\begin{figure*}[!ht]
    \centering
    \begin{subfigure}{0.26\textwidth} % width of the subfigure
        \centering
        \includegraphics[width=\textwidth]{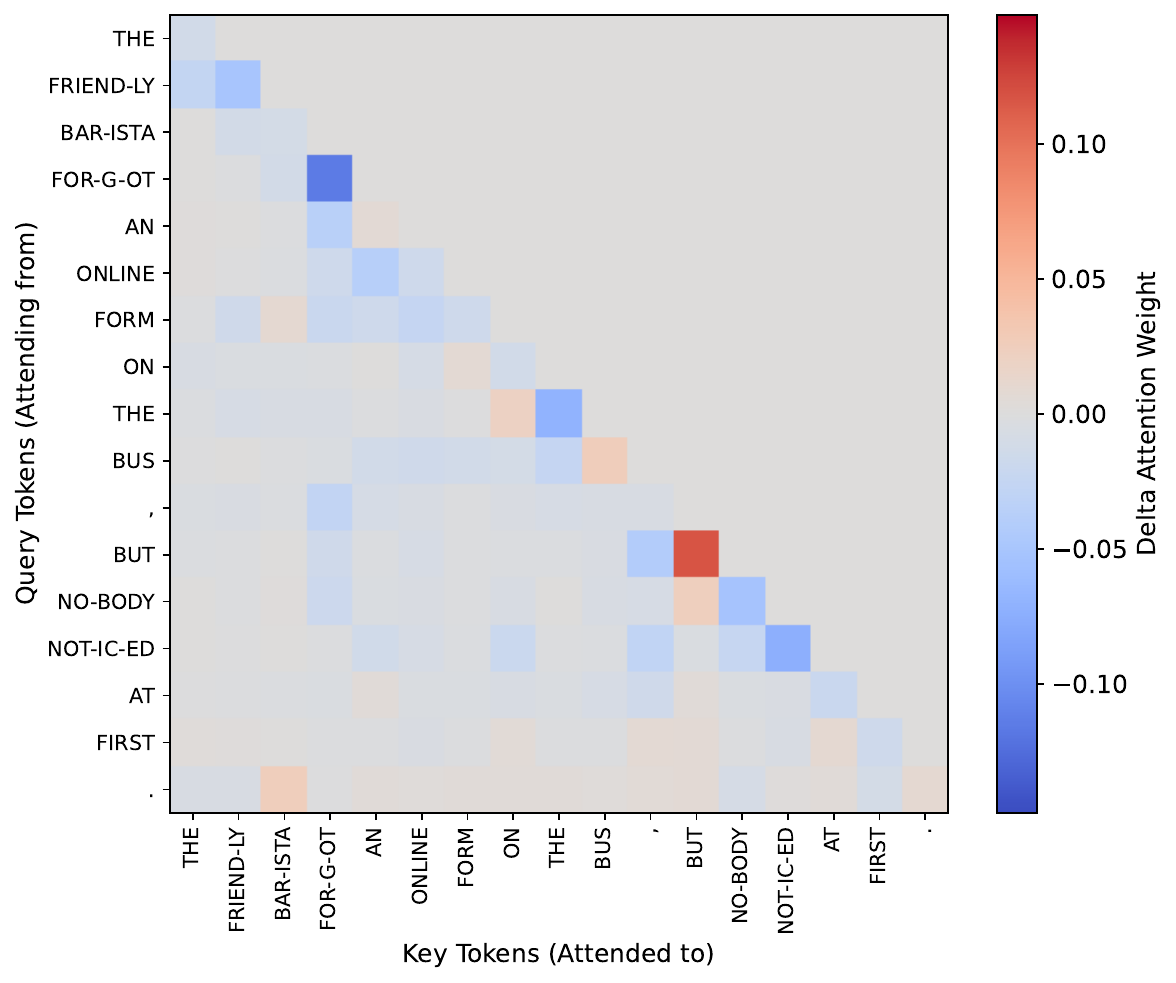}
    \end{subfigure}% 
    \hspace{2em}
    \begin{subfigure}{0.26\textwidth} % width of the subfigure
        \centering
        \includegraphics[width=\textwidth]{Figures/attention_shifts_sentence_355_pattern_02.pdf}
    \end{subfigure}% 
    \hspace{2em}
    \begin{subfigure}{0.26\textwidth} % width of the subfigure
        \centering
        \includegraphics[width=\textwidth]{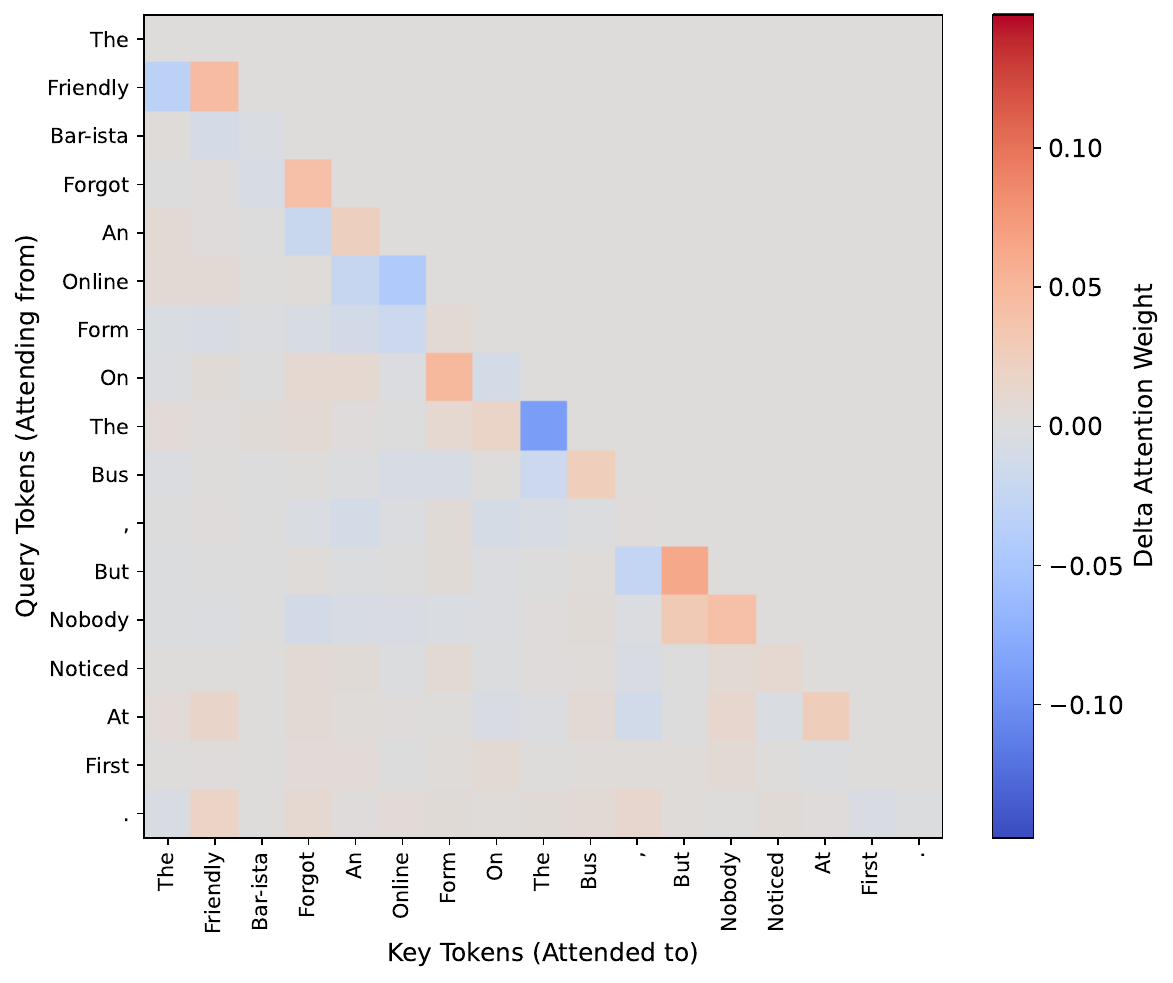}
    \end{subfigure}% 
    \vspace{0.5em}
    \begin{subfigure}{0.26\textwidth} % width of the subfigure
        \centering
        \includegraphics[width=\textwidth]{Figures/attention_shifts_sentence_355_pattern_04.pdf}
    \end{subfigure}% 
    \hspace{2em}
    \begin{subfigure}{0.26\textwidth} % width of the subfigure
        \centering
        \includegraphics[width=\textwidth]{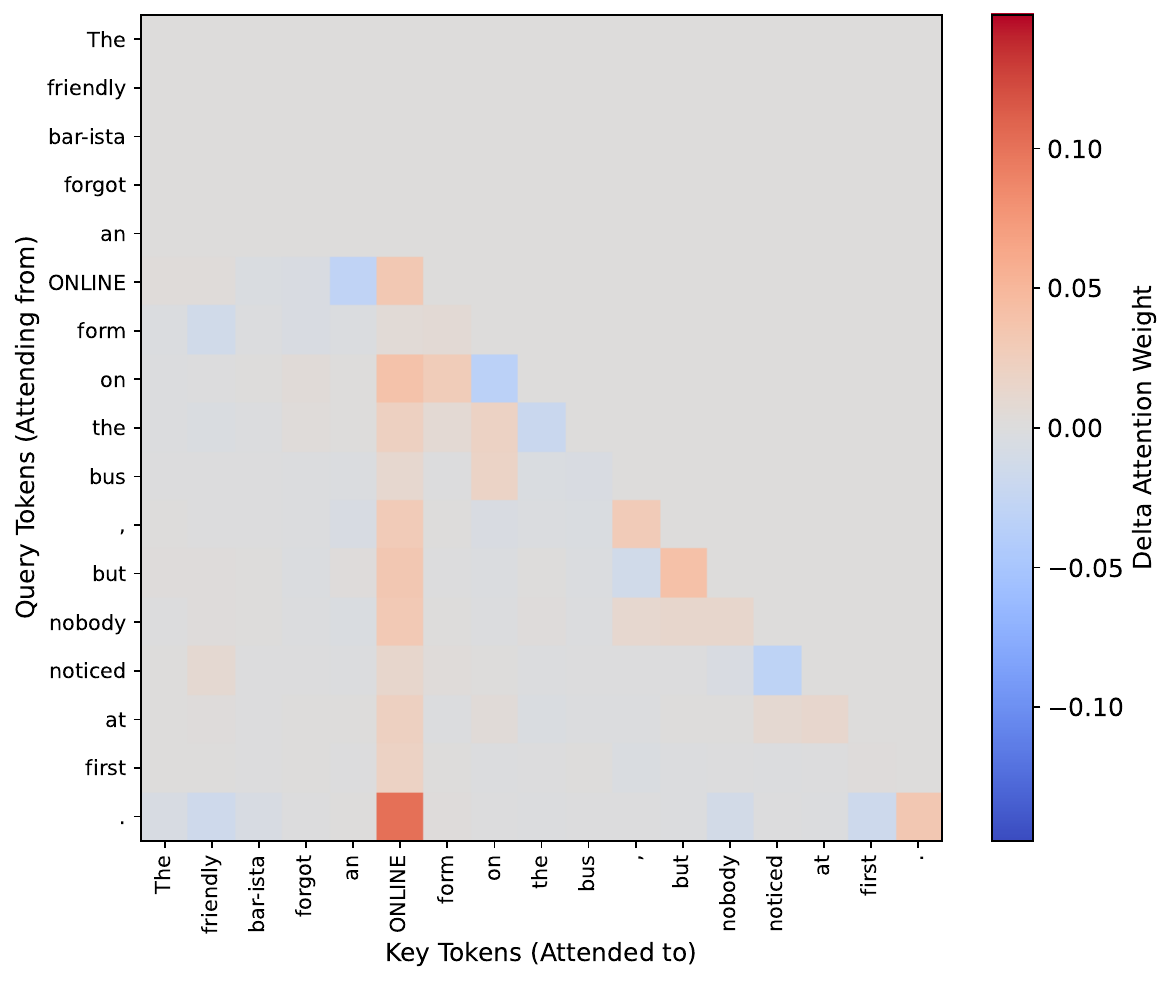}
    \end{subfigure}% 
    \hspace{2em}
    \begin{subfigure}{0.26\textwidth} % width of the subfigure
        \centering
        \includegraphics[width=\textwidth]{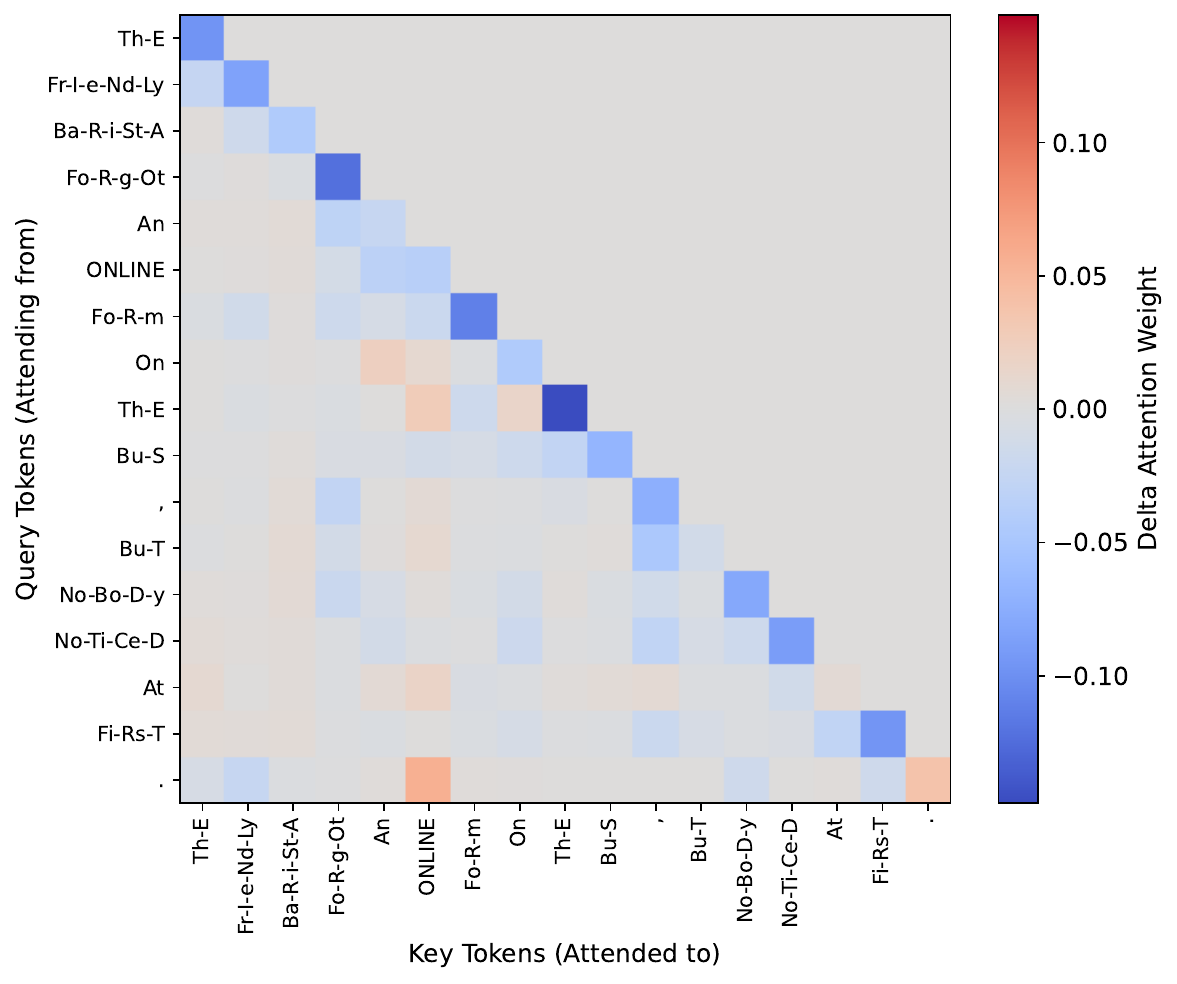}
    \end{subfigure}% 
    \vspace{0.5em}
    \begin{subfigure}{0.26\textwidth} % width of the subfigure
        \centering
        \includegraphics[width=\textwidth]{Figures/attention_shifts_sentence_355_pattern_08.pdf}
    \end{subfigure}% 
    \hspace{2em}
    \begin{subfigure}{0.26\textwidth} % width of the subfigure
        \centering
        \includegraphics[width=\textwidth]{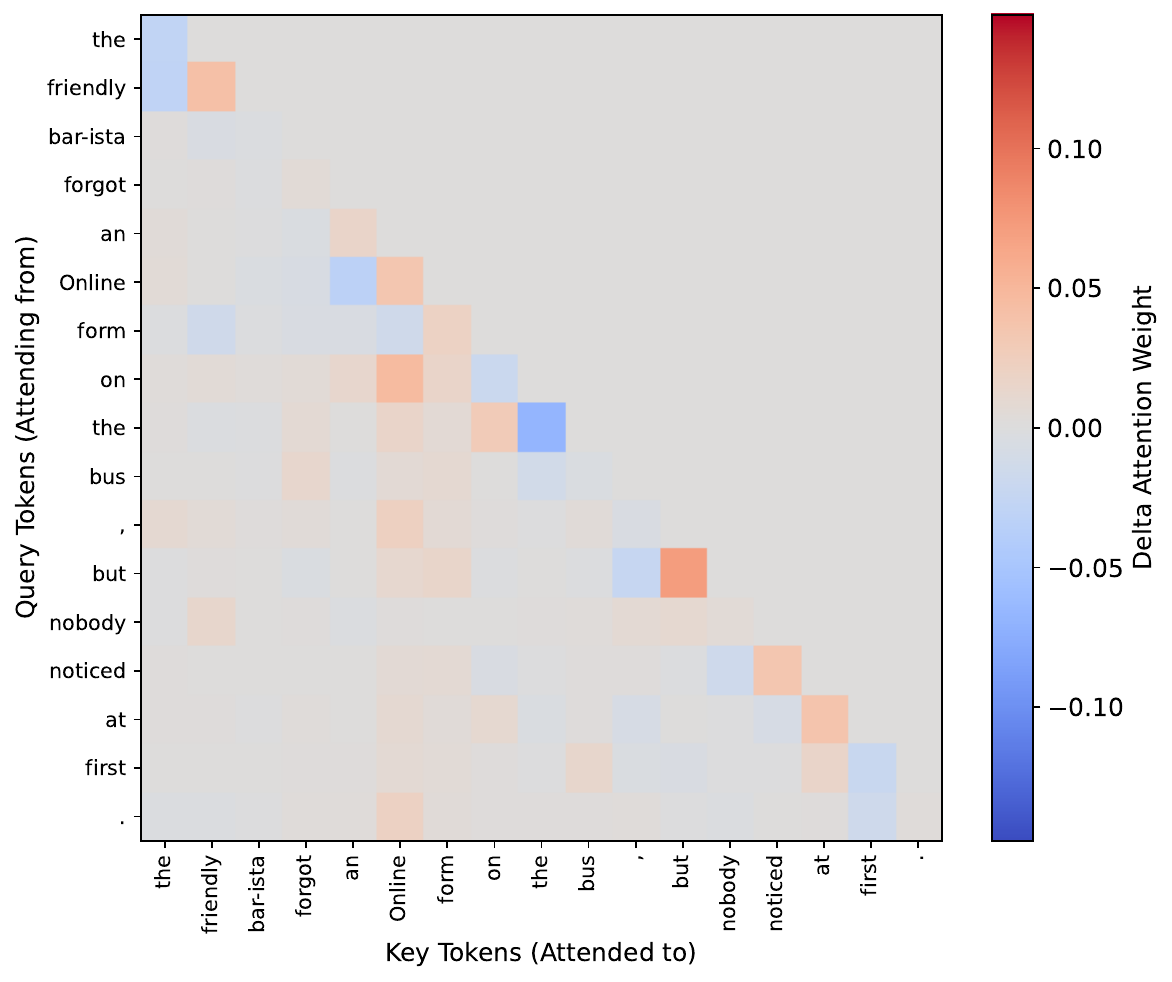}
    \end{subfigure}% 
    \hspace{2em}
    \begin{subfigure}{0.26\textwidth} % width of the subfigure
        \centering
        \includegraphics[width=\textwidth]{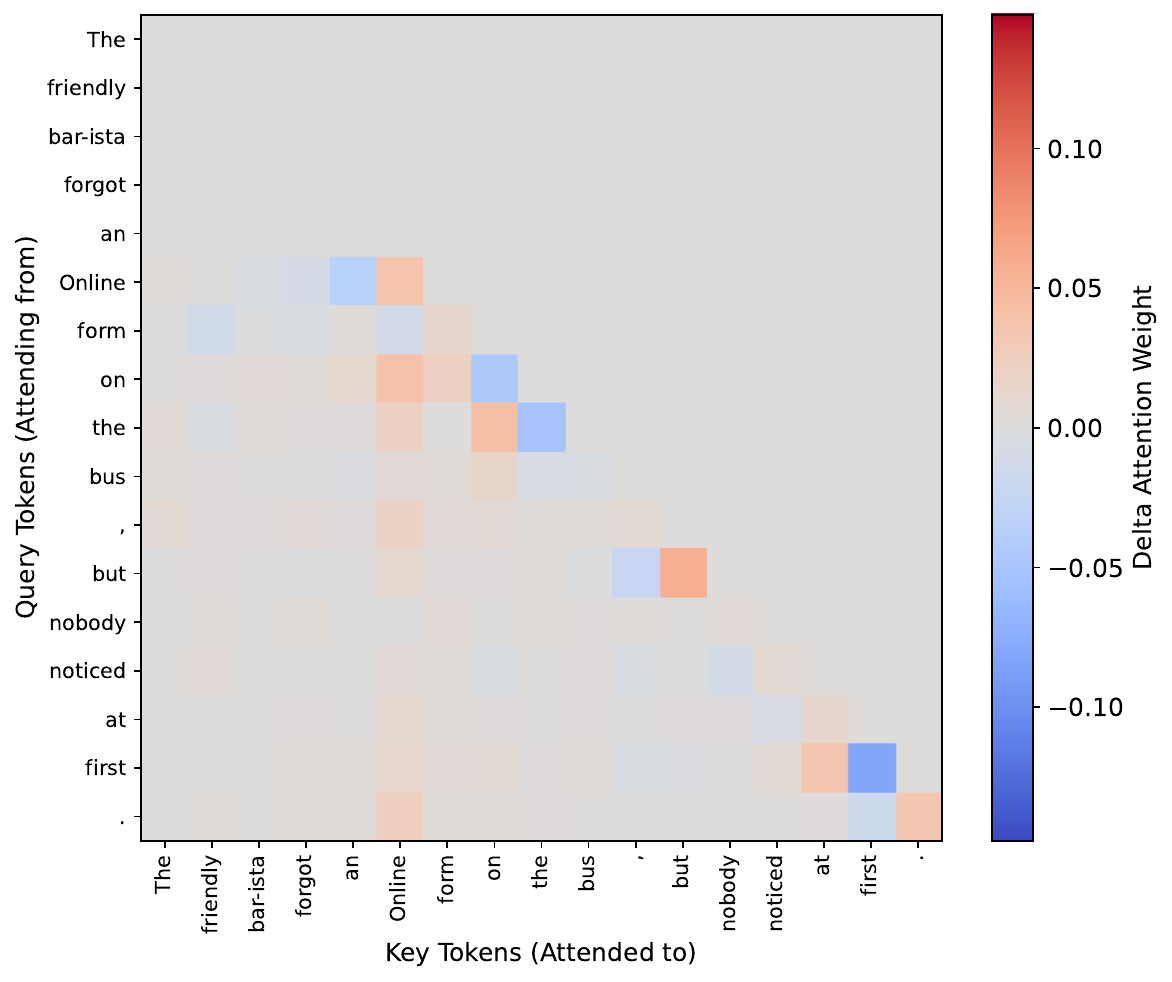}
    \end{subfigure}% 
    \vspace{0.5em}
    \begin{subfigure}{0.26\textwidth} % width of the subfigure
        \centering
        \includegraphics[width=\textwidth]{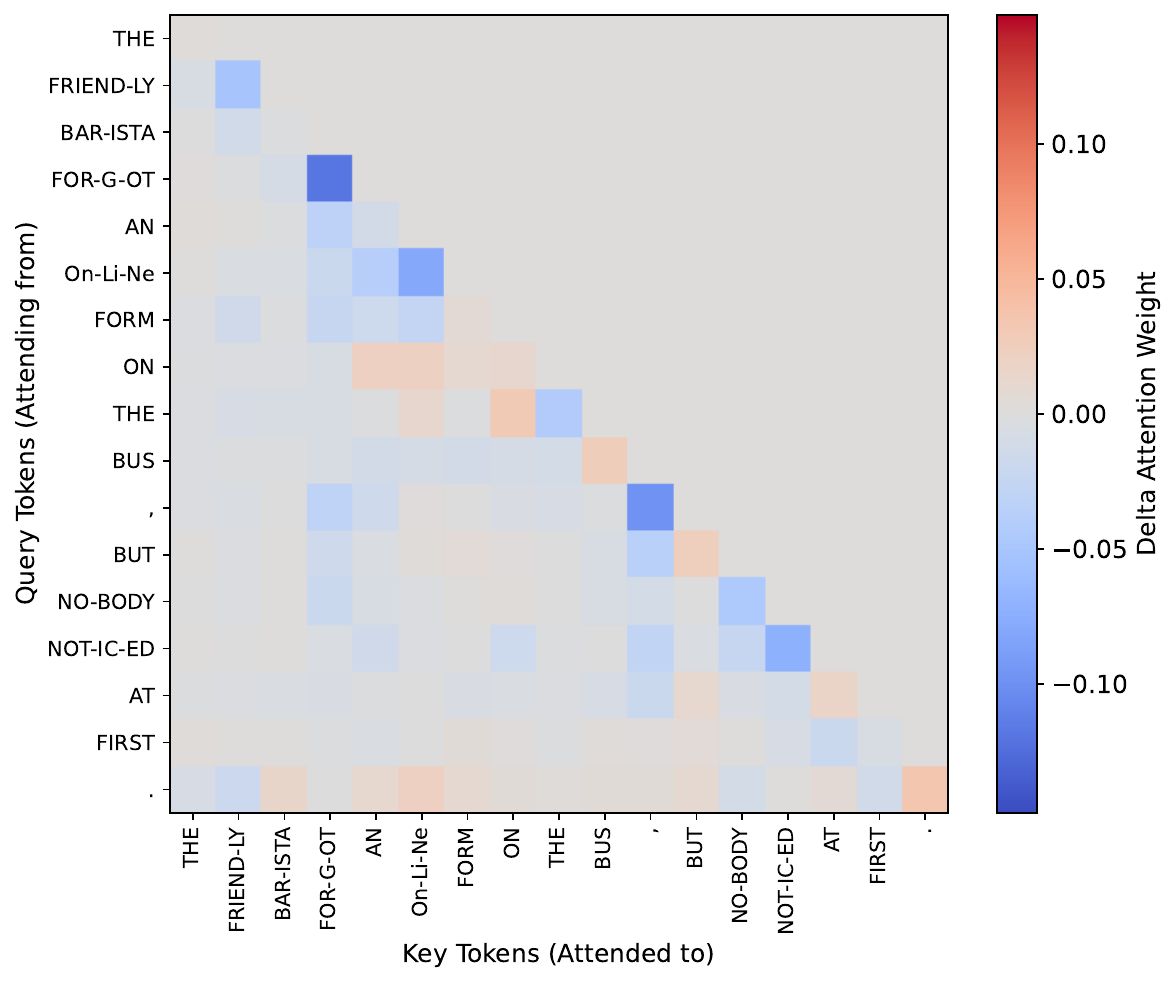}
    \end{subfigure}% 
    \hspace{2em}
    \begin{subfigure}{0.26\textwidth} % width of the subfigure
        \centering
        \includegraphics[width=\textwidth]{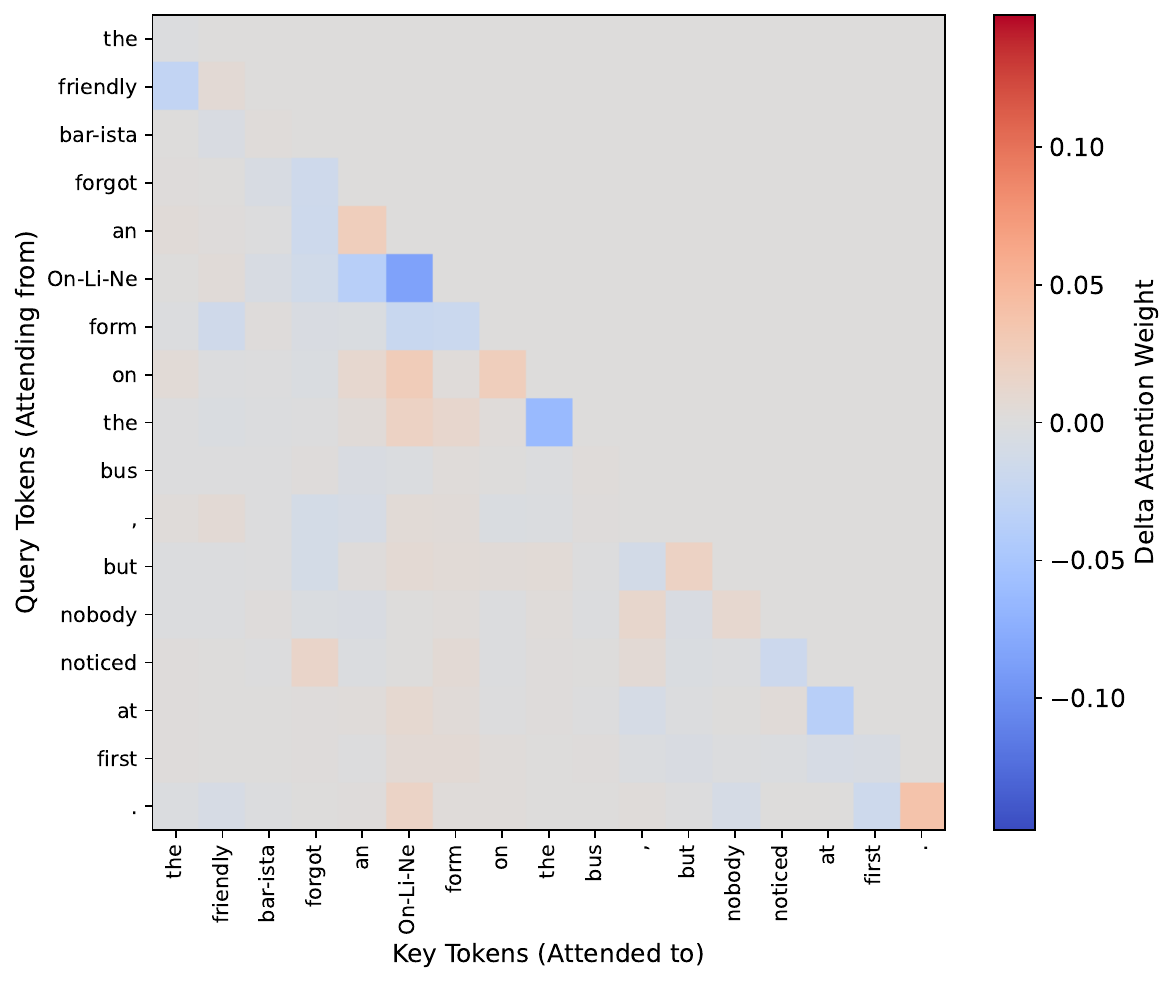}
    \end{subfigure}% 
    \hspace{2em}
    \begin{subfigure}{0.26\textwidth} % width of the subfigure
        \centering
        \includegraphics[width=\textwidth]{Figures/attention_shifts_sentence_355_pattern_11.pdf}
    \end{subfigure}% 
    \vspace{0.5em}
    \begin{subfigure}{0.26\textwidth} % width of the subfigure
        \centering
        \includegraphics[width=\textwidth]{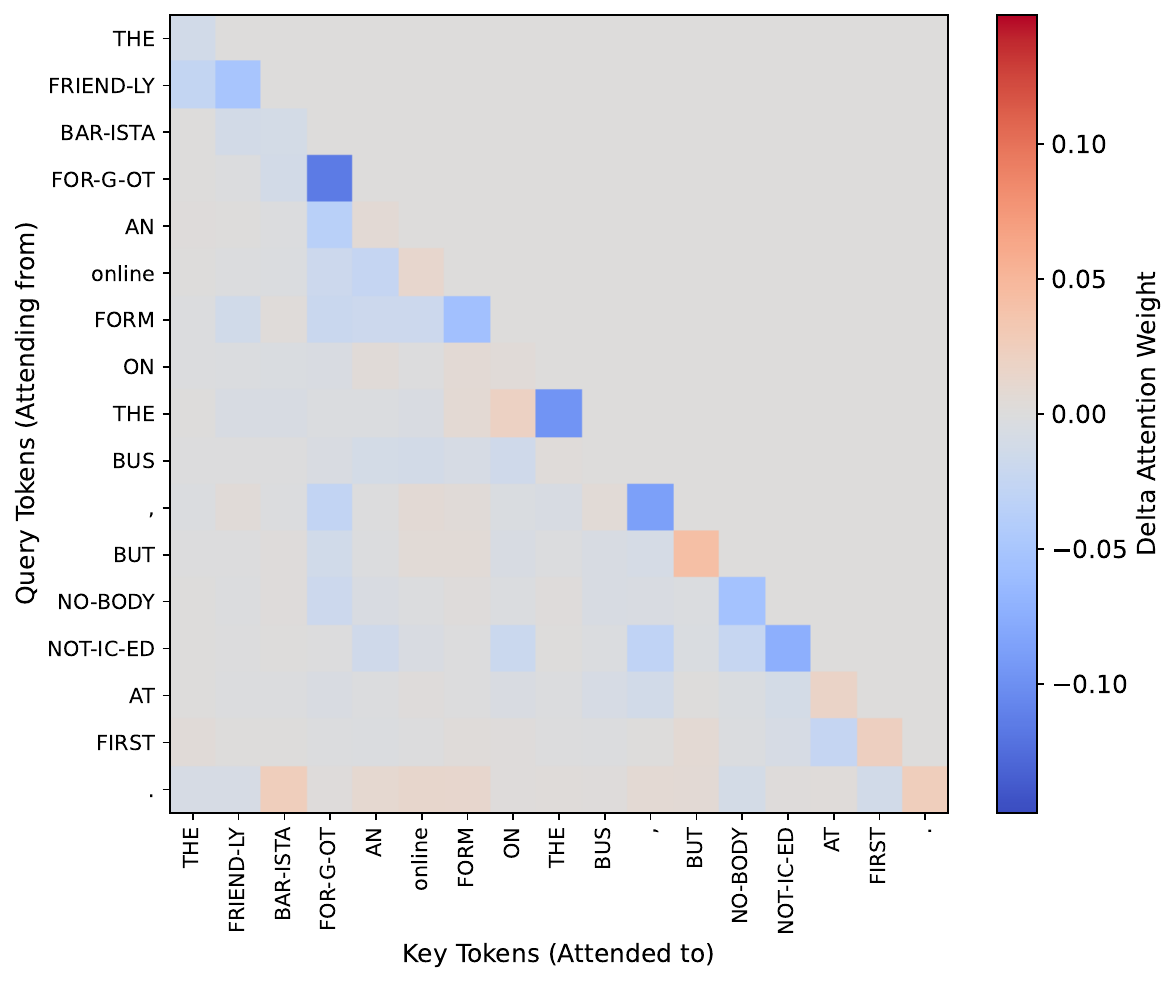}
    \end{subfigure}% 
    \hspace{2em}
    \begin{subfigure}{0.26\textwidth} % width of the subfigure
        \centering
        \includegraphics[width=\textwidth]{Figures/attention_shifts_sentence_355_pattern_14.pdf}
    \end{subfigure}% 
    \hspace{2em}
    \begin{subfigure}{0.26\textwidth} % width of the subfigure
        \centering
        \includegraphics[width=\textwidth]{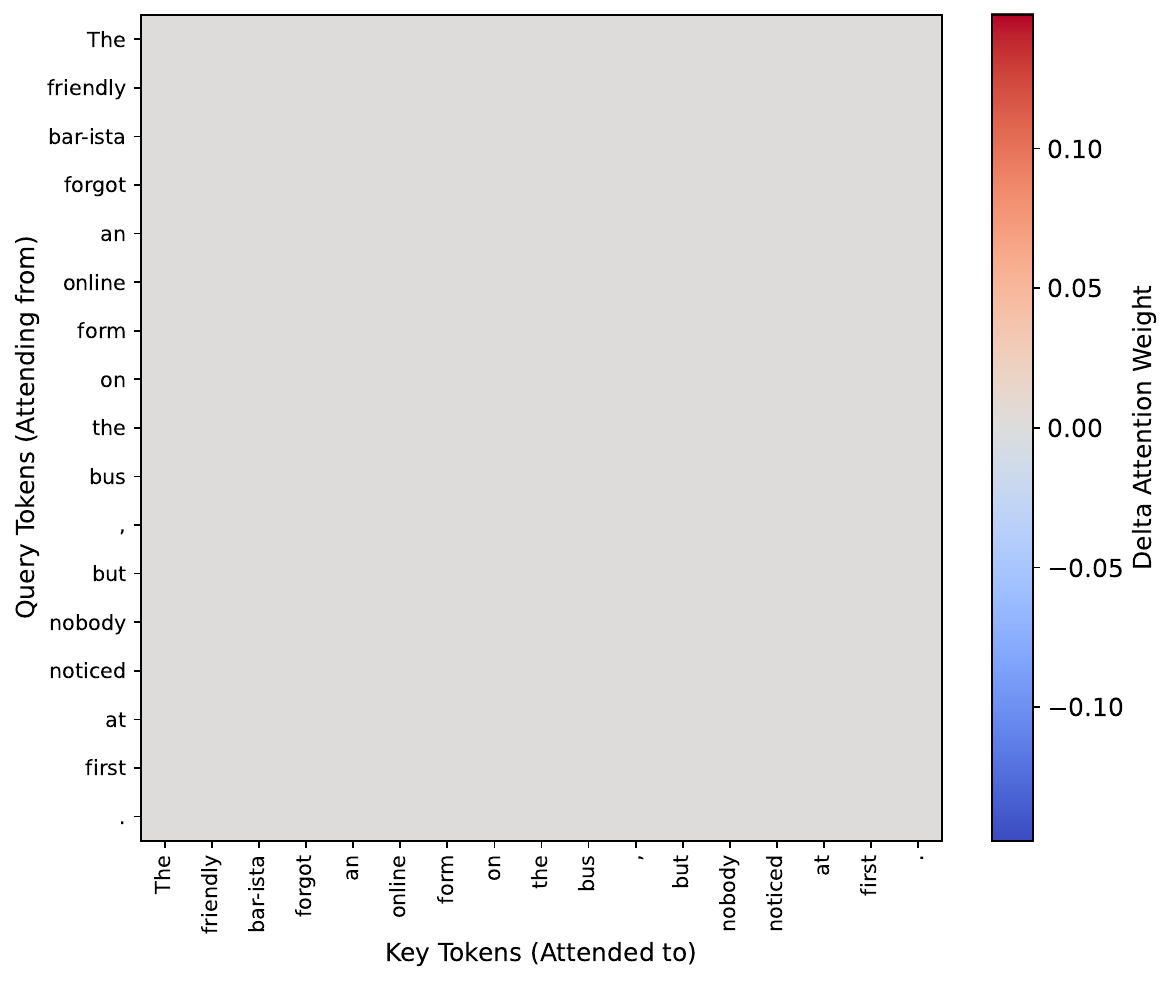}
    \end{subfigure}% 
    \caption{\textbf{Relative attention shifts for the 500-sentence self-reported dataset.} The target word in this example is \textbf{\textit{online}}. Increases in relative attention mass compared to the natural cased baseline sentence are shown in red, while decreases are shown in blue. The panels, from top-left to bottom-right, correspond to schemes U1--3, TE1--3, TT1--3, TA1--3, and ADE1--3.}
    \label{fig:attention_maps_355}
\end{figure*}

\begin{figure*}[!ht]
    \centering
    \begin{subfigure}{0.26\textwidth} % width of the subfigure
        \centering
        \includegraphics[width=\textwidth]{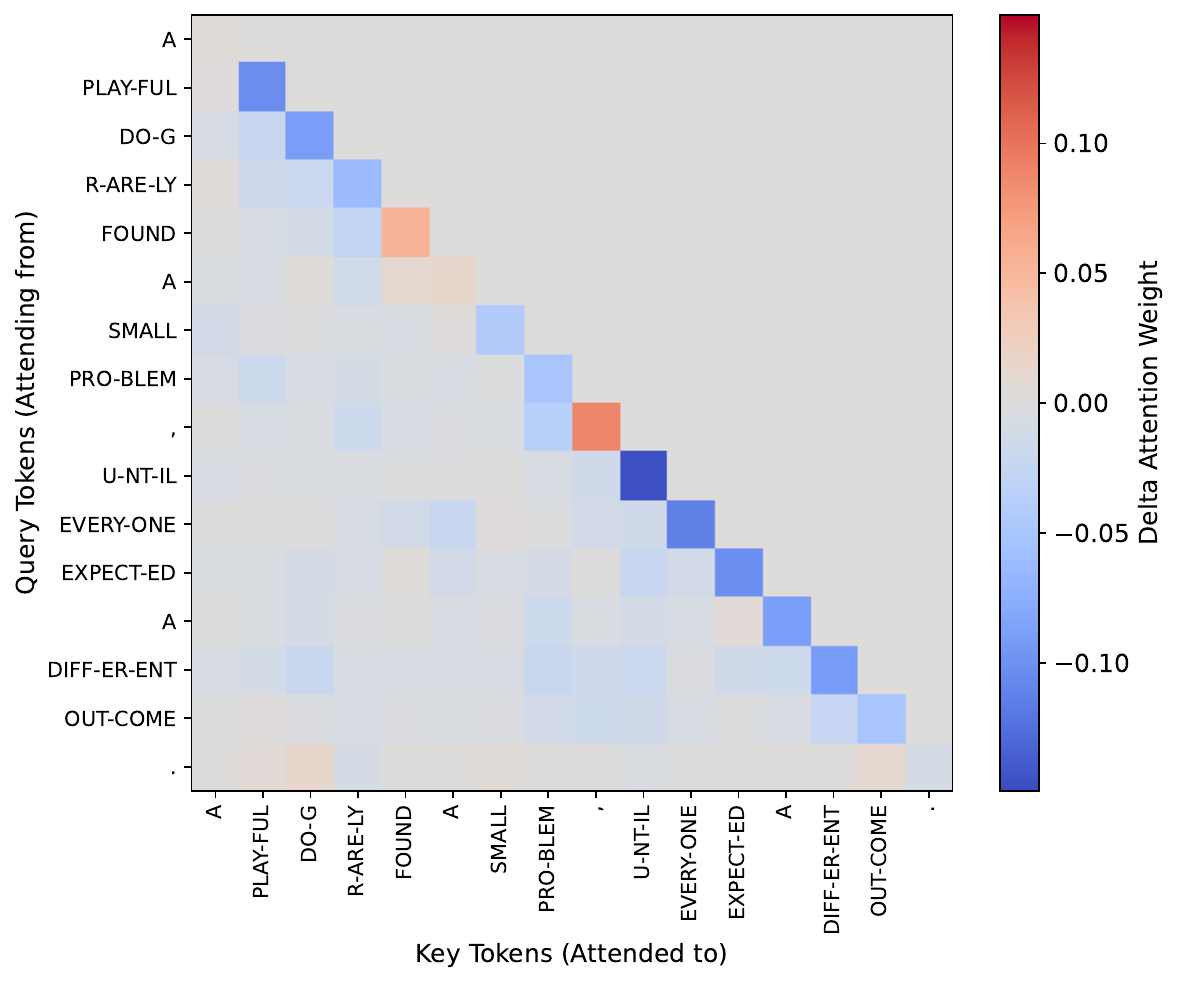}
    \end{subfigure}% 
    \hspace{2em}
    \begin{subfigure}{0.26\textwidth} % width of the subfigure
        \centering
        \includegraphics[width=\textwidth]{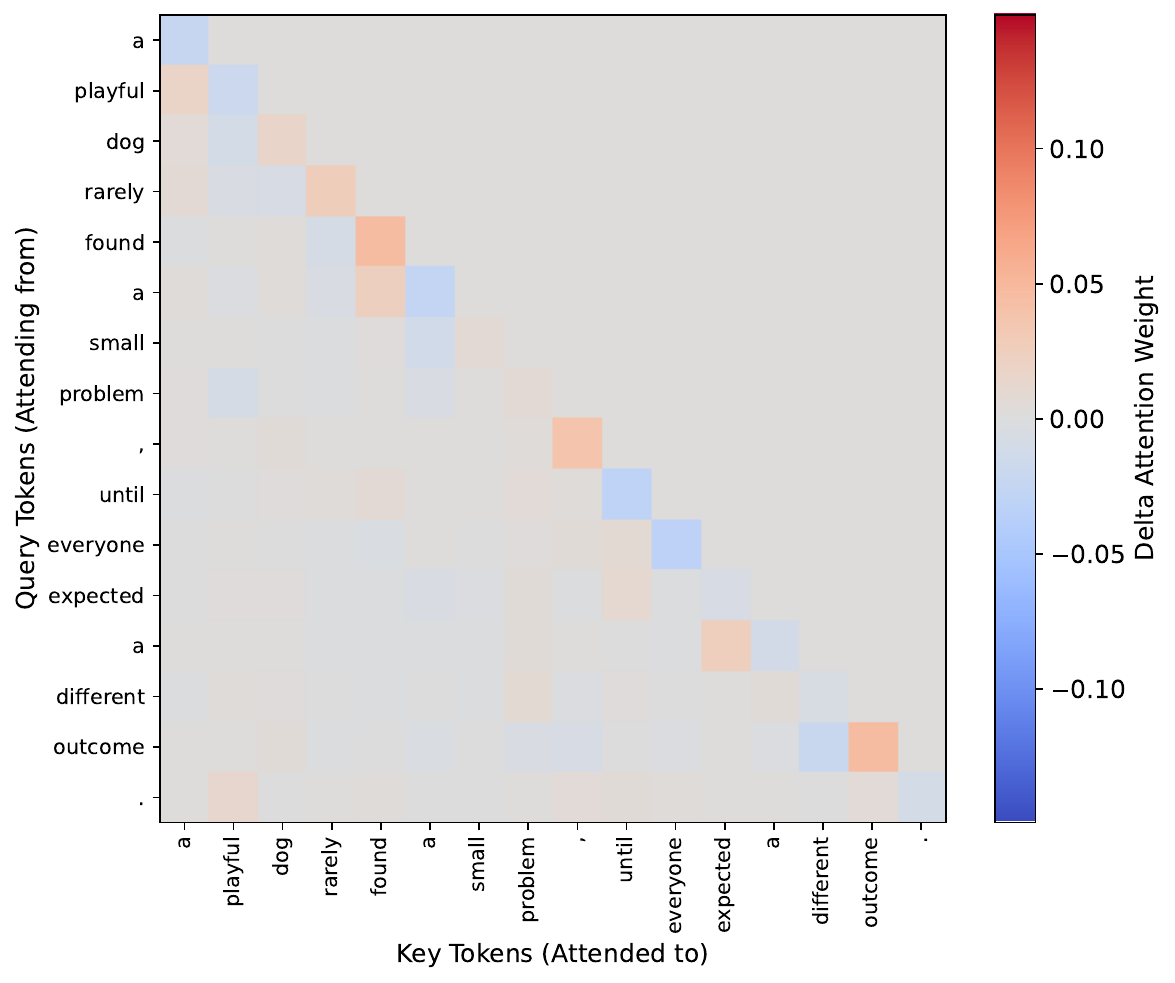}
    \end{subfigure}% 
    \hspace{2em}
    \begin{subfigure}{0.26\textwidth} % width of the subfigure
        \centering
        \includegraphics[width=\textwidth]{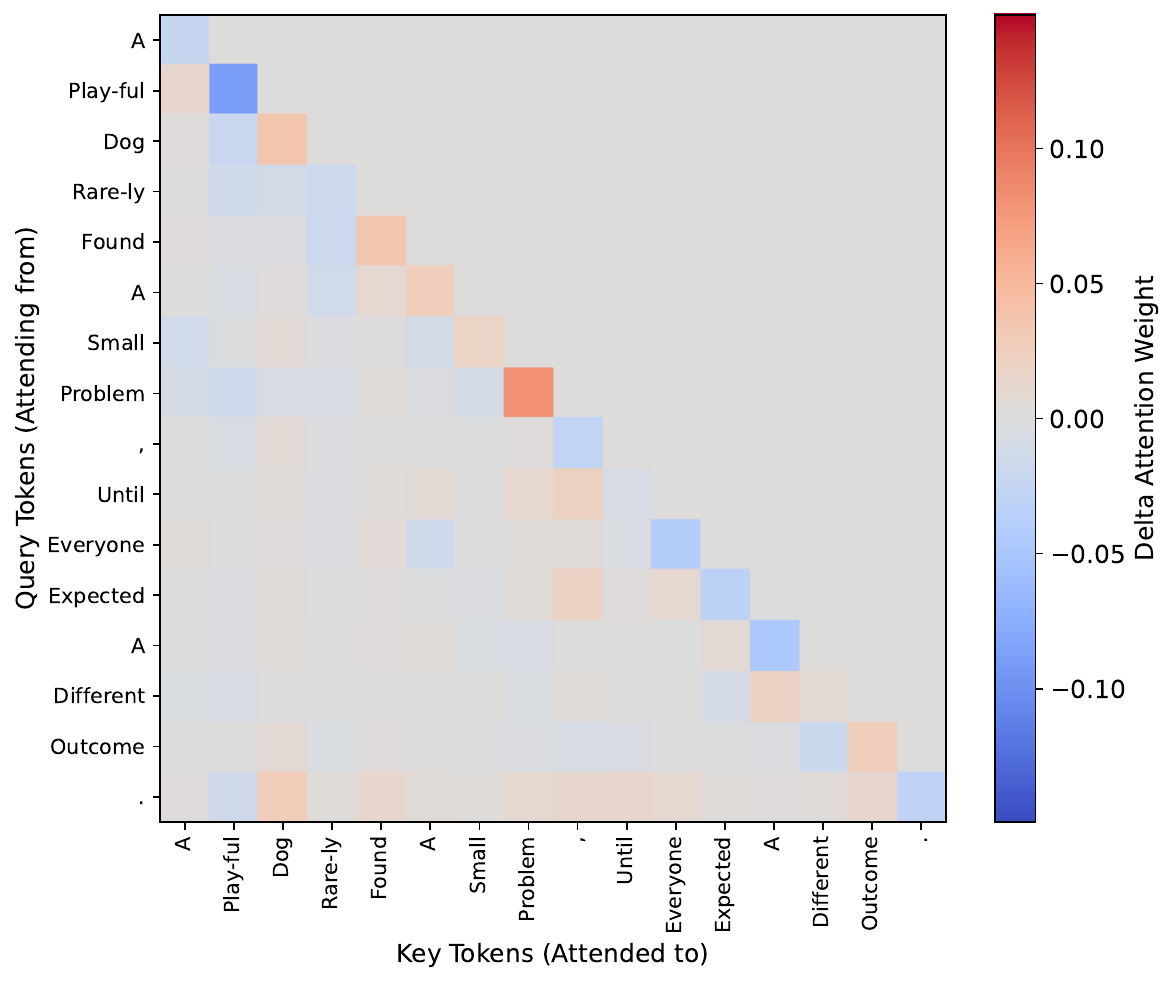}
    \end{subfigure}% 
    \vspace{0.5em}
    \begin{subfigure}{0.26\textwidth} % width of the subfigure
        \centering
        \includegraphics[width=\textwidth]{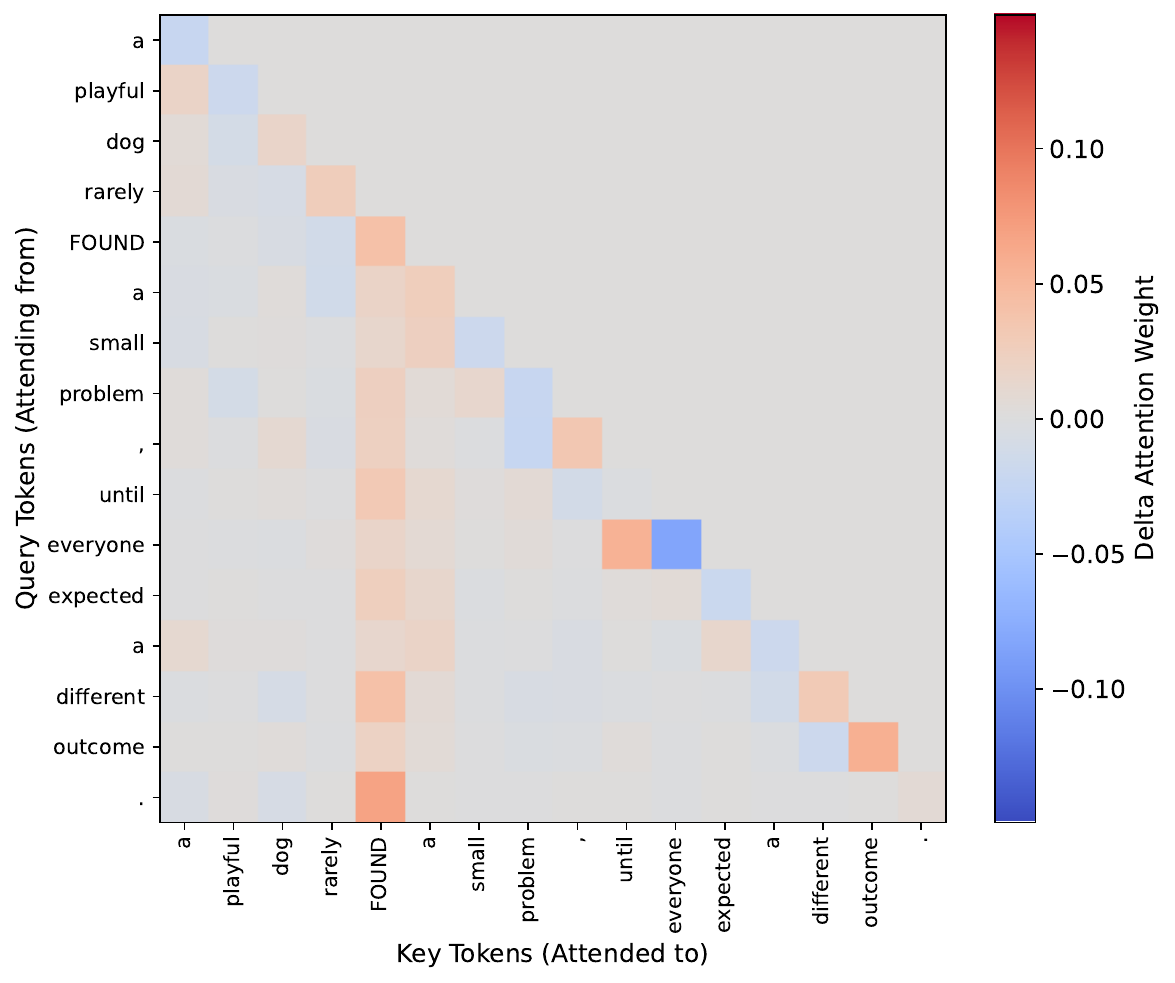}
    \end{subfigure}% 
    \hspace{2em}
    \begin{subfigure}{0.26\textwidth} % width of the subfigure
        \centering
        \includegraphics[width=\textwidth]{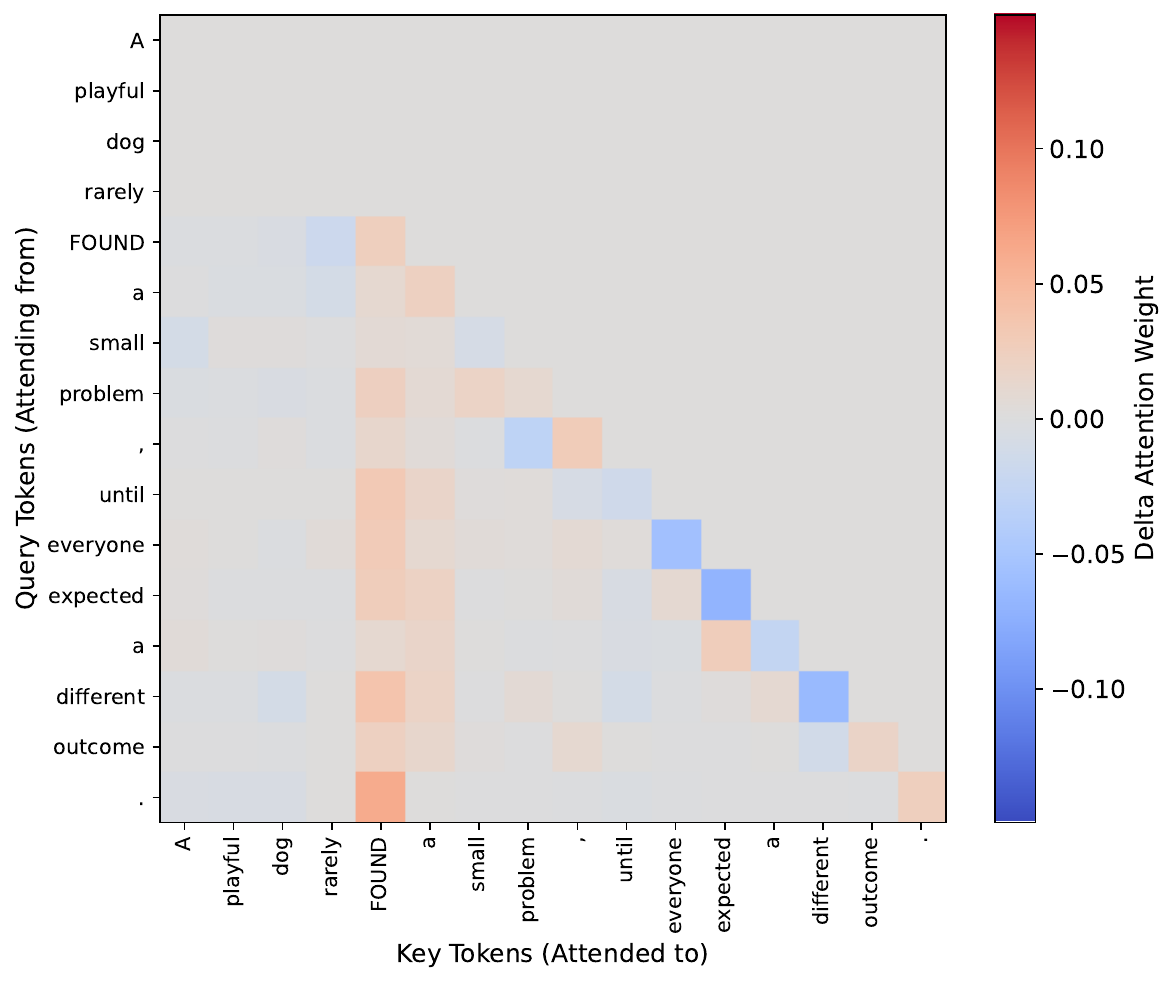}
    \end{subfigure}% 
    \hspace{2em}
    \begin{subfigure}{0.26\textwidth} % width of the subfigure
        \centering
        \includegraphics[width=\textwidth]{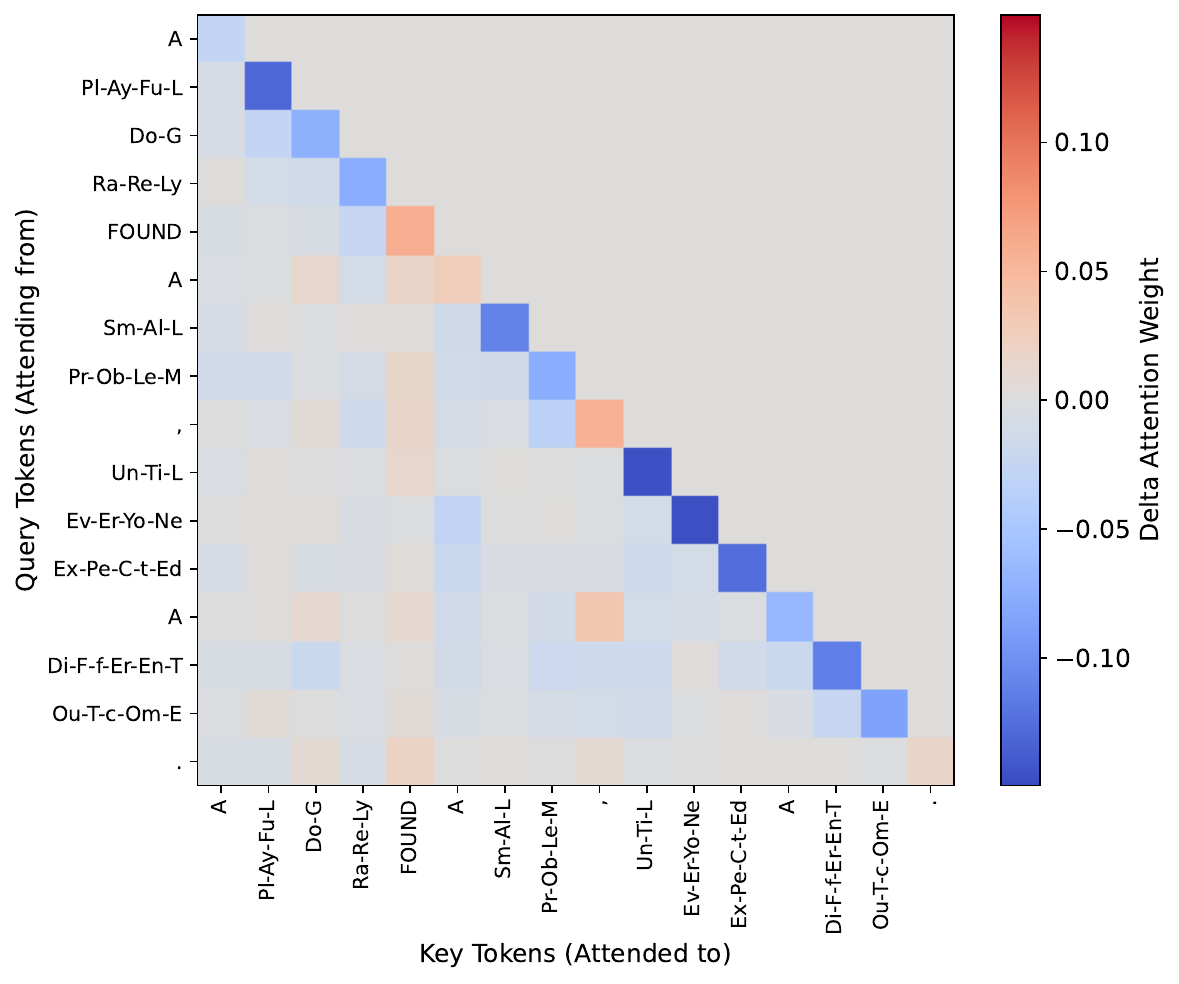}
    \end{subfigure}% 
    \vspace{0.5em}
    \begin{subfigure}{0.26\textwidth} % width of the subfigure
        \centering
        \includegraphics[width=\textwidth]{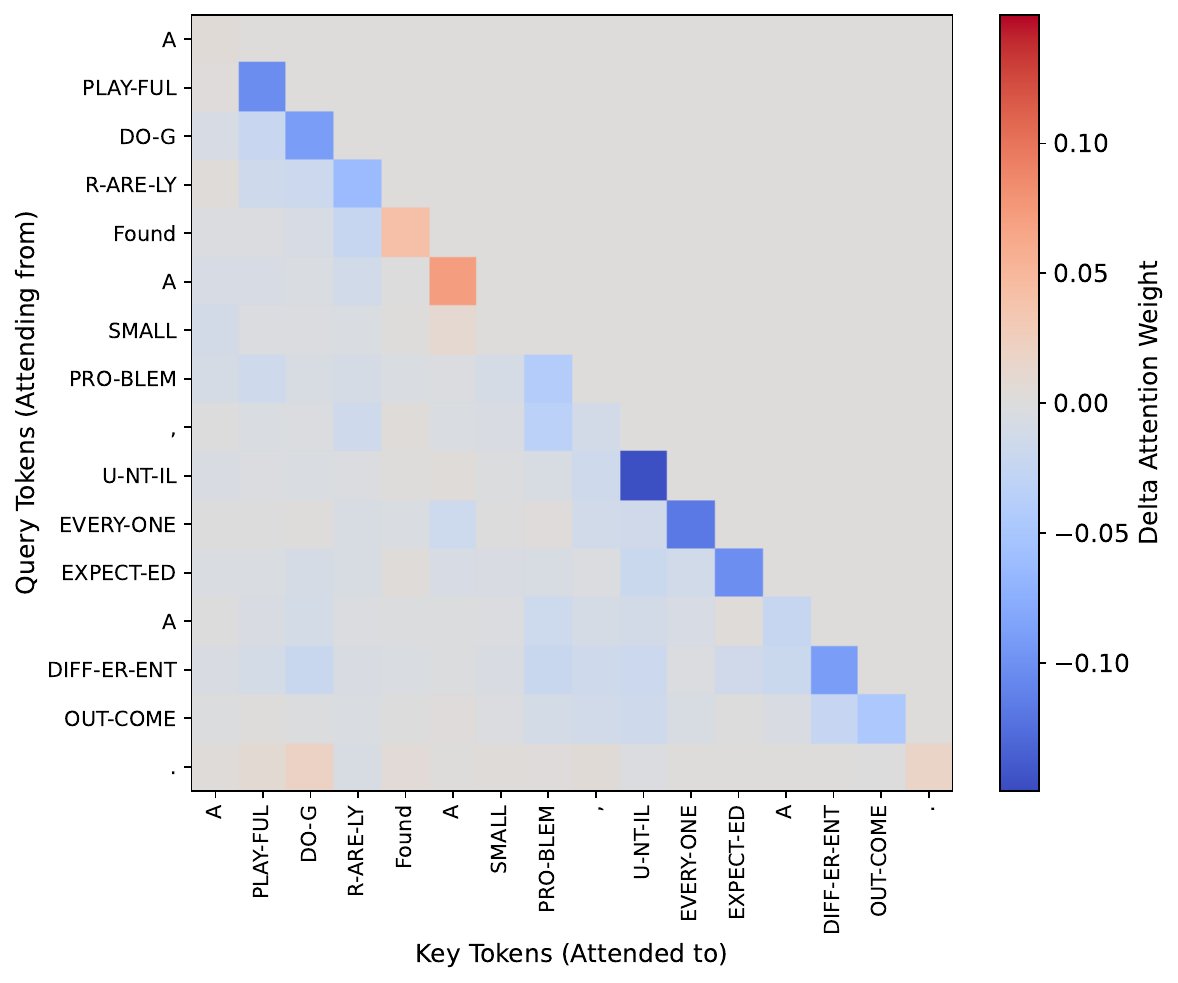}
    \end{subfigure}% 
    \hspace{2em}
    \begin{subfigure}{0.26\textwidth} % width of the subfigure
        \centering
        \includegraphics[width=\textwidth]{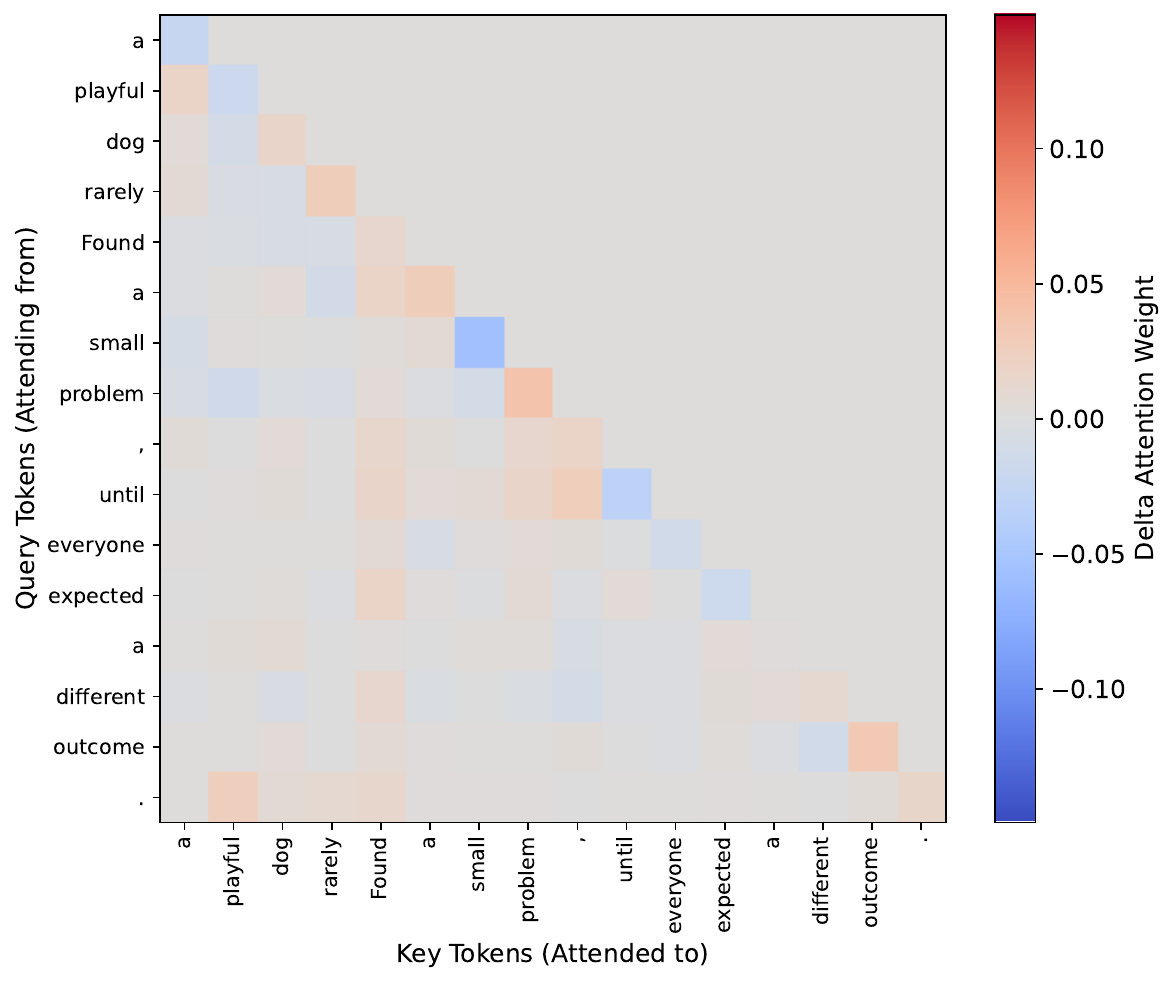}
    \end{subfigure}% 
    \hspace{2em}
    \begin{subfigure}{0.26\textwidth} % width of the subfigure
        \centering
        \includegraphics[width=\textwidth]{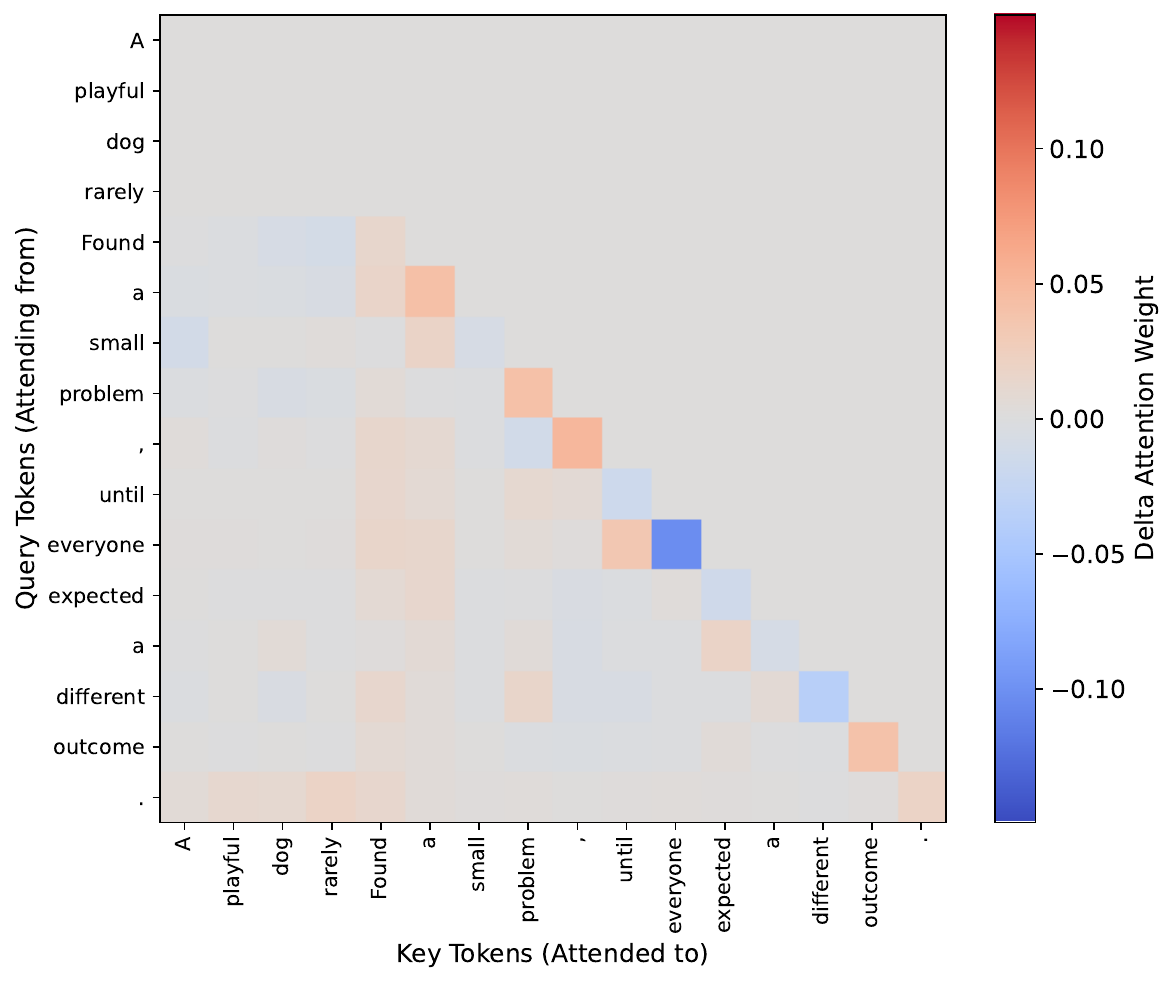}
    \end{subfigure}% 
    \vspace{0.5em}
    \begin{subfigure}{0.26\textwidth} % width of the subfigure
        \centering
        \includegraphics[width=\textwidth]{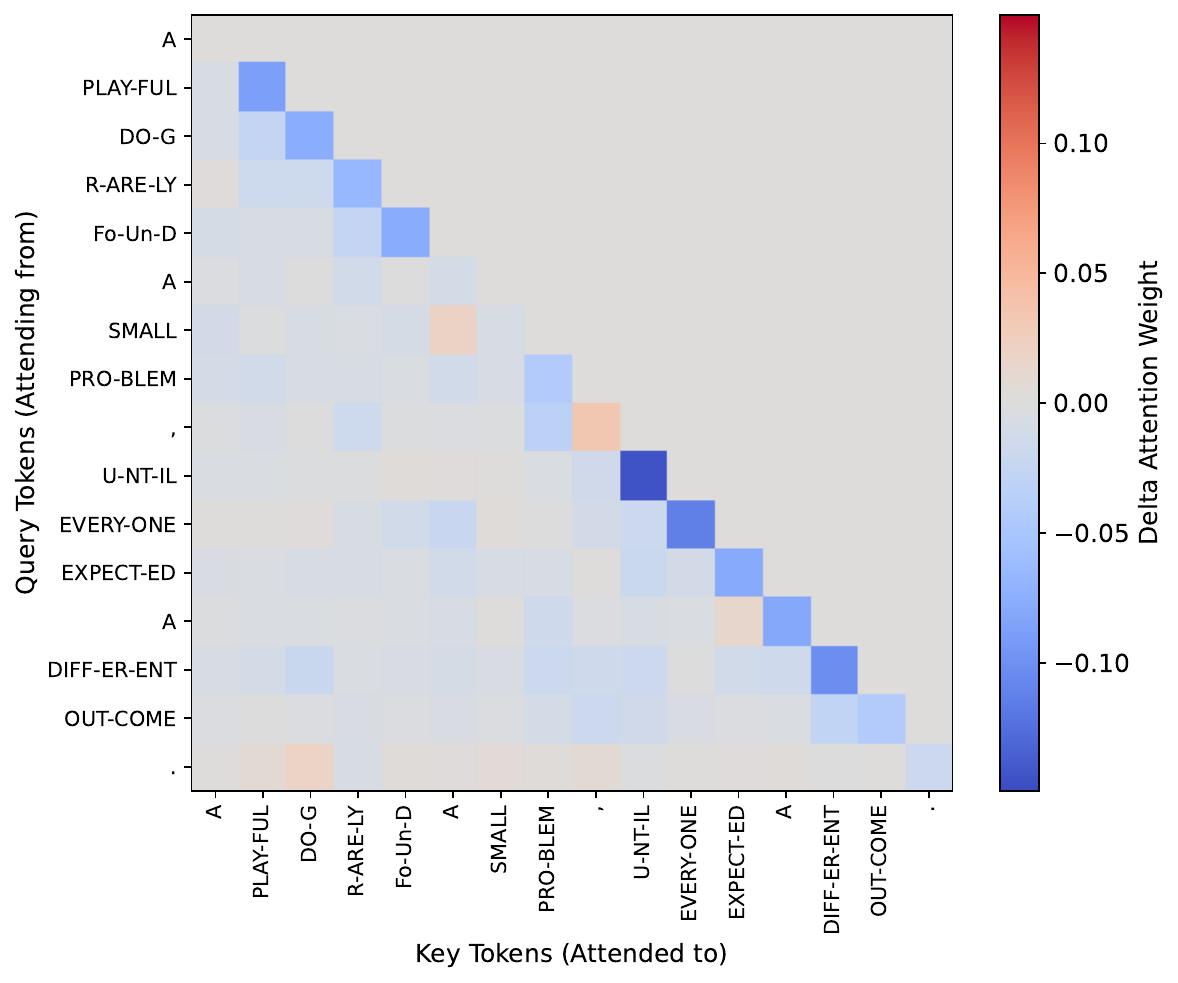}
    \end{subfigure}% 
    \hspace{2em}
    \begin{subfigure}{0.26\textwidth} % width of the subfigure
        \centering
        \includegraphics[width=\textwidth]{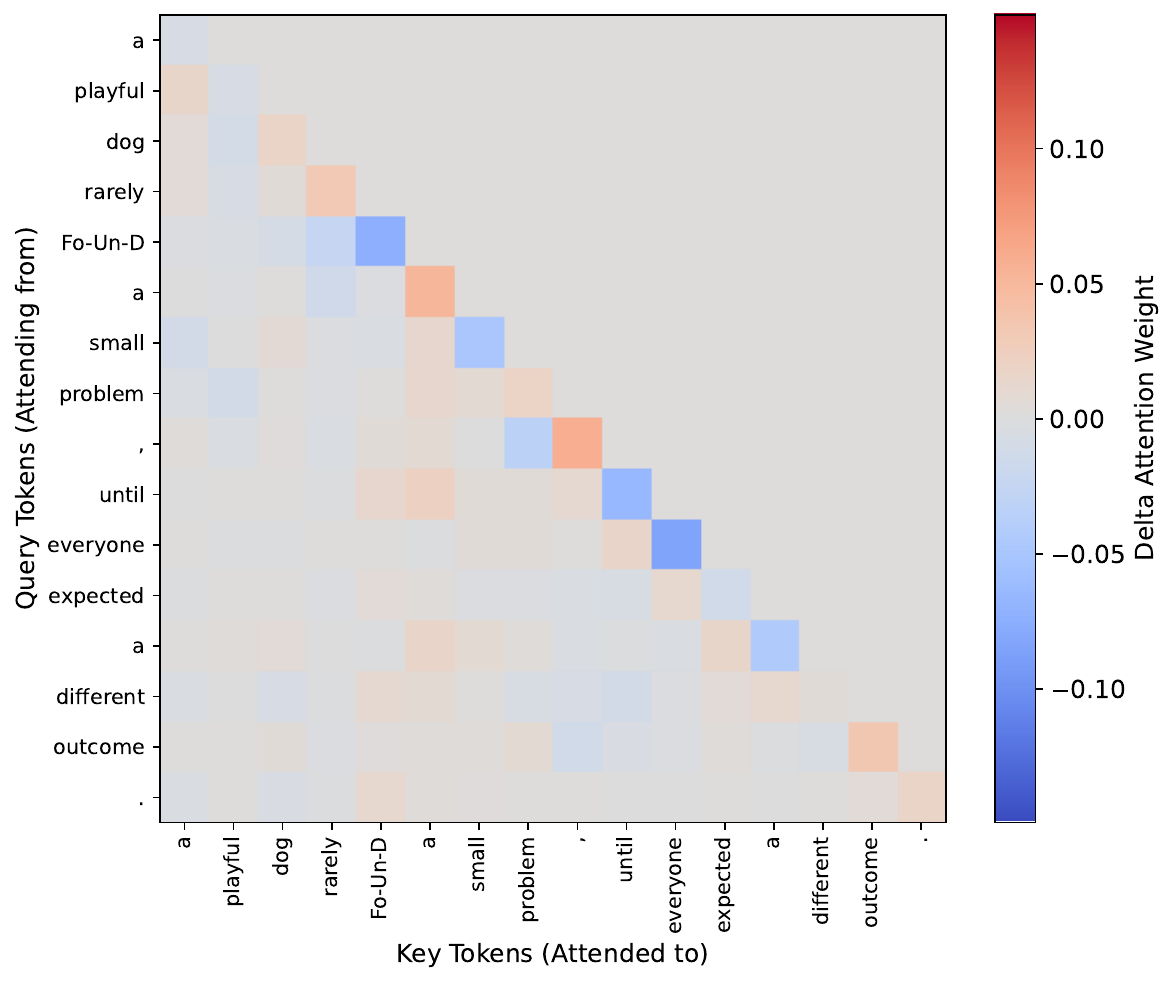}
    \end{subfigure}% 
    \hspace{2em}
    \begin{subfigure}{0.26\textwidth} % width of the subfigure
        \centering
        \includegraphics[width=\textwidth]{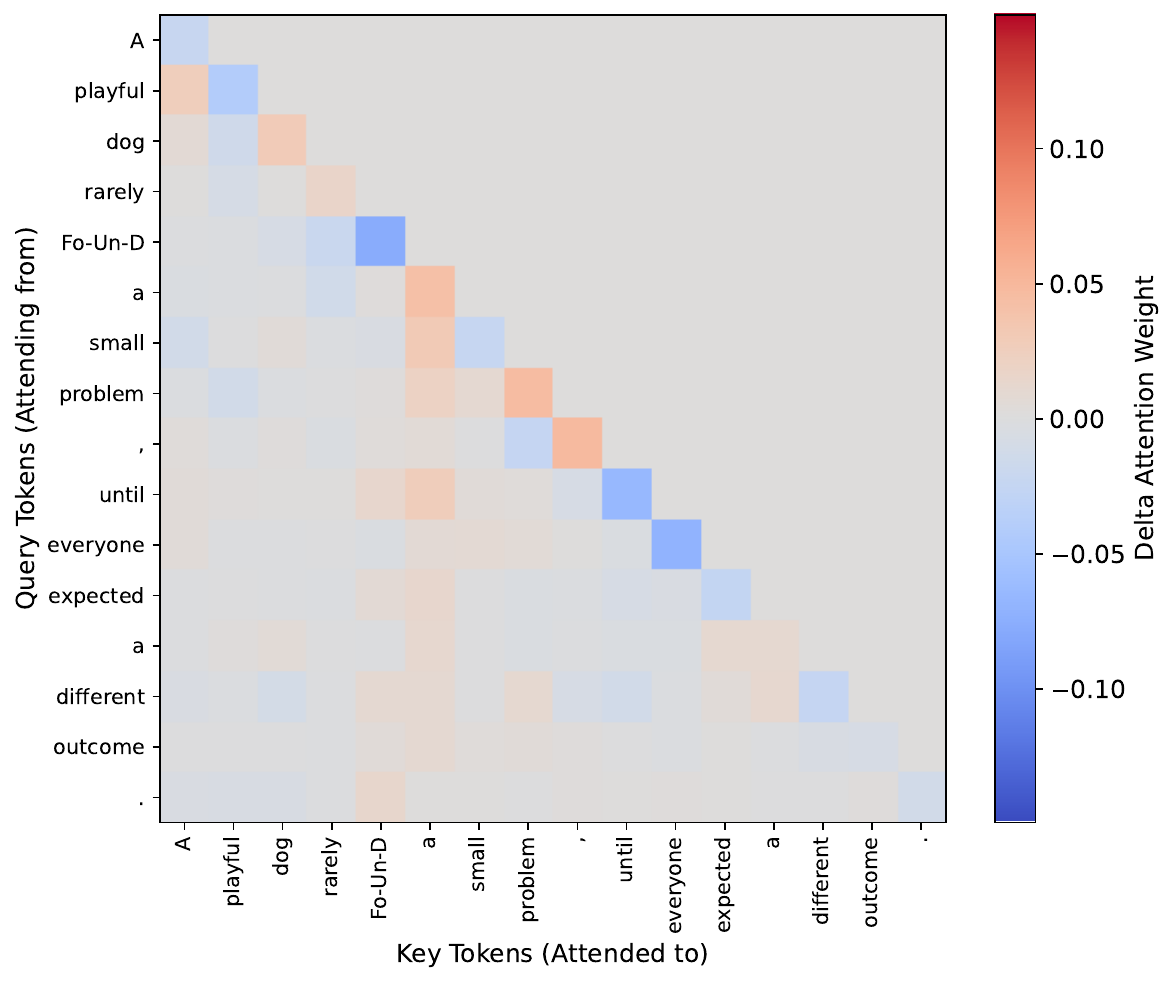}
    \end{subfigure}% 
    \vspace{0.5em}
    \begin{subfigure}{0.26\textwidth} % width of the subfigure
        \centering
        \includegraphics[width=\textwidth]{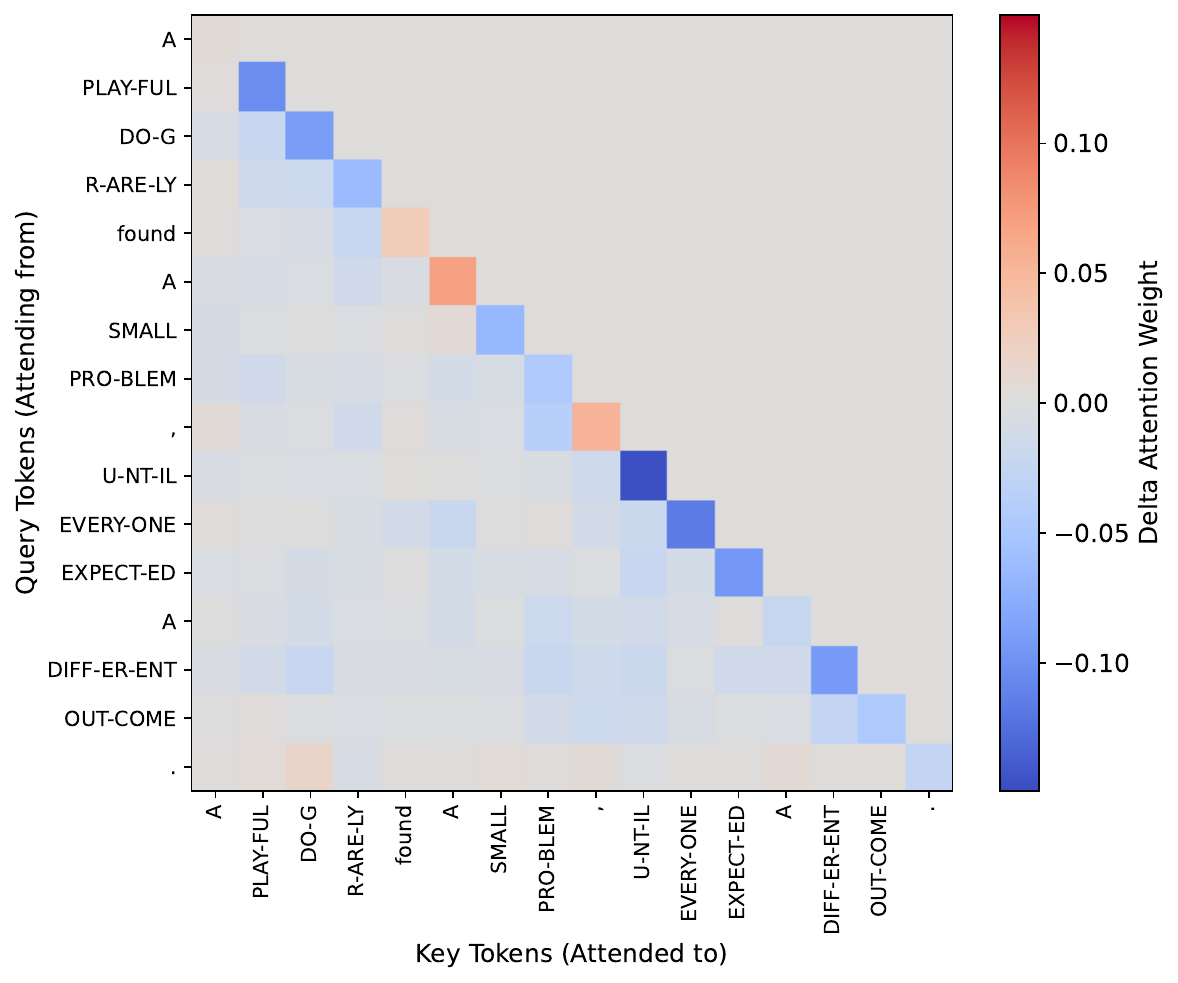}
    \end{subfigure}% 
    \hspace{2em}
    \begin{subfigure}{0.26\textwidth} % width of the subfigure
        \centering
        \includegraphics[width=\textwidth]{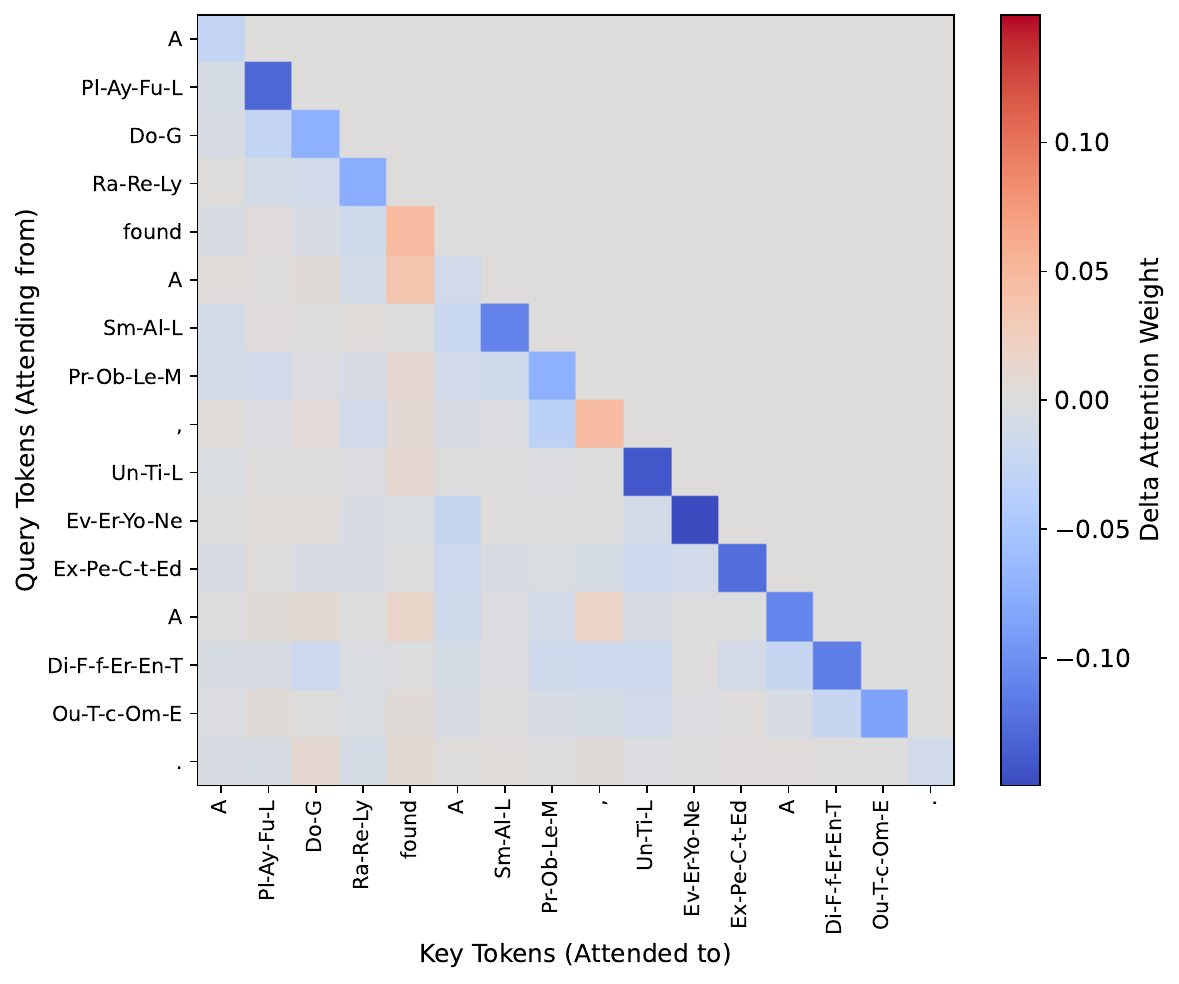}
    \end{subfigure}% 
    \hspace{2em}
    \begin{subfigure}{0.26\textwidth} % width of the subfigure
        \centering
        \includegraphics[width=\textwidth]{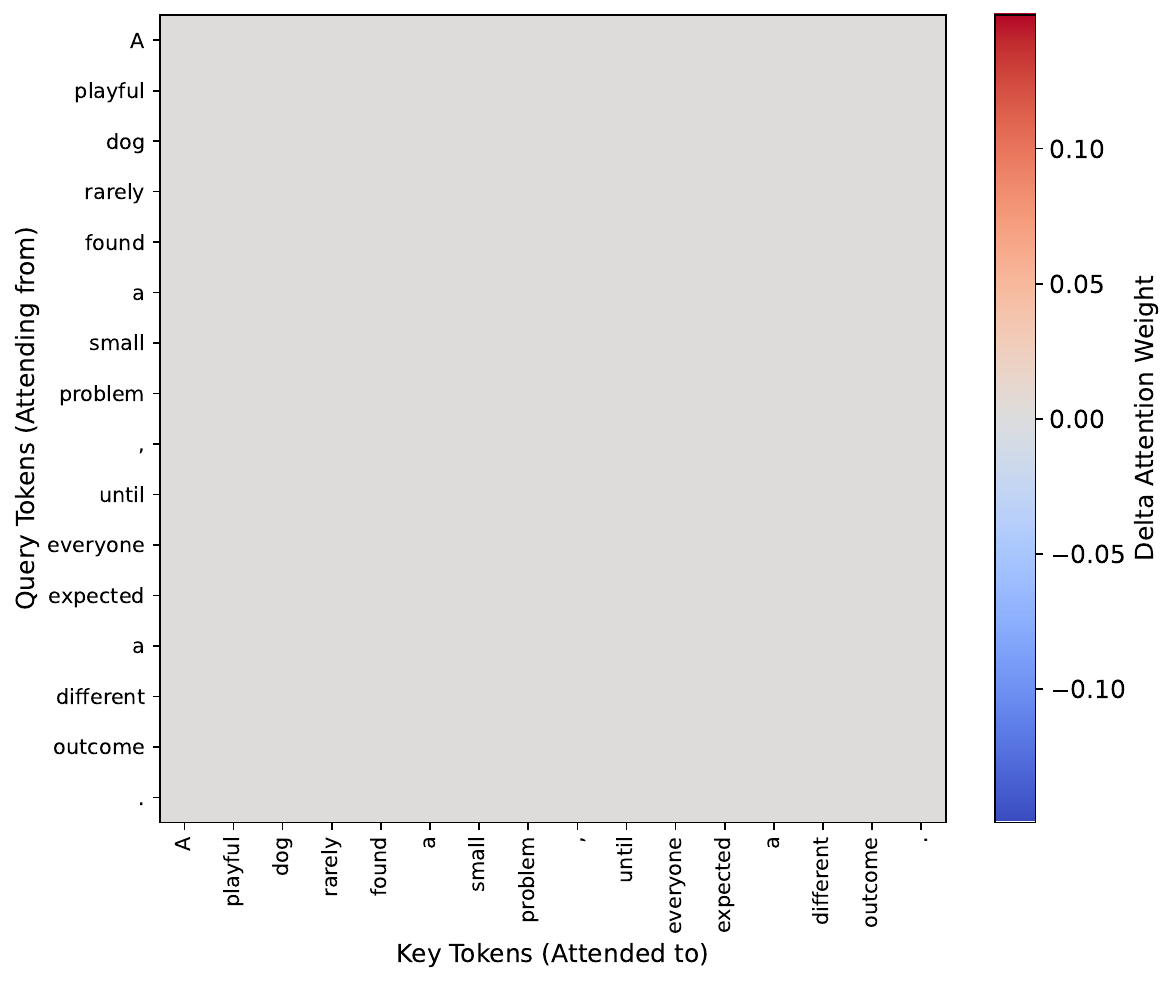}
    \end{subfigure}% 
    \caption{\textbf{Relative attention shifts for the 500-sentence self-reported dataset.} The target word in this example is \textbf{\textit{found}}. Increases in relative attention mass compared to the natural cased baseline sentence are shown in red, while decreases are shown in blue. The panels, from top-left to bottom-right, correspond to schemes U1--3, TE1--3, TT1--3, TA1--3, and ADE1--3.}
    \label{fig:attention_maps_384}
\end{figure*}

\begin{figure*}[!ht]
    \centering
    \begin{subfigure}{0.26\textwidth} % width of the subfigure
        \centering
        \includegraphics[width=\textwidth]{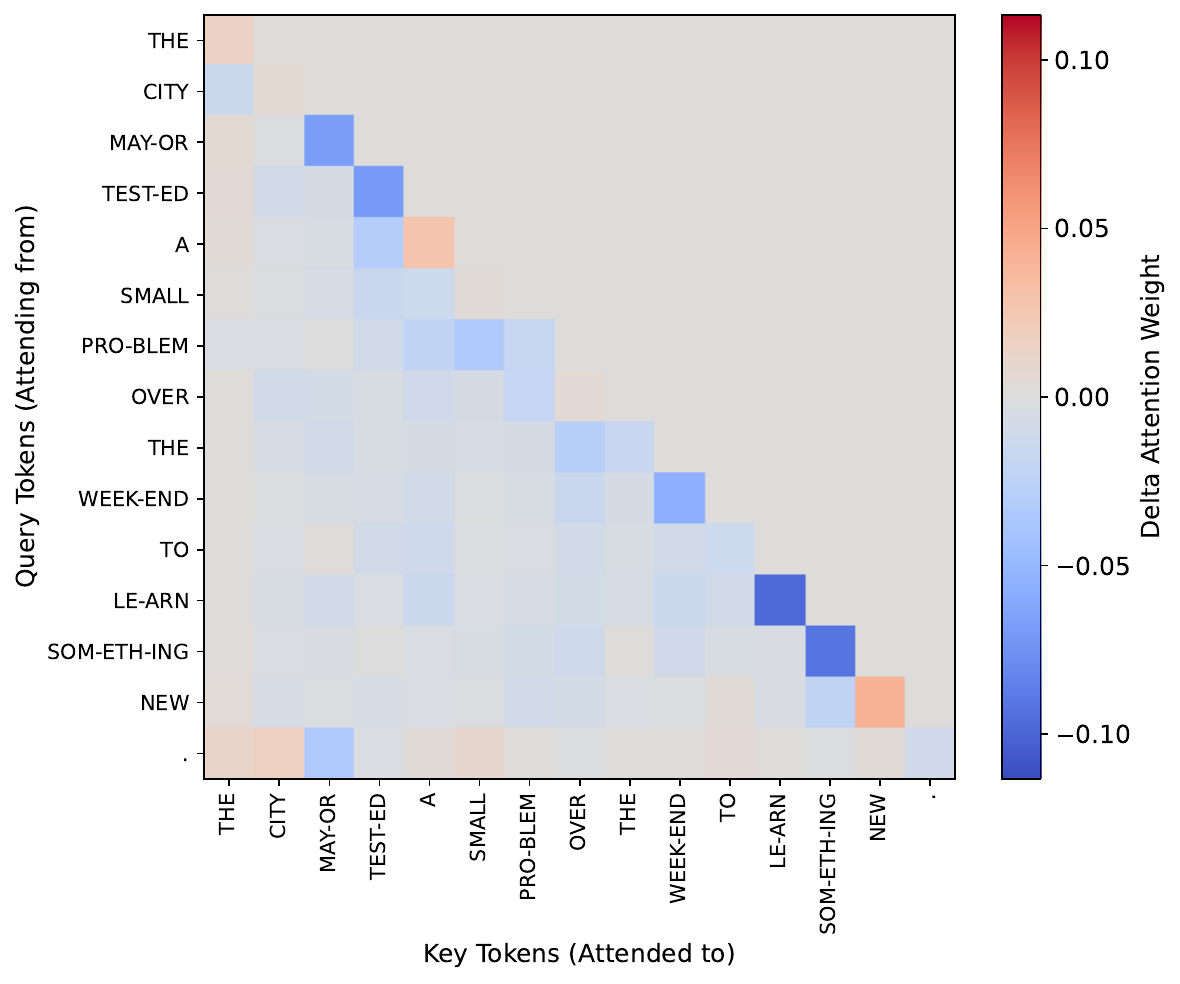}
    \end{subfigure}% 
    \hspace{2em}
    \begin{subfigure}{0.26\textwidth} % width of the subfigure
        \centering
        \includegraphics[width=\textwidth]{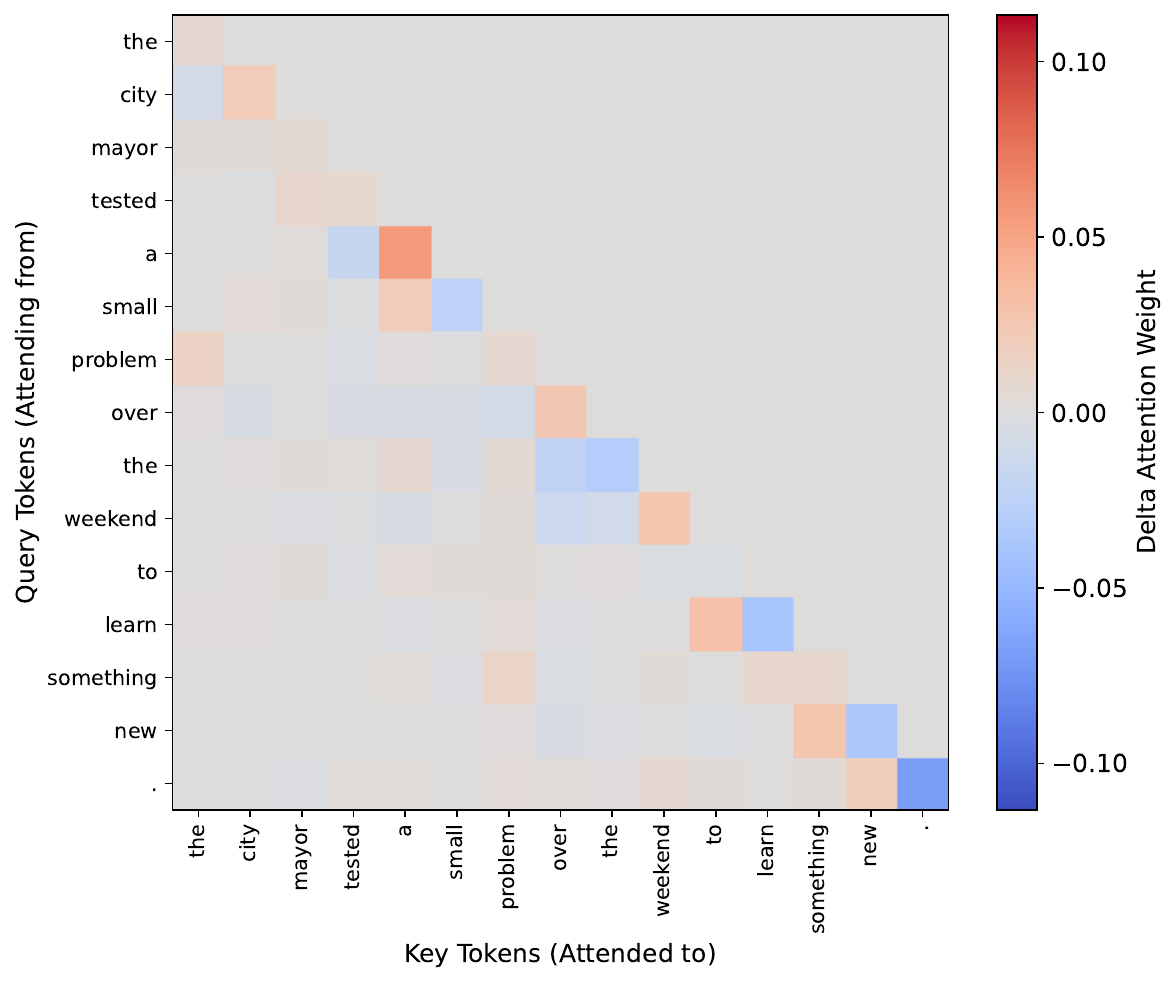}
    \end{subfigure}% 
    \hspace{2em}
    \begin{subfigure}{0.26\textwidth} % width of the subfigure
        \centering
        \includegraphics[width=\textwidth]{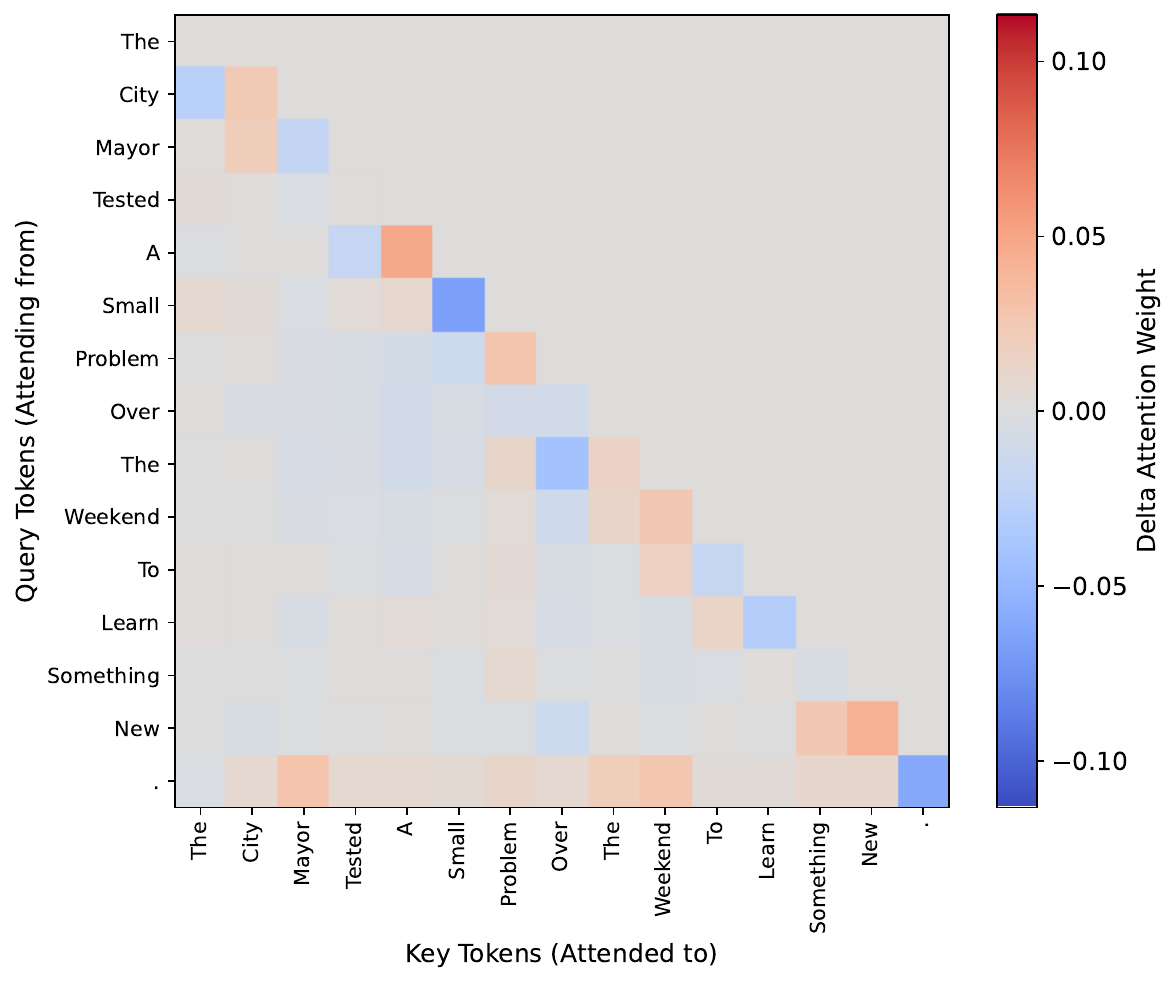}
    \end{subfigure}% 
    \vspace{0.5em}
    \begin{subfigure}{0.26\textwidth} % width of the subfigure
        \centering
        \includegraphics[width=\textwidth]{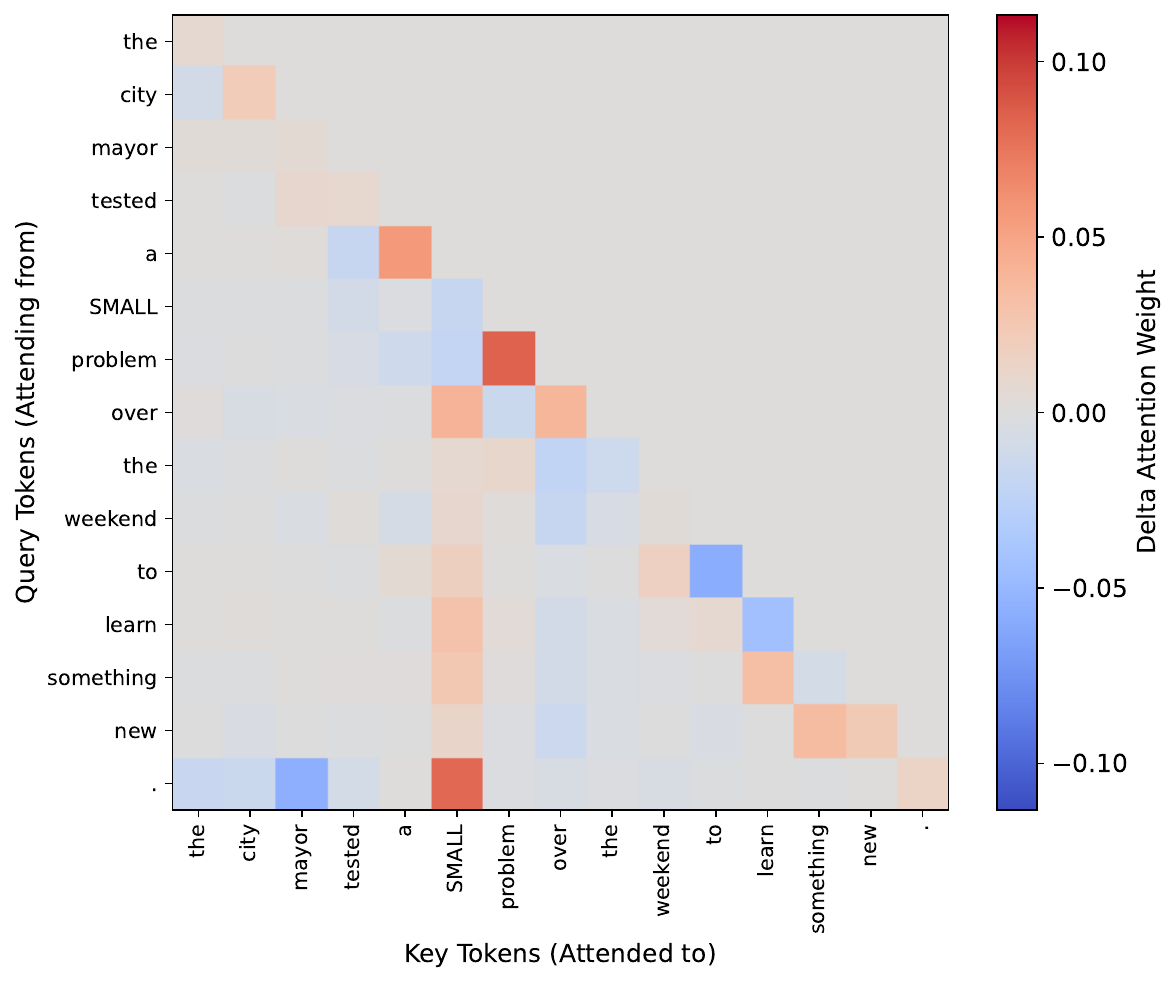}
    \end{subfigure}% 
    \hspace{2em}
    \begin{subfigure}{0.26\textwidth} % width of the subfigure
        \centering
        \includegraphics[width=\textwidth]{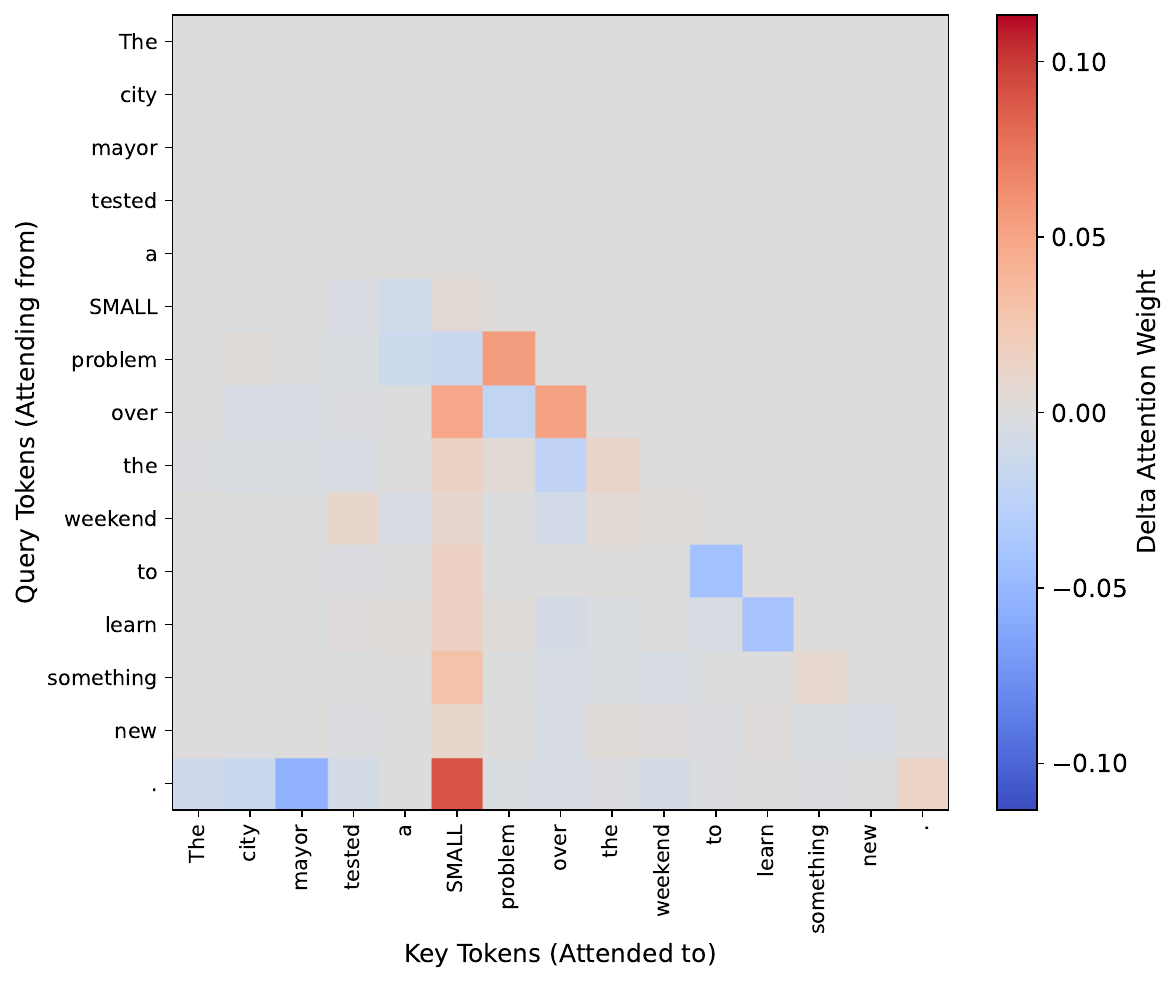}
    \end{subfigure}% 
    \hspace{2em}
    \begin{subfigure}{0.26\textwidth} % width of the subfigure
        \centering
        \includegraphics[width=\textwidth]{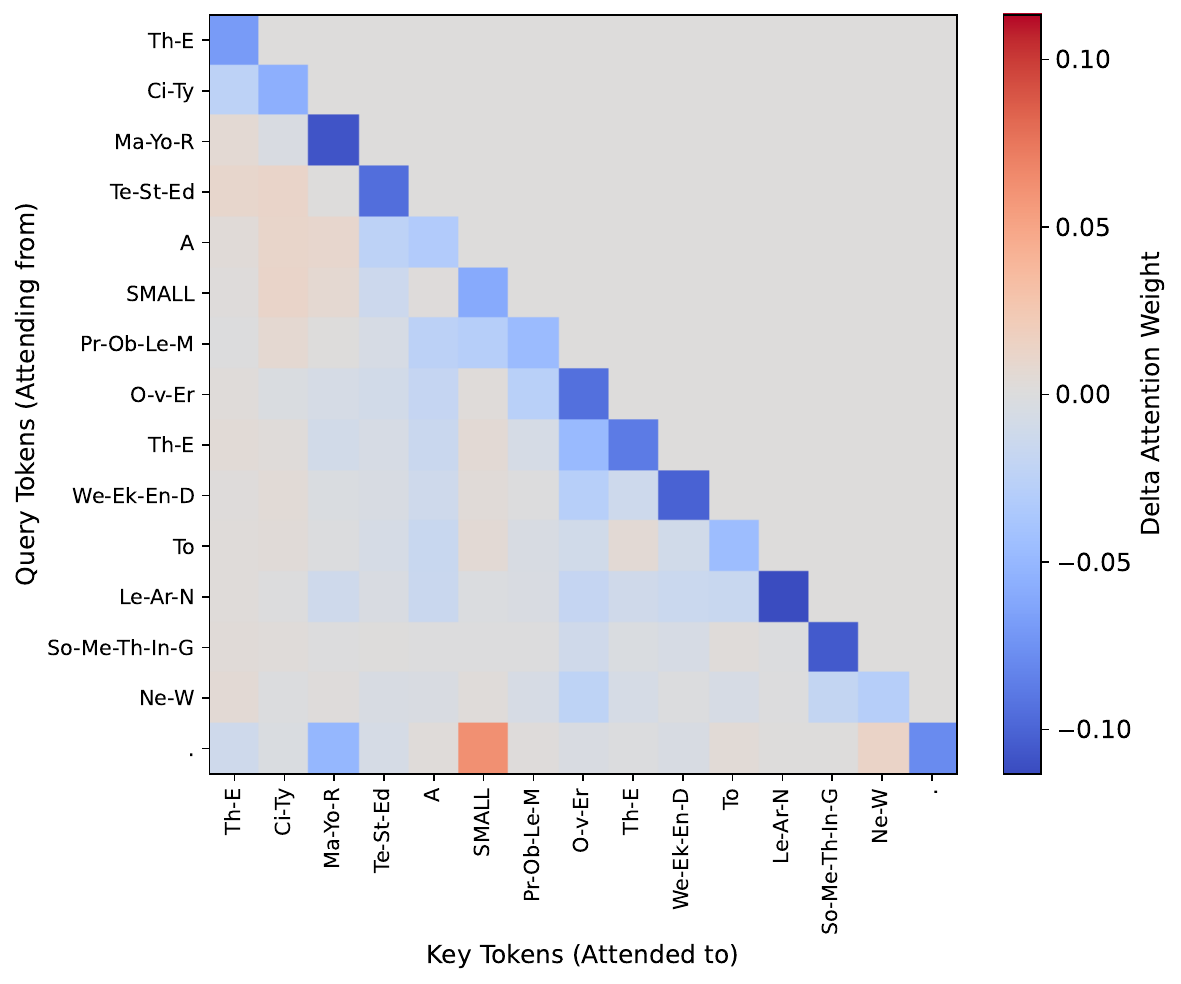}
    \end{subfigure}% 
    \vspace{0.5em}
    \begin{subfigure}{0.26\textwidth} % width of the subfigure
        \centering
        \includegraphics[width=\textwidth]{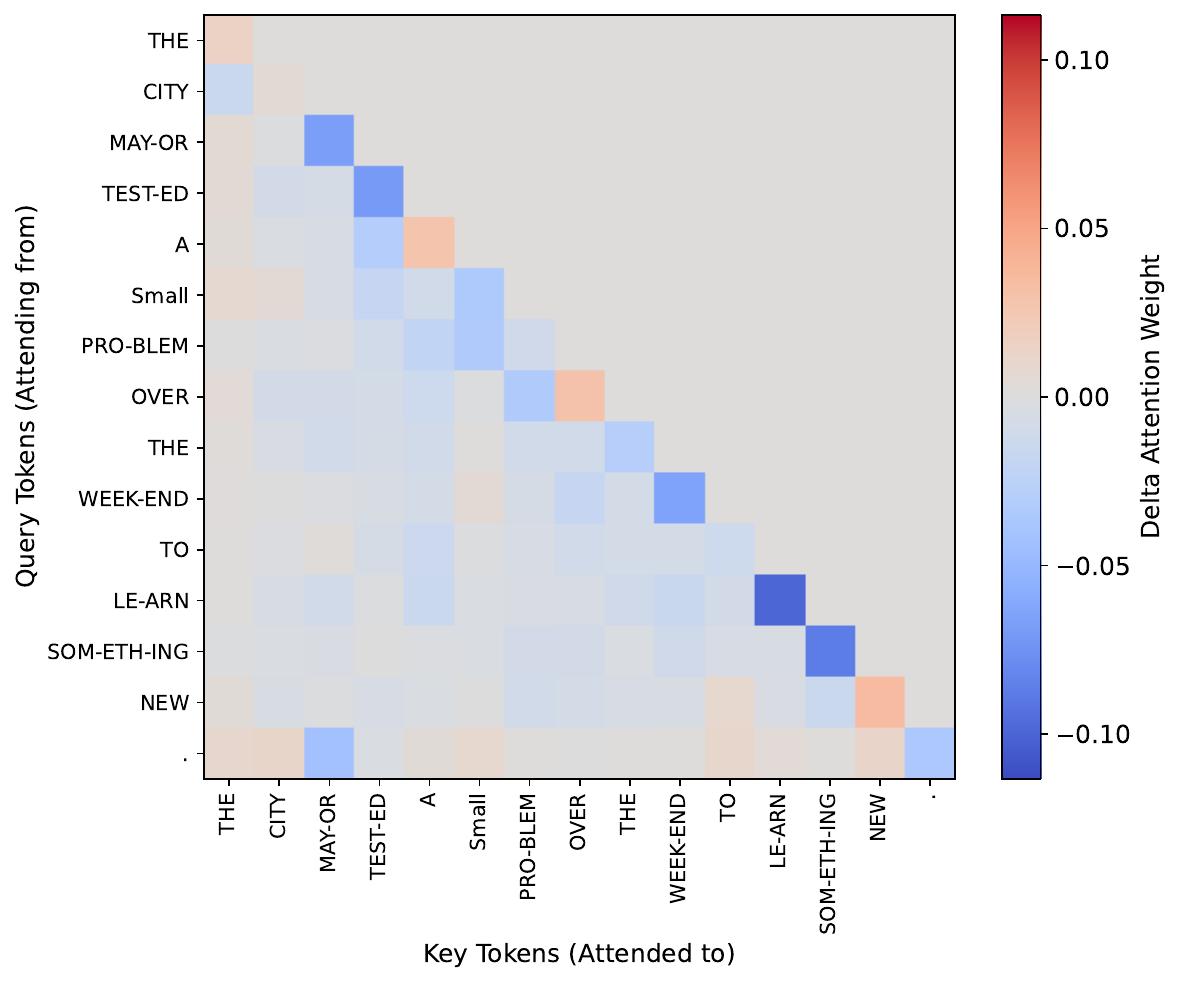}
    \end{subfigure}% 
    \hspace{2em}
    \begin{subfigure}{0.26\textwidth} % width of the subfigure
        \centering
        \includegraphics[width=\textwidth]{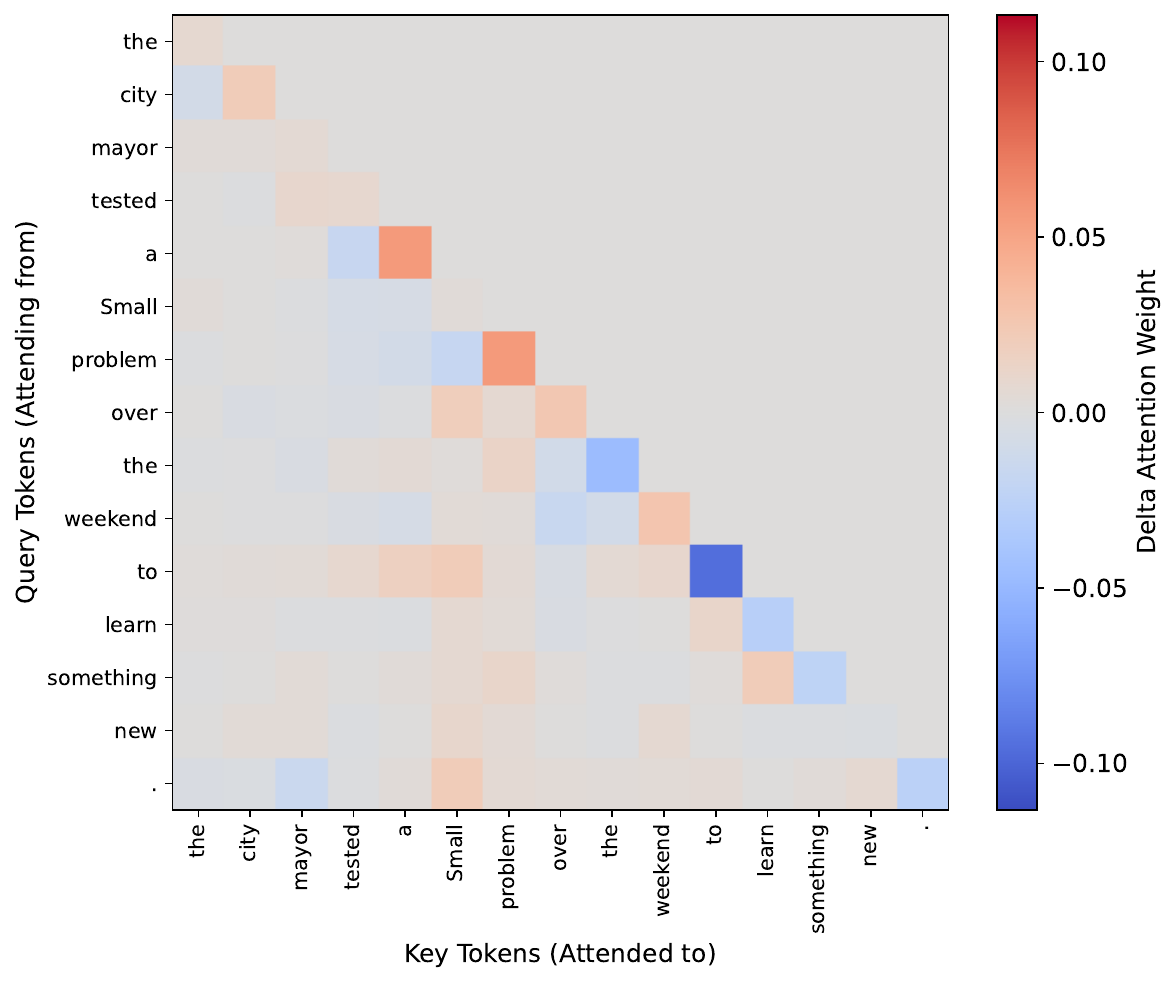}
    \end{subfigure}% 
    \hspace{2em}
    \begin{subfigure}{0.26\textwidth} % width of the subfigure
        \centering
        \includegraphics[width=\textwidth]{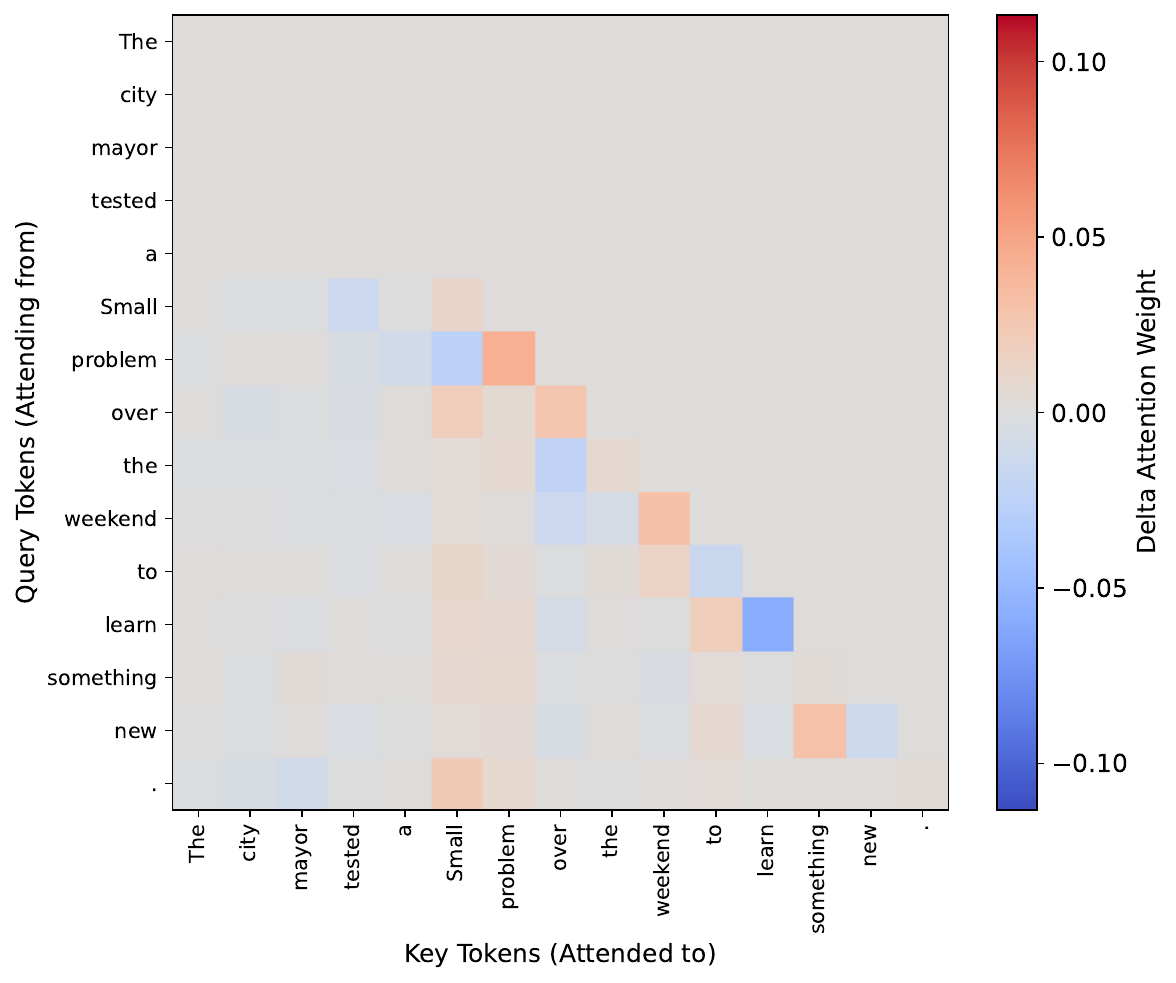}
    \end{subfigure}% 
    \vspace{0.5em}
    \begin{subfigure}{0.26\textwidth} % width of the subfigure
        \centering
        \includegraphics[width=\textwidth]{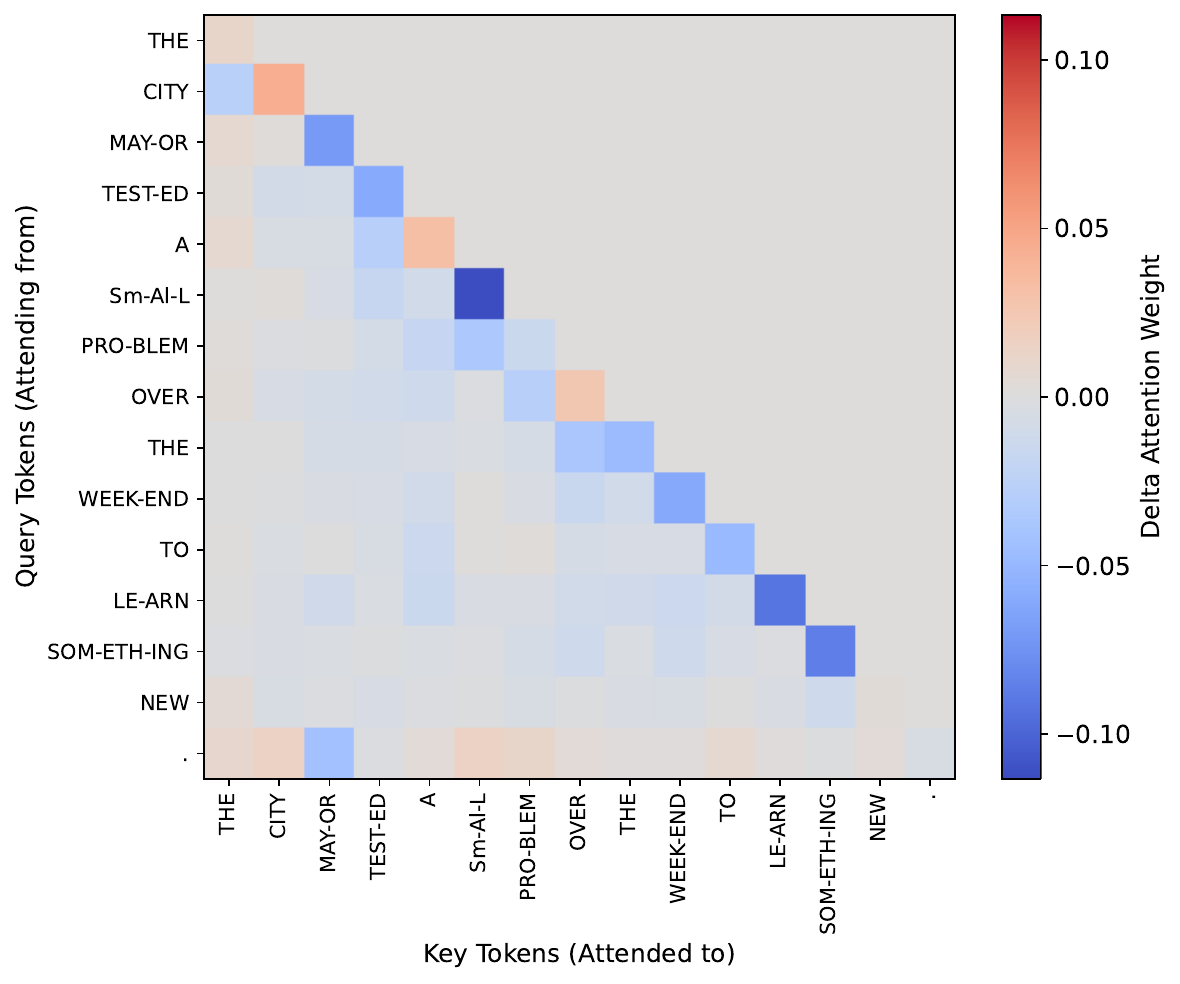}
    \end{subfigure}% 
    \hspace{2em}
    \begin{subfigure}{0.26\textwidth} % width of the subfigure
        \centering
        \includegraphics[width=\textwidth]{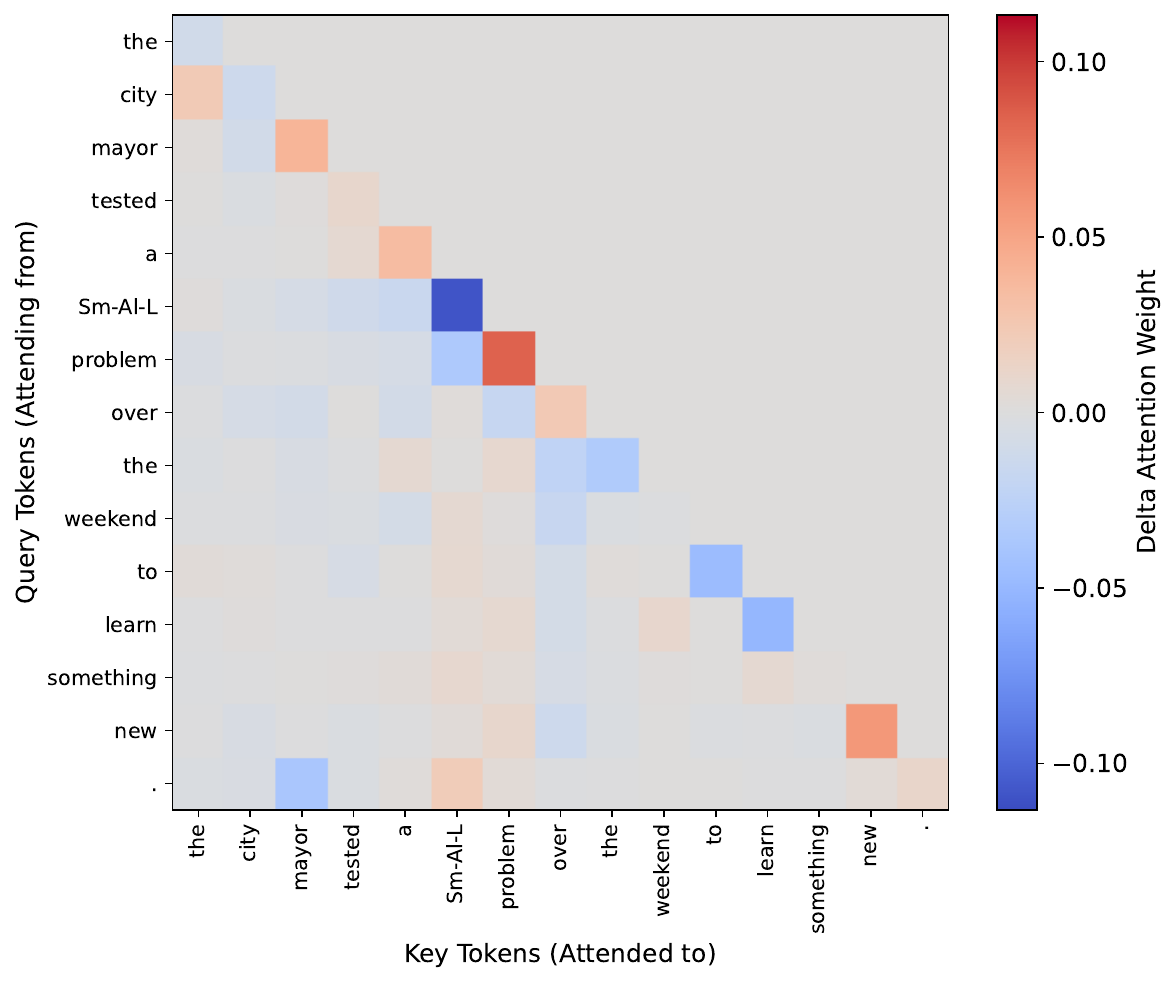}
    \end{subfigure}% 
    \hspace{2em}
    \begin{subfigure}{0.26\textwidth} % width of the subfigure
        \centering
        \includegraphics[width=\textwidth]{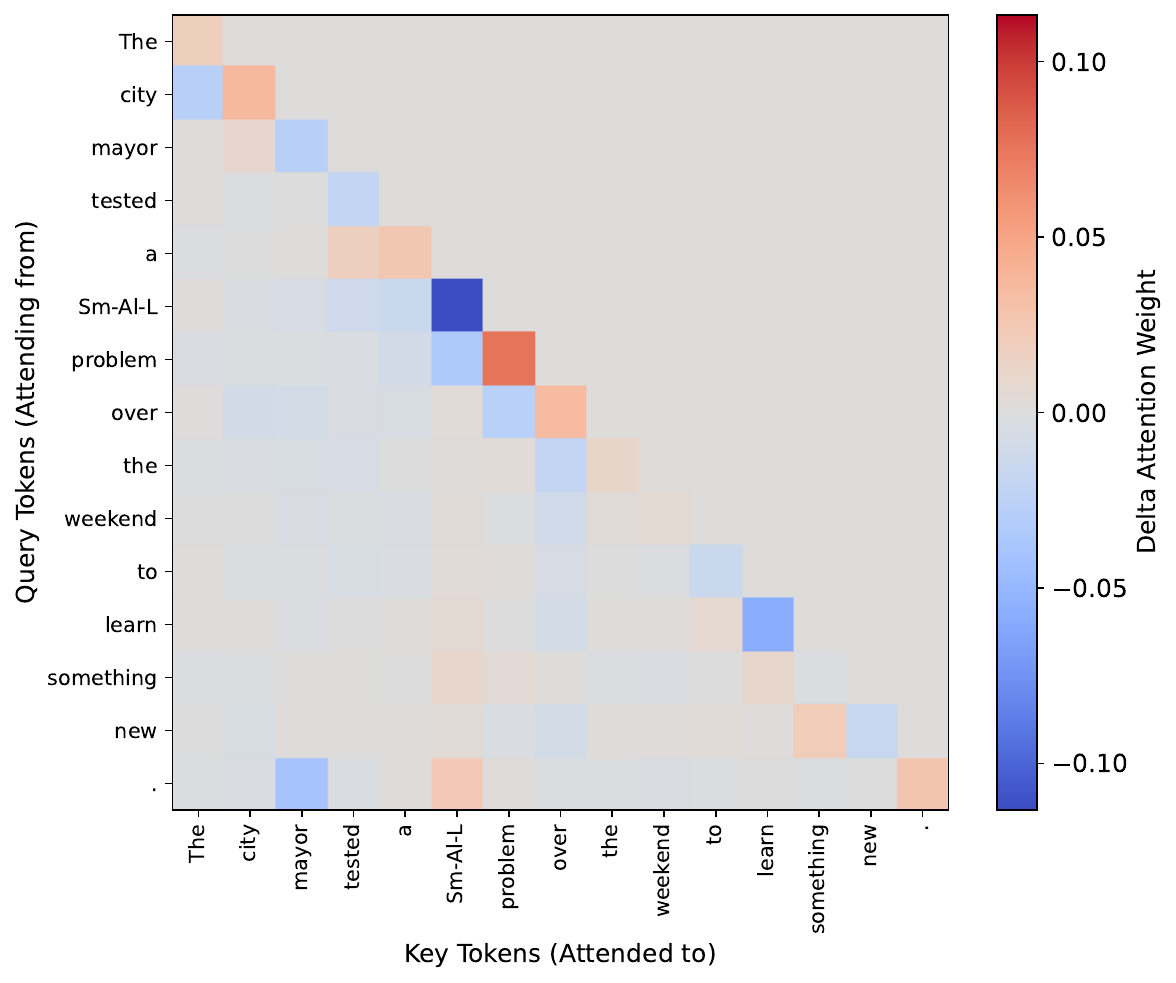}
    \end{subfigure}% 
    \vspace{0.5em}
    \begin{subfigure}{0.26\textwidth} % width of the subfigure
        \centering
        \includegraphics[width=\textwidth]{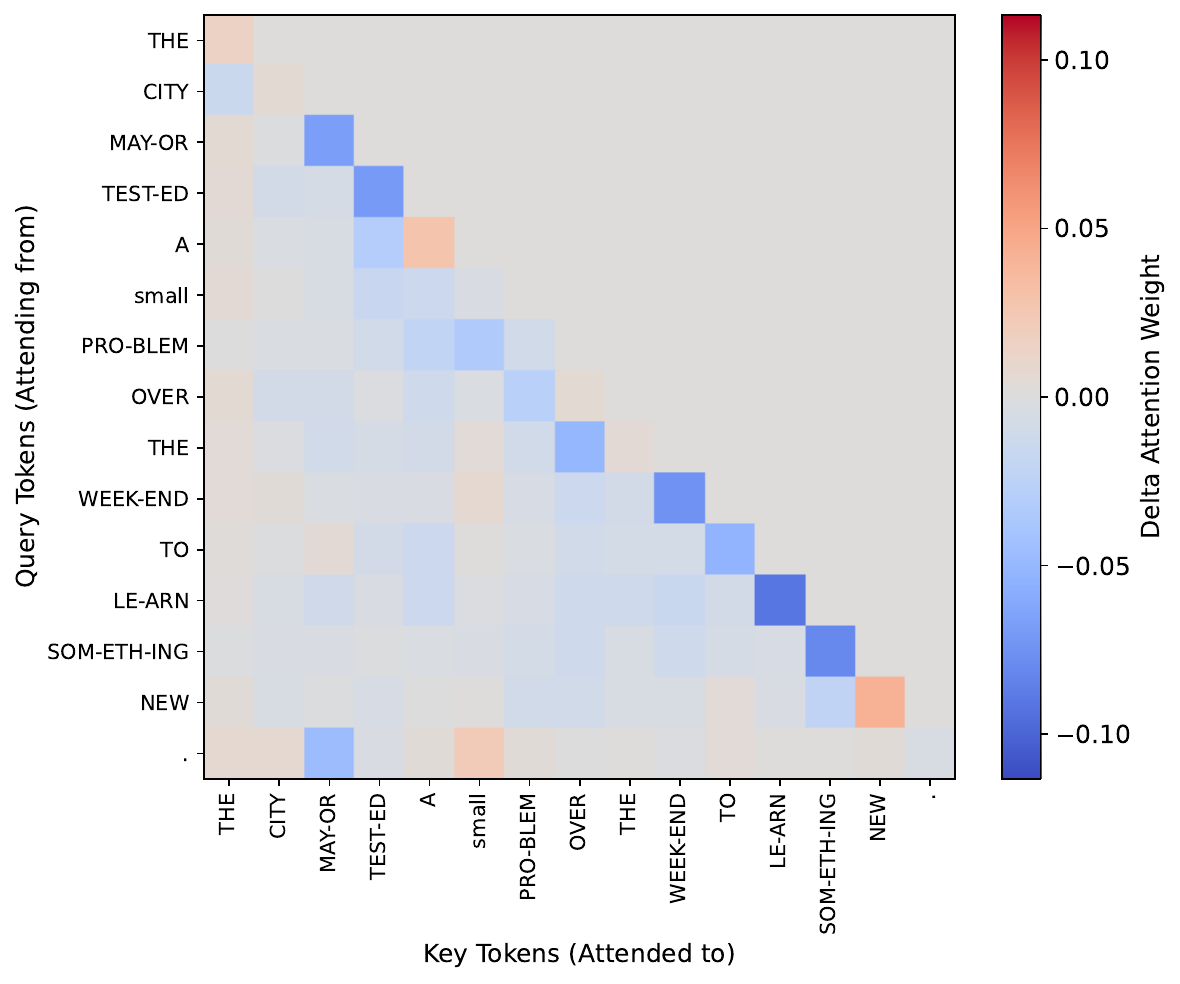}
    \end{subfigure}% 
    \hspace{2em}
    \begin{subfigure}{0.26\textwidth} % width of the subfigure
        \centering
        \includegraphics[width=\textwidth]{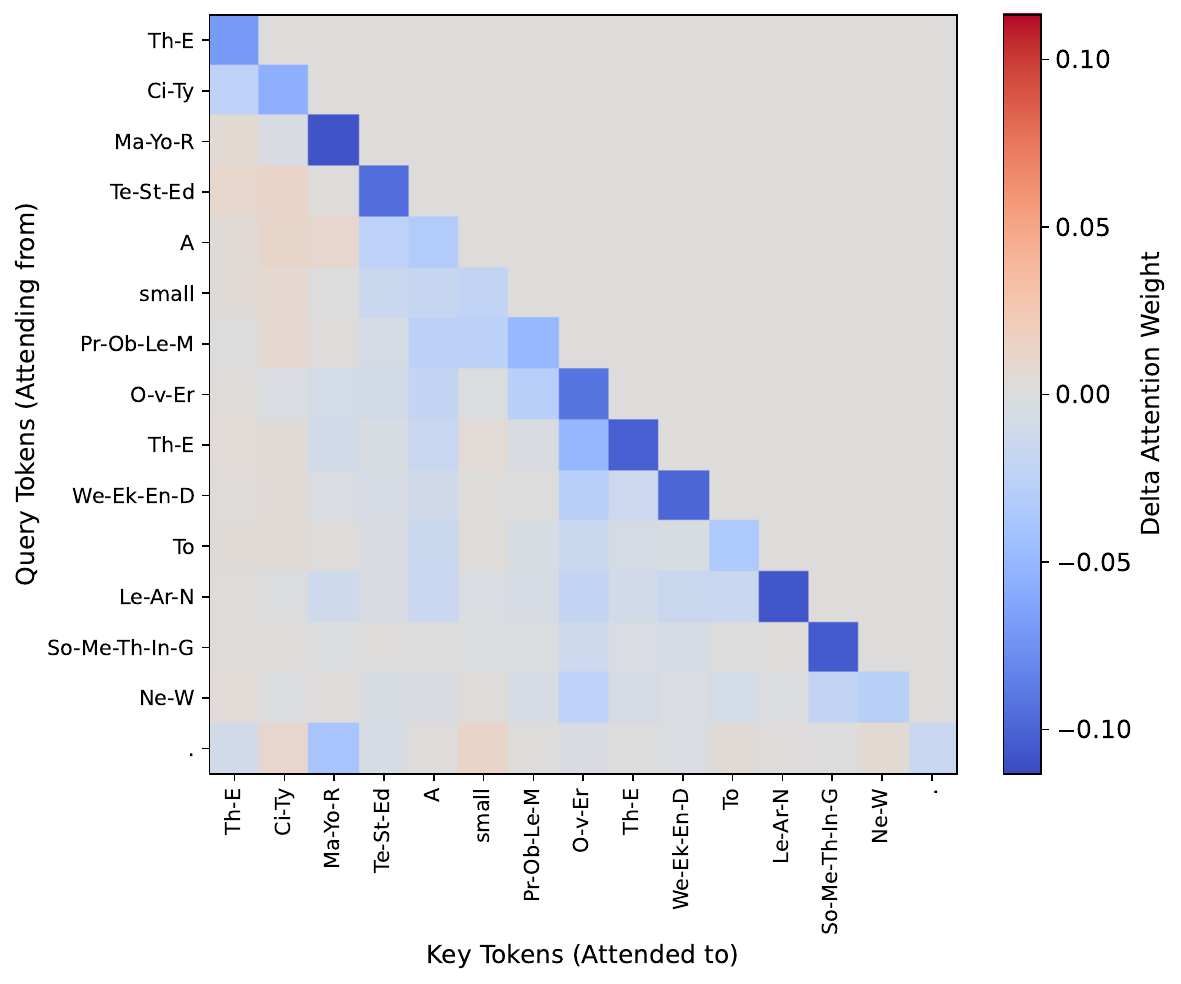}
    \end{subfigure}% 
    \hspace{2em}
    \begin{subfigure}{0.26\textwidth} % width of the subfigure
        \centering
        \includegraphics[width=\textwidth]{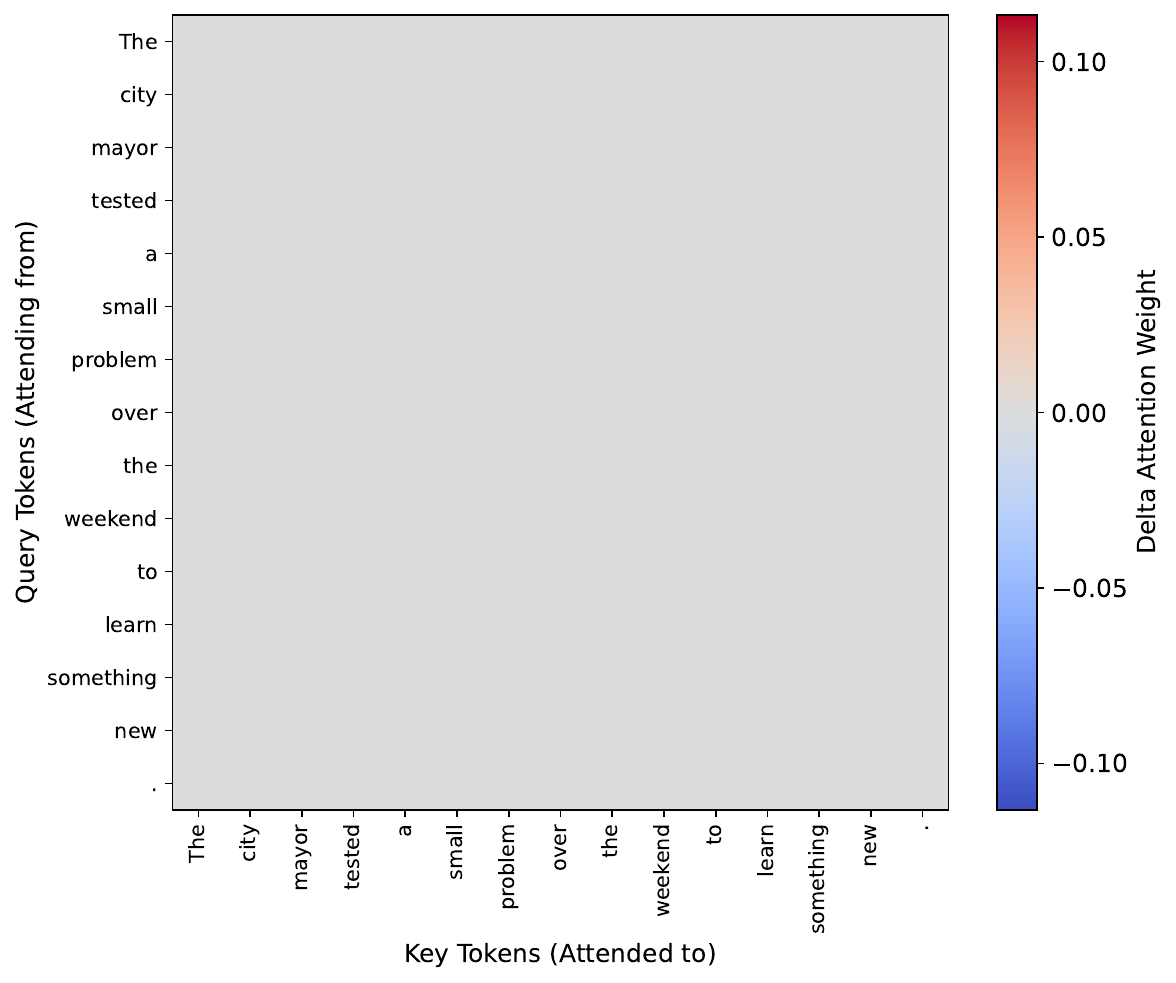}
    \end{subfigure}% 
    \caption{\textbf{Relative attention shifts for the 500-sentence self-reported dataset.} The target word in this example is \textbf{\textit{small}}. Increases in relative attention mass compared to the natural cased baseline sentence are shown in red, while decreases are shown in blue. The panels, from top-left to bottom-right, correspond to schemes U1--3, TE1--3, TT1--3, TA1--3, and ADE1--3.}
    \label{fig:attention_maps_119}
\end{figure*}

To isolate the model's contextual focus, special tokens---including system prompts and control characters---were explicitly excluded from the visualizations. This ensures that the attention matrices reflect the distribution of weights solely across the core lexical text (\cf \cref{fig:attention_maps_355,fig:attention_maps_384,fig:attention_maps_119}). Furthermore, to ensure comparability across letter casing schemes, scattered subword tokens representing a single semantic unit were aggregated by computing the mean of their respective attention weights. Analyzing these contextualized, word-level attention maps provides visual evidence that corroborates the quantitative shifts observed in our benchmark experiments.

When evaluating the model on prompts containing altered letter casing, we analyze the delta attention plots for the emergence of vertical structures, which serve as structural signatures of attention pooling. The model's behavior exhibits high sensitivity to the specific perturbation scheme applied. In the Global Uniformity (U) schemes, no substantial structural shifts are observable; deviations from the baseline are primarily confined to the main diagonal, indicating minor fluctuations in self-attention with no localized contextual redirection.

Conversely, the Target Emphasis (TE) interventions reveal a notable shift in attention toward the uppercased target span. This pooling manifests as a distinct vertical band across the attention plots, demonstrating that subsequent tokens systematically attend back to the emphasized word regardless of their relative position. Interestingly, in the TE3 configuration (target uppercase, context alternating case), all tokens self-attention decreases markedly. This phenomenon likely arises from the high orthographic entropy of the alternating-case context; the target itself becomes less salient, thereby diffusing the internal attention.

For the Target Title (TT) schemes, shifts in TT1 are again restricted almost exclusively to self-attention along the main diagonal, whereas TT2 and TT3 show small deviations from the natural-case baseline. This suggests that standard title-casing is a too subtle signal to trigger significant cross-token redirection.

In the Target Alternating (TA) schemes, a localized small increase in attention on the target token is visible, particularly in the TA2 and TA3 configurations. While these results align with our broader benchmark findings, the magnitude of the attention shift in this 500-sentence study is less pronounced than that observed in long-form datasets. This attenuation is likely a function of sequence length: in shorter sentences, the ``attentional budget'' is distributed across fewer tokens, resulting in smaller absolute deltas even when the underlying sensitivity remains consistent.

Finally, the Adversarial De-Emphasis (ADE) schemes are characterized by a notable decrease in self-attention along the main diagonal. Within this group, ADE3 serves as a unique case, displaying zero change. This is mathematically expected, as the ADE3 protocol (target lowercase, context natural) replicates the original sentence casing exactly in our test set, resulting in an identical attention distribution and confirming the validity of our experimental pipeline.

We acknowledge that LLMs may hallucinate meta-cognitive focus. However, our Alignment Score demonstrates that our self-reports are not random, but consistently reflect observed numerical attention shifts, suggesting explanations align with the model's actual internal processing state.

\subsection{Extended Multimodal Experimental Evaluation}
\label{appendix:multimodal_results}

\begin{figure*}[!ht]
    \centering
    \begin{subfigure}{0.18\textwidth} % width of the subfigure
        \centering
        \includegraphics[width=\textwidth]{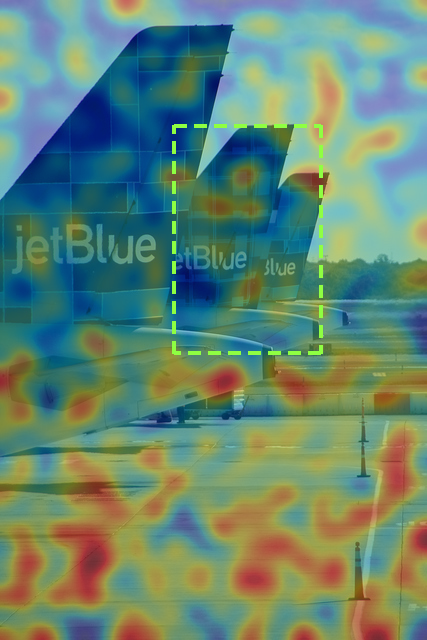}
        \caption{U1}
    \end{subfigure}% 
    \hspace{0.5em}
    \begin{subfigure}{0.18\textwidth} % width of the subfigure
        \centering
        \includegraphics[width=\textwidth]{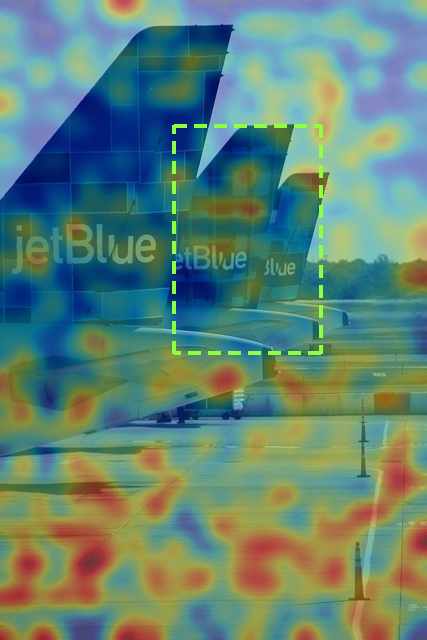}
        \caption{U2}
    \end{subfigure}% 
    \hspace{0.5em}
    \begin{subfigure}{0.18\textwidth} % width of the subfigure
        \centering
        \includegraphics[width=\textwidth]{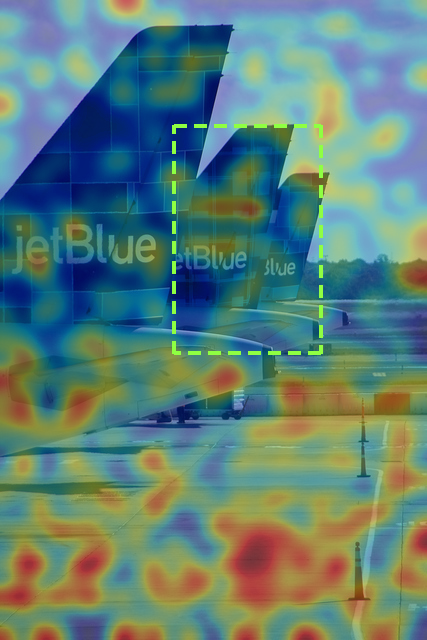}
        \caption{U3}
    \end{subfigure}% 
    \hspace{0.5em}
    \begin{subfigure}{0.18\textwidth} % width of the subfigure
        \centering
        \includegraphics[width=\textwidth]{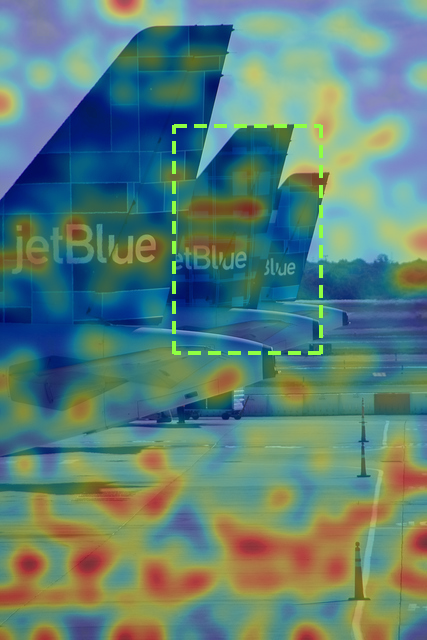}
        \caption{TE1}
    \end{subfigure}% 
    
    \vspace{0.5em}
    
    \begin{subfigure}{0.18\textwidth} % width of the subfigure
        \centering
        \includegraphics[width=\textwidth]{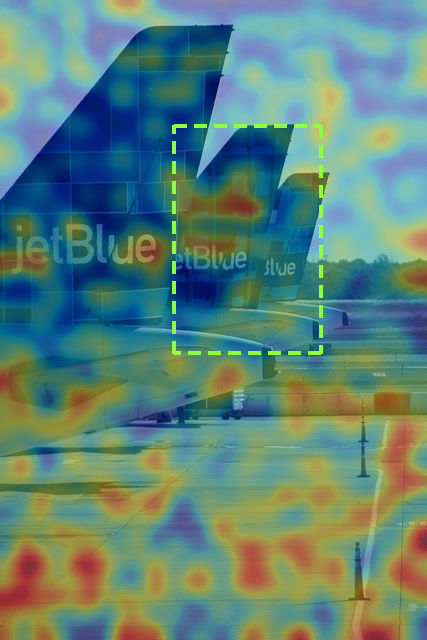}
        \caption{TE2}
    \end{subfigure}% 
    \hspace{0.5em}
    \begin{subfigure}{0.18\textwidth} % width of the subfigure
        \centering
        \includegraphics[width=\textwidth]{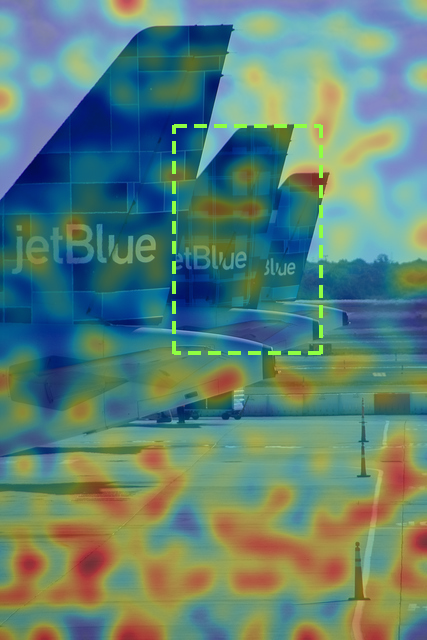}
        \caption{TE3}
    \end{subfigure}% 
    \hspace{0.5em}
    \begin{subfigure}{0.18\textwidth} % width of the subfigure
        \centering
        \includegraphics[width=\textwidth]{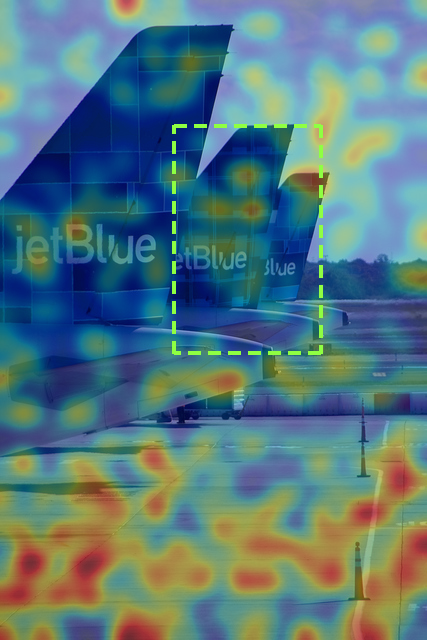}
        \caption{TT1}
    \end{subfigure}% 
    \hspace{0.5em}
    \begin{subfigure}{0.18\textwidth} % width of the subfigure
        \centering
        \includegraphics[width=\textwidth]{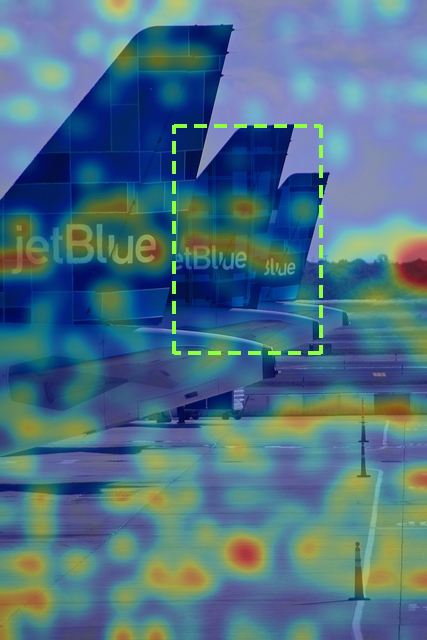}
        \caption{TT2}
    \end{subfigure}% 
    
    \vspace{0.5em}
    
    \begin{subfigure}{0.18\textwidth} % width of the subfigure
        \centering
        \includegraphics[width=\textwidth]{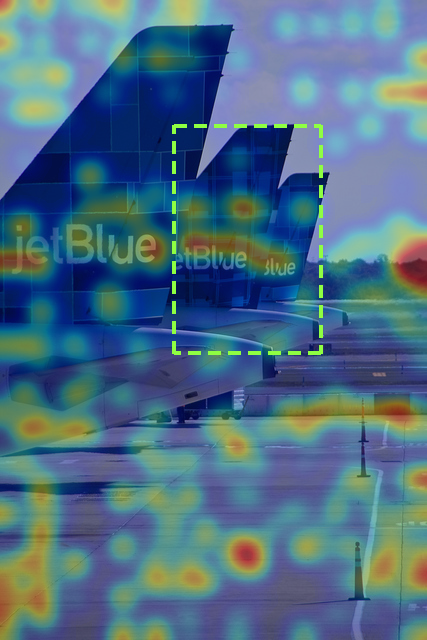}
        \caption{TT3}
    \end{subfigure}% 
    \hspace{0.5em}
    \begin{subfigure}{0.18\textwidth} % width of the subfigure
        \centering
        \includegraphics[width=\textwidth]{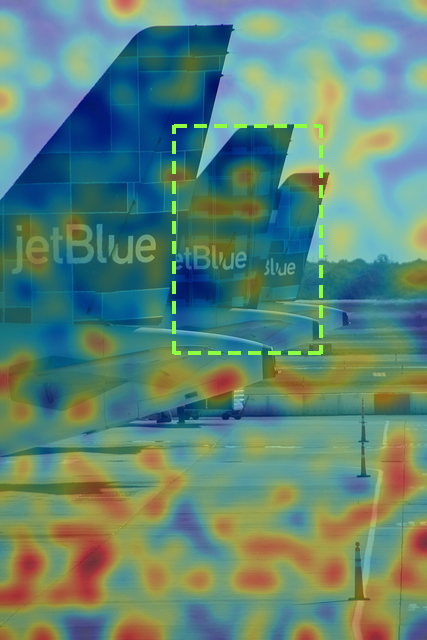}
        \caption{TA1}
    \end{subfigure}% 
    \hspace{0.5em}
    \begin{subfigure}{0.18\textwidth} % width of the subfigure
        \centering
        \includegraphics[width=\textwidth]{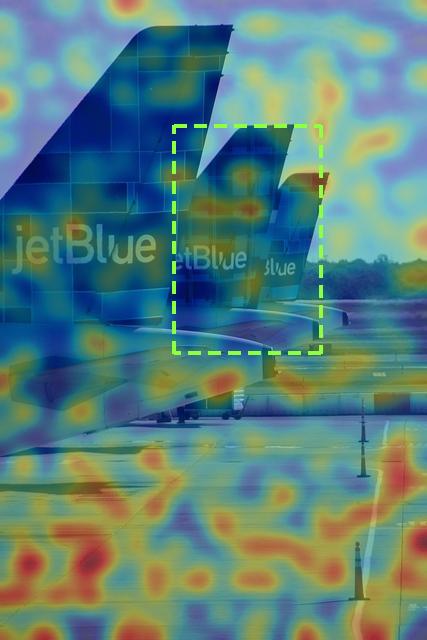}
        \caption{TA2}
    \end{subfigure}% 
    \hspace{0.5em}
    \begin{subfigure}{0.18\textwidth} % width of the subfigure
        \centering
        \includegraphics[width=\textwidth]{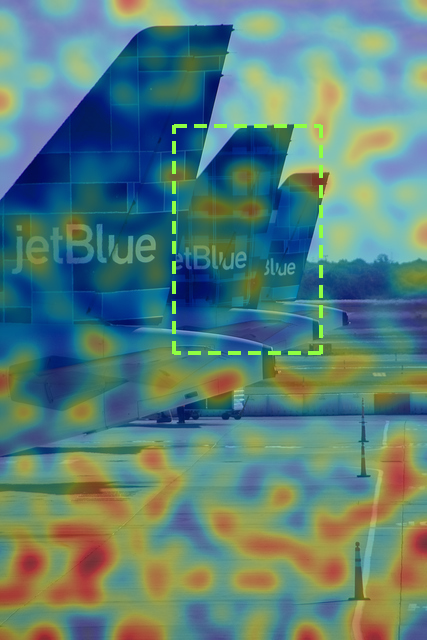}
        \caption{TA3}
    \end{subfigure}% 
    
    \vspace{0.5em}
    
    \begin{subfigure}{0.18\textwidth} % width of the subfigure
        \centering
        \includegraphics[width=\textwidth]{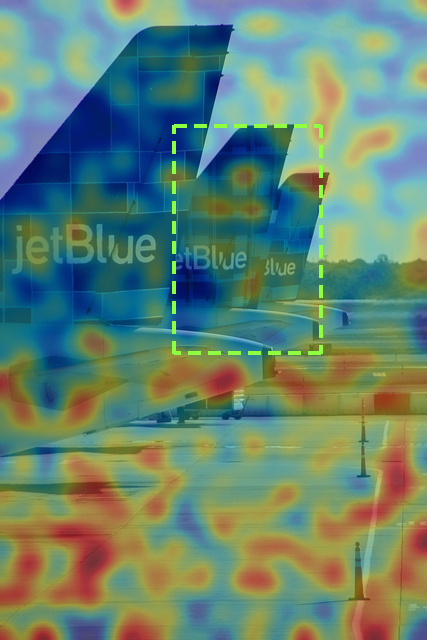}
        \caption{ADE1}
    \end{subfigure}% 
    \hspace{0.5em}
    \begin{subfigure}{0.18\textwidth} % width of the subfigure
        \centering
        \includegraphics[width=\textwidth]{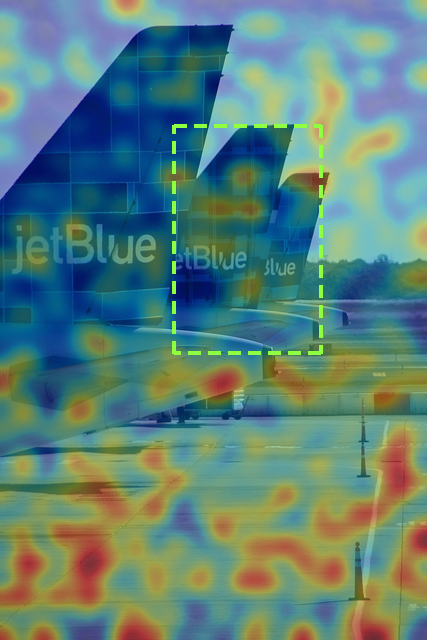}
        \caption{ADE2}
    \end{subfigure}% 
    \hspace{0.5em}
    \begin{subfigure}{0.18\textwidth} % width of the subfigure
        \centering
        \includegraphics[width=\textwidth]{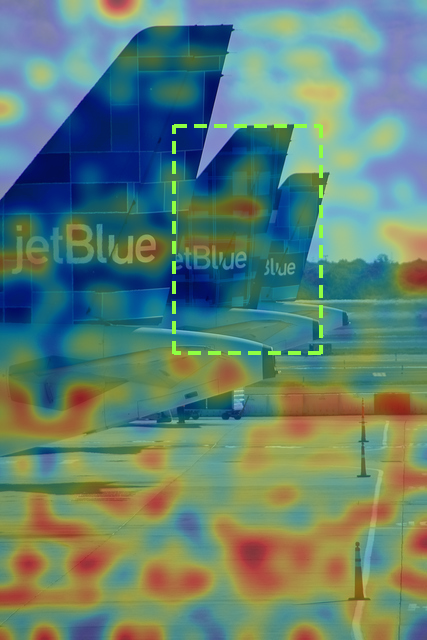}
        \caption{ADE3}
    \end{subfigure}% 
    \hspace{0.5em}
    \begin{subfigure}{0.18\textwidth} % width of the subfigure
        \centering
        \includegraphics[width=\textwidth]{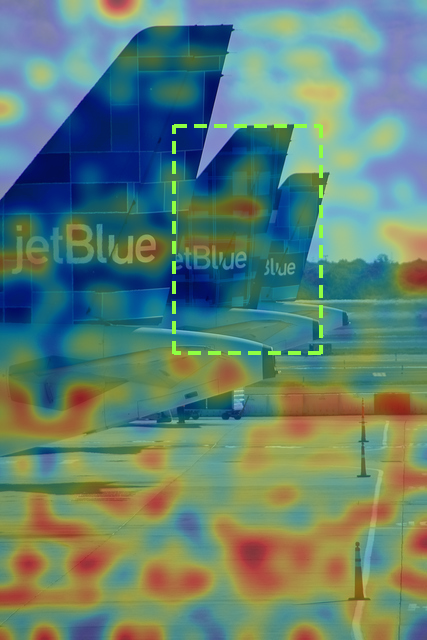}
        \caption{Baseline}
    \end{subfigure}% 
    \caption{\textbf{Impact of typographic interventions on image attention in Qwen3-VL-4B-Instruct.} We report the absolute image attention maps of the final cross attention layer for a single image via Grad-CAM.}  
    \label{fig:visual_qwen3vl_4b_it}
\end{figure*}

\begin{figure*}[!ht]
    \centering
    \begin{subfigure}{0.26\textwidth} % width of the subfigure
        \centering
        \includegraphics[width=\textwidth]{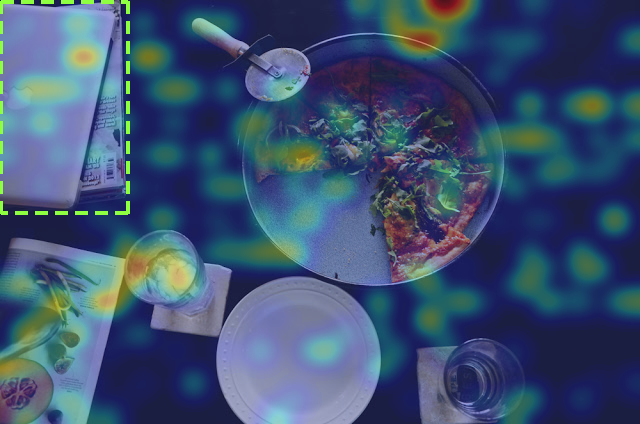}
        \caption{U1}
    \end{subfigure}% 
    \hspace{0.5em}
    \begin{subfigure}{0.26\textwidth} % width of the subfigure
        \centering
        \includegraphics[width=\textwidth]{Figures/Qwen3_Think_U2_gradcam.png}
        \caption{U2}
    \end{subfigure}% 
    \hspace{0.5em}
    \begin{subfigure}{0.26\textwidth} % width of the subfigure
        \centering
        \includegraphics[width=\textwidth]{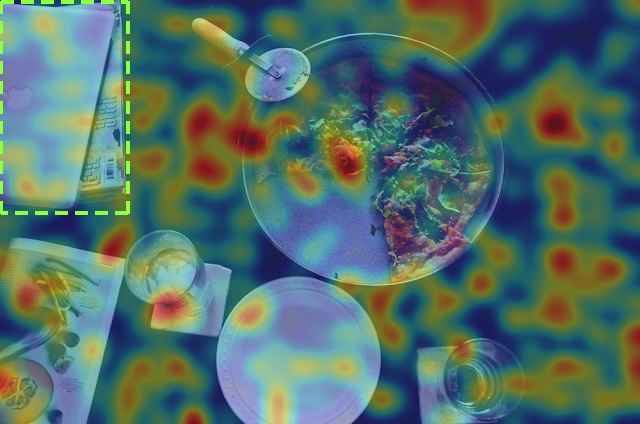}
        \caption{U3}
    \end{subfigure}% 
    \vspace{0.5em}
    \begin{subfigure}{0.26\textwidth} % width of the subfigure
        \centering
        \includegraphics[width=\textwidth]{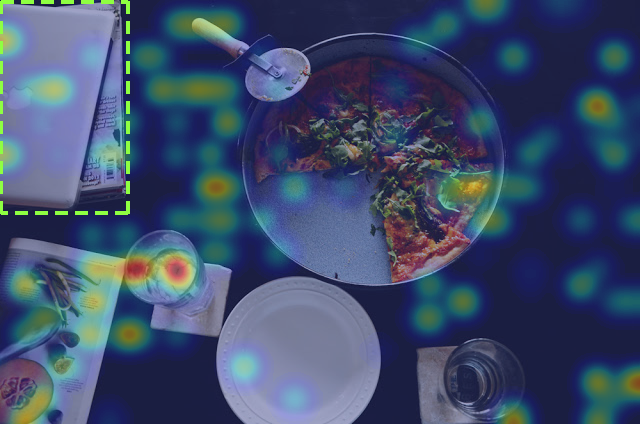}
        \caption{TE1}
    \end{subfigure}% 
    \hspace{0.5em}
    \begin{subfigure}{0.26\textwidth} % width of the subfigure
        \centering
        \includegraphics[width=\textwidth]{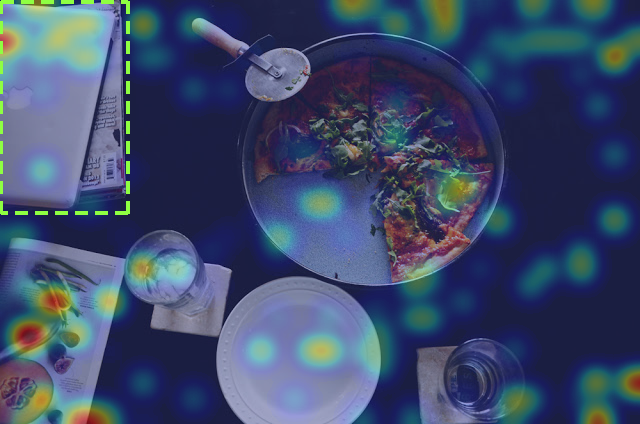}
        \caption{TE2}
    \end{subfigure}% 
    \hspace{0.5em}
    \begin{subfigure}{0.26\textwidth} % width of the subfigure
        \centering
        \includegraphics[width=\textwidth]{Figures/Qwen3_Think_TE3_gradcam.png}
        \caption{TE3}
    \end{subfigure}% 
    \vspace{0.5em}
    \begin{subfigure}{0.26\textwidth} % width of the subfigure
        \centering
        \includegraphics[width=\textwidth]{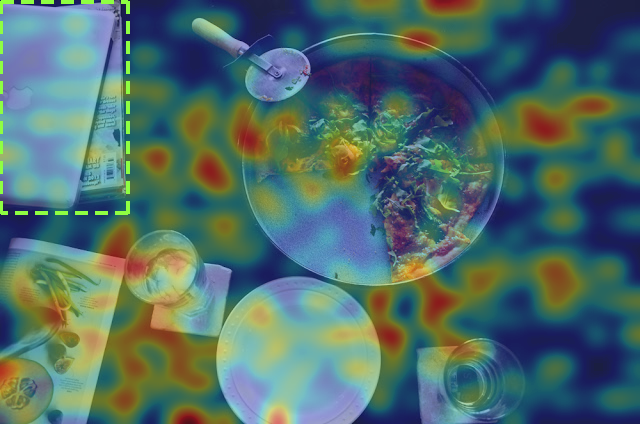}
        \caption{TT1}
    \end{subfigure}% 
    \hspace{0.5em}
    \begin{subfigure}{0.26\textwidth} % width of the subfigure
        \centering
        \includegraphics[width=\textwidth]{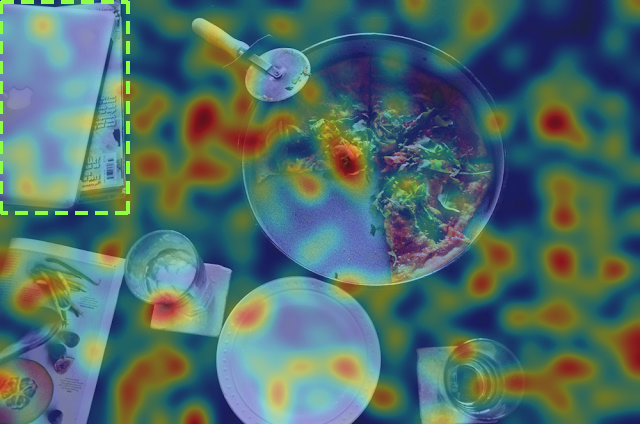}
        \caption{TT2}
    \end{subfigure}% 
    \hspace{0.5em}
    \begin{subfigure}{0.26\textwidth} % width of the subfigure
        \centering
        \includegraphics[width=\textwidth]{Figures/Qwen3_Think_TT3_gradcam.png}
        \caption{TT3}
    \end{subfigure}% 
    \vspace{0.5em}
    \begin{subfigure}{0.26\textwidth} % width of the subfigure
        \centering
        \includegraphics[width=\textwidth]{Figures/Qwen3_Think_TA1_gradcam.png}
        \caption{TA1}
    \end{subfigure}% 
    \hspace{0.5em}
    \begin{subfigure}{0.26\textwidth} % width of the subfigure
        \centering
        \includegraphics[width=\textwidth]{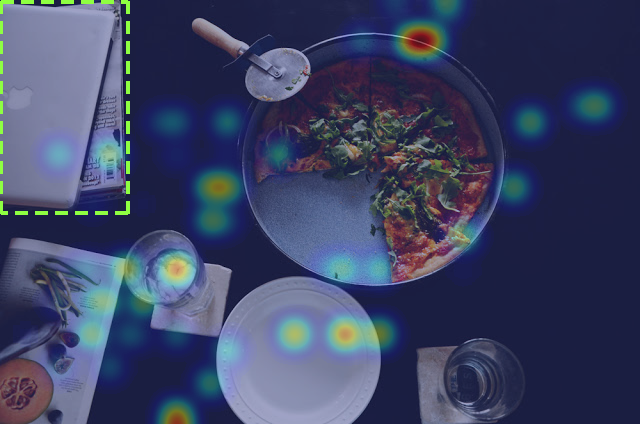}
        \caption{TA2}
    \end{subfigure}% 
    \hspace{0.5em}
    \begin{subfigure}{0.26\textwidth} % width of the subfigure
        \centering
        \includegraphics[width=\textwidth]{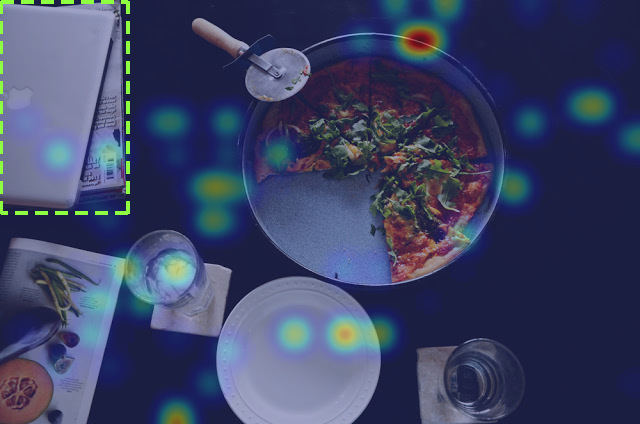}
        \caption{TA3}
    \end{subfigure}% 
    \vspace{0.5em}
    \begin{subfigure}{0.26\textwidth} % width of the subfigure
        \centering
        \includegraphics[width=\textwidth]{Figures/Qwen3_Think_ADE1_gradcam.png}
        \caption{ADE1}
    \end{subfigure}% 
    \hspace{0.5em}
    \begin{subfigure}{0.26\textwidth} % width of the subfigure
        \centering
        \includegraphics[width=\textwidth]{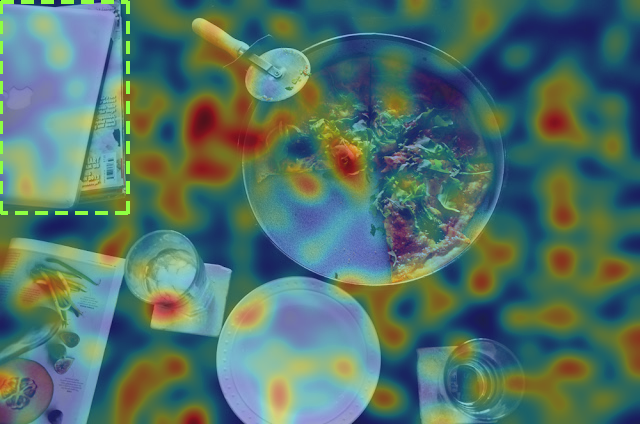}
        \caption{ADE2}
    \end{subfigure}% 
    \hspace{0.5em}
    \begin{subfigure}{0.26\textwidth} % width of the subfigure
        \centering
        \includegraphics[width=\textwidth]{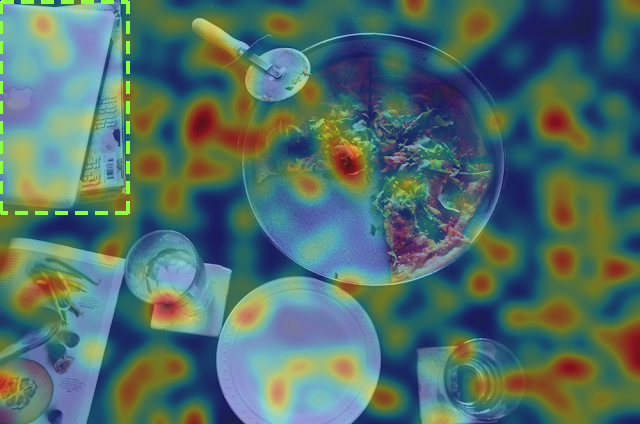}
        \caption{ADE3}
    \end{subfigure}% 
    \vspace{0.5em}
    \begin{subfigure}{0.26\textwidth} % width of the subfigure
        \centering
        \includegraphics[width=\textwidth]{Figures/Qwen3_Think_baseline_gradcam.png}
        \caption{Baseline}
    \end{subfigure}% 
    \caption{\textbf{Impact of typographic interventions on image attention in Qwen3-VL-2B-Thinking.} We report the absolute image attention maps of the final cross attention layer for a single image via Grad-CAM.}  
    \label{fig:visual_qwen3vl_2b_think}
\end{figure*}

\begin{figure*}[!ht]
    \centering
    \begin{subfigure}{0.225\textwidth} % width of the subfigure
        \centering
        \includegraphics[width=\textwidth]{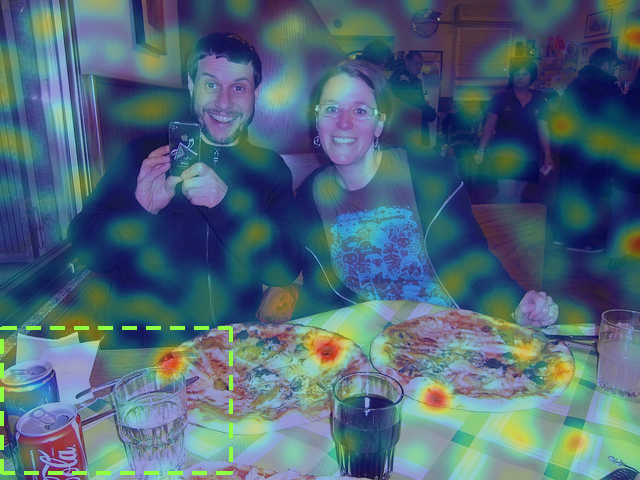}
        \caption{U1}
    \end{subfigure}% 
    \hspace{0.5em}
    \begin{subfigure}{0.225\textwidth} % width of the subfigure
        \centering
        \includegraphics[width=\textwidth]{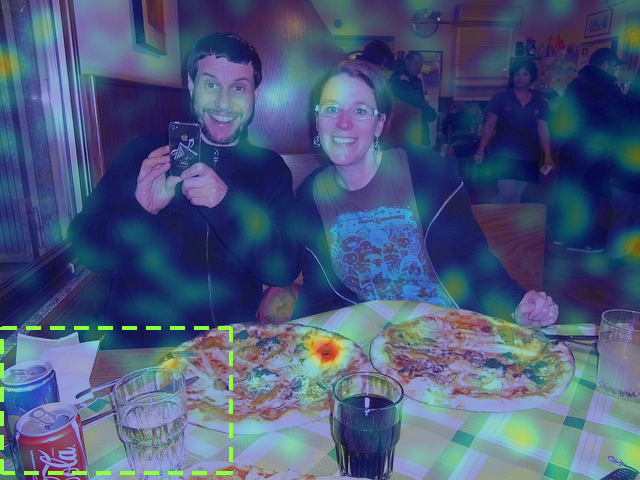}
        \caption{U2}
    \end{subfigure}% 
    \hspace{0.5em}
    \begin{subfigure}{0.225\textwidth} % width of the subfigure
        \centering
        \includegraphics[width=\textwidth]{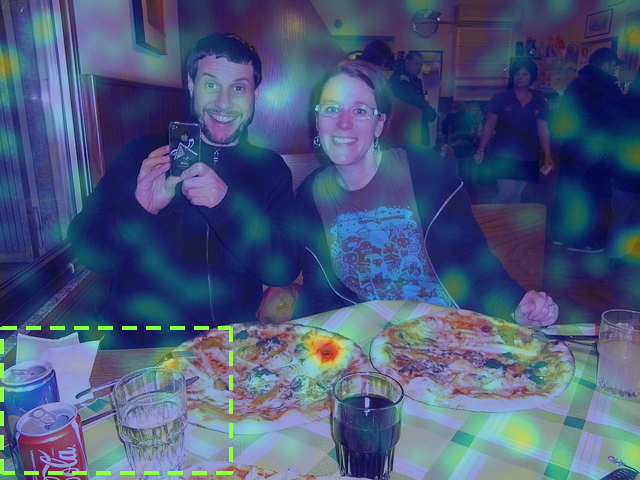}
        \caption{U3}
    \end{subfigure}% 

    \vspace{0.5em}
    \begin{subfigure}{0.225\textwidth} % width of the subfigure
        \centering
        \includegraphics[width=\textwidth]{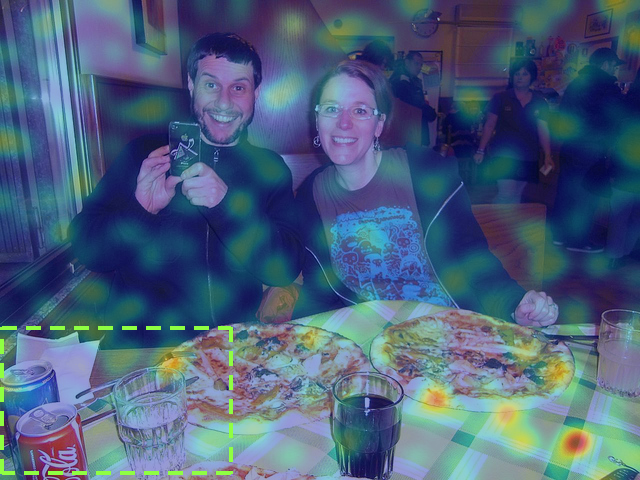}
        \caption{TE1}
    \end{subfigure}% 
    \hspace{0.5em}
    \begin{subfigure}{0.225\textwidth} % width of the subfigure
        \centering
        \includegraphics[width=\textwidth]{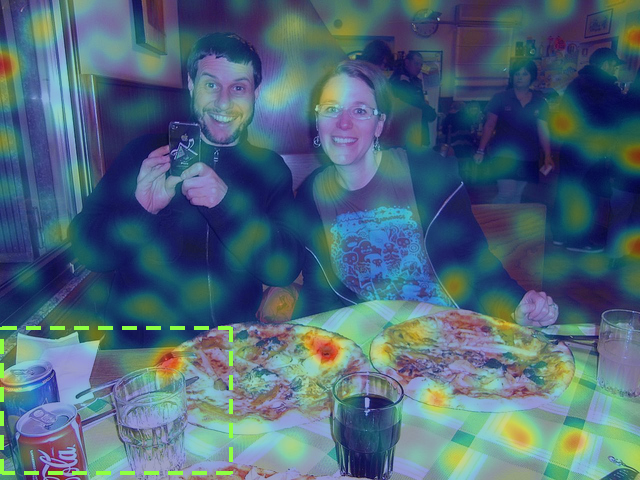}
        \caption{TE2}
    \end{subfigure}% 
    \hspace{0.5em}
    \begin{subfigure}{0.225\textwidth} % width of the subfigure
        \centering
        \includegraphics[width=\textwidth]{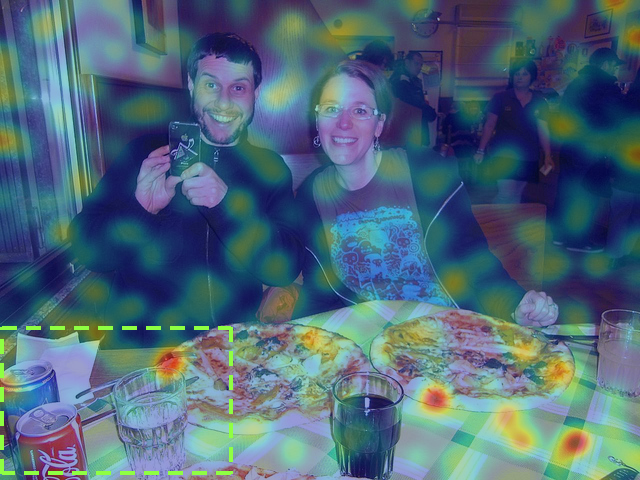}
        \caption{TE3}
    \end{subfigure}% 
    
    \vspace{0.5em}
    
    \begin{subfigure}{0.225\textwidth} % width of the subfigure
        \centering
        \includegraphics[width=\textwidth]{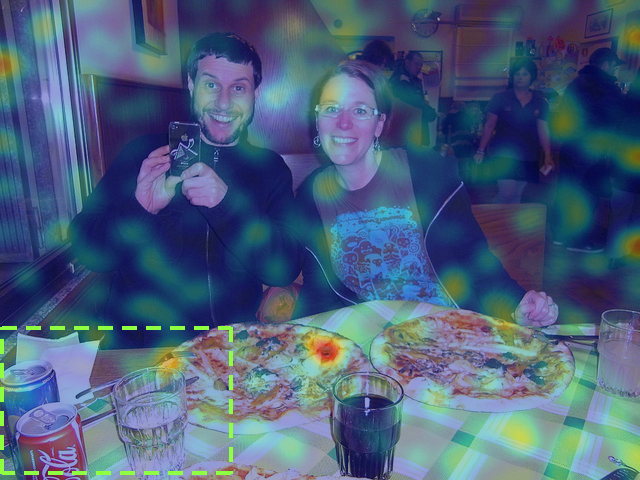}
        \caption{TT1}
    \end{subfigure}% 
    \hspace{0.5em}
    \begin{subfigure}{0.225\textwidth} % width of the subfigure
        \centering
        \includegraphics[width=\textwidth]{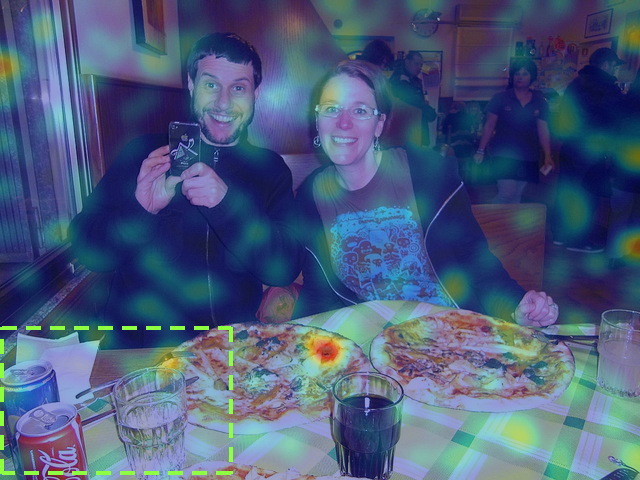}
        \caption{TT2}
    \end{subfigure}% 
    \hspace{0.5em}
    \begin{subfigure}{0.225\textwidth} % width of the subfigure
        \centering
        \includegraphics[width=\textwidth]{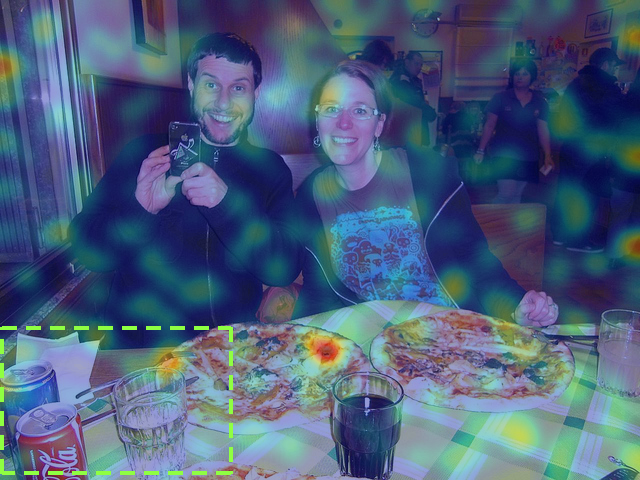}
        \caption{TT3}
    \end{subfigure}% 
    
    \vspace{0.5em}
    
    \begin{subfigure}{0.225\textwidth} % width of the subfigure
        \centering
        \includegraphics[width=\textwidth]{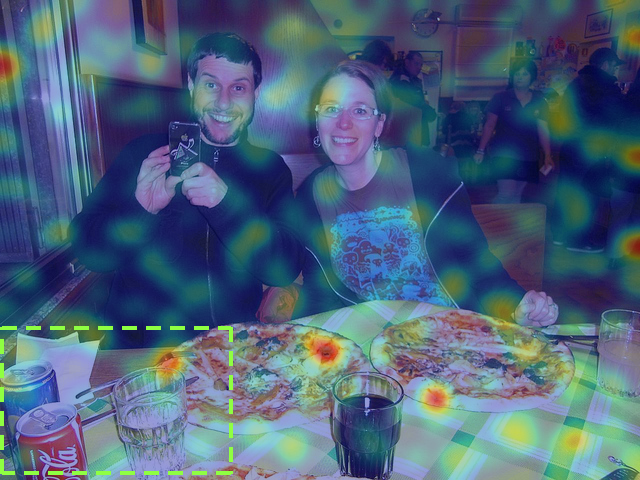}
        \caption{TA1}
    \end{subfigure}% 
    \hspace{0.5em}
    \begin{subfigure}{0.225\textwidth} % width of the subfigure
        \centering
        \includegraphics[width=\textwidth]{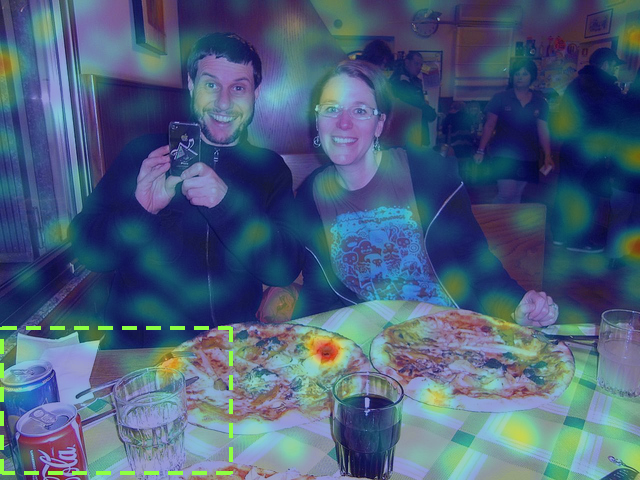}
        \caption{TA2}
    \end{subfigure}% 
    \hspace{0.5em}
    \begin{subfigure}{0.225\textwidth} % width of the subfigure
        \centering
        \includegraphics[width=\textwidth]{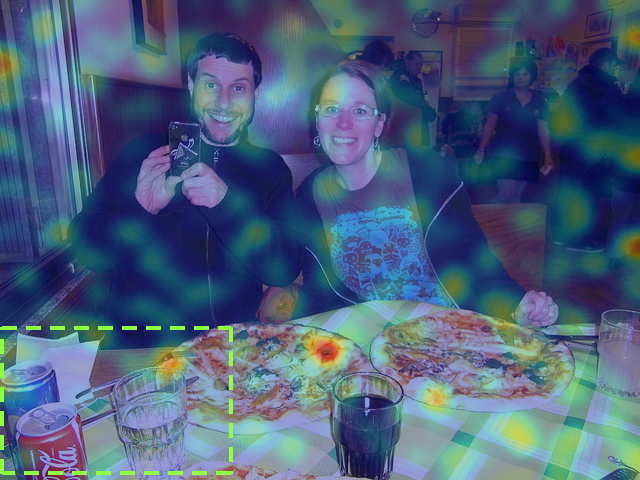}
        \caption{TA3}
    \end{subfigure}% 
    
    \vspace{0.5em}
    
    \begin{subfigure}{0.225\textwidth} % width of the subfigure
        \centering
        \includegraphics[width=\textwidth]{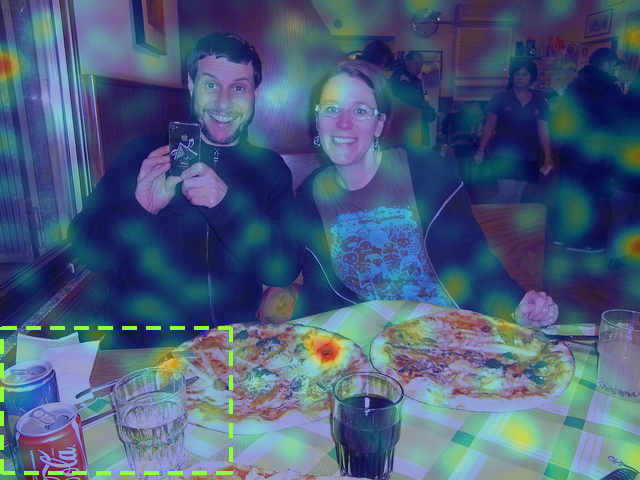}
        \caption{ADE1}
    \end{subfigure}% 
    \hspace{0.5em}
    \begin{subfigure}{0.225\textwidth} % width of the subfigure
        \centering
        \includegraphics[width=\textwidth]{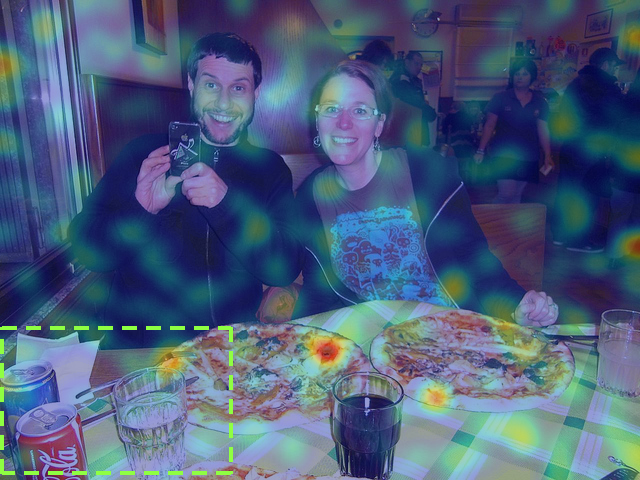}
        \caption{ADE2}
    \end{subfigure}% 
    \hspace{0.5em}
    \begin{subfigure}{0.225\textwidth} % width of the subfigure
        \centering
        \includegraphics[width=\textwidth]{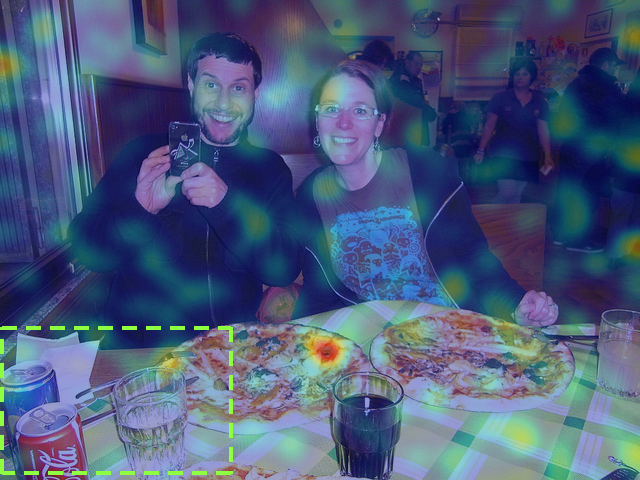}
        \caption{ADE3}
    \end{subfigure}% 
    
    \vspace{0.5em}
    
    \begin{subfigure}{0.225\textwidth} % width of the subfigure
        \centering
        \includegraphics[width=\textwidth]{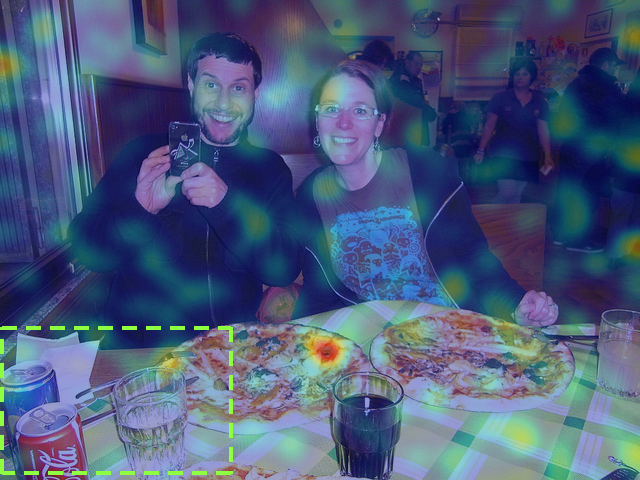}
        \caption{Baseline}
    \end{subfigure}% 
    \caption{\textbf{Impact of typographic interventions on image attention in Gemma-3-4B-IT.} We report the absolute image attention maps of the final cross attention layer for a single image via Grad-CAM.}
    \label{fig:visual_gemma3_4b}
\end{figure*}

\begin{figure*}[!ht]
    \centering
    \begin{subfigure}{0.18\textwidth} % width of the subfigure
        \centering
        \includegraphics[width=\textwidth]{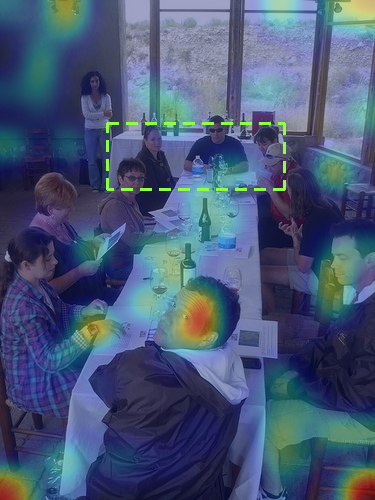}
        \caption{U1}
    \end{subfigure}% 
    \hspace{0.5em}
    \begin{subfigure}{0.18\textwidth} % width of the subfigure
        \centering
        \includegraphics[width=\textwidth]{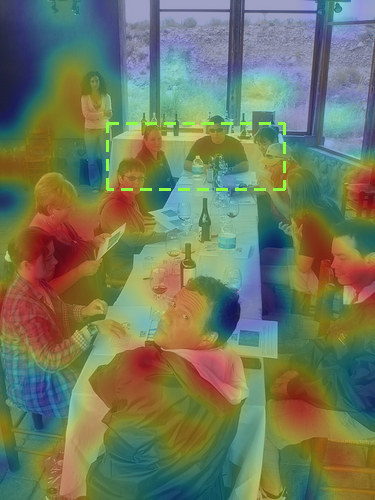}
        \caption{U2}
    \end{subfigure}% 
    \hspace{0.5em}
    \begin{subfigure}{0.18\textwidth} % width of the subfigure
        \centering
        \includegraphics[width=\textwidth]{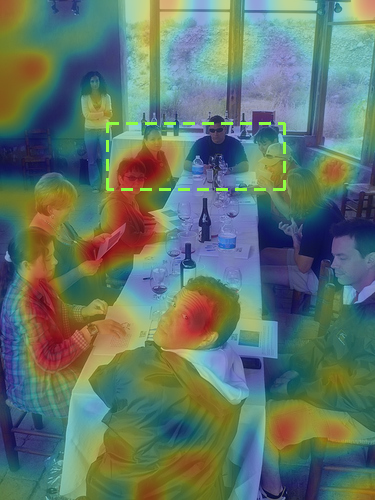}
        \caption{U3}
    \end{subfigure}% 
    \hspace{0.5em}
    \begin{subfigure}{0.18\textwidth} % width of the subfigure
        \centering
        \includegraphics[width=\textwidth]{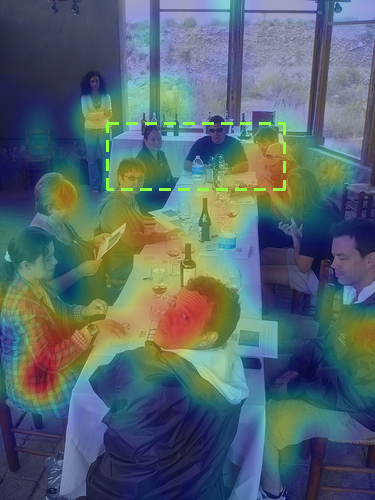}
        \caption{TE1}
    \end{subfigure}% 
    
    \vspace{0.5em}
    
    \begin{subfigure}{0.18\textwidth} % width of the subfigure
        \centering
        \includegraphics[width=\textwidth]{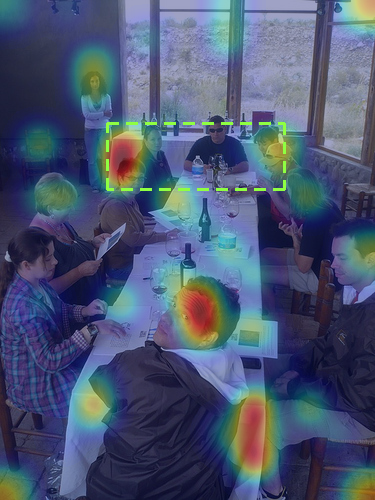}
        \caption{TE2}
    \end{subfigure}% 
    \hspace{0.5em}
    \begin{subfigure}{0.18\textwidth} % width of the subfigure
        \centering
        \includegraphics[width=\textwidth]{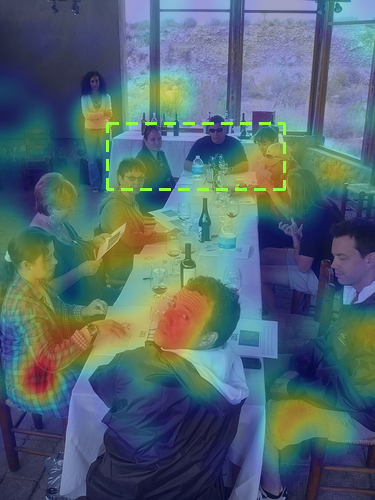}
        \caption{TE3}
    \end{subfigure}% 
    \hspace{0.5em}
    \begin{subfigure}{0.18\textwidth} % width of the subfigure
        \centering
        \includegraphics[width=\textwidth]{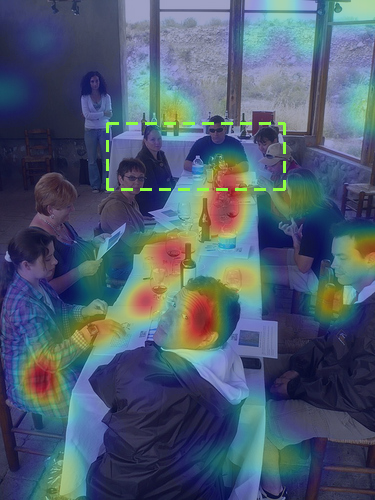}
        \caption{TT1}
    \end{subfigure}% 
    \hspace{0.5em}
    \begin{subfigure}{0.18\textwidth} % width of the subfigure
        \centering
        \includegraphics[width=\textwidth]{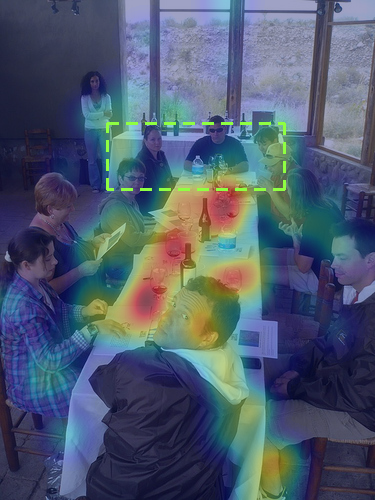}
        \caption{TT2}
    \end{subfigure}% 
    
    \vspace{0.5em}
    
    \begin{subfigure}{0.18\textwidth} % width of the subfigure
        \centering
        \includegraphics[width=\textwidth]{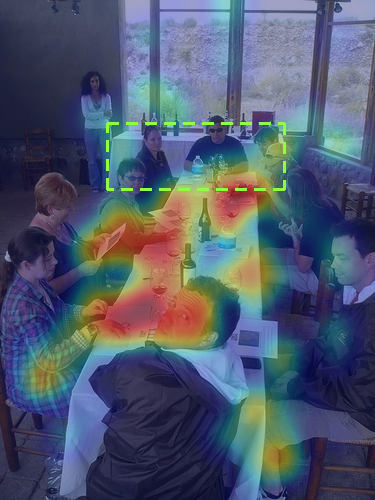}
        \caption{TT3}
    \end{subfigure}% 
    \hspace{0.5em}
    \begin{subfigure}{0.18\textwidth} % width of the subfigure
        \centering
        \includegraphics[width=\textwidth]{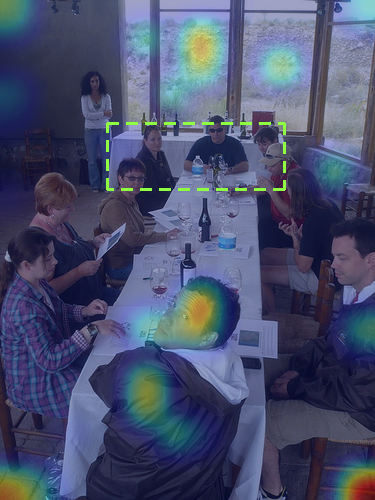}
        \caption{TA1}
    \end{subfigure}% 
    \hspace{0.5em}
    \begin{subfigure}{0.18\textwidth} % width of the subfigure
        \centering
        \includegraphics[width=\textwidth]{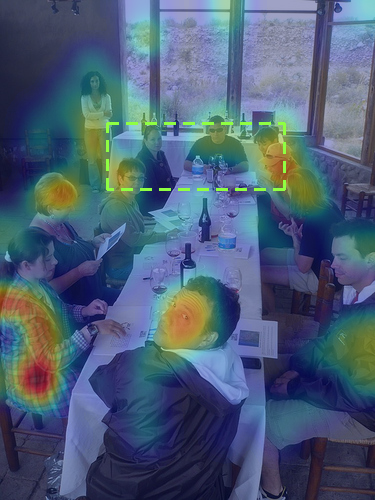}
        \caption{TA2}
    \end{subfigure}% 
    \hspace{0.5em}
    \begin{subfigure}{0.18\textwidth} % width of the subfigure
        \centering
        \includegraphics[width=\textwidth]{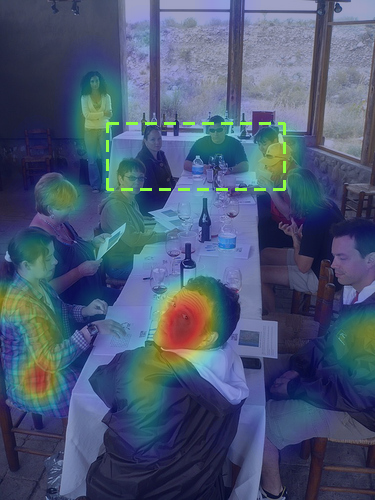}
        \caption{TA3}
    \end{subfigure}% 
    
    \vspace{0.5em}
    
    \begin{subfigure}{0.18\textwidth} % width of the subfigure
        \centering
        \includegraphics[width=\textwidth]{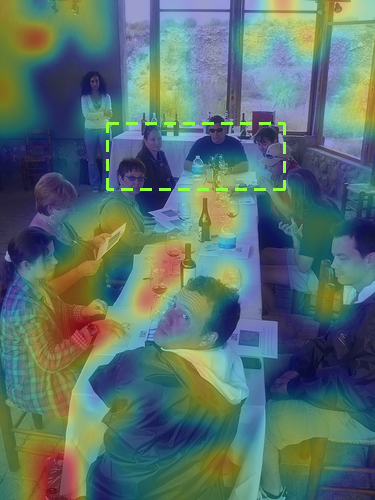}
        \caption{ADE1}
    \end{subfigure}% 
    \hspace{0.5em}
    \begin{subfigure}{0.18\textwidth} % width of the subfigure
        \centering
        \includegraphics[width=\textwidth]{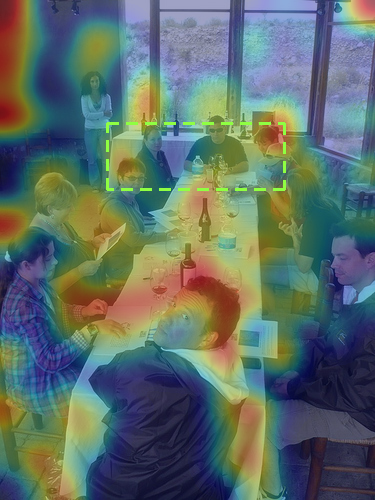}
        \caption{ADE2}
    \end{subfigure}% 
    \hspace{0.5em}
    \begin{subfigure}{0.18\textwidth} % width of the subfigure
        \centering
        \includegraphics[width=\textwidth]{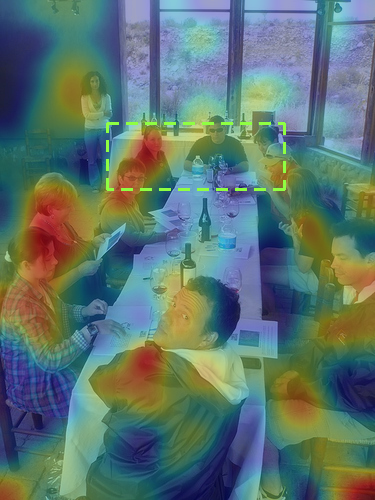}
        \caption{ADE3}
    \end{subfigure}% 
    \hspace{0.5em}
    \begin{subfigure}{0.18\textwidth} % width of the subfigure
        \centering
        \includegraphics[width=\textwidth]{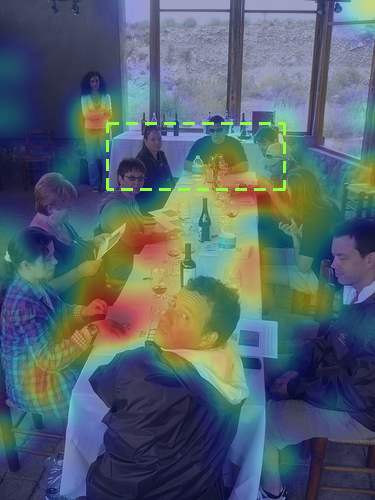}
        \caption{Baseline}
    \end{subfigure}% 
    \caption{\textbf{Impact of typographic interventions on image attention in Gemma-4-E4B-IT.} We report the absolute image attention maps of the final cross attention layer for a single image via Grad-CAM.}
    \label{fig:visual_gemma4_e4b}
\end{figure*}

In this section, we provide detailed tabular printouts compiling the performance metrics across our vision-language evaluation pipeline on the RefCOCOg benchmark~\cite{Mao_2016_RefCOCO}. The results below substantiate the dual-action mechanism of \Cref{sec:exp:vlms} and make explicit that the multimodal effect is a \emph{partial} transfer of the text-only phenomenon: the family-level ordering of attractors recurs, but along two distinct axes (local spatial concentration and global modality re-allocation) whose relative strength is model-dependent.
 Qualitative examples can be seen in \Cref{fig:visual_qwen3vl_4b_it,fig:visual_qwen3vl_2b_think,fig:visual_gemma3_4b,fig:visual_gemma4_e4b}.

\paragraph{Methodological Setup and Attention Extraction}
To quantify attention routing within our evaluated VLM architectures, we extract attention weight matrices directly from the cross-attention blocks responsible for fusing visual token embeddings with textual token embeddings. For each sample, we construct an spatial attention mask matching the ground-truth bounding box coordinates of the target object. 
Let $A_{i,j}$ represent the attention weight assigned from textual query token $i$ to visual patch token $j$. The spatial attention mass allocated to the target bounding box is calculated by summing weights over all patches falling inside the target mask boundaries, averaged across all attention heads and multi-modal alignment layers. Baseline evaluations are conducted using standard sentence-case templates, against which our 15 typographic casing schemes are systematically applied.

\paragraph{On the microscopic--macroscopic coupling.}
Because microscopic steering is measured as the share of a \emph{re-normalized} visual budget falling inside the target box, part of the observed concentration is a mechanical consequence of the simultaneous macroscopic contraction of whole-image attention: as total visual mass shrinks, a fixed-size target region can occupy a larger fraction even without genuine sharpening of spatial focus. We therefore report the macroscopic (absolute $\Delta$pp whole-image) and microscopic (in-box $\Delta$pp) axes \emph{separately} rather than as a single ratio, so the two effects are not conflated. The qualitative maps in \Cref{fig:visual_qwen3vl_4b_it,fig:visual_qwen3vl_2b_think,fig:visual_gemma3_4b,fig:visual_gemma4_e4b} corroborate that the residual attention is spatially re-organized and not merely uniformly scaled.

\paragraph{Granular model evaluations.}
To substantiate the distinction between direct-inference and reasoning VLMs,
\Cref{tab:appendix_micro_spatial_steering,tab:appendix_macro_modality_shifts} report the full per-model
microscopic ($\Delta$pp) and macroscopic ($\Delta\%$) tracking matrices, and \Cref{fig:visual_qwen3vl_4b_it,fig:visual_qwen3vl_2b_think,fig:visual_gemma3_4b,fig:visual_gemma4_e4b} visualise the corresponding final-layer image-attention maps for each model on exemplary images. The granular view exposes the per-model heterogeneity that the cross-model aggregates conceal: direct-inference models exhibit the sharpest \emph{spatial} variance (Gemma-3-4B-IT peaking at $+6.23$ pp under TA1), whereas multi-step reasoning models exhibit the largest and most persistent \emph{macroscopic} re-allocation toward text (Qwen3-VL-2B-Thinking maintaining double-digit visual disengagement across all high-entropy interventions). On the Qwen3-VL pair, the microscopic ordering departs markedly from the family hierarchy---uniform (U1), title (TT1), and de-emphasis (ADE1) schemes are comparable to TA1---underscoring that the cross-modal hierarchy is reliable only in aggregate and along the macroscopic axis.

\subsection{Attention Extraction and Normalization Protocol}
\label{sec:attention_extraction}

To quantify the ``casing effect,'' we extract the model's internal attention distribution during the prefill forward pass, prior to the generation of the first token. This captures the model's initial encoding of the relationship between the target span and the context. Our extraction protocol follows three steps to ensure that the measured attention mass is isolated to the semantic content of the input:

\paragraph{Head and Layer Aggregation.} 
While attention patterns vary across transformer blocks, we focus our analysis on the \textbf{final attention layer}, averaged across all attention heads. The final layer serves as the most representative proxy for the information bottleneck immediately preceding the prediction head. For each sample, we compute the mean attention weight $\alpha_{i,j}$ for all tokens $i$ attending to tokens $j$, providing a comprehensive map of the model's internal focus.

\paragraph{Token-Level Masking and Filtering.}
To isolate the pure effect of the typographic intervention, we apply a multi-stage masking process to remove \textit{non-semantic} attentional sinks:
\begin{itemize}
    \item \textbf{System and Special Tokens}: We exclude attention weights associated with system prompts, and special control tokens (\eg, \texttt{<|BOS|>}, \texttt{<|EOS|>}, \texttt{<|PAD|>}). System prompt removal is performed via character-span matching to ensure only the user-provided context and target remain.
    \item \textbf{Softmax Re-normalization}: After masking these tokens, we re-normalize the remaining attention weights so they sum to unity. This ensures that the calculated Attention Mass reflects the relative distribution of focus strictly within the lexical input, preventing the \textit{sink effect} of special tokens~\cite{Li_2025_arXiv_CTR_SinkTokens} from skewing the results.
\end{itemize}

\paragraph{Span-to-Token Mapping.}
Because typographic changes can alter tokenization boundaries (\cf App.~\ref{appendix:tokenization_analysis}), we map the character-level ground-truth span $A$ to its corresponding token indices using the model's specific offsets. The attention mass is then calculated as the sum of all weights directed toward the tokens comprising $A$. By using this normalized, filtered, and aggregated metric, we ensure that our measurements of \textit{attention concentration} are robust to variations in sequence length and architectural scale.

\clearpage \newpage

\clearpage \newpage

\clearpage \newpage

\clearpage \newpage
\begin{figure*}[!ht]
    \centering
    \begin{subfigure}{0.425\textwidth} % width of the subfigure
        \centering
        \includegraphics[width=\textwidth]{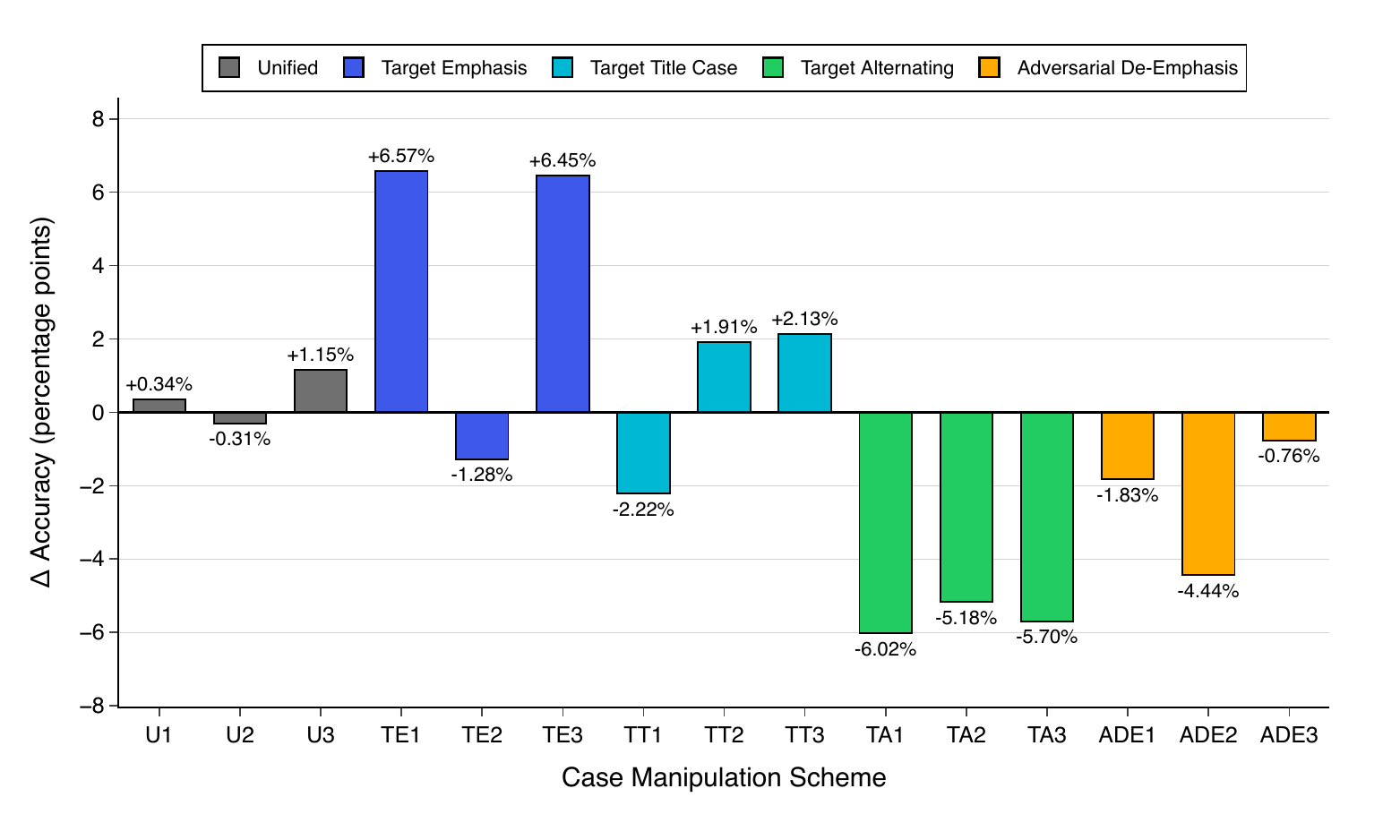}
        \caption{LLaMA-3.2-3B-Instruct}
    \end{subfigure}% 
    \hspace{2em}
    \begin{subfigure}{0.425\textwidth} % width of the subfigure
        \centering
        \includegraphics[width=\textwidth]{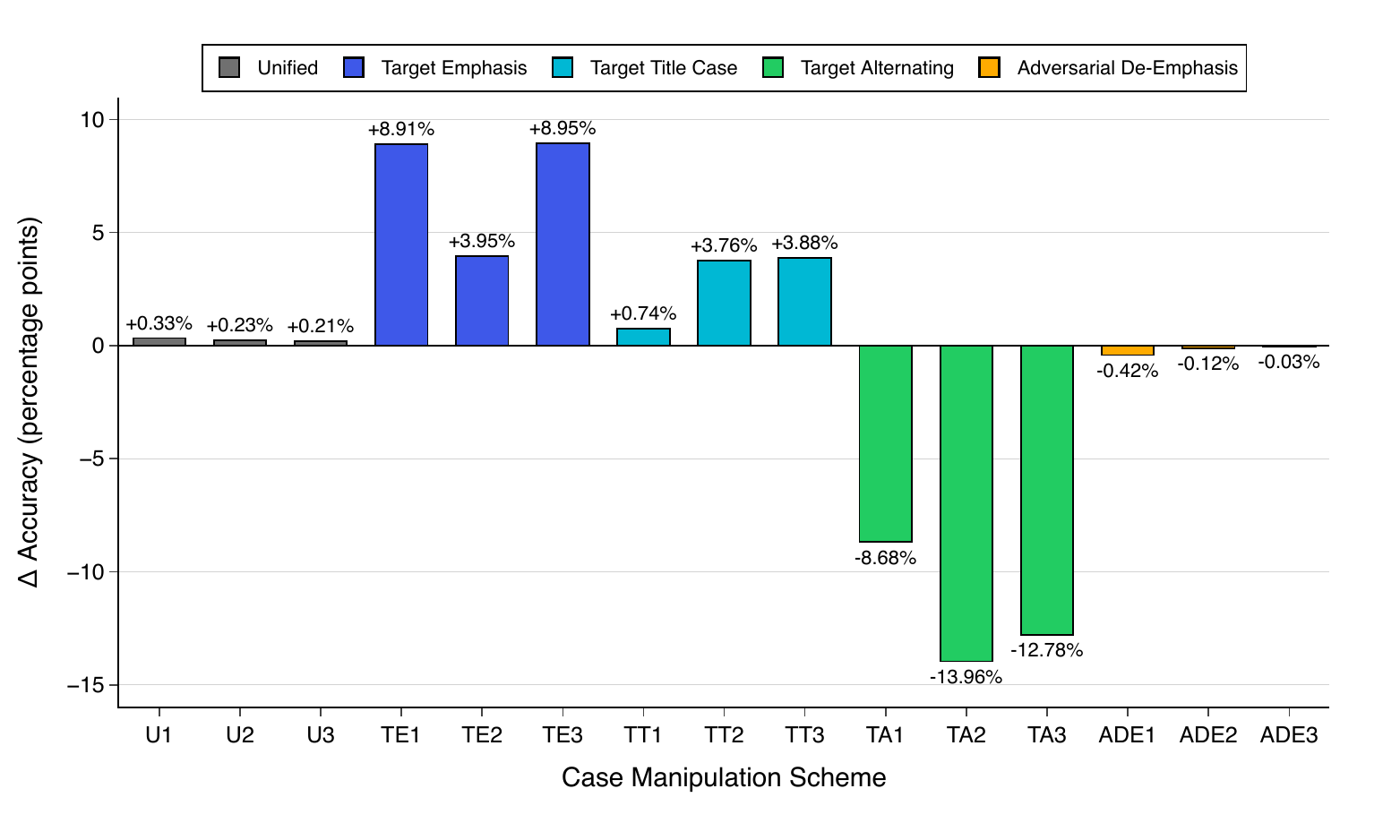}
        \caption{LLaMA-3.1-8B-Instruct}
    \end{subfigure}% 
    \vspace{0.5em}
    \begin{subfigure}{0.425\textwidth} % width of the subfigure
        \centering
        \includegraphics[width=\textwidth]{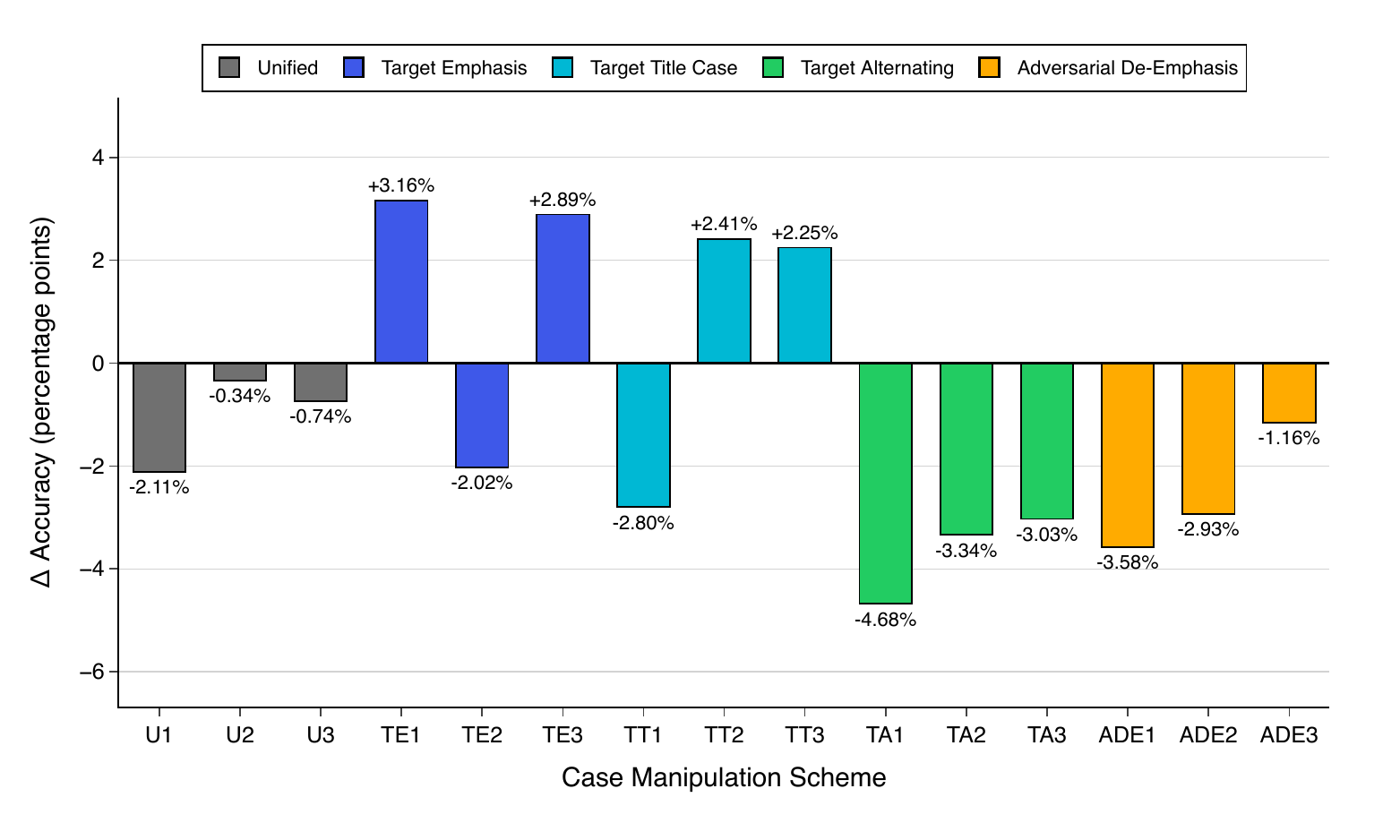}
        \caption{Gemma-3-1B-IT}
    \end{subfigure}% 
    \hspace{2em}
    \begin{subfigure}{0.425\textwidth} % width of the subfigure
        \centering
        \includegraphics[width=\textwidth]{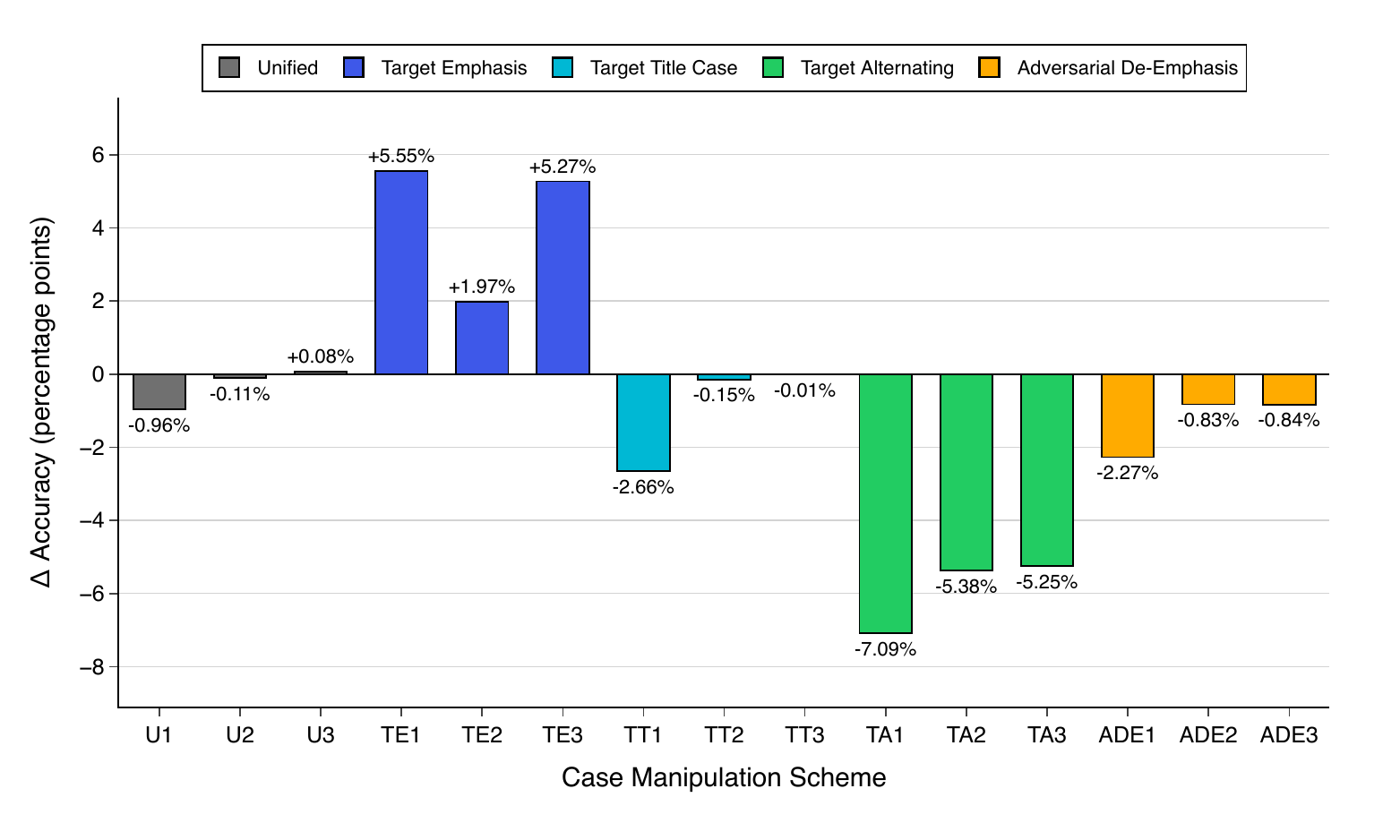}
        \caption{Gemma-2-2B-IT}
    \end{subfigure}% 
    \vspace{0.5em}
    \begin{subfigure}{0.425\textwidth} % width of the subfigure
        \centering
        \includegraphics[width=\textwidth]{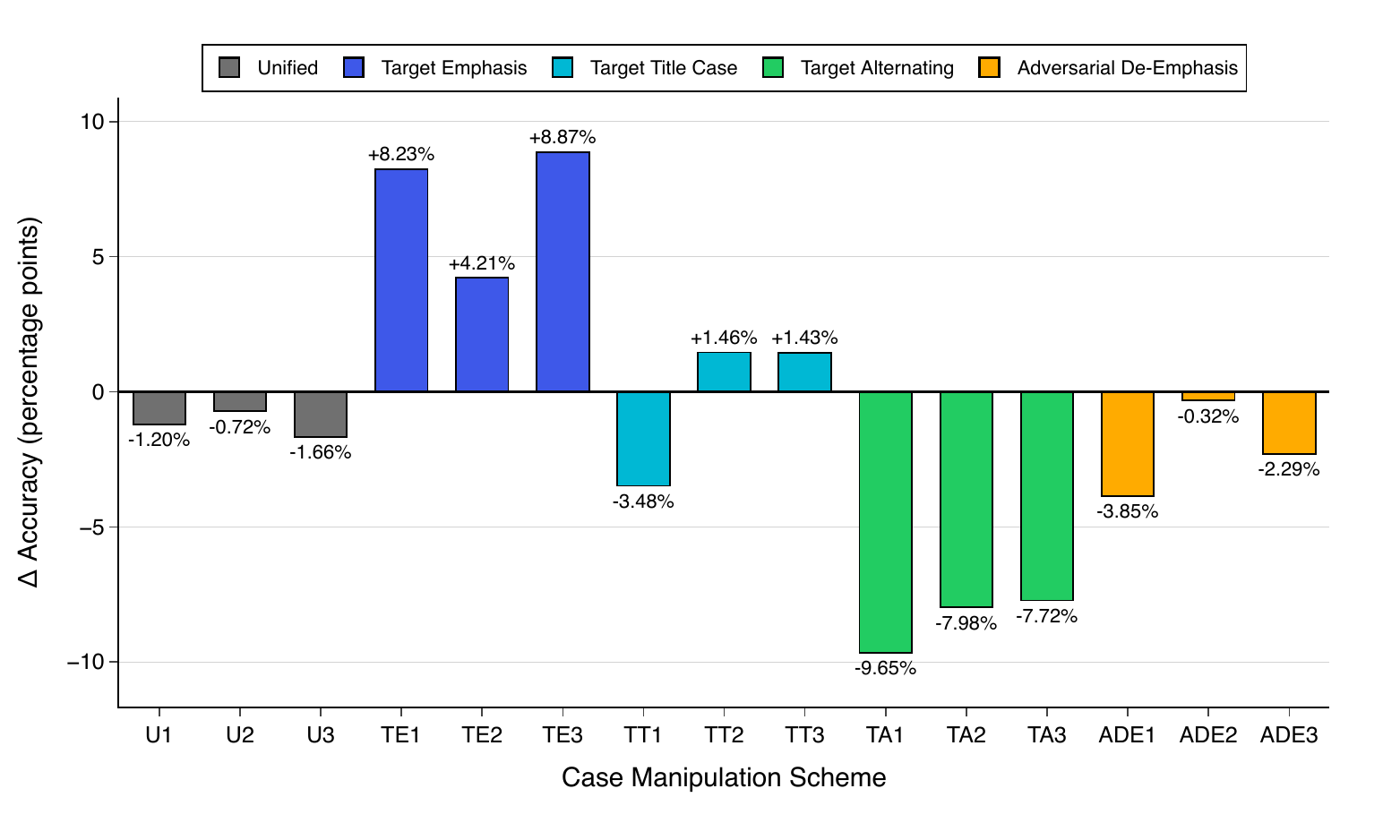}
        \caption{Mistral-7B-Instruct-v0.3}
    \end{subfigure}% 
    \hspace{2em}
    \begin{subfigure}{0.425\textwidth} % width of the subfigure
        \centering
        \includegraphics[width=\textwidth]{Figures/case_steering_accuracy_Qwen_7b.pdf}
        \caption{Qwen2.5-7B-Instruct}
    \end{subfigure}% 
    \vspace{0.5em}
    \begin{subfigure}{0.425\textwidth} % width of the subfigure
        \centering
        \includegraphics[width=\textwidth]{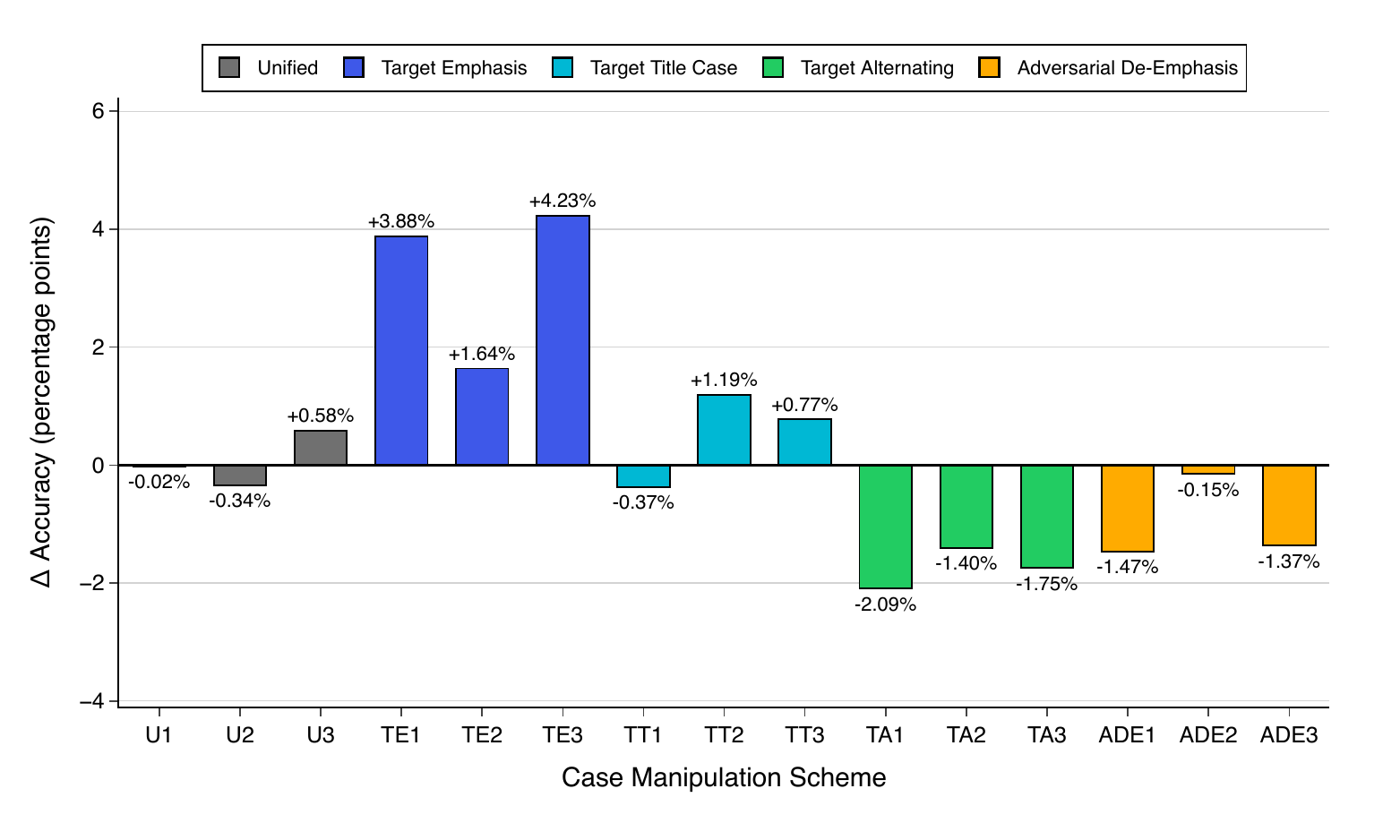}
        \caption{Qwen2.5-14B-Instruct}
    \end{subfigure}% 
    \caption{\textbf{Impact of typographic interventions on downstream task accuracy.} We report the relative change in mean accuracy across three benchmarks compared to naturally cased baselines.}    
    \label{fig:accuracy_per_scheme}
\end{figure*}

\clearpage \newpage

\newpage \clearpage

\newpage \clearpage

\newpage \clearpage

\newpage \clearpage
\begin{table*}[!ht]
    \caption{\textbf{Impact of letter case schemes on performance and target attention allocation for the LLaMA-3.2-3B-Instruct baseline.} Task metrics (Acc., EM, F1) and attention mass (AM) are reported across three benchmarks.}
    \centering
    \small
    \setlength{\tabcolsep}{4.5pt}
    \begin{tabular}{lccccccc}
    \toprule
    \multirow{2}{*}{\textbf{Method}}
    & \multicolumn{2}{c}{\textbf{MMLU-Pro}}
    & \multicolumn{2}{c}{\textbf{ARC}}
    & \multicolumn{3}{c}{\textbf{SQuADv2}} \\
    \cmidrule(lr){2-3} \cmidrule(lr){4-5} \cmidrule(lr){6-8}
    & Acc. $\uparrow$ & AM $\uparrow$
    & Acc. $\uparrow$ & AM $\uparrow$
    & EM $\uparrow$ & F1 $\uparrow$ & AM $\uparrow$ \\
    \midrule
    LLaMA-3.2-3B~\cite{llama3.2_3b_instruct}
        & 21.53 & 0.028 & 71.79 & 0.043 & 8.47 & 38.32 & 0.009 \\
    \addlinespace
    \quad + U1
        & 21.70 & 0.029 & 70.94 & 0.047 & 7.63 & 40.04 & 0.014 \\
    \quad + U2
        & 21.45 & 0.028 & 71.79 & 0.043 & 8.47 & 37.49 & 0.009 \\
    \quad + U3
        & 21.70 & 0.028 & 72.65 & 0.045 & 8.47 & 40.76 & 0.010 \\
    \quad + TE1
        & 29.51 & 0.036 & 82.05 & 0.054 & 8.47 & 39.80 & 0.014 \\
    \quad + TE2
        & 22.94 & 0.031 & 64.96 & 0.050 & 7.63 & 39.90 & 0.013 \\
    \quad + TE3
        & 30.01 & 0.036 & 81.20 & 0.054 & 8.47 & 39.80 & 0.014 \\
    \quad + TT1
        & 20.28 & 0.029 & 66.67 & 0.046 & 7.63 & 38.05 & 0.010 \\
    \quad + TT2
        & 22.11 & 0.032 & 76.07 & 0.049 & 8.47 & 39.19 & 0.010 \\
    \quad + TT3
        & 22.78 & 0.031 & 76.07 & 0.049 & 8.47 & 39.19 & 0.010 \\
    \quad + TA1
        & 16.96 & 0.039 & 58.12 & 0.065 & 7.63 & 38.50 & 0.019 \\
    \quad + TA2
        & 18.45 & 0.044 & 58.97 & 0.071 & 8.47 & 38.69 & 0.019 \\
    \quad + TA3
        & 18.45 & 0.044 & 58.12 & 0.071 & 7.63 & 37.96 & 0.019 \\
    \quad + ADE1
        & 20.78 & 0.028 & 66.67 & 0.043 & 8.47 & 38.71 & 0.008 \\
    \quad + ADE2
        & 21.03 & 0.028 & 58.97 & 0.043 & 8.47 & 38.32 & 0.009 \\
    \quad + ADE3
        & 20.86 & 0.028 & 70.94 & 0.043 & 8.47 & 37.56 & 0.009 \\
    \bottomrule
    \end{tabular}
    \label{tab:schemes_absolutevals_llama_3b}

    \vspace{20pt}
    
    \caption{\textbf{Impact of letter case schemes on performance and target attention allocation for the LLaMA-3.1-8B-Instruct baseline.} Task metrics (Acc., EM, F1) and attention mass (AM) are reported across three benchmarks.}
    \centering
    \small
    \setlength{\tabcolsep}{4.5pt}
    \begin{tabular}{lccccccc}
    \toprule
    \multirow{2}{*}{\textbf{Method}}
    & \multicolumn{2}{c}{\textbf{MMLU-Pro}}
    & \multicolumn{2}{c}{\textbf{ARC}}
    & \multicolumn{3}{c}{\textbf{SQuADv2}} \\
    \cmidrule(lr){2-3} \cmidrule(lr){4-5} \cmidrule(lr){6-8}
    & Acc. $\uparrow$ & AM $\uparrow$
    & Acc. $\uparrow$ & AM $\uparrow$
    & EM $\uparrow$ & F1 $\uparrow$ & AM $\uparrow$ \\
    \midrule
    LLaMA-3.1-8B~\cite{llama3.1_8b_instruct}
        & 29.59 & 0.017 & 72.65 & 0.034 & 29.66 & 63.19 & 0.003 \\
    \addlinespace
    \quad + U1
        & 28.10 & 0.018 & 75.21 & 0.035 & 28.81 & 63.11 & 0.007 \\
    \quad + U2
        & 29.43 & 0.017 & 73.50 & 0.034 & 29.66 & 63.19 & 0.004 \\
    \quad + U3
        & 28.43 & 0.018 & 73.50 & 0.035 & 29.66 & 64.12 & 0.004 \\
    \quad + TE1
        & 44.39 & 0.022 & 83.76 & 0.044 & 29.66 & 64.02 & 0.007 \\
    \quad + TE2
        & 34.41 & 0.019 & 79.49 & 0.037 & 28.81 & 63.38 & 0.007 \\
    \quad + TE3
        & 44.22 & 0.022 & 83.76 & 0.044 & 29.66 & 64.29 & 0.007 \\
    \quad + TT1
        & 30.09 & 0.018 & 73.50 & 0.036 & 28.81 & 64.05 & 0.004 \\
    \quad + TT2
        & 34.41 & 0.020 & 76.92 & 0.039 & 30.51 & 65.38 & 0.005 \\
    \quad + TT3
        & 33.92 & 0.020 & 77.78 & 0.039 & 30.51 & 65.38 & 0.005 \\
    \quad + TA1
        & 22.44 & 0.025 & 53.85 & 0.048 & 29.66 & 63.08 & 0.008 \\
    \quad + TA2
        & 21.11 & 0.027 & 41.88 & 0.053 & 27.97 & 60.55 & 0.008 \\
    \quad + TA3
        & 20.37 & 0.027 & 43.59 & 0.053 & 30.51 & 63.13 & 0.008 \\
    \quad + ADE1
        & 29.01 & 0.017 & 71.79 & 0.033 & 29.66 & 63.36 & 0.004 \\
    \quad + ADE2
        & 28.01 & 0.017 & 74.36 & 0.032 & 30.51 & 62.71 & 0.004 \\
    \quad + ADE3
        & 29.93 & 0.017 & 72.65 & 0.034 & 29.66 & 62.75 & 0.004 \\
    \bottomrule
    \end{tabular}
    \label{tab:schemes_absolutevals_llama_8b}
\end{table*}

\newpage \clearpage
\begin{table*}[!ht]
    \caption{\textbf{Impact of letter case schemes on performance and target attention allocation for the Gemma-3-1B-IT baseline.} Task metrics (Acc., EM, F1) and attention mass (AM) are reported across three benchmarks.}
    \centering
    \small
    \setlength{\tabcolsep}{4.5pt}
    \begin{tabular}{lccccccc}
    \toprule
    \multirow{2}{*}{\textbf{Method}}
    & \multicolumn{2}{c}{\textbf{MMLU-Pro}}
    & \multicolumn{2}{c}{\textbf{ARC}}
    & \multicolumn{3}{c}{\textbf{SQuADv2}} \\
    \cmidrule(lr){2-3} \cmidrule(lr){4-5} \cmidrule(lr){6-8}
    & Acc. $\uparrow$ & AM $\uparrow$
    & Acc. $\uparrow$ & AM $\uparrow$
    & EM $\uparrow$ & F1 $\uparrow$ & AM $\uparrow$ \\
    \midrule
    Gemma-3-1B-IT~\cite{gemma_3_1b_it}
        & 14.46 & 0.012 & 47.86 & 0.027 & 7.63 & 26.88 & 0.002 \\
    \addlinespace
    \quad + U1
        & 13.30 & 0.011 & 41.88 & 0.028 & 9.32 & 27.70 & 0.007 \\
    \quad + U2
        & 14.05 & 0.012 & 48.72 & 0.026 & 7.63 & 25.43 & 0.002 \\
    \quad + U3
        & 14.71 & 0.012 & 45.30 & 0.028 & 7.63 & 26.98 & 0.003 \\
    \quad + TE1
        & 16.96 & 0.013 & 52.99 & 0.030 & 10.17 & 28.73 & 0.008 \\
    \quad + TE2
        & 11.80 & 0.010 & 44.44 & 0.026 & 10.17 & 26.89 & 0.007 \\
    \quad + TE3
        & 17.29 & 0.014 & 52.99 & 0.030 & 9.32 & 27.60 & 0.008 \\
    \quad + TT1
        & 14.05 & 0.012 & 41.88 & 0.029 & 8.47 & 24.88 & 0.003 \\
    \quad + TT2
        & 16.29 & 0.013 & 52.99 & 0.030 & 8.47 & 27.15 & 0.003 \\
    \quad + TT3
        & 15.79 & 0.013 & 52.99 & 0.030 & 8.47 & 27.16 & 0.003 \\
    \quad + TA1
        & 13.88 & 0.016 & 35.04 & 0.039 & 10.17 & 26.25 & 0.006 \\
    \quad + TA2
        & 17.21 & 0.019 & 36.75 & 0.042 & 7.63 & 25.23 & 0.006 \\
    \quad + TA3
        & 17.12 & 0.019 & 36.75 & 0.042 & 8.47 & 26.24 & 0.006 \\
    \quad + ADE1
        & 12.88 & 0.011 & 39.32 & 0.028 & 7.63 & 26.27 & 0.002 \\
    \quad + ADE2
        & 11.64 & 0.010 & 43.59 & 0.026 & 7.63 & 25.19 & 0.002 \\
    \quad + ADE3
        & 13.97 & 0.012 & 46.15 & 0.027 & 7.63 & 25.62 & 0.002 \\
    \bottomrule
    \end{tabular}
    \label{tab:schemes_absolutevals_gemma_1b}

    \vspace{20pt}
    
    \caption{\textbf{Impact of letter case schemes on performance and target attention allocation for the Gemma-2-2B-IT baseline.} Task metrics (Acc., EM, F1) and attention mass (AM) are reported across three benchmarks.}
    \centering
    \small
    \setlength{\tabcolsep}{4.5pt}
    \begin{tabular}{lccccccc}
    \toprule
    \multirow{2}{*}{\textbf{Method}}
    & \multicolumn{2}{c}{\textbf{MMLU-Pro}}
    & \multicolumn{2}{c}{\textbf{ARC}}
    & \multicolumn{3}{c}{\textbf{SQuADv2}} \\
    \cmidrule(lr){2-3} \cmidrule(lr){4-5} \cmidrule(lr){6-8}
    & Acc. $\uparrow$ & AM $\uparrow$
    & Acc. $\uparrow$ & AM $\uparrow$
    & EM $\uparrow$ & F1 $\uparrow$ & AM $\uparrow$ \\
    \midrule
    Gemma-2-2B-IT~\cite{gemma_2_2b_it}
        & 24.94 & 0.021 & 70.09 & 0.035 & 40.68 & 51.68 & 0.004 \\
    \addlinespace
    \quad + U1
        & 23.86 & 0.023 & 69.23 & 0.040 & 39.83 & 50.73 & 0.005 \\
    \quad + U2
        & 25.27 & 0.021 & 70.09 & 0.034 & 38.98 & 51.02 & 0.004 \\
    \quad + U3
        & 25.77 & 0.023 & 69.23 & 0.040 & 41.53 & 51.93 & 0.004 \\
    \quad + TE1
        & 32.17 & 0.029 & 78.63 & 0.046 & 43.22 & 52.56 & 0.005 \\
    \quad + TE2
        & 28.68 & 0.026 & 71.79 & 0.042 & 43.22 & 52.14 & 0.005 \\
    \quad + TE3
        & 32.17 & 0.029 & 78.63 & 0.046 & 42.37 & 51.71 & 0.005 \\
    \quad + TT1
        & 20.95 & 0.026 & 66.67 & 0.042 & 40.68 & 51.12 & 0.004 \\
    \quad + TT2
        & 22.53 & 0.026 & 71.79 & 0.041 & 41.53 & 51.93 & 0.004 \\
    \quad + TT3
        & 22.11 & 0.025 & 71.79 & 0.041 & 42.37 & 52.78 & 0.004 \\
    \quad + TA1
        & 18.54 & 0.034 & 56.41 & 0.062 & 39.83 & 50.47 & 0.008 \\
    \quad + TA2
        & 19.29 & 0.038 & 62.39 & 0.065 & 38.98 & 48.89 & 0.008 \\
    \quad + TA3
        & 18.79 & 0.038 & 62.39 & 0.065 & 38.98 & 49.77 & 0.008 \\
    \quad + ADE1
        & 20.62 & 0.024 & 68.38 & 0.039 & 38.98 & 50.90 & 0.004 \\
    \quad + ADE2
        & 25.85 & 0.025 & 68.38 & 0.038 & 38.14 & 49.99 & 0.004 \\
    \quad + ADE3
        & 23.28 & 0.022 & 70.09 & 0.035 & 38.98 & 50.83 & 0.004 \\
    \bottomrule
    \end{tabular}
    \label{tab:schemes_absolutevals_gemma_2b}
\end{table*}

\newpage \clearpage
\begin{table*}[!ht]
    \caption{\textbf{Impact of letter case schemes on performance and target attention allocation for the Mistral-7B-Instruct-v0.3 baseline.} Task metrics (Acc., EM, F1) and attention mass (AM) are reported across three benchmarks.}
    \centering
    \small
    \setlength{\tabcolsep}{4.5pt}
    \begin{tabular}{lccccccc}
    \toprule
    \multirow{2}{*}{\textbf{Method}}
    & \multicolumn{2}{c}{\textbf{MMLU-Pro}}
    & \multicolumn{2}{c}{\textbf{ARC}}
    & \multicolumn{3}{c}{\textbf{SQuADv2}} \\
    \cmidrule(lr){2-3} \cmidrule(lr){4-5} \cmidrule(lr){6-8}
    & Acc. $\uparrow$ & AM $\uparrow$
    & Acc. $\uparrow$ & AM $\uparrow$
    & EM $\uparrow$ & F1 $\uparrow$ & AM $\uparrow$ \\
    \midrule
    Mistral-7B~\cite{mistral_7b_instruct_v03}
        & 29.84 & 0.020 & 75.21 & 0.030 & 21.19 & 43.52 & 0.012 \\
    \addlinespace
    \quad + U1
        & 29.68 & 0.023 & 71.79 & 0.037 & 20.34 & 43.50 & 0.031 \\
    \quad + U2
        & 29.59 & 0.020 & 73.50 & 0.030 & 21.19 & 43.31 & 0.018 \\
    \quad + U3
        & 28.93 & 0.021 & 70.09 & 0.032 & 20.34 & 44.56 & 0.014 \\
    \quad + TE1
        & 45.14 & 0.037 & 84.62 & 0.049 & 19.49 & 43.51 & 0.032 \\
    \quad + TE2
        & 38.49 & 0.028 & 77.78 & 0.043 & 21.19 & 44.94 & 0.030 \\
    \quad + TE3
        & 44.64 & 0.037 & 85.47 & 0.050 & 21.19 & 45.09 & 0.033 \\
    \quad + TT1
        & 28.35 & 0.021 & 66.67 & 0.032 & 19.49 & 43.12 & 0.014 \\
    \quad + TT2
        & 33.00 & 0.024 & 76.92 & 0.035 & 19.49 & 43.02 & 0.014 \\
    \quad + TT3
        & 30.92 & 0.023 & 76.92 & 0.034 & 21.19 & 45.01 & 0.014 \\
    \quad + TA1
        & 20.78 & 0.029 & 58.12 & 0.041 & 19.49 & 40.71 & 0.023 \\
    \quad + TA2
        & 22.03 & 0.035 & 61.54 & 0.046 & 19.49 & 41.08 & 0.023 \\
    \quad + TA3
        & 21.78 & 0.034 & 62.39 & 0.046 & 19.49 & 41.23 & 0.023 \\
    \quad + ADE1
        & 28.35 & 0.023 & 64.96 & 0.035 & 21.19 & 43.71 & 0.018 \\
    \quad + ADE2
        & 33.42 & 0.024 & 70.94 & 0.036 & 21.19 & 43.26 & 0.018 \\
    \quad + ADE3
        & 27.35 & 0.022 & 71.79 & 0.031 & 20.34 & 42.55 & 0.017 \\
    \bottomrule
    \end{tabular}
    \label{tab:schemes_absolutevals_mistral_7b}

    \vspace{20pt}
    
    \caption{\textbf{Impact of letter case schemes on performance and target attention allocation for the Qwen2.5-7B-Instruct baseline.} Task metrics (Acc., EM, F1) and attention mass (AM) are reported across three benchmarks.}
    \centering
    \small
    \setlength{\tabcolsep}{4.5pt}
    \begin{tabular}{lccccccc}
    \toprule
    \multirow{2}{*}{\textbf{Method}}
    & \multicolumn{2}{c}{\textbf{MMLU-Pro}}
    & \multicolumn{2}{c}{\textbf{ARC}}
    & \multicolumn{3}{c}{\textbf{SQuADv2}} \\
    \cmidrule(lr){2-3} \cmidrule(lr){4-5} \cmidrule(lr){6-8}
    & Acc. $\uparrow$ & AM $\uparrow$
    & Acc. $\uparrow$ & AM $\uparrow$
    & EM $\uparrow$ & F1 $\uparrow$ & AM $\uparrow$ \\
    \midrule
    Qwen2.5-7B~\cite{qwen2.5_7b_instruct}
        & 39.07 & 0.037 & 88.03 & 0.061 & 32.20 & 62.64 & 0.029 \\
    \addlinespace
    \quad + U1
        & 39.57 & 0.039 & 88.03 & 0.068 & 33.05 & 64.50 & 0.055 \\
    \quad + U2
        & 39.32 & 0.038 & 87.18 & 0.061 & 32.20 & 62.52 & 0.032 \\
    \quad + U3
        & 39.40 & 0.037 & 87.18 & 0.063 & 32.20 & 62.06 & 0.030 \\
    \quad + TE1
        & 41.65 & 0.050 & 88.89 & 0.082 & 33.05 & 64.76 & 0.056 \\
    \quad + TE2
        & 39.48 & 0.043 & 87.18 & 0.074 & 33.05 & 64.43 & 0.058 \\
    \quad + TE3
        & 40.90 & 0.050 & 88.89 & 0.083 & 32.20 & 63.92 & 0.056 \\
    \quad + TT1
        & 37.57 & 0.038 & 86.32 & 0.066 & 32.20 & 64.60 & 0.030 \\
    \quad + TT2
        & 39.24 & 0.042 & 88.89 & 0.067 & 32.20 & 62.25 & 0.031 \\
    \quad + TT3
        & 38.57 & 0.041 & 88.03 & 0.068 & 32.20 & 62.25 & 0.031 \\
    \quad + TA1
        & 32.09 & 0.054 & 86.32 & 0.099 & 32.20 & 62.70 & 0.047 \\
    \quad + TA2
        & 32.59 & 0.059 & 87.18 & 0.104 & 31.36 & 62.16 & 0.047 \\
    \quad + TA3
        & 31.92 & 0.059 & 87.18 & 0.104 & 31.36 & 62.16 & 0.047 \\
    \quad + ADE1
        & 36.91 & 0.039 & 85.47 & 0.065 & 33.05 & 63.47 & 0.032 \\
    \quad + ADE2
        & 38.15 & 0.037 & 81.20 & 0.062 & 31.36 & 62.01 & 0.032 \\
    \quad + ADE3
        & 37.49 & 0.039 & 87.18 & 0.063 & 32.20 & 62.71 & 0.032 \\
    \bottomrule
    \end{tabular}
    \label{tab:schemes_absolutevals_qwen_7b}
\end{table*}

\newpage \clearpage
\begin{table*}[!ht]
    \caption{\textbf{Impact of letter case schemes on performance and target attention allocation for the Qwen2.5-14B-Instruct baseline.} Task metrics (Acc., EM, F1) and attention mass (AM) are reported across three benchmarks.}
    \centering
    \small
    \setlength{\tabcolsep}{4.5pt}
    \begin{tabular}{lccccccc}
    \toprule
    \multirow{2}{*}{\textbf{Method}}
    & \multicolumn{2}{c}{\textbf{MMLU-Pro}}
    & \multicolumn{2}{c}{\textbf{ARC}}
    & \multicolumn{3}{c}{\textbf{SQuADv2}} \\
    \cmidrule(lr){2-3} \cmidrule(lr){4-5} \cmidrule(lr){6-8}
    & Acc. $\uparrow$ & AM $\uparrow$
    & Acc. $\uparrow$ & AM $\uparrow$
    & EM $\uparrow$ & F1 $\uparrow$ & AM $\uparrow$ \\
    \midrule
    Qwen2.5-14B~\cite{qwen2.5_14b_instruct}
        & 52.70 & 0.036 & 94.02 & 0.058 & 41.53 & 73.76 & 0.005 \\
    \addlinespace
    \quad + U1
        & 51.70 & 0.038 & 93.16 & 0.063 & 40.68 & 75.55 & 0.014 \\
    \quad + U2
        & 52.37 & 0.036 & 93.16 & 0.058 & 41.53 & 73.92 & 0.006 \\
    \quad + U3
        & 52.12 & 0.037 & 94.02 & 0.061 & 41.53 & 76.07 & 0.007 \\
    \quad + TE1
        & 61.76 & 0.046 & 95.73 & 0.071 & 40.68 & 74.63 & 0.014 \\
    \quad + TE2
        & 56.53 & 0.040 & 94.02 & 0.063 & 41.53 & 74.85 & 0.014 \\
    \quad + TE3
        & 61.10 & 0.046 & 95.73 & 0.071 & 42.37 & 76.33 & 0.014 \\
    \quad + TT1
        & 51.62 & 0.038 & 91.45 & 0.061 & 42.37 & 76.29 & 0.007 \\
    \quad + TT2
        & 54.70 & 0.041 & 93.16 & 0.065 & 41.53 & 76.17 & 0.007 \\
    \quad + TT3
        & 53.45 & 0.041 & 93.16 & 0.064 & 41.53 & 76.19 & 0.007 \\
    \quad + TA1
        & 50.29 & 0.049 & 90.60 & 0.079 & 41.53 & 73.30 & 0.017 \\
    \quad + TA2
        & 52.87 & 0.053 & 89.74 & 0.084 & 41.53 & 73.66 & 0.018 \\
    \quad + TA3
        & 51.12 & 0.053 & 89.74 & 0.085 & 42.37 & 74.37 & 0.018 \\
    \quad + ADE1
        & 51.62 & 0.037 & 90.60 & 0.057 & 41.53 & 73.86 & 0.006 \\
    \quad + ADE2
        & 52.37 & 0.036 & 92.31 & 0.056 & 42.37 & 75.35 & 0.006 \\
    \quad + ADE3
        & 50.12 & 0.037 & 93.16 & 0.059 & 40.68 & 73.09 & 0.006 \\
    \bottomrule
    \end{tabular}
    \label{tab:schemes_absolutevals_qwen_14b}

    \vspace{20pt}
    
    \caption{\textbf{Impact of letter case schemes on performance and target attention allocation for the Qwen3-4B-Thinking-2507 baseline.} Task metrics (Acc., EM, F1) and attention mass (AM) are reported across three benchmarks.}
    \centering
    \small
    \setlength{\tabcolsep}{4.5pt}
    \begin{tabular}{lccccccc}
    \toprule
    \multirow{2}{*}{\textbf{Method}}
    & \multicolumn{2}{c}{\textbf{MMLU-Pro}}
    & \multicolumn{2}{c}{\textbf{ARC}}
    & \multicolumn{3}{c}{\textbf{SQuADv2}} \\
    \cmidrule(lr){2-3} \cmidrule(lr){4-5} \cmidrule(lr){6-8}
    & Acc. $\uparrow$ & AM $\uparrow$
    & Acc. $\uparrow$ & AM $\uparrow$
    & EM $\uparrow$ & F1 $\uparrow$ & AM $\uparrow$ \\
    \midrule
    Qwen3-4B-Thinking~\cite{qwen3_4b_thinking_2507}
        & 3.74 & 0.004 & 6.84 & 0.006 & 8.47 & 9.80 & 0.000 \\
    \addlinespace
    \quad + U1
        & 3.57 & 0.004 & 6.84 & 0.008 & 7.63 & 8.98 & 0.001 \\
    \quad + U2
        & 3.66 & 0.004 & 6.84 & 0.006 & 8.47 & 9.92 & 0.000 \\
    \quad + U3
        & 3.66 & 0.004 & 6.84 & 0.007 & 7.63 & 9.07 & 0.000 \\
    \quad + TE1
        & 3.41 & 0.005 & 6.84 & 0.009 & 9.32 & 10.84 & 0.001 \\
    \quad + TE2
        & 3.49 & 0.004 & 6.84 & 0.007 & 10.17 & 11.64 & 0.001 \\
    \quad + TE3
        & 3.49 & 0.005 & 6.84 & 0.009 & 7.63 & 9.28 & 0.001 \\
    \quad + TT1
        & 3.66 & 0.004 & 6.84 & 0.007 & 5.08 & 6.63 & 0.000 \\
    \quad + TT2
        & 3.57 & 0.005 & 6.84 & 0.008 & 8.47 & 9.98 & 0.000 \\
    \quad + TT3
        & 3.66 & 0.004 & 6.84 & 0.008 & 8.47 & 9.98 & 0.000 \\
    \quad + TA1
        & 3.57 & 0.005 & 6.84 & 0.010 & 6.78 & 8.47 & 0.001 \\
    \quad + TA2
        & 3.57 & 0.006 & 6.84 & 0.010 & 7.63 & 9.28 & 0.001 \\
    \quad + TA3
        & 3.57 & 0.006 & 6.84 & 0.010 & 7.63 & 9.30 & 0.001 \\
    \quad + ADE1
        & 3.74 & 0.004 & 6.84 & 0.006 & 7.63 & 9.30 & 0.000 \\
    \quad + ADE2
        & 3.66 & 0.004 & 6.84 & 0.006 & 7.63 & 9.14 & 0.000 \\
    \quad + ADE3
        & 3.82 & 0.004 & 6.84 & 0.006 & 8.47 & 9.82 & 0.000 \\
    \bottomrule
    \end{tabular}
    \label{tab:schemes_absolutevals_qwen_thinking_4b}
\end{table*}

\newpage \clearpage
\begin{table*}[!ht]
    \caption{\textbf{Impact of letter case schemes on performance and target attention allocation for the gpt-oss-20B baseline.} Task metrics (Acc., EM, F1) and attention mass (AM) are reported across three benchmarks.}
    \centering
    \small
    \setlength{\tabcolsep}{4.5pt}
    \begin{tabular}{lccccccc}
    \toprule
    \multirow{2}{*}{\textbf{Method}}
    & \multicolumn{2}{c}{\textbf{MMLU-Pro}}
    & \multicolumn{2}{c}{\textbf{ARC}}
    & \multicolumn{3}{c}{\textbf{SQuADv2}} \\
    \cmidrule(lr){2-3} \cmidrule(lr){4-5} \cmidrule(lr){6-8}
    & Acc. $\uparrow$ & AM $\uparrow$
    & Acc. $\uparrow$ & AM $\uparrow$
    & EM $\uparrow$ & F1 $\uparrow$ & AM $\uparrow$ \\
    \midrule
    gpt-oss-20B~\cite{openai_2025_gptoss20b}
        & 0.33 & 0.004 & 0.85 & 0.004 & 0.85 & 7.22 & 0.000 \\
    \addlinespace
    \quad + U1
        & 0.42 & 0.004 & 0.85 & 0.005 & 1.69 & 8.10 & 0.000 \\
    \quad + U2
        & 0.33 & 0.004 & 0.85 & 0.004 & 0.85 & 7.13 & 0.000 \\
    \quad + U3
        & 0.33 & 0.004 & 0.85 & 0.004 & 0.85 & 7.16 & 0.000 \\
    \quad + TE1
        & 0.33 & 0.004 & 0.00 & 0.005 & 0.85 & 7.21 & 0.000 \\
    \quad + TE2
        & 0.25 & 0.004 & 0.00 & 0.005 & 0.85 & 7.02 & 0.000 \\
    \quad + TE3
        & 0.33 & 0.004 & 0.00 & 0.005 & 0.85 & 7.21 & 0.000 \\
    \quad + TT1
        & 0.42 & 0.004 & 0.85 & 0.004 & 1.69 & 8.07 & 0.000 \\
    \quad + TT2
        & 0.33 & 0.004 & 0.85 & 0.004 & 0.85 & 7.24 & 0.000 \\
    \quad + TT3
        & 0.33 & 0.004 & 0.85 & 0.004 & 0.85 & 7.24 & 0.000 \\
    \quad + TA1
        & 0.42 & 0.004 & 0.00 & 0.005 & 2.54 & 8.79 & 0.000 \\
    \quad + TA2
        & 0.33 & 0.004 & 0.00 & 0.005 & 0.85 & 7.16 & 0.000 \\
    \quad + TA3
        & 0.33 & 0.004 & 0.00 & 0.005 & 0.85 & 7.07 & 0.000 \\
    \quad + ADE1
        & 0.42 & 0.004 & 0.00 & 0.004 & 0.85 & 7.19 & 0.000 \\
    \quad + ADE2
        & 0.25 & 0.004 & 0.00 & 0.004 & 0.85 & 7.39 & 0.000 \\
    \quad + ADE3
        & 0.33 & 0.004 & 0.85 & 0.004 & 0.85 & 7.16 & 0.000 \\
    \bottomrule
    \end{tabular}
    \label{tab:schemes_absolutevals_gptoss_20b}
\end{table*}

\begin{table*}[!ht]
    \caption{\textbf{Microscopic spatial steering.} Absolute percentage point ($\Delta$pp) focus increases inside target object bounding boxes across typographic manipulation schemes.}
    \label{tab:appendix_micro_spatial_steering}
    \centering
    \small
    \setlength{\tabcolsep}{4.5pt}
    \begin{tabular}{lcccc}
    \toprule
    \begin{tabular}[c]{@{}l@{}}\textbf{Method}\end{tabular} &
    \begin{tabular}[c]{@{}c@{}}\textbf{Gemma-3}\\\textbf{4B-IT}\end{tabular} & 
    \begin{tabular}[c]{@{}c@{}}\textbf{Gemma-4}\\\textbf{E4B-Think}\end{tabular} & 
    \begin{tabular}[c]{@{}c@{}}\textbf{Qwen3-VL}\\\textbf{4B-Instruct}\end{tabular} &
    \begin{tabular}[c]{@{}c@{}}\textbf{Qwen3-VL}\\\textbf{2B-Think}\end{tabular} \\
    \midrule
    \addlinespace
    \quad U1   & +1.607 & +1.746 & +1.876 & +2.214  \\
    \quad U2   & +0.333 & +1.069 & +0.214 & +0.297  \\
    \quad U3   & +0.593 & +1.432 & +0.386 & +0.478  \\
    \addlinespace
    \quad TE1  & +1.692 & +1.881 & +0.682 & +0.921  \\
    \quad TE2  & +5.742 & +1.610 & +1.000 & +2.535  \\
    \quad TE3  & +1.612 & +1.849 & +0.668 & +1.001  \\
    \addlinespace
    \quad TT1  & +0.676 & +1.371 & +1.851 & +2.181  \\
    \quad TT2  & +0.471 & +1.342 & +0.393 & +0.404  \\
    \quad TT3  & +0.435 & +1.299 & +0.349 & +0.422  \\
    \addlinespace
    \quad TA1  & +6.229 & +2.419 & +1.902 & +2.055  \\
    \quad TA2  & +5.997 & +2.271 & +1.331 & +1.290  \\
    \quad TA3  & +5.675 & +2.151 & +1.348 & +1.346  \\
    \addlinespace
    \quad ADE1 & +0.573 & +1.093 & +1.769 & +2.327  \\
    \quad ADE2 & +2.407 & +1.033 & +0.961 & +1.789  \\
    \quad ADE3 & +0.464 & +1.234 & +0.292 & +0.335  \\
    \bottomrule
    \end{tabular}

    \vspace{20pt}

    \caption{\textbf{Macroscopic modality shifts.} Relative percentage shifts ($\Delta\%$) in attention redirection away from visual frames and image token routing features. Positive values denote a shift toward the text modality, while negative values signify a shift toward the image modality.}
    \label{tab:appendix_macro_modality_shifts}
    \centering
    \small
    \setlength{\tabcolsep}{4.5pt}
    \begin{tabular}{lcccc}
    \toprule
    \begin{tabular}[c]{@{}l@{}}\textbf{Method}\end{tabular} &
    \begin{tabular}[c]{@{}c@{}}\textbf{Gemma-3}\\\textbf{4B-IT}\end{tabular} & 
    \begin{tabular}[c]{@{}c@{}}\textbf{Gemma-4}\\\textbf{E4B-Think}\end{tabular} & 
    \begin{tabular}[c]{@{}c@{}}\textbf{Qwen3-VL}\\\textbf{4B-Instruct}\end{tabular} &
    \begin{tabular}[c]{@{}c@{}}\textbf{Qwen3-VL}\\\textbf{2B-Think}\end{tabular} \\
    \midrule
    \addlinespace
    \quad U1   & +8.903  & +25.237 & +29.025 & +24.016  \\
    \quad U2   & $-$0.656  & +0.118  & $-$0.862 & $-$9.236    \\
    \quad U3   & +6.950  & +9.228  & +4.511 & +9.765    \\
    \addlinespace
    \quad TE1  & +5.196  & +16.296 & +18.806 & +21.829  \\
    \quad TE2  & +31.831 & +22.560 & +27.481 & +43.735  \\
    \quad TE3  & +4.175  & +18.331 & +20.186 & +23.053  \\
    \addlinespace
    \quad TT1  & +8.035  & +10.009 & +3.392 & +7.687    \\
    \quad TT2  & +6.521  & +0.606  & $-$3.538 & +8.610    \\
    \quad TT3  & +4.126  & +2.029  & $-$2.508 & +8.203    \\
    \addlinespace
    \quad TA1  & +26.403 & +28.828 & +27.318 & +28.805  \\
    \quad TA2  & +23.066 & +23.945 & +17.895 & +24.783  \\
    \quad TA3  & +22.348 & +25.450 & +17.710 & +24.079  \\
    \addlinespace
    \quad ADE1 & +2.554  & +1.017  & $-$7.045 & $-$8.707    \\
    \quad ADE2 & +21.415 & +7.616  & $-$0.773 & +25.977   \\
    \quad ADE3 & $-$2.815  & +2.592  & +7.980 & $-$9.402    \\
    \bottomrule
    \end{tabular}
\end{table*}

\clearpage \newpage

\clearpage \newpage

\clearpage \newpage

\clearpage \newpage

\clearpage \newpage
\section{Robustness and Generalization across Languages and Tasks}
\label{appendix:vision_code_language_results}

\subsection{Case Sensitivity across Different Languages}
While our primary analysis focuses on English text, orthographic characteristics vary significantly across global languages. Certain scripts, such as Chinese (which is logographic) and Arabic (a unicase abjad script), do not feature an orthographic case distinction. Consequently, case-based typographic perturbations are inherently undefined in these typographic regimes. To determine whether the observed casing effect is an inherent structural property of cased Transformer models or merely an artifact of English training distributions, we validate our framework on XQuAD~\cite{Artetxe_2020_XQuAD} across a diverse set of cased languages: German, Spanish, and Russian (Cyrillic).

\begin{figure}[t!]
    \centering
    \includegraphics[width=\linewidth]{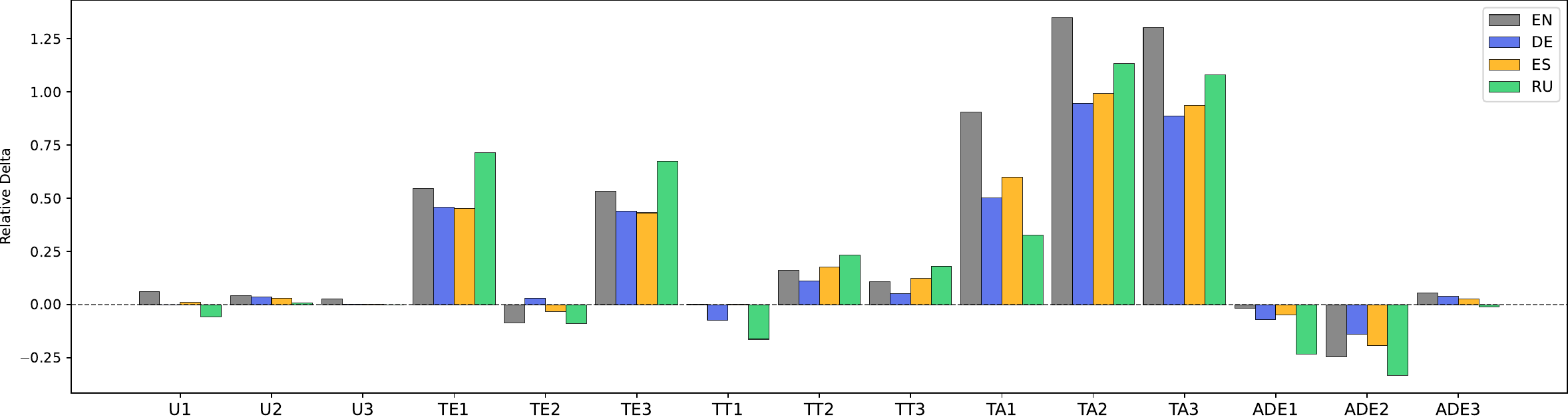}
    \caption{\textbf{Cross-lingual generalization of case-sensitivity effects across diverse typographic scripts.} Evaluated on XQuAD, delta attention mass tracking demonstrates that casing salience operates as a universal representational property across English (EN), German (DE), Spanish (ES), and Russian (RU).}
    \label{fig:xquad}
\end{figure}

As illustrated in \Cref{fig:xquad} and \Cref{tab:schemes_cross_lingual_deltas} (with English provided for baseline reference), the empirical attention dynamics across different languages closely mirror our primary findings. Productive Target Emphasis (TE) and high-entropy Target Alternating (TA) patterns notably shift latent attention across all tested cased scripts up to $+1.133$ pp in Russian (TA2), while Adversarial De-emphasis (ADE) consistently suppresses it. This cross-lingual consistency confirms that casing-salience operates as a universal representational property within cased linguistic regimes rather than an English-specific artifact. Extending this paradigm to characterize typographic salience in non-cased scripts via alternative formatting mechanisms (\eg Markdown bolding) remains an important direction for future work.

\subsection{Case Sensitivity for Code Generation}
\begin{figure}[!t]
    \centering
    \begin{subfigure}{0.45\linewidth} % width of the subfigure
        \centering
        \includegraphics[width=\textwidth]{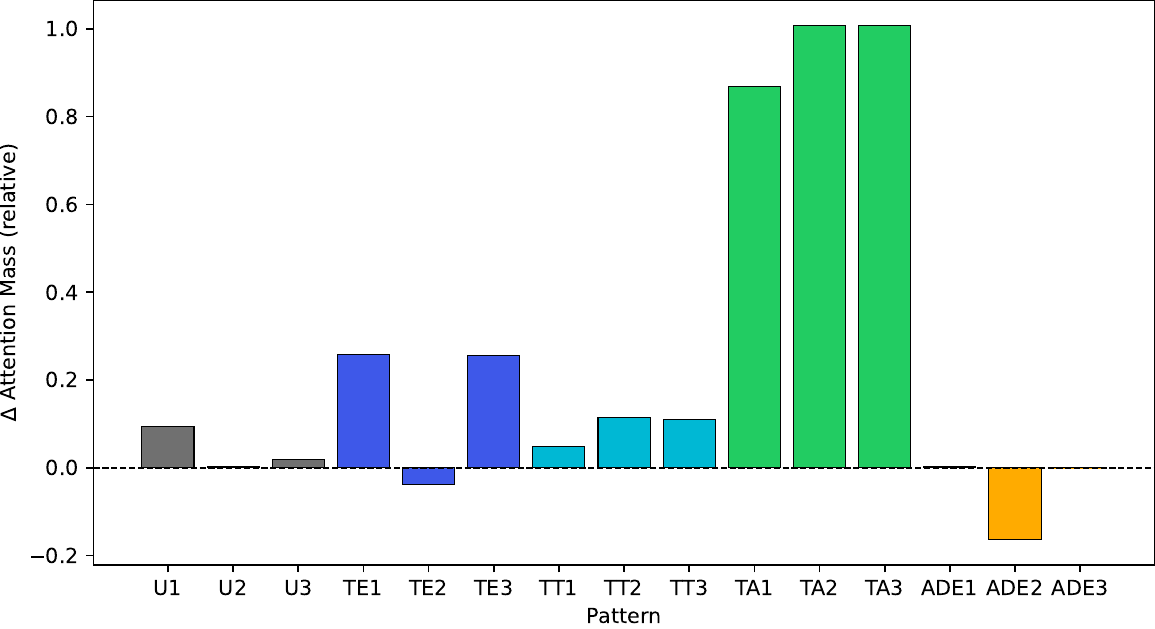}
        \caption{Relative Attention vs. Baseline}
    \end{subfigure}% 
    \hspace{2em}
    \begin{subfigure}{0.45\linewidth} % width of the subfigure
        \centering
        \includegraphics[width=\textwidth]{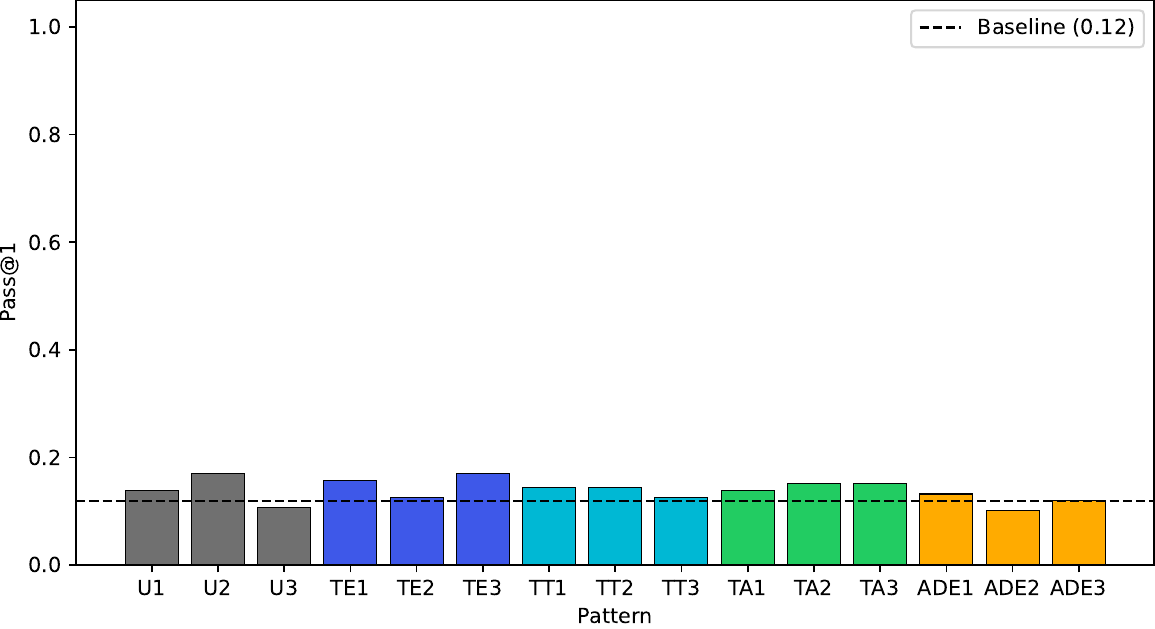}
        \caption{Pass@1 per Pattern}
    \end{subfigure}%
    \caption{\textbf{Casing-induced attention mass variations and corresponding downstream task performance on HumanEval.} Relative attention mass and performance deltas illustrate high syntactic resilience in coding tasks.}
    \label{fig:humaneval}
\end{figure}

We further investigate the task-dependent nature of typographic attractors by evaluating model performance on the HumanEval coding benchmark~\cite{Chen_2021_HumanEval} (\cf \cref{fig:humaneval,tab:schemes_humaneval_full}). In generative programming tasks, the operational dynamics shift compared to standard natural language reading comprehension or question-answering formats. Logical code generation demonstrates a unique resilience to semantic disruption caused by irregular casing patterns. Because code synthesis relies heavily on rigid syntax, structural declarations, and exact identifier matching, increased target attention directly benefits code output token probabilities rather than introducing semantic ambiguity.

Uniquely, the high-entropy target alternating (TA) pattern yields the highest relative attention gains (up to $+1.007$ pp) while simultaneously preserving or marginally improving generation accuracy. Furthermore, productive target emphasis attractors maximize output quality, resulting in an increase in Pass@1 accuracy over the baseline of up to $+5.1$ pp (specifically for TE3). Conversely, Adversarial De-emphasis (ADE2) shifts attention away from the target tokens ($-0.163$ pp) and induces the highest degradation to coding performance among the targeted schemes, dropping accuracy by $-1.8$ pp relative to the baseline. These results corroborate our hypothesis that in functional, structured text regimes, casing-induced attention allocation directly steers generative success.

\begin{table*}[!ht]
    \caption{\textbf{Cross-lingual generalization of typographic interventions on internal attention distribution.} Mean relative attention mass shifts ($\Delta$pp) are evaluated across German (DE), English (EN), Spanish (ES), and Russian (RU) scripts.}
    \centering
    \small
    \setlength{\tabcolsep}{4.5pt}
    \begin{tabular}{lcccc}
    \toprule
    \textbf{Method} & \textbf{EN} & \textbf{DE} & \textbf{ES} & \textbf{RU} \\
    \midrule
    Qwen2.5-7B~\cite{qwen2.5_7b_instruct} 
        & -- & -- & -- & -- \\
    \addlinespace
        \quad + U1   & +0.062 & $-$0.002 & +0.010 & $-$0.057 \\
        \quad + U2   & +0.042 & +0.035 & +0.030 & +0.007 \\
        \quad + U3   & +0.028 & +0.003 & +0.001 & $-$0.001 \\
    \addlinespace
        \quad + TE1  & +0.547 & +0.457 & +0.453 & +0.714 \\
        \quad + TE2  & $-$0.084 & +0.031 & $-$0.032 & $-$0.089 \\
        \quad + TE3  & +0.535 & +0.440 & +0.432 & +0.675 \\
    \addlinespace
        \quad + TT1  & +0.002 & $-$0.074 & +0.001 & $-$0.162 \\
        \quad + TT2  & +0.160 & +0.111 & +0.177 & +0.232 \\
        \quad + TT3  & +0.107 & +0.052 & +0.123 & +0.181 \\
    \addlinespace
        \quad + TA1  & +0.904 & +0.501 & +0.600 & +0.326 \\
        \quad + TA2  & +1.348 & +0.946 & +0.993 & +1.133 \\
        \quad + TA3  & +1.304 & +0.887 & +0.937 & +1.080 \\
    \addlinespace
        \quad + ADE1 & $-$0.016 & $-$0.072 & $-$0.047 & $-$0.232 \\
        \quad + ADE2 & $-$0.247 & $-$0.139 & $-$0.193 & $-$0.332 \\
        \quad + ADE3 & +0.055 & +0.039 & +0.025 & $-$0.010 \\
    \bottomrule
    \end{tabular}
    \label{tab:schemes_cross_lingual_deltas}
\end{table*}

\begin{table*}[!ht]
    \caption{\textbf{Impact of letter case schemes on attention distribution and code generation performance on the HumanEval benchmark.} The mean and standard deviation of relative attention mass (AM) shifts ($\Delta$pp) are tracked alongside downstream functional programming accuracy (\text{Pass@1}).}
    \centering
    \small
    \setlength{\tabcolsep}{4.5pt}
    \begin{tabular}{lccc}
    \toprule
    \multirow{2}{*}{\textbf{Method}} & \multicolumn{2}{c}{\textbf{Relative AM $\Delta$pp}} & \multirow{2}{*}{\textbf{Pass@1}} \\
    \cmidrule(lr){2-3}
    & \textbf{Mean} & \textbf{Std.} & \\
    \midrule
    Qwen2.5-7B~\cite{qwen2.5_7b_instruct} 
        & -- & -- & 0.119 \\
    \addlinespace
    \quad + U1   & +0.094 & 0.146 & 0.138 \\
    \quad + U2   & +0.004 & 0.020 & 0.170 \\
    \quad + U3   & +0.019 & 0.072 & 0.107 \\
    \quad + TE1  & +0.258 & 0.169 & 0.157 \\
    \quad + TE2  & $-$0.038 & 0.176 & 0.126 \\
    \quad + TE3  & +0.255 & 0.171 & 0.170 \\
    \quad + TT1  & +0.048 & 0.100 & 0.145 \\
    \quad + TT2  & +0.114 & 0.091 & 0.145 \\
    \quad + TT3  & +0.110 & 0.089 & 0.126 \\
    \quad + TA1  & +0.868 & 0.521 & 0.138 \\
    \quad + TA2  & +1.007 & 0.557 & 0.151 \\
    \quad + TA3  & +1.007 & 0.568 & 0.151 \\
    \quad + ADE1 & +0.002 & 0.071 & 0.132 \\
    \quad + ADE2 & $-$0.163 & 0.103 & 0.101 \\
    \quad + ADE3 & +0.000 & 0.011 & 0.119 \\
    \bottomrule
    \end{tabular}
    \label{tab:schemes_humaneval_full}
\end{table*}

\newpage \clearpage
\section{Tokenization Dynamics and Computational Overhead}
\label{appendix:tokenization_analysis}

A potential concern in typographic intervention studies is the confounding role of sub-word tokenization. Since BPE and SentencePiece algorithms are sensitive to character-level casing, altering the surface form can lead to different sub-word splits and variations in total sequence length. In this section, we empirically quantify these shifts and evaluate their impact on computational latency.

\paragraph{Impact on Token Fragmentation.}
To address whether the ``casing effect'' is merely a byproduct of increased token density, we measure the change in token count ($\Delta$ Tokens) across all intervention schemes relative to the naturally cased baseline (\cref{tab:pattern_comparison}). For the Qwen2.5-7B-Instruct model~\cite{qwen2.5_7b_instruct} (mean baseline length: 202.12 tokens), we observe that while conventional schemes like TE1 (uppercase target, lowercase context) induce only a marginal increase of $+1.25$ pp, high-entropy schemes such as TE2 and ADE2 trigger significant fragmentation, increasing token counts by up to $+22.72$ pp.

Crucially, our findings in \Cref{sec:results_attention} show that attention concentration does not scale linearly with token count. For example, while TA1 ($+10.82\%$ tokens) and TA2 ($+4.36\%$ tokens) differ in fragmentation, they both yield nearly identical gains in attention mass ($+2.44$ pp vs $+2.75$ pp). This suggests that while casing changes BPE boundaries, the model's attentional response is driven by the orthographic signal rather than the number of sub-word units.

\paragraph{Computational Latency.}
As our interventions are applied purely at the pre-tokenization string level, they require zero architectural modification. To verify the efficiency of this approach, we measured the per-strategy latency on a single NVIDIA H100 GPU. As shown in \Cref{tab:pattern_comparison}, the mean latency across all patterns is a negligible $0.0015$ ms. Even the most complex alternating schemes (TA3) add only $0.0052$ ms to the processing time, confirming that case-based steering is an exceptionally lightweight mechanism for influencing model behavior compared to alternative methods like adapter inference or runtime activation steering.

\begin{table}[!t]
    \centering
    \small
    \caption{\textbf{Performance comparison across different letter casing schemes for the Qwen2.5-7B-Instruct baseline.} Results indicate a negligible latency impact and slight variations in token count due to different tokenization of letter casing schemes.}
    \begin{tabular}{lcc c lcc}
        \toprule
        \textbf{Pattern} & \textbf{Latency $\downarrow$} & \textbf{$\Delta$ Tokens} & \hspace{0.5em} & \textbf{Pattern} & \textbf{Latency $\downarrow$} & \textbf{$\Delta$ Tokens} \\
        & \textit{(ms)} & \textit{(\% vs base)} & & & \textit{(ms)} & \textit{(\% vs base)} \\
        \midrule
        \multicolumn{3}{l}{\textit{Unified}} & & \multicolumn{3}{l}{\textit{Target Alternating}} \\
        U1 & 0.0007 & +7.74\% & & TA1 & 0.0013 & +10.82\% \\
        U2 & 0.0007 & $-$0.06\% & & TA2 & 0.0014 & +4.36\% \\
        U3 & 0.0011 & +1.85\% & & TA3 & 0.0052 & +4.44\% \\
        \addlinespace[0.6em]
        \multicolumn{3}{l}{\textit{Target Emphasis}} & & \multicolumn{3}{l}{\textit{Adversarial De-Emphasis}} \\
        TE1 & 0.0007 & +1.25\% & & ADE1 & 0.0007 & +6.42\% \\
        TE2 & 0.0025 & +22.72\% & & ADE2 & 0.0025 & +21.43\% \\
        TE3 & 0.0013 & +1.32\% & & ADE3 & 0.0013 & $\pm$0.00\% \\
        \addlinespace[0.6em]
        \multicolumn{3}{l}{\textit{Target Title Case}} & & & & \\
        TT1 & 0.0009 & +6.74\% & & & & \\
        TT2 & 0.0009 & +0.25\% & & & & \\
        TT3 & 0.0025 & +0.32\% & & & & \\
        \bottomrule
    \end{tabular}
    \label{tab:pattern_comparison}
\end{table}

\newpage \clearpage
\section{Tokenizer-Specific Letter Casing Effects}
\label{appendix:tokenizer_details}

A critical consideration in our study is whether the observed attention shifts are merely artifacts of tokenization density---the phenomenon where uncommon casing patterns (\eg, alternating case) cause a word to be fragmented into a larger number of sub-word tokens. To investigate this, we analyze the behavior of four distinct tokenization schemes: Byte-Pair Encoding (BPE) (LLaMA-3 and GPT), byte-level BPE (BBPE) (Qwen2.5, Qwen3, and Qwen3-VL), and two variants of SentencePiece (Gemma and Mistral). Qualitative examples of these behaviors are provided in \Cref{tab:tokenizer_comparison,tab:tokenizer_comparison_reasoning,tab:tokenizer_comparison_VLM}.

\paragraph{Variation in Token Fragmentation.} 
As shown in the comparative tables, all evaluated tokenizers exhibit significant structural sensitivity to case variation. Standard lowercase or capitalized forms (\eg, \textit{running}) are typically represented as single tokens. However, transitioning to full uppercase or alternating case (\textit{RuNnInG}) frequently triggers aggressive fragmentation. For instance, in LLaMA-3's BPE, \texttt{UNBELIEVABLY} is split into six tokens, whereas its lowercase counterpart requires only two. Mistral's SentencePiece implementation is even more sensitive, often splitting uppercase tokens into near-character-level fragments. 
We observe that the reasoning-specialized models, Qwen3 and GPT, follow nearly identical fragmentation trajectories as their non-reasoning counterparts (\cf \cref{tab:tokenizer_comparison_reasoning}). Despite the architectural shift toward reasoning, the initial tokenization layer still decomposes high-entropy casing (\eg, \textit{UnBeLiEvAbLy}) into character-level or near-character-level sub-words, confirming that the ``reasoning buffer'' observed in these models (\cf \cref{appendix:reasoning_models}) operates on top of standard, case-sensitive input processing.

\paragraph{Case Sensitivity as Representational Property.}
Our empirical results (\cf \cref{sec:results_attention}) show that the ``casing effect'' remains robust across all models, despite their different tokenization budgets for the same lexical content. While one might expect that a higher token count naturally attracts more attention mass simply by providing more keys and values for the attention mechanism to attend to, we find that:
\begin{enumerate}
    \item Models like Gemma-2, which can represent \texttt{RUNNING} as a single token, still exhibit the same directionality in attention shifts as models that fragment the word.
    \item The performance degradation observed in alternating case (TA) persists even when comparing models with significantly different fragmentation ratios for the same alternating span.
\end{enumerate}
To fully isolate typographic salience from subword segmentation boundaries, we isolated a strict \textit{token-stable} subset of our 500 sentences dataset where sequence lengths and token IDs remain invariant to the baseline. Even with zero token fragmentation variation, the casing effect persists with statistical significance: TA3 induces a localized attention mass shift of $+2.35$ pp and TE3 retains a $+0.54$ pp shift, confirming that embedding shifts drive this phenomenon independent of segmentation artifacts.

\begin{figure*}[!t]
    \centering
    \begin{subfigure}{0.45\textwidth}
        \centering
        \includegraphics[width=\textwidth]{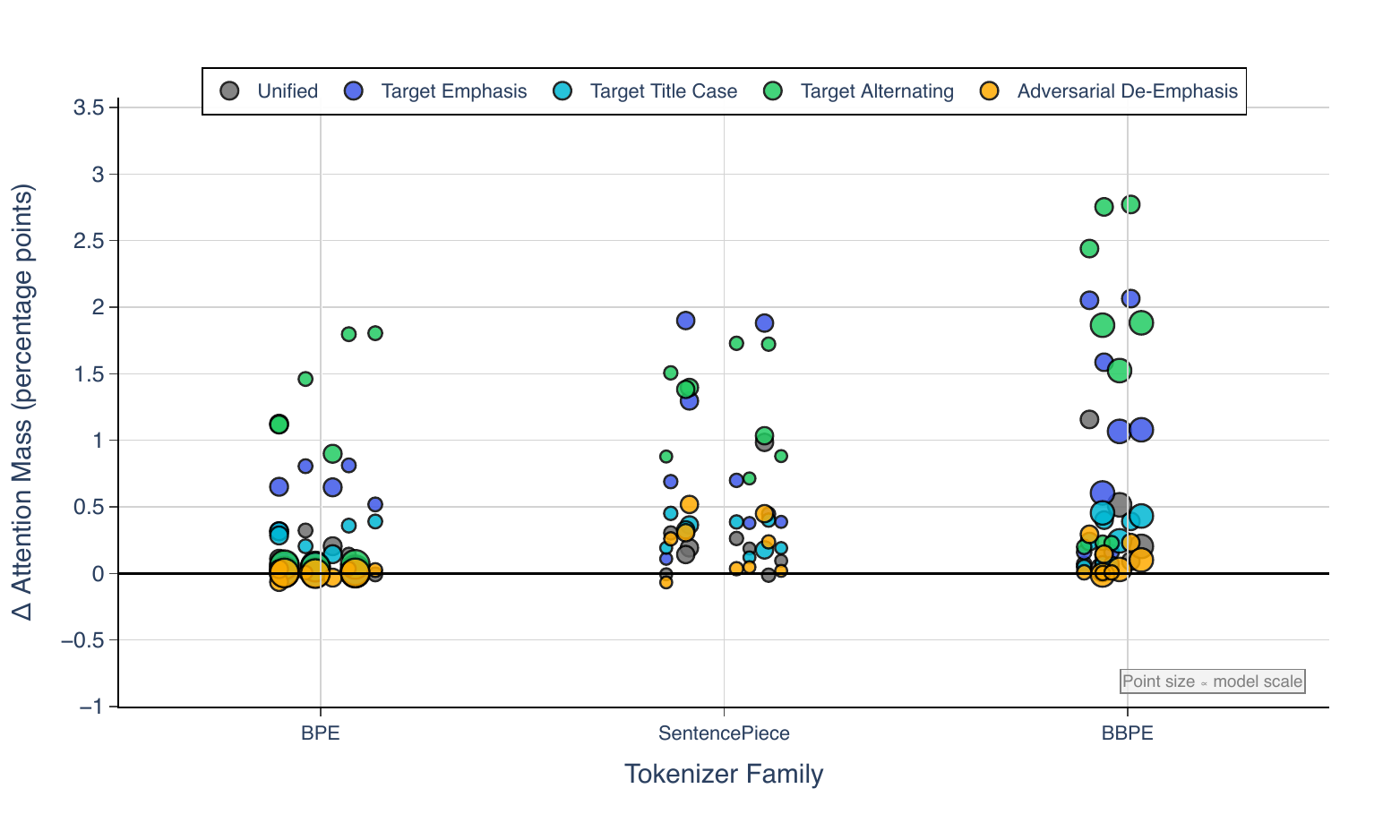}
        \caption{Relative Attention Mass on Target}
         \label{fig:tokenizer_family_attention}
    \end{subfigure}% 
    \hspace{2em}
    \begin{subfigure}{0.45\textwidth}
        \centering
        \includegraphics[width=\textwidth]{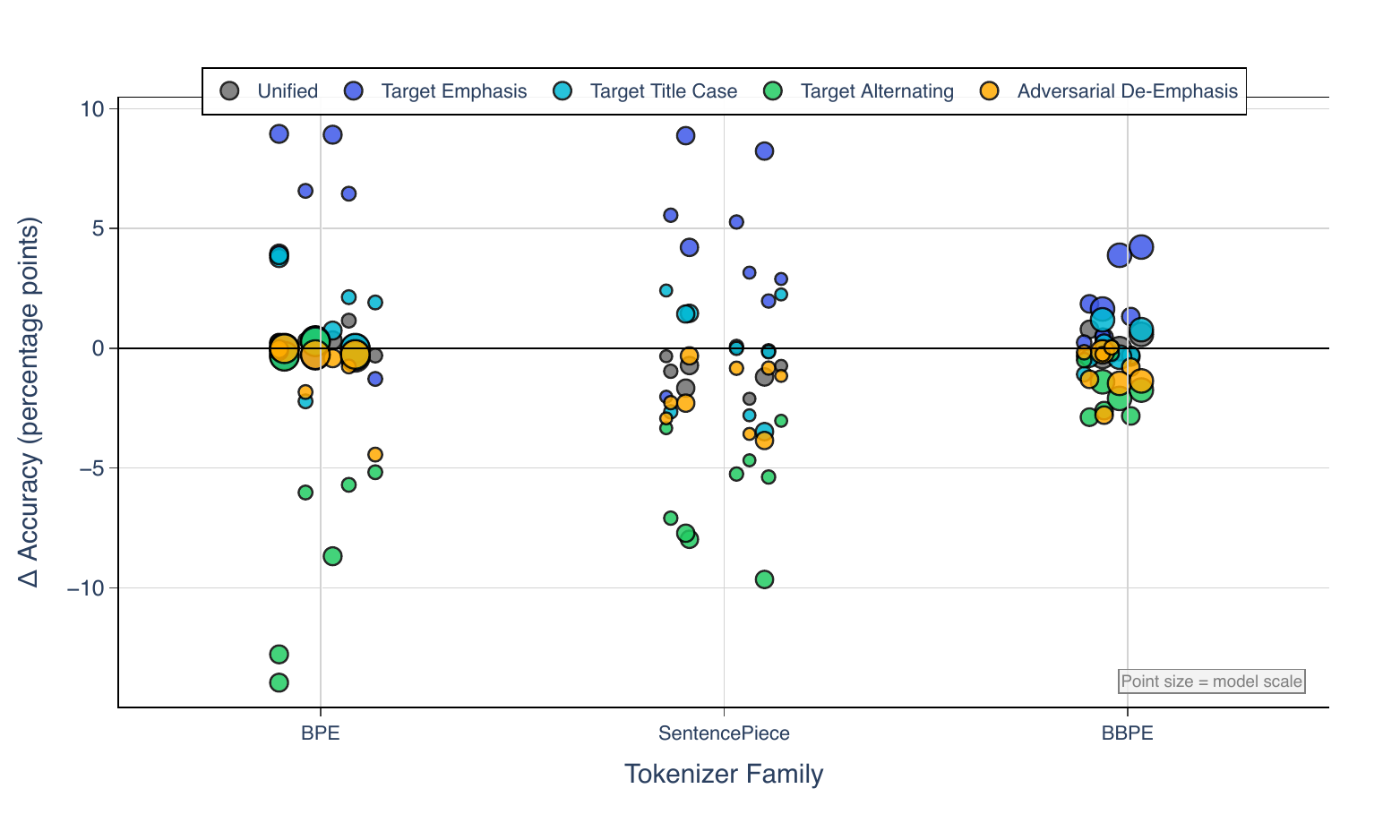}
        \caption{Relative Task Accuracy}
         \label{fig:tokenizer_family_accuracy}
    \end{subfigure}% 
    \caption{\textbf{Aggregate impact across tokenizer families.} Comparison of BPE, SentencePiece, and BBPE shows that while the magnitude of the casing effect varies, the directionality of attention shifts and performance sensitivity remains robust across different sub-word segmentation philosophies.}
    \label{fig:tokenizer_family_comparison}
\end{figure*}

\paragraph{Cross-Family Statistical Robustness.}
To further validate the independence of this property from tokenization logic, we aggregate the effects across three major tokenizer families: BPE, BBPE, and SentencePiece (\cref{fig:tokenizer_family_comparison}). Our statistical analysis reveals that the attentional shift is a universal feature across all families, with BBPE showing the highest mean gain in attention mass ($+0.626$ pp) and the highest stability in accuracy ($\sigma = 1.444$). 
As illustrated in \Cref{fig:tokenizer_family_attention}, every tokenizer family exhibits a positive mean shift in attention mass, confirming that typographic casing functions as a cross-architecture attractor. While BPE shows a lower mean accuracy drop ($-0.344$ pp) compared to SentencePiece ($-0.963$ pp), it exhibits significantly higher variance ($\sigma = 5.422$), as shown in \Cref{fig:tokenizer_family_accuracy}. This indicates that while the casing effect is universal, the underlying tokenization scheme influences the \textit{predictability} of the model's behavioral response to these shifts. The persistence of both attention concentration and accuracy fluctuations across these distinct families confirms that LLMs treat case as a high-level importance signal rather than a low-level tokenization artifact.

\clearpage \newpage
\begin{table}[!ht]
    \caption{\textbf{Qualitative comparison on letter case sensitivity of tokenization schemes used in our LLM experiments.} SP (SentencePiece), BPE (Byte-Pair Encoding), BBPE (Byte-level BPE). The symbol $|$ denotes a token boundary.}
    \centering
    \small
    \setlength{\tabcolsep}{4.5pt} % Snug padding for single-column fit
    \begin{tabular}{lll}
        \toprule
        \textbf{Model (Tokenizer)} & \textbf{Input String} & \textbf{Tokenized Output} \\
        \midrule
        LLaMA-3 (BPE)& Running & Running \\
                     & RUNNING & RUN $|$ NING \\
                     & running & running \\
                     & RuNnInG & Ru $|$ N $|$ n $|$ In $|$ G \\
                     \addlinespace
                     & Unbelievably & Un $|$ belie $|$ vably \\
                     & UNBELIEVABLY & UN $|$ BEL $|$ IE $|$ V $|$ AB $|$ LY \\
                     & unbelievably & un $|$ belie $|$ vably \\
                     & UnBeLiEvAbLy & Un $|$ Be $|$ Li $|$ Ev $|$ Ab $|$ Ly \\
        \midrule
        Gemma-2 (SP) & Running & Running \\
                     & RUNNING & RUNNING \\
                     & running & running \\
                     & RuNnInG & Ru $|$ N $|$ n $|$ In $|$ G \\
                     \addlinespace
                     & Unbelievably & Un $|$ bel $|$ iev $|$ ably \\
                     & UNBELIEVABLY & UN $|$ BEL $|$ IEV $|$ AB $|$ LY \\
                     & unbelievably & un $|$ bel $|$ iev $|$ ably \\
                     & UnBeLiEvAbLy & Un $|$ Be $|$ Li $|$ Ev $|$ Ab $|$ Ly \\
        \midrule
        Gemma-3 (SP) & Running & Running \\
                     & RUNNING & RUN $|$ NING \\
                     & running & running \\
                     & RuNnInG & Ru $|$ N $|$ n $|$ In $|$ G \\
                     \addlinespace
                     & Unbelievably & Un $|$ bel $|$ iev $|$ ably \\
                     & UNBELIEVABLY & UN $|$ BEL $|$ IE $|$ V $|$ AB $|$ LY \\
                     & unbelievably & un $|$ bel $|$ iev $|$ ably \\
                     & UnBeLiEvAbLy & Un $|$ Be $|$ Li $|$ Ev $|$ Ab $|$ Ly \\
        \midrule
        Mistral (SP) & Running & Running \\
                     & RUNNING & R $|$ UN $|$ NING \\
                     & running & running \\
                     & RuNnInG & Ru $|$ N $|$ n $|$ In $|$ G \\
                     \addlinespace
                     & Unbelievably & Un $|$ bel $|$ iev $|$ ably \\
                     & UNBELIEVABLY & UN $|$ B $|$ EL $|$ I $|$ EV $|$ AB $|$ LY \\
                     & unbelievably & un $|$ bel $|$ iev $|$ ably \\
                     & UnBeLiEvAbLy & Un $|$ Be $|$ Li $|$ Ev $|$ Ab $|$ L $|$ y \\
        \midrule
        Qwen2.5 (BBPE) & Running & Running \\
                     & RUNNING & RUN $|$ NING \\
                     & running & running \\
                     & RuNnInG & Ru $|$ N $|$ n $|$ In $|$ G \\
                     \addlinespace
                     & Unbelievably & Un $|$ belie $|$ vably \\
                     & UNBELIEVABLY & UN $|$ BEL $|$ IE $|$ V $|$ AB $|$ LY \\
                     & unbelievably & un $|$ belie $|$ vably \\
                     & UnBeLiEvAbLy & n $|$ Be $|$ Li $|$ Ev $|$ Ab $|$ Ly \\
        \bottomrule
        \end{tabular}
    \label{tab:tokenizer_comparison}
\end{table}

\clearpage \newpage
\begin{table}[!ht]
    \caption{\textbf{Qualitative comparison on letter case sensitivity of tokenization schemes of both used reasoning models.} BPE (Byte-Pair Encoding), BBPE (Byte-level BPE). The symbol $|$ denotes a token boundary.}
    \centering
    \small
    \setlength{\tabcolsep}{4.5pt} % Snug padding for single-column fit
    \begin{tabular}{lll}
        \toprule
        \textbf{Model (Tokenizer)} & \textbf{Input String} & \textbf{Tokenized Output} \\
        \midrule
        Qwen3 (BBPE) & Running & Running \\
                     & RUNNING & RUN $|$ NING \\
                     & running & running \\
                     & RuNnInG & Ru $|$ N $|$ n $|$ In $|$ G \\
                     \addlinespace
                     & Unbelievably & Un $|$ bel $|$ ievably \\
                     & UNBELIEVABLY & UN $|$ BEL $|$ IE $|$ V $|$ AB $|$ LY \\
                     & unbelievably & un $|$ bel $|$ ievably \\
                     & UnBeLiEvAbLy & Un $|$ Be $|$ Li $|$ Ev $|$ Ab $|$ Ly \\
        \midrule
        GPT (BPE)    & Running & Running \\
                     & RUNNING & RUN $|$ NING \\
                     & running & running \\
                     & RuNnInG & Ru $|$ N $|$ n $|$ In $|$ G \\
                     \addlinespace
                     & Unbelievably & Un $|$ belie $|$ vably \\
                     & UNBELIEVABLY & UN $|$ BEL $|$ IE $|$ V $|$ AB $|$ LY \\
                     & unbelievably & un $|$ belie $|$ vably \\
                     & UnBeLiEvAbLy & Un $|$ Be $|$ Li $|$ Ev $|$ Ab $|$ Ly \\
            \bottomrule
        \end{tabular}
    \label{tab:tokenizer_comparison_reasoning}
\end{table}

\begin{table}[!ht]
    \caption{\textbf{Qualitative comparison on letter case sensitivity of tokenization schemes of VLMs.} SP (SentencePiece), BPE (Byte-Pair Encoding), BBPE (Byte-level BPE). The symbol $|$ denotes a token boundary.}
    \centering
    \small
    \setlength{\tabcolsep}{4.5pt} % Snug padding for single-column fit
    \begin{tabular}{lll}
        \toprule
        \textbf{Model (Tokenizer)} & \textbf{Input String} & \textbf{Tokenized Output} \\
        \midrule
        Qwen3-VL (BBPE)     & Running & Running \\
                            & RUNNING & RUN $|$ NING \\
                            & running & running \\
                            & RuNnInG & Ru $|$ N $|$ n $|$ In $|$ G \\
                            \addlinespace
                            & Unbelievably & Un $|$ belie $|$ vably \\
                            & UNBELIEVABLY & UN $|$ BEL $|$ IE $|$ V $|$ AB $|$ LY \\
                            & unbelievably & un $|$ belie $|$ vably \\
                            & UnBeLiEvAbLy & Un $|$ Be $|$ Li $|$ Ev $|$ Ab $|$ Ly \\
        \midrule
        Gemma-3 (SP) & & see \Cref{tab:tokenizer_comparison} \\
        \midrule
        Gemma-4 (SP) & Running & Running \\
                     & RUNNING & RUN $|$ NING \\
                     & running & running \\
                     & RuNnInG & Ru $|$ N $|$ n $|$ In $|$ G \\
                     \addlinespace
                     & Unbelievably & Un $|$ bel $|$ iev $|$ ably \\
                     & UNBELIEVABLY & UN $|$ BEL $|$ IE $|$ V $|$ AB $|$ LY \\
                     & unbelievably & un $|$ bel $|$ iev $|$ ably \\
                     & UnBeLiEvAbLy & Un $|$ Be $|$ Li $|$ Ev $|$ Ab $|$ Ly \\
            \bottomrule
        \end{tabular}
    \label{tab:tokenizer_comparison_VLM}
\end{table}

%% file: main.bib
@misc{Mao_2016_RefCOCO,
      title={Generation and Comprehension of Unambiguous Object Descriptions}, 
      author={Junhua Mao and Jonathan Huang and Alexander Toshev and Oana Camburu and Alan Yuille and Kevin Murphy},
      year={2016},
      eprint={1511.02283},
      archivePrefix={arXiv},
      primaryClass={cs.CV},
      url={https://arxiv.org/abs/1511.02283}, 
}

@misc{Wang_2024_arXiv_MMLUPro,
      title={MMLU-Pro: A More Robust and Challenging Multi-Task Language Understanding Benchmark}, 
      author={Yubo Wang and Xueguang Ma and Ge Zhang and Yuansheng Ni and Abhranil Chandra and Shiguang Guo and Weiming Ren and Aaran Arulraj and Xuan He and Ziyan Jiang and Tianle Li and Max Ku and Kai Wang and Alex Zhuang and Rongqi Fan and Xiang Yue and Wenhu Chen},
      year={2024},
      eprint={2406.01574},
      archivePrefix={arXiv},
      primaryClass={cs.CL},
      url={https://arxiv.org/abs/2406.01574}, 
}

@misc{Kudo_2018_arXiv_SentencePiece,
      title={SentencePiece: A simple and language independent subword tokenizer and detokenizer for Neural Text Processing}, 
      author={Taku Kudo and John Richardson},
      year={2018},
      eprint={1808.06226},
      archivePrefix={arXiv},
      primaryClass={cs.CL},
      url={https://arxiv.org/abs/1808.06226}, 
}

@misc{Grattafiori_2024_arXiv_LLaMA3,
      title={The Llama 3 Herd of Models}, 
      author={Aaron Grattafiori and Abhimanyu Dubey and Abhinav Jauhri and others},
      year={2024},
      eprint={2407.21783},
      archivePrefix={arXiv},
      primaryClass={cs.AI},
      url={https://arxiv.org/abs/2407.21783}, 
}

@misc{llama3.2_3b_instruct,
  author = {Meta},
  title = {Llama-3.2-3B-Instruct},
  year = {2024},
  publisher = {Hugging Face},
  journal = {Hugging Face Repository},
  howpublished = {\url{https://huggingface.co/meta-llama/Llama-3.2-3B-Instruct}},
}

@misc{Jiang_2023_arXiv_Mistral7b,
      title={Mistral 7B}, 
      author={Albert Q. Jiang and Alexandre Sablayrolles and Arthur Mensch and Chris Bamford and Devendra Singh Chaplot and Diego de las Casas and Florian Bressand and Gianna Lengyel and Guillaume Lample and Lucile Saulnier and Lélio Renard Lavaud and Marie-Anne Lachaux and Pierre Stock and Teven Le Scao and Thibaut Lavril and Thomas Wang and Timothée Lacroix and William El Sayed},
      year={2023},
      eprint={2310.06825},
      archivePrefix={arXiv},
      primaryClass={cs.CL},
      url={https://arxiv.org/abs/2310.06825}, 
}

@misc{mistral_7b_instruct_v03,
  author = {{Mistral AI}},
  title = {Mistral-7B-Instruct-v0.3},
  year = {2024},
  publisher = {Hugging Face},
  journal = {Hugging Face Repository},
  howpublished = {\url{https://huggingface.co/mistralai/Mistral-7B-Instruct-v0.3}},
}

@misc{Qwen_2024_arXiv_Qwen2,
      title={Qwen2.5 Technical Report}, 
      author={{Qwen Team}},
      year={2025},
      eprint={2412.15115},
      archivePrefix={arXiv},
      primaryClass={cs.CL},
      url={https://arxiv.org/abs/2412.15115}, 
}

@misc{Qwen_2025_arXiv_Qwen3,
  title={Qwen3 Technical Report}, 
  author={{Qwen Team}},
  year={2025},
  eprint={2505.09388},
  archivePrefix={arXiv},
  primaryClass={cs.CL},
  url={https://arxiv.org/abs/2505.09388}, 
}

@misc{qwen2.5_7b_instruct,
  author = {{Qwen Team}},
  title = {Qwen2.5-7B-Instruct},
  year = {2024},
  publisher = {Hugging Face},
  journal = {Hugging Face Repository},
  howpublished = {\url{https://huggingface.co/Qwen/Qwen2.5-7B-Instruct}},
}

@misc{qwen3_4b_thinking_2507,
  author = {{Qwen Team}},
  title = {Qwen3-4B-Thinking-2507},
  year = {2025},
  publisher = {Hugging Face},
  journal = {Hugging Face Repository},
  howpublished = {\url{https://huggingface.co/Qwen/Qwen3-4B-Thinking-2507}},
}

@misc{Gemma2_arXiv_TechnicalReport,
      title={Gemma 2: Improving Open Language Models at a Practical Size}, 
      author={Gemma Team and Morgane Riviere and Shreya Pathak and others},
      year={2024},
      eprint={2408.00118},
      archivePrefix={arXiv},
      primaryClass={cs.CL},
      url={https://arxiv.org/abs/2408.00118}, 
}

@misc{Gemma3_arXiv_TechnicalReport,
      title={Gemma 3 Technical Report}, 
      author={Gemma Team and Aishwarya Kamath and Johan Ferret and others},
      year={2025},
      eprint={2503.19786},
      archivePrefix={arXiv},
      primaryClass={cs.CL},
      url={https://arxiv.org/abs/2503.19786}, 
}

@misc{gemma_3_1b_it,
  author = {{Google Team}},
  title = {Gemma-3-1B-IT},
  year = {2025},
  publisher = {Hugging Face},
  journal = {Hugging Face Repository},
  howpublished = {\url{https://huggingface.co/google/gemma-3-1b-it}},
}

@misc{gemma_2_2b_it,
  author = {{Google Team}},
  title = {Gemma-2-2B-IT},
  year = {2024},
  publisher = {Hugging Face},
  journal = {Hugging Face Repository},
  howpublished = {\url{https://huggingface.co/google/gemma-2-2b-it}},
}

@misc{openai_2025_gptoss20b,
      title={gpt-oss-120b \& gpt-oss-20b Model Card}, 
      author={{OpenAI}},
      year={2025},
      eprint={2508.10925},
      archivePrefix={arXiv},
      primaryClass={cs.CL},
      url={https://arxiv.org/abs/2508.10925}, 
}

@misc{llama3.1_8b_instruct,
  author = {Meta},
  title = {Llama-3.1-8B-Instruct},
  year = {2024},
  publisher = {Hugging Face},
  journal = {Hugging Face Repository},
  howpublished = {\url{https://huggingface.co/meta-llama/Llama-3.1-8B-Instruct}},
}

@misc{qwen2.5_14b_instruct,
  author = {{Qwen Team}},
  title = {Qwen2.5-14B-Instruct},
  year = {2024},
  publisher = {Hugging Face},
  journal = {Hugging Face Repository},
  howpublished = {\url{https://huggingface.co/Qwen/Qwen2.5-14B-Instruct}},
}

@misc{Bai_2023_arXiv_QwensTokenizer,
      title={Qwen Technical Report}, 
      author={Jinze Bai and Shuai Bai and Yunfei Chu and others},
      year={2023},
      eprint={2309.16609},
      archivePrefix={arXiv},
      primaryClass={cs.CL},
      url={https://arxiv.org/abs/2309.16609}, 
}

@misc{Chollet_2019_arXiv_ARC,
      title={On the Measure of Intelligence}, 
      author={François Chollet},
      year={2019},
      eprint={1911.01547},
      archivePrefix={arXiv},
      primaryClass={cs.AI},
      url={https://arxiv.org/abs/1911.01547}, 
}

@misc{Rajpurkar_2018_ArXiv_SQuADv2,
      title={Know What You Don't Know: Unanswerable Questions for SQuAD}, 
      author={Pranav Rajpurkar and Robin Jia and Percy Liang},
      year={2018},
      eprint={1806.03822},
      archivePrefix={arXiv},
      primaryClass={cs.CL},
      url={https://arxiv.org/abs/1806.03822}, 
}

@inproceedings{Meng_2022_NIPS_ROME,
 author = {Meng, Kevin and Bau, David and Andonian, Alex and Belinkov, Yonatan},
 booktitle = {Advances in Neural Information Processing Systems},
 editor = {S. Koyejo and S. Mohamed and A. Agarwal and D. Belgrave and K. Cho and A. Oh},
 pages = {17359--17372},
 publisher = {Curran Associates, Inc.},
 title = {Locating and Editing Factual Associations in GPT},
 url = {https://proceedings.neurips.cc/paper_files/paper/2022/file/6f1d43d5a82a37e89b0665b33bf3a182-Paper-Conference.pdf},
 volume = {35},
 year = {2022}
}

@inproceedings{Tamayo_2024_ACL_MEMAT,
    title = "Mass-Editing Memory with Attention in Transformers: A cross-lingual exploration of knowledge",
    author = "Tamayo, Daniel  and Gonzalez-Agirre, Aitor  and Hernando, Javier  and Villegas, Marta",
    editor = "Ku, Lun-Wei  and Martins, Andre  and Srikumar, Vivek",
    booktitle = "Findings of the Association for Computational Linguistics: ACL 2024",
    month = "8",
    year = "2024",
    address = "Bangkok, Thailand",
    publisher = "Association for Computational Linguistics",
    url = "https://aclanthology.org/2024.findings-acl.347/",
    doi = "10.18653/v1/2024.findings-acl.347",
    pages = "5831--5847",
}

@misc{Sinii_2025_arXiv_SteeringLLMReasonging,
      title={Steering LLM Reasoning Through Bias-Only Adaptation}, 
      author={Viacheslav Sinii and Alexey Gorbatovski and Artem Cherepanov and Boris Shaposhnikov and Nikita Balagansky and Daniil Gavrilov},
      year={2025},
      eprint={2505.18706},
      archivePrefix={arXiv},
      primaryClass={cs.LG},
      url={https://arxiv.org/abs/2505.18706}, 
}

@misc{Bu_2026_arXiv_VGA,
      title={Value-State Gated Attention for Mitigating Extreme-Token Phenomena in Transformers}, 
      author={Rui Bu and Haofeng Zhong and Wenzheng Chen and Yangyan Li},
      year={2026},
      eprint={2510.09017},
      archivePrefix={arXiv},
      primaryClass={cs.LG},
      url={https://arxiv.org/abs/2510.09017}, 
}

@misc{Zhang_2024_arXiv_PASTA,
      title={Tell Your Model Where to Attend: Post-hoc Attention Steering for LLMs}, 
      author={Qingru Zhang and Chandan Singh and Liyuan Liu and Xiaodong Liu and Bin Yu and Jianfeng Gao and Tuo Zhao},
      year={2024},
      eprint={2311.02262},
      archivePrefix={arXiv},
      primaryClass={cs.CL},
      url={https://arxiv.org/abs/2311.02262}, 
}

@misc{Zhang_2024_arXiv_AutoPASTA,
      title={Model Tells Itself Where to Attend: Faithfulness Meets Automatic Attention Steering}, 
      author={Qingru Zhang and Xiaodong Yu and Chandan Singh and Xiaodong Liu and Liyuan Liu and Jianfeng Gao and Tuo Zhao and Dan Roth and Hao Cheng},
      year={2024},
      eprint={2409.10790},
      archivePrefix={arXiv},
      primaryClass={cs.CL},
      url={https://arxiv.org/abs/2409.10790}, 
}

@misc{Guardieiro_2025_arXiv_InstABoost,
      title={Instruction Following by Boosting Attention of Large Language Models}, 
      author={Vitoria Guardieiro and Adam Stein and Avishree Khare and Eric Wong},
      year={2025},
      eprint={2506.13734},
      archivePrefix={arXiv},
      primaryClass={cs.CL},
      url={https://arxiv.org/abs/2506.13734}, 
}

@misc{Davarmanesh_2026_arXiv_EfficientSteering,
      title={Efficient and accurate steering of Large Language Models through attention-guided feature learning}, 
      author={Parmida Davarmanesh and Ashia Wilson and Adityanarayanan Radhakrishnan},
      year={2026},
      eprint={2602.00333},
      archivePrefix={arXiv},
      primaryClass={cs.LG},
      url={https://arxiv.org/abs/2602.00333}, 
}

@misc{Li_2025_arXiv_CTR_SinkTokens,
      title={CTR-Sink: Attention Sink for Language Models in Click-Through Rate Prediction}, 
      author={Zixuan Li and Binzong Geng and Jing Xiong and Yong He and Yuxuan Hu and Jian Chen and Dingwei Chen and Xiyu Chang and Liang Zhang and Linjian Mo and Chengming Li and Chuan Yuan and Zhenan Sun},
      year={2025},
      eprint={2508.03668},
      archivePrefix={arXiv},
      primaryClass={cs.CL},
      url={https://arxiv.org/abs/2508.03668}, 
}

@misc{Phan_2024_arXiv_TokenBias,
      title={Understanding and Mitigating Tokenization Bias in Language Models}, 
      author={Buu Phan and Marton Havasi and Matthew Muckley and Karen Ullrich},
      year={2024},
      eprint={2406.16829},
      archivePrefix={arXiv},
      primaryClass={cs.CL},
      url={https://arxiv.org/abs/2406.16829}, 
}

@misc{Wu_2025_arXiv_PositionBias,
      title={On the Emergence of Position Bias in Transformers}, 
      author={Xinyi Wu and Yifei Wang and Stefanie Jegelka and Ali Jadbabaie},
      year={2025},
      eprint={2502.01951},
      archivePrefix={arXiv},
      primaryClass={cs.LG},
      url={https://arxiv.org/abs/2502.01951}, 
}

@misc{Alpay_2025_arXiv_XLMPrompting,
      title={XML Prompting as Grammar-Constrained Interaction: Fixed-Point Semantics, Convergence Guarantees, and Human-AI Protocols}, 
      author={Faruk Alpay and Taylan Alpay},
      year={2025},
      eprint={2509.08182},
      archivePrefix={arXiv},
      primaryClass={cs.PL},
      url={https://arxiv.org/abs/2509.08182}, 
}

@article{Sinclair_2022_TACL_StructuralPriming,
    author = {Sinclair, Arabella and Jumelet, Jaap and Zuidema, Willem and Fernández, Raquel},
    title = {Structural Persistence in Language Models: Priming as a Window into Abstract Language Representations},
    journal = {Transactions of the Association for Computational Linguistics},
    volume = {10},
    pages = {1031-1050},
    year = {2022},
    month = {09},
    issn = {2307-387X},
    doi = {10.1162/tacl_a_00504},
    url = {https://doi.org/10.1162/tacl_a_00504},
}

@misc{Stolfo_2025_arXiv_InstructionSteering,
      title={Improving Instruction-Following in Language Models through Activation Steering}, 
      author={Alessandro Stolfo and Vidhisha Balachandran and Safoora Yousefi and Eric Horvitz and Besmira Nushi},
      year={2025},
      eprint={2410.12877},
      archivePrefix={arXiv},
      primaryClass={cs.CL},
      url={https://arxiv.org/abs/2410.12877}, 
}

@misc{vonRuette_2024_arXiv_ActivationSteering,
      title={A Language Model's Guide Through Latent Space}, 
      author={Dimitri von Rütte and Sotiris Anagnostidis and Gregor Bachmann and Thomas Hofmann},
      year={2024},
      eprint={2402.14433},
      archivePrefix={arXiv},
      primaryClass={cs.CL},
      url={https://arxiv.org/abs/2402.14433}, 
}

@misc{Lee_2025_arXiv_CAST,
      title={Programming Refusal with Conditional Activation Steering}, 
      author={Bruce W. Lee and Inkit Padhi and Karthikeyan Natesan Ramamurthy and Erik Miehling and Pierre Dognin and Manish Nagireddy and Amit Dhurandhar},
      year={2025},
      eprint={2409.05907},
      archivePrefix={arXiv},
      primaryClass={cs.LG},
      url={https://arxiv.org/abs/2409.05907}, 
}

@misc{Venkateswaran_2026_arXiv_SpotLight,
      title={Spotlight Your Instructions: Instruction-following with Dynamic Attention Steering}, 
      author={Praveen Venkateswaran and Danish Contractor},
      year={2026},
      eprint={2505.12025},
      archivePrefix={arXiv},
      primaryClass={cs.LG},
      url={https://arxiv.org/abs/2505.12025}, 
}

@misc{Turner_2024_arXiv_ActAdd,
      title={Steering Language Models With Activation Engineering}, 
      author={Alexander Matt Turner and Lisa Thiergart and Gavin Leech and David Udell and Juan J. Vazquez and Ulisse Mini and Monte MacDiarmid},
      year={2024},
      eprint={2308.10248},
      archivePrefix={arXiv},
      primaryClass={cs.CL},
      url={https://arxiv.org/abs/2308.10248}, 
}

@misc{Hernandez_2024_arXiv_REMEDI,
      title={Inspecting and Editing Knowledge Representations in Language Models}, 
      author={Evan Hernandez and Belinda Z. Li and Jacob Andreas},
      year={2024},
      eprint={2304.00740},
      archivePrefix={arXiv},
      primaryClass={cs.CL},
      url={https://arxiv.org/abs/2304.00740}, 
}

@misc{Mitchell_2022_arXiv_MEND,
      title={Fast Model Editing at Scale}, 
      author={Eric Mitchell and Charles Lin and Antoine Bosselut and Chelsea Finn and Christopher D. Manning},
      year={2022},
      eprint={2110.11309},
      archivePrefix={arXiv},
      primaryClass={cs.LG},
      url={https://arxiv.org/abs/2110.11309}, 
}

@inproceedings{Hsieh_2024_ACL_FoundInTheMiddle,
    title = "Found in the middle: Calibrating Positional Attention Bias Improves Long Context Utilization",
    author = "Hsieh, Cheng-Yu  and Chuang, Yung-Sung  and Li, Chun-Liang  and Wang, Zifeng  and Le, Long  and Kumar, Abhishek  and Glass, James  and Ratner, Alexander  and Lee, Chen-Yu  and Krishna, Ranjay  and Pfister, Tomas",
    editor = "Ku, Lun-Wei  and Martins, Andre  and Srikumar, Vivek",
    booktitle = "Findings of the Association for Computational Linguistics: ACL 2024",
    month = "08",
    year = "2024",
    address = "Bangkok, Thailand",
    publisher = "Association for Computational Linguistics",
    doi = "10.18653/v1/2024.findings-acl.890",
    pages = "14982--14995",
}

@inproceedings{Anonymous_2026_OpenReview_SEKA,
    title={Spectral Attention Steering for Prompt Highlighting},
    author={Anonymous},
    booktitle={The Fourteenth International Conference on Learning Representations},
    year={2026},
    url={https://openreview.net/forum?id=XfLvGIFmAN}
}

@article{Klinke_2024_Advertising,
    author = {Tobias Klinke and Malte Christ and Nader Fadl and Charlotte Lamerz and Tobias Langner},
    title = {The effects of letter capitalization in advertising headlines},
    journal = {Journal of Marketing Communications},
    volume = {0},
    number = {0},
    pages = {1--23},
    year = {2024},
    publisher = {Routledge},
    doi = {10.1080/13527266.2024.2401393},
    eprint = {https://doi.org/10.1080/13527266.2024.2401393}
}

@article{Fournet_2022_APA_Effects,
  title={Effects of letter case on processing sequences of written words.},
  author={Fournet, Colas and Mirault, Jonathan and Perea, Manuel and Grainger, Jonathan},
  journal={Journal of experimental psychology: learning, memory, and cognition},
  volume={48},
  number={12},
  pages={1995},
  year={2022},
  publisher={American Psychological Association}
}

@article{Vergara_2020_Lettercase,
    title = {The time course of the lowercase advantage in visual word recognition: An ERP investigation},
    journal = {Neuropsychologia},
    volume = {146},
    pages = {107556},
    year = {2020},
    issn = {0028-3932},
    doi = {https://doi.org/10.1016/j.neuropsychologia.2020.107556},
    author = {Marta Vergara-Martínez and Manuel Perea and Barbara Leone-Fernandez},
}

@article{Cutter_2020_APA_Capitalization,
  title={Capitalization interacts with syntactic complexity.},
  author={Cutter, Michael G and Martin, Andrea E and Sturt, Patrick},
  journal={Journal of Experimental Psychology: Learning, Memory, and Cognition},
  volume={46},
  number={6},
  pages={1146},
  year={2020},
  publisher={American Psychological Association}
}

@article{Slattery_2011_APA_UppercaseFixation,
  title={Parafoveal and foveal processing of abbreviations during eye fixations in reading: Making a case for case.},
  author={Slattery, Timothy J and Schotter, Elizabeth R and Berry, Raymond W and Rayner, Keith},
  journal={Journal of Experimental Psychology: Learning, Memory, and Cognition},
  volume={37},
  number={4},
  pages={1022},
  year={2011},
  publisher={American Psychological Association}
}

@article{Inhoff_2000_Springer_Uppercase,
  author  = {Inhoff, Albrecht W. and Starr, Matthew and Shindler, Kelley L.},
  title   = {Is the processing of words during eye fixations in reading strictly serial?},
  journal = {Perception \& Psychophysics},
  year    = {2000},
  volume  = {62},
  number  = {7},
  pages   = {1474--1484},
  doi     = {10.3758/BF03212147},
  url     = {https://doi.org/10.3758/BF03212147},
  issn    = {1532-5962},
}

@misc{singh2025openaigpt5card,
      title={OpenAI GPT-5 System Card}, 
      author={Aaditya Singh and Adam Fry and Adam Perelman and others},
      year={2025},
      eprint={2601.03267},
      archivePrefix={arXiv},
      primaryClass={cs.CL},
      url={https://arxiv.org/abs/2601.03267}, 
}

@misc{tiktoken,
  author = {{OpenAI}},
  title = {tiktoken: A fast BPE tokeniser for use with OpenAI's models},
  year = {2023},
  publisher = {GitHub},
  journal = {GitHub repository},
  howpublished = {\url{https://github.com/openai/tiktoken}},
}

@inproceedings{Artetxe_2020_XQuAD,
   title={On the Cross-lingual Transferability of Monolingual Representations},
   url={http://dx.doi.org/10.18653/v1/2020.acl-main.421},
   DOI={10.18653/v1/2020.acl-main.421},
   booktitle={Proceedings of the 58th Annual Meeting of the Association for Computational Linguistics},
   publisher={Association for Computational Linguistics},
   author={Artetxe, Mikel and Ruder, Sebastian and Yogatama, Dani},
   year={2020},
   pages={4623–4637} 
}

@misc{Chen_2021_HumanEval,
      title={Evaluating Large Language Models Trained on Code}, 
      author={Mark Chen and Jerry Tworek and Heewoo Jun and Qiming Yuan and Henrique Ponde de Oliveira Pinto and Jared Kaplan and Harri Edwards and Yuri Burda and Nicholas Joseph and Greg Brockman and Alex Ray and Raul Puri and Gretchen Krueger and Michael Petrov and Heidy Khlaaf and Girish Sastry and Pamela Mishkin and Brooke Chan and Scott Gray and Nick Ryder and Mikhail Pavlov and Alethea Power and Lukasz Kaiser and Mohammad Bavarian and Clemens Winter and Philippe Tillet and Felipe Petroski Such and Dave Cummings and Matthias Plappert and Fotios Chantzis and Elizabeth Barnes and Ariel Herbert-Voss and William Hebgen Guss and Alex Nichol and Alex Paino and Nikolas Tezak and Jie Tang and Igor Babuschkin and Suchir Balaji and Shantanu Jain and William Saunders and Christopher Hesse and Andrew N. Carr and Jan Leike and Josh Achiam and Vedant Misra and Evan Morikawa and Alec Radford and Matthew Knight and Miles Brundage and Mira Murati and Katie Mayer and Peter Welinder and Bob McGrew and Dario Amodei and Sam McCandlish and Ilya Sutskever and Wojciech Zaremba},
      year={2021},
      eprint={2107.03374},
      archivePrefix={arXiv},
      primaryClass={cs.LG},
      url={https://arxiv.org/abs/2107.03374}, 
}

@misc{wu2025textdominance,
      title={When Language Overrules: Revealing Text Dominance in Multimodal Large Language Models}, 
      author={Huyu Wu and Meng Tang and Xinhan Zheng and Haiyun Jiang},
      year={2025},
      eprint={2508.10552},
      archivePrefix={arXiv},
      primaryClass={cs.CL},
      url={https://arxiv.org/abs/2508.10552}, 
}

@misc{bai2025qwen3vltechnicalreport,
      title={Qwen3-VL Technical Report}, 
      author={Shuai Bai and Yuxuan Cai and Ruizhe Chen and Keqin Chen and Xionghui Chen and Zesen Cheng and Lianghao Deng and Wei Ding and Chang Gao and Chunjiang Ge and Wenbin Ge and Zhifang Guo and Qidong Huang and Jie Huang and Fei Huang and Binyuan Hui and Shutong Jiang and Zhaohai Li and Mingsheng Li and Mei Li and Kaixin Li and Zicheng Lin and Junyang Lin and Xuejing Liu and Jiawei Liu and Chenglong Liu and Yang Liu and Dayiheng Liu and Shixuan Liu and Dunjie Lu and Ruilin Luo and Chenxu Lv and Rui Men and Lingchen Meng and Xuancheng Ren and Xingzhang Ren and Sibo Song and Yuchong Sun and Jun Tang and Jianhong Tu and Jianqiang Wan and Peng Wang and Pengfei Wang and Qiuyue Wang and Yuxuan Wang and Tianbao Xie and Yiheng Xu and Haiyang Xu and Jin Xu and Zhibo Yang and Mingkun Yang and Jianxin Yang and An Yang and Bowen Yu and Fei Zhang and Hang Zhang and Xi Zhang and Bo Zheng and Humen Zhong and Jingren Zhou and Fan Zhou and Jing Zhou and Yuanzhi Zhu and Ke Zhu},
      year={2025},
      eprint={2511.21631},
      archivePrefix={arXiv},
      primaryClass={cs.CV},
      url={https://arxiv.org/abs/2511.21631}, 
}

@misc{google_gemma_core_en,
  author       = {{Google Developers}},
  title        = {{Gemma core models}},
  howpublished = {\url{https://ai.google.dev/gemma/docs/core}},
  year         = {2026},
  note         = {Accessed: 2026-06-21}
}

@misc{qwen3vl_4b_instruct,
  author = {{Qwen Team}},
  title = {Qwen3-VL-4B-Instruct},
  year = {2025},
  publisher = {Hugging Face},
  journal = {Hugging Face Repository},
  howpublished = {\url{https://huggingface.co/Qwen/Qwen3-VL-4B-Instruct}},
}

@misc{qwen3vl_2b_thinking,
  author = {{Qwen Team}},
  title = {Qwen3-VL-2B-Thinking},
  year = {2025},
  publisher = {Hugging Face},
  journal = {Hugging Face Repository},
  howpublished = {\url{https://huggingface.co/Qwen/Qwen3-VL-2B-Thinking}},
}

@misc{gemma_3_4b_it,
  author = {{Google Team}},
  title = {Gemma-3-4B-IT},
  year = {2025},
  publisher = {Hugging Face},
  journal = {Hugging Face Repository},
  howpublished = {\url{https://huggingface.co/google/gemma-3-4b-it}},
}

@misc{gemma_4_e4b_it,
  author = {{Google Team}},
  title = {Gemma-4-E4B-IT},
  year = {2026},
  publisher = {Hugging Face},
  journal = {Hugging Face Repository},
  howpublished = {\url{https://huggingface.co/google/gemma-4-E4B-it}},
}
